%% file: main.tex
\documentclass{article}

\usepackage{PRIMEarxiv}

\usepackage[utf8]{inputenc} 
\usepackage[T1]{fontenc}    
\usepackage{hyperref}       
\usepackage{url}            
\usepackage{booktabs}       
\usepackage{amsfonts}       
\usepackage{nicefrac}       
\usepackage{microtype}      
\usepackage{lipsum}
\usepackage{fancyhdr}       
\usepackage{graphicx}       
\graphicspath{{media/}}     

\usepackage{acronym}
\usepackage{tikz}
\usetikzlibrary{positioning,shapes.geometric}
\usepackage{subcaption}
\usepackage{xltabular}
\usepackage{algorithm}		  
\usepackage{algorithmic}	  
\usepackage{import}

\newcommand{\othertab}[2]{
    \begin{xltabular}{\textwidth}{|c|X|c|}
        \multicolumn{3}{@{}c}{Short overview of other usages of #1:}\\
        \hline
        \textbf{Approach} & \textbf{Description} & \textbf{Year} \\\hline
        \endfirsthead
        \multicolumn{3}{@{}l}{\ldots\ \small Continuation}\\\hline
        \endhead
        \hline
        \multicolumn{3}{@{}r}{\small Continuation\ldots}\\
        \endfoot
        \hline
        \endlastfoot
        #2
    \end{xltabular}
}

\usepackage{amsmath}
\usepackage{amsfonts}
\usepackage{amssymb}

\DeclareMathOperator*{\argmin}{arg\,min}
\DeclareMathOperator*{\argmax}{arg\,max}
\DeclareMathOperator*{\softmax}{softmax}

\usepackage{pgf}
\pgfdeclarelayer{bg}    
\pgfsetlayers{bg,main}  

\title{Comprehensive Exploration of Synthetic Data Generation: A Survey
}

\author{
  André Bauer \\
  University of Chicago \\
  Chicago, United States of America\\
  \texttt{andrebauer@uchicago.edu} \\
   \And
  Simon Trapp, Michael Stenger, Robert Leppich, Samuel Kounev \\
  University of Würzburg \\
  Würzburg, Germany\\
  \texttt{\{firstname\}.\{lastname\}@uni-wuerzburg.de} \\
   \And
  Mark Leznik \\
  University of Ulm \\
  Ulm, Germany\\
  \texttt{mark.leznik@uni-ulm.de} \\
  \And
  Kyle Chard, Ian Foster \\
  Argonne National Laboratory \\
  Lemont, United States of America\\
  \texttt{\{lastname\}@anl.gov}
}

\newcommand\works{417}

\begin{document}
\maketitle

\begin{abstract}
Recent years have witnessed a surge in the popularity of \ac{ML}, applied across diverse domains. However, progress is impeded by the scarcity of training data due to expensive acquisition and privacy legislation. Synthetic data emerges as a solution, but the abundance of released models and limited overview literature pose challenges for decision-making. This work surveys \works{} \ac{SDG} models over the last decade, providing a comprehensive overview of model types, functionality, and improvements. Common attributes are identified, leading to a classification and trend analysis. The findings reveal increased model performance and complexity, with neural network-based approaches prevailing, except for privacy-preserving data generation. Computer vision dominates, with \acsp{GAN} as primary generative models, while diffusion models, transformers, and \acsp{RNN} compete. Implications from our performance evaluation highlight the scarcity of common metrics and datasets, making comparisons challenging. Additionally, the neglect of training and computational costs in literature necessitates attention in future research. This work serves as a guide for \ac{SDG} model selection and identifies crucial areas for future exploration.
\end{abstract}

\keywords{Survey \and Synthesis \and Synthetic Data Generation}

\input{introduction}
\input{models}
\input{classification}
\input{relatedwork}
\input{conclusion}

\bibliographystyle{unsrt}  
\bibliography{combined}  

\appendix

\include{acronyms_sorted}

\end{document}

%% file: introduction.tex
\section{Introduction}
\label{ch:introduction}

In recent years, \acf{AI}, particularly in subfields like \acf{ML} and \acf{DL}, has experienced significant growth and popularity \cite{stathoulopoulos2018eating, nikolenko2021synthetic}. As \ac{ML} and \ac{DL} models have evolved in complexity and efficiency, a persistent challenge has been the limited size of training datasets. This limitation stems from high labeling costs and privacy concerns, hindering the model's ability to generalize effectively. To address this issue, \acf{SDG} emerges as a viable solution, providing substantial amounts of artificial data for model training. This artificial data includes novel, diverse, and realistic samples, alleviating the constraints imposed by traditional datasets~\cite{nikolenko2021synthetic}.

Broadly speaking, \ac{SDG} involves generating artificial data and labels, aiming to emulate authentic samples closely. This process is automated by utilizing generative models that estimate the probability distribution of their training data. This sets it apart from data augmentation, which manipulates existing data, as \ac{SDG} generates new data by sampling from learned distributions. The advantages of synthetic data go beyond mere cost reduction, with its on-the-fly generation contributing to reduced computational time and addressing bias in data distribution. Synthetic data proves highly valuable when real data is insufficient, costly to label, or exhibits biased distributions.

The concept of synthetic data dates back to the 1960s and has evolved alongside the broader AI landscape. However, the dynamic nature of \ac{SDG}, with new approaches emerging annually, poses a challenge for researchers, especially those new to the field. Existing literature on \ac{SDG} often lacks a comprehensive overview and classification of approaches, making it difficult to stay current with generative models, their applications, and the relationships between them. Recent studies have summarized the literature on specific model types such as \acp{GAN} \cite{goodfellow2016nips, hong2019generative} and computer-rendered virtual 3D environments \cite{korakakis2018short, seib2020mixing}. Other works concentrate on distinct domains, including graph generation \cite{guo2020systematic}, computer vision \cite{gaidon2018reasonable, tsirikoglou2020survey}, text generation \cite{iqbal2020text}, music \cite{briot2017deep}, privacy \cite{dankar2022multi}, and molecular science \cite{jorgensen2018deep}. Additionally, some related literature has undertaken the task of compiling, comparing, or classifying various \ac{SDG} approaches comprehensively \cite{turhan2018recent, oussidi2018deep, harshvardhan2020comprehensive, eigenschink2021deep, nikolenko2021synthetic}. However, these works have limitations in scope, often overlooking recent advancements such as self-attention \cite{vaswani2017attention}. They may focus exclusively on a single domain, classify models based on only a few aspects, or provide coarse coverage of literature and model architecture.

To address these gaps in the existing literature, this work endeavors to provide a comprehensive overview of \ac{SDG}. More precisely, the contributions of this work include:
\begin{enumerate}
    \item \textbf{Surveying Literature:} We conduct an extensive survey of the literature from the last decade, aiming to comprehensively cover all model types suitable for \ac{SDG}. Our investigation involves scrutinizing \works{} models, revealing 20 distinct model types, further categorized into 42 subtypes.
    \item \textbf{Exploring Applications and Enhancements:} Within this survey, we delve into the applications and enhancements of the identified \ac{SDG} model types, providing insights into their respective practical implementations.
    \item \textbf{Introducing Classification Categories:} In addition to model identification, we introduce various categories for classifying the collected generative models. These categories include generated data types, performance, privacy considerations, and training processes.
    \item \textbf{Knowledge Foundation:} The acquired knowledge from this exploration serves as a solid foundation, providing a comprehensive understanding of the diverse landscape of \ac{SDG} model types.
    \item \textbf{Guideline Development:} Building upon this foundation, we develop a practical guideline. This guideline is tailored to facilitate the selection of an appropriate \ac{SDG} model type, offering valuable insights for researchers and practitioners in the field.
\end{enumerate}
 
Our survey of \works{} \ac{SDG} models reveals notable trends. We classified these models based on over ten criteria, demonstrating an evident increase in complexity and performance over the years. The evolution is marked by the shift from simpler probabilistic models like Markov chains and \acp{BN} to more sophisticated neural network-based approaches. Notably, in the realm of \ac{SDG}, computer vision stands out as the most popular application field. \acp{GAN} and diffusion models have emerged as top performers in this domain. For handling sequential data, such as music or text, \acp{RNN} dominate. Additionally, privacy-preserving data generation commonly employs models like Markov chains, \acp{BN}, and more advanced \acp{GAN}. The work also highlights challenges such as non-standardized evaluation metrics and datasets, offering solutions like building graphs of predecessors based on the models' performance evaluations \cite{borji2022pros}.

We are convinced that our work can serve as a valuable resource for newcomers and experienced researchers, providing an updated overview of the \ac{SDG} landscape and aiding in identifying suitable models and datasets for specific tasks. Furthermore, we present the first comprehensive classification of numerous model implementations spanning multiple domains, facilitating the identification of strengths and weaknesses in these models.

The remainder of this work is structured as follows: In \autoref{ch:models}, we introduce \textcolor{red}{42} different \ac{SDG} model types and highlight their usage in literature. In \autoref{ch:classification}, we classify and compare the found generative models according to different categories before discussing a guideline for choosing which \ac{SDG} model type is suitable for a given scenario. In \autoref{ch:relatedwork}, review existing surveys and distinguish the scope of our study from existing surveys. In \autoref{ch:conclusion}, we conclude our paper. In \autoref{sec:acronyms}, we list all used acronyms.

%% file: models.tex
\section{Overview of Generative Models}\label{ch:models}

Various approaches exist to generate synthetic data, ranging from models based on graphs or simple probabilistic assumptions to deep neural networks. The following sections describe the different available model architectures in general and provide a chronological overview of the recent literature regarding the usage of these models for \ac{SDG}. The structure is inspired by the work done by Harshvardhan et al.~\cite{harshvardhan2020comprehensive} and extended further to include missing models and some novel and important implementations we found. We focus on approaches released in the last ten years to keep this work within viable and reasonable bounds. We mainly restrict ourselves to models referenced by the works mentioned in \autoref{ch:relatedwork}, extended by some of their often-cited literature and additional material from our research on Google Scholar.

\subsection{Gaussian Mixture Models}

\acfp{GMM} are density estimation algorithms mostly used for data clustering but can also be used as a generative probabilistic model. A \ac{GMM} consists of $N$ Gaussian distributions (components), which are normal distributions that are continuous for a real-valued random variable and symmetric about their mean. Each Gaussian can be characterized by a mean $\mu_{i}$ and a variance $\rho_{i}$ and has a probability/weight $\pi_{i}$ in the mixture model so that $\sum_{i=1}^{N}\pi_{i}=1$. For one-dimensional data, $\pi_{i}$ and $\mu_{i}$ are numbers, but in two-dimensional space, $\pi_{i}$ is a vector, and $\mu_{i}$ is a covariance matrix. The probability of a data point $d$ belonging to the cluster represented by Gaussian $i$ is $P(d=i)=\pi_{i}$ and the observation likelihood of $d$ in Gaussian $i$ is $P(d\vert d=i,\mu_{i},\rho_{i})=N(d\vert \mu_{i},\rho_{i})$ where $N(.)$ is the normal distribution function. \cite{harshvardhan2020comprehensive}

\begin{figure} [ht]
    \centering
    \begin{subfigure}[t]{0.49\textwidth}
        \centering
        \includegraphics[width=\textwidth]{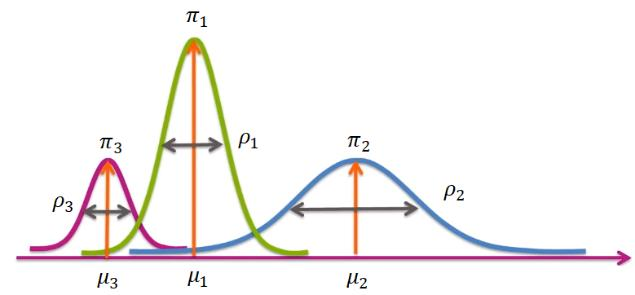}
        \caption{One-dimensional \ac{GMM} with three components. (Source:~\cite{harshvardhan2020comprehensive})}
    \end{subfigure} \hfill
    \begin{subfigure}[t]{0.49\textwidth}
        \centering
        \includegraphics[width=\textwidth]{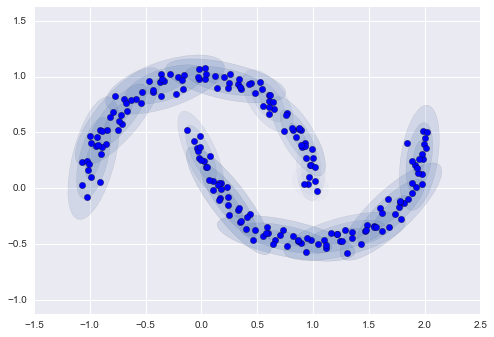}
        \caption{A \ac{GMM} with 16 components mapped to two-dimensional data. (Source: \cite{vanderplas2016python})}
        \label{fig:gmm_16_dim}
    \end{subfigure}
    \caption{Illustrations of \acp{GMM} used for one and two-dimensional data.}
    \label{fig:gmms}
\end{figure}

To map data points to $N$ clusters/components of a \ac{GMM} during training, an expectation-maximization (EM) algorithm is used:

\begin{enumerate}
    \item Choose (random) locations $\mu_{i}$ and shapes $\rho_{i}$ for each component
    \item Repeat until convergence:
    \begin{enumerate}
        \item \textit{E-step}: For each data point, find $\pi_{i}$ encoding the membership probability
        \item \textit{M-step}: For each cluster, update $\mu_{i}$, $\rho_{i}$ and $\pi_{i}$ based on all data points
    \end{enumerate}
\end{enumerate}

Multiple \acp{GMM} can be initialized with different random values to reduce the chance of missing the globally optimal solution. \cite{vanderplas2016python}.

\begin{figure} [htb!]
    \centering
    \begin{subfigure}[t]{0.3\textwidth}
        \centering
        \includegraphics[height=3cm]{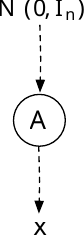}
        \caption{Gaussian}
    \end{subfigure}
    \begin{subfigure}[t]{0.3\textwidth}
        \centering
        \includegraphics[height=3cm]{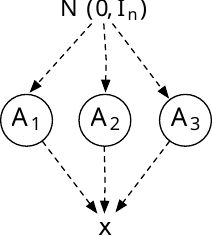}
        \caption{\ac{GMM}}
    \end{subfigure}
    \begin{subfigure}[t]{0.3\textwidth}
        \centering
        \includegraphics[height=3cm]{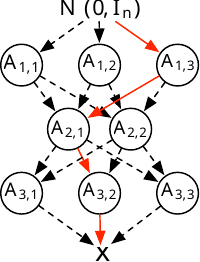}
        \caption{Deep \ac{GMM}}
    \end{subfigure}
    \caption{Comparison between a Gaussian, \ac{GMM} and a deep \ac{GMM} with transformation biases $b_{i,j}$ not shown. (Source:~\cite{oord2014factoring})}
    \label{fig:deep_gmm}
\end{figure}

Van der Oord \cite{oord2014factoring} propose a deep \ac{GMM}, which contains multiple layers of linear transformations $x=Az+b$ applied to the normal variable $z\sim\mathcal{N}(0,I_{n})$. For each sample, a random path through the graph is taken (see \autoref{fig:deep_gmm}), and the transformations are concatenated. In theory, deep \acp{GMM} can be represented by normal \acp{GMM}, but training would be more complex, and the deep models generalize better. The model is trained with an EM algorithm and parallelizable by design. Its capabilities are demonstrated by generating low-resolution greyscale images.

Zen and Senior~\cite{zen2014deep} applied mixture density networks for speech synthesis to enable multimodal regression and the prediction of variances. To this end, the authors model the conditional probability distribution using a \ac{GMM}, with parameters predicted by a fully connected multi-layer neural network.

VanderPlas \cite{vanderplas2016python} demonstrates how \acp{GMM} can be fitted well on two-dimensional data (see \autoref{fig:gmm_16_dim}) by trying different component sizes $N$ and searching for the model with the optimal \ac{AIC} or \ac{BIC}. Synthetic data can be obtained from this trained model by randomly selecting a Gaussian according to probability $\pi_{i}$ and sampling from its normal distribution. The author uses this approach on binary black-and-white images of handwritten digits and generates authentic artificial samples.

\subsection{Kernel Density Estimators}

\acfp{KDE} are models that approximate the probability density function $p(X=x)$ of a random variable $X$ in between a set of observations. For observations $x_{i}$ and an optimizable smoothing parameter $h>0$, the density estimation is defined as

\begin{equation}
    \hat{p}(x)=\frac{1}{n}\sum_{i=1}^{n}K(x-x_{i},h)
\end{equation}

with a positive kernel function $K(y,h)$, for which various implementations exist. One kernel often used is the Gaussian kernel $K(y,h)=\exp(-\frac{y^{2}}{2h^{2}})$, but many others are available to model different kinds of data. \cite{bhattacharya2020synthetic}

Bhattacharya et al. \cite{bhattacharya2020synthetic} synthesize \acp{PPG}, which are time series over volumetric blood flow in the human body, by decomposing it into components (pulse length, peak position, amplitude, etc.), training a \ac{KDE} for each component and sampling from the resulting probability distributions, from which new \acp{PPG} can be constructed.

\subsection{Markov Chain Models}

\begin{figure} [ht]
    \centering
    \begin{subfigure}[t]{0.3\textwidth}
        \centering
        \begin{tikzpicture}[node distance={2cm}, minimum size=10mm, font=\tiny, thick, action/.style = {draw, circle}]
            \node[action] (1) {A};
            \node[action] (2) [right of=1] {B};
            \node[action] (3) [below of=1] {C};
            \node[action] (4) [below of=2] {D};
            \path[]
                (1) edge [loop above] (1)
                (2) edge [loop above] (2)
                (3) edge [loop below] (3)
                (4) edge [loop below] (4)
                (1) [->] edge [bend right] (2)
                (1) [->] edge [bend right] (3)
                (1) [->] edge [bend right] (4)
                (2) [->] edge [bend right] (1)
                (2) [->] edge [bend right] (3)
                (2) [->] edge [bend right] (4)
                (3) [->] edge [bend right] (2)
                (3) [->] edge [bend right] (1)
                (3) [->] edge [bend right] (4)
                (4) [->] edge [bend right] (2)
                (4) [->] edge [bend right] (3)
                (4) [->] edge [bend right] (1);
        \end{tikzpicture}
        \caption{An order $1$ Markov chain.}
    \end{subfigure}
    \hspace{1cm}
    \begin{subfigure}[t]{0.45\textwidth}
        \centering
        \begin{tikzpicture}[node distance={2cm}, minimum size=10mm, font=\tiny, thick, action/.style = {draw, circle}]
            \node[action] (1) {A};
            \node[action] (2) [right of=1] {B};
            \node[action] (3) [below of=1] {C};
            \node[action] (4) [below of=2] {D};
            \node[action] (5) [double, left of=1, below=2.5mm of 1] {$\mathcal{B}$};
            \node[action] (6) [double, right of=2,below=2.5mm of 2] {$\mathcal{E}$};
            \path[]
                (1) edge [loop above] (1)
                (2) edge [loop above] (2)
                (3) edge [loop below] (3)
                (4) edge [loop below] (4)
                (1) [->] edge [bend right] (2)
                (1) [->] edge [bend right] (3)
                (1) [->] edge [bend right] (4)
                (2) [->] edge [bend right] (1)
                (2) [->] edge [bend right] (3)
                (2) [->] edge [bend right] (4)
                (3) [->] edge [bend right] (2)
                (3) [->] edge [bend right] (1)
                (3) [->] edge [bend right] (4)
                (4) [->] edge [bend right] (2)
                (4) [->] edge [bend right] (3)
                (4) [->] edge [bend right] (1)
                (5) [->] edge [draw=black!50, out=90, in=140] (2)
                (5) [->] edge [draw=black!50, out=290, in=180] (3)
                (5) [->] edge [draw=black!50, out=270, in=220] (4)
                (5) [->] edge [draw=black!50, out=70, in=180] (1)
                (1) [->] edge [draw=black!50, out=90, in=100] (6)
                (2) [->] edge [draw=black!50, out=0, in=120] (6)
                (3) [->] edge [draw=black!50, out=270, in=260] (6)
                (4) [->] edge [draw=black!50, out=0, in=240] (6);
        \end{tikzpicture}
        \caption{An order $1$ Markov chain with start and end state.}
    \end{subfigure}
    \caption{Illustrations of graphs of Markov chains with symbols as nodes and transition probabilities as edges. (Adapted from: \cite{durbin1998biological})}
    \label{fig:markov_chains}
\end{figure}
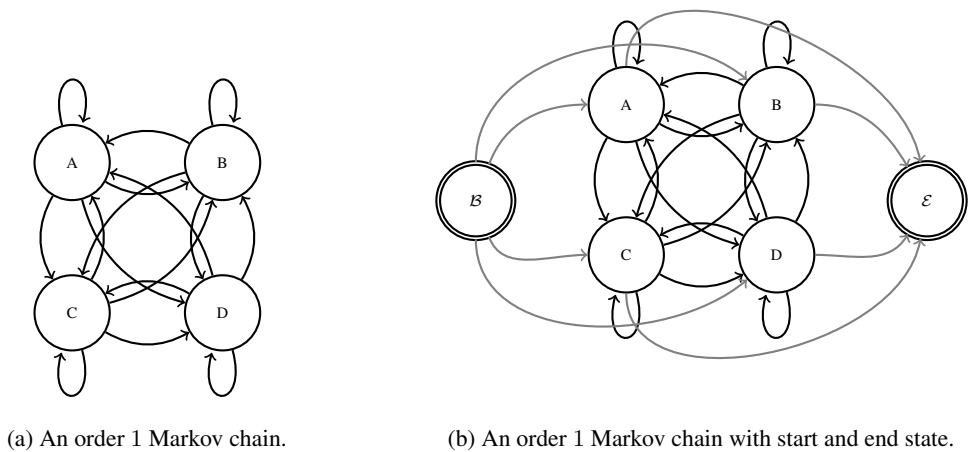

Order $n$ Markov chains are probabilistic models for infinite sequences of symbols where the probability for each symbol only depends on the previous $n$ symbols \cite{briot2017deep}. For $n=1$, they can be drawn as a graph with the states (symbols) as nodes and the transition probabilities $a_{st}=P(x_{i}=t\vert x_{i-1}=s)$ as edges (see \autoref{fig:markov_chains}). The probability of a whole sequence can be computed by applying $P(X,Y)=P(X\vert Y)P(Y)$ many times; that is, the probability of a sequence $x$ with length $L$ is $P(x)=P(x_{1})\prod_{i=2}^{L}a_{x_{i-1}x_{i}}$. \cite{durbin1998biological}

The advantage of Markov chains is their simplicity. They can be easily understood because the conditional probabilities can be computed by counting relative symbol appearances. Further, they can be interpreted as automata on which additional control mechanisms like Markov constraints and factor graphs can be imposed. On the downside, simple order $1$ models can not capture long-term temporal structures, and order $n$ models tend to overfit, require significantly bigger amounts of training data, and need to compute exponentially more conditional probabilities for large $n$. \cite{briot2017deep,trivino2001using}

Pachet et al. \cite{pachet2004beyond} introduce the Continuator, a system for interactive music generation in the user's style without a priori musical knowledge, allowing a musician to ``jam'' in real-time with the computer. The Continuator is powered by an \textit{analysis} module and a \textit{generator}. The analysis module first detects the ends of musical phrases, then builds a Markovian model of these phrases and detects global properties like tempo, meter, and note density. The generator uses the Markov model and the properties to generate music in the input style and continue it note-by-note.

\subsubsection{Hidden Markov Models}

\begin{figure} [ht]
    \centering
    \begin{subfigure}{0.55\textwidth}
        \centering
        \includegraphics[width=\textwidth]{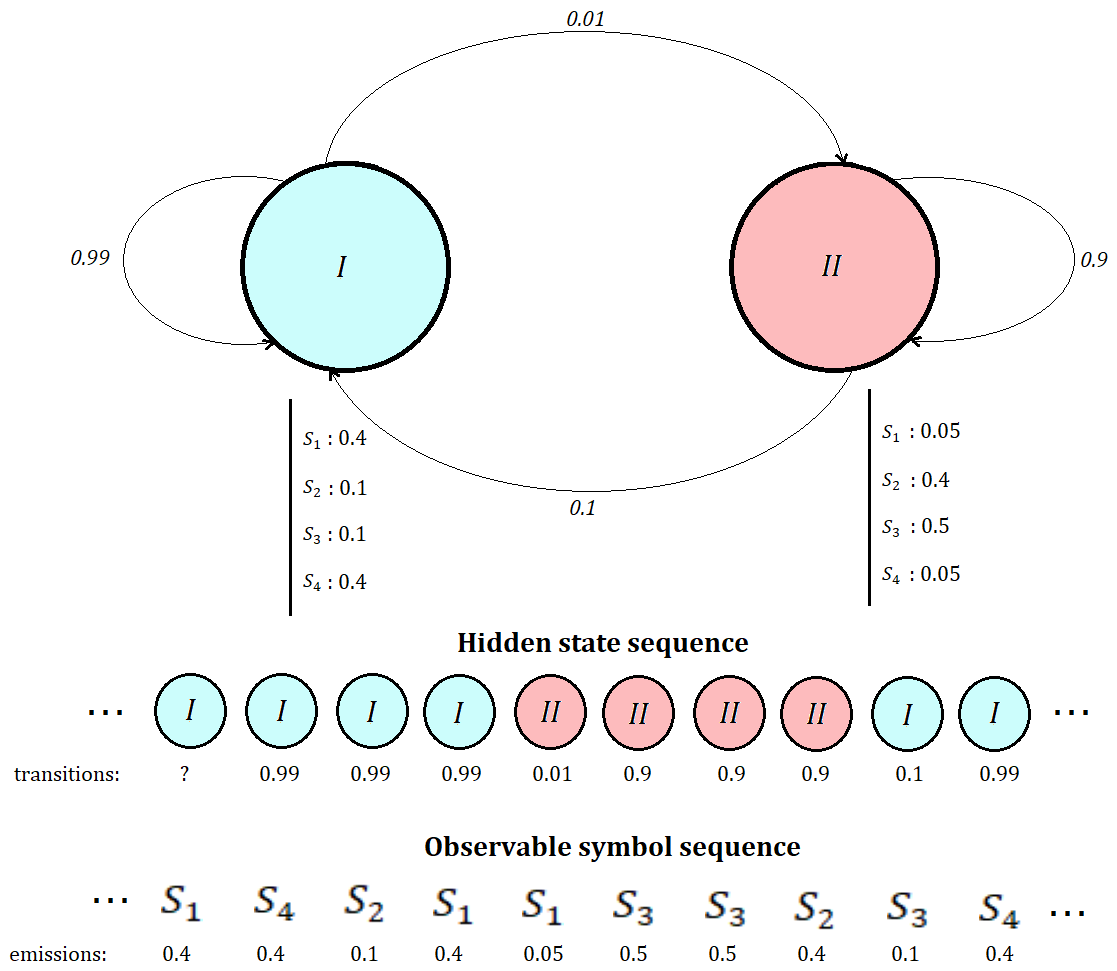}
        \caption{Architecture of an \ac{HMM}. (Source: \cite{harshvardhan2020comprehensive})}
        \label{fig:hmm_architecture}
    \end{subfigure}
    \hfill
    \begin{subfigure}{0.4\textwidth}
        \centering
        \includegraphics[width=\textwidth]{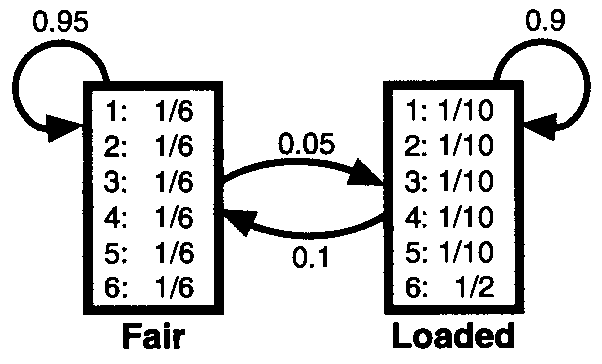}
        \caption{\ac{HMM} of a dishonest casino switching between fair and unfair dices. (Source: \cite{durbin1998biological})}
        \label{fig:hmm_casino}
    \end{subfigure}
    \caption{Illustrations of \acp{HMM}.}
\end{figure}

\acfp{HMM} are extensions of Markov chains where the Markov chain state sequence $\pi$ is \textit{hidden} and each hidden state $k$ has emission probabilities $e_{k}(b)=P(x_{i}=b\vert \pi_{i}=k)$ for the observable symbols $x$ (see \autoref{fig:hmm_architecture}). This model can depict many issues with reduced complexity compared to simple Markov chains, but the inference of hidden states $\pi$ for observation $x$ with joint probability $P(x,\pi)=a_{0\pi_{1}}\prod_{i=1}^{L}e_{\pi_{i}}(x_{i})a_{\pi_{i}\pi_{i+1}}$ is more complex (see for example \autoref{fig:hmm_casino}). The most probable state sequence $\pi^{*}=\argmax_{\pi}P(x,\pi)$ for an observation $x$ can be recursively computed with the Viterbi algorithm. \cite{harshvardhan2020comprehensive,durbin1998biological}

Durbin et al. \cite{durbin1998biological} use \acp{HMM} to model biological (genetic) sequences and demonstrate the labeling of unannotated and generation of new data for this topic. They show that these simple models can learn truthful models even from observations where the hidden paths are unknown using the Baum-Welch and Viterbi algorithms. They also discuss different model topologies for different sequence lengths and find the careful topology construction of a \ac{HMM} validated by human experts to be essential for good model performance.

Racyzński et al. \cite{raczynski2013melody} interpolate results using multiple learned sub-models (namely, a bigram Markov chain model $P(C_{t}\vert C_{t-1})$, a tonality relation $P(C_{t}\vert T_{t})$, and a melody relation model $P(C_{t}\vert M_{t})$) to sequentially generate chords $C_{i}$ as accompaniment for music.

Kaliakatsos-Papakostas et al. \cite{kaliakatsos2014probabilistic} propose the constrained \ac{HMM} (CHMM), which allows intermediate states of the sequence to be fixed to a specific value (anchor chords) with probability $1$. Then, the Viterbi algorithm is used to find the most likely path between these checkpoints. The model is used to generate harmonic chord sequences for a melody by mapping the hidden note states between the anchors defined by the melody to chords.

Bindschaedler and Shokri \cite{bindschaedler2016synthesizing} use \acp{HMM} to generate plausible and private location traces by clustering the available locations and synthesizing a sequence of locations from the same cluster label sequence as a real seed trace. The generated trace is further evaluated using a plausibility and a privacy (geographic similarity to the seed trace) test.

\subsubsection{N-Grams}

The \textit{n-gram} is a tuple of $n$ values, for instance, a sequence of $n$ words of a sentence $(w_{1},...,w_{n})$. They are usually used to efficiently model the probabilities of words based on the already written text:

\begin{equation}
    p(X_{t}\vert X_{t-1},...,X_{t-n+1})=\frac{count(X_{t-n+1},...,X_{t})+1}{count(X_{t-n+1},...,X_{t-1})+V}
\end{equation}

under the Markovian assumption $p(X_{t}\vert X_{t-1},...,X_{1})=p(X_{t}\vert X_{t-1},...,X_{t-n+1})$, that is, the probability of a word at position $t$ only depends on the previous $n-1$ words. Adding Laplace smoothing with the $+1$ in the numerator and the word vocabulary size $V$ in the denominator also allows previously unseen word combinations to be created with low probability. N-grams are normally used for language modeling and synthetic text generation. \cite{ranzato2014video}

Bengio et al. \cite{bengio2000neural} extend the idea of n-grams with a \ac{MLP}: First, each word $X_{k}$ is encoded into a 1-hot vector $\mathbf{1}(X_{k})$. Then, the vectors are linearly embedded using matrix $W_{x}$, concatenated, and fed into the \ac{MLP}, which is trained to predict the next word probability. Finally, a softmax function $SM$ is applied, resulting in the neural network language model:

\begin{equation}
    p(X_{t}\vert X_{t-1},...,X_{t-n+1})=SM(MLP[W_{x}\mathbf{1}(X_{t-1}),...,W_{x}\mathbf{1}(W_{t-n+1})]).
\end{equation}

Barbieri et al. \cite{barbieri2012markov} implement Markov constraints to generate lyrics in a given style and rhythm for music. They train a Markov process using relative frequencies of n-grams of lyrics, for example, of a specific author and then restructure it as a finite-length sequence of constrained variables with assigned probability for each value. The constraints are related to rhyme, rhythm, syntax (part-of-speech templates), and semantics (relations between words).

\othertab{n-grams}{
    \cite{roy2013enforcing} & Extension of the constrained Markov model from \cite{barbieri2012markov} to allow a global meter constraint to be efficiently implemented. & 2013 \\\hline 
    \cite{forsyth2013generating} & Construction of a probabilistic finite state automat from n-grams of preprocessed MIDI files to harmonically accompany a melody. & 2013 \\\hline 
    \cite{papadopoulos2014avoiding} & Avoidance of plagiarism in generated sequences in high-order Markov chains with a \textit{MaxOrder} global Markov constraint that prevents chunks longer than \textit{MaxOrder} from being replicated from the training data by building Markov automatons with restricted maximum path lengths. & 2014 \\\hline 
    \cite{papadopoulos2016assisted} & FlowComposer: A web tool consisting of two collaborating constrained Markov models (melody and chords) for generation, re-harmonization, and interactive composition of music lead sheets. & 2016 \\\hline 
    \cite{whorley2016music} & A multiple viewpoint system consisting of Markov chains obtained from n-grams generates music. The agents are responsible for different aspects of the music and are ordered sequentially and hierarchically. & 2016 \\ 
}

\subsection{Bayesian Networks}\label{sec:bayes_nets}

\begin{figure} [ht]
    \centering
    \includegraphics[width=0.5\textwidth]{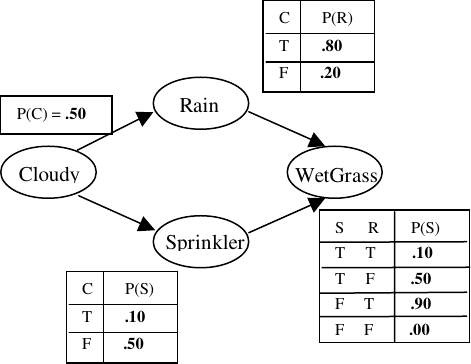}
    \caption{Example of a \acl{BN} with discrete random variables. (Source: \cite{guo2002survey})}
    \label{fig:bayes_network}
\end{figure}

\cite{draghi2021bayesboost})
A \acf{BN} is a \ac{DAG}, a special type of graphical model in which random variables are the nodes and dependencies between these variables are the edges. The directed edges run from the ``cause'' or parent node to the ``effect'' or child node and define the conditional dependency of the nodes. Each random variable has a continuous or discrete probability distribution function. An example for a \acl{BN} can be seen in \autoref{fig:bayes_network}. \cite{stephenson2000introduction}

If we define $\textbf{V}=\{V_{j}:j\in \{1,...,N\}\}$ as the set of random variables of a Bayesian network, the probability distribution of $V_{j}$ as $p(V_{j}\vert\textbf{Pa}_{j})$ and $\textbf{Pa}_{j}$ as the set of parents of $V_{j}$, the joint distribution over $\textbf{V}$ is defined as \cite{kingma2013fast}:

\begin{equation}
    p(\textbf{V})=p(V_{1},...,V_{N})=\prod_{j=1}^{N}p(V_{j}\vert\textbf{Pa}_{j}).
\end{equation}

There are many algorithms to train \acp{BN} under different conditions where the net structure is known or unknown in advance, and the reference data is fully or partially observable. In the context of synthetic data, we usually either have the trivial case of available expert knowledge, from which the structure of a \ac{BN} can be constructed, or available real-world data, but no information about the structure of a \ac{BN} that corresponds well to the data. For the latter case, two problems need to be solved:

\begin{enumerate}
    \item Finding a metric to compare potential structures of \acp{BN}.
    \item Searching for potential \ac{BN} structures algorithmically.
\end{enumerate}

A solution for the first problem is provided by the joint probability $p(D,S^{h})$ for the data $D$ and a hypothetical structure $S^{h}$ of a \ac{BN}:

\begin{equation}
    \log p(D,S^{h}) = \log p(D\vert S^{h}) + \log p(S^{h}).
\end{equation}

The \acf{BIC} \cite{schwarz1978estimating} can be used to calculate $\log p(D\vert S^{h})$ while the prior probability $p(S^{h})$ of a structure can be determined for example by assigning probabilities to a predefined set of possible structures or defining a prior network and measuring the deviation of $S^{h}$ from it. \cite{stephenson2000introduction}

The second problem, searching for structures, is NP-hard if done for all possible combinations, so different greedy algorithms are employed. In general, these algorithms increase $p(S^{h})$ step-wise by performing actions on the graph (adding, removing, or reversing an edge) until a maximum is reached. \cite{stephenson2000introduction}

In recent years, more sophisticated learning algorithms have been developed. They use dynamic programming \cite{koivisto2004exact,silander2012simple,singh2005finding} to split the global learning problem into small subproblems. Others define the learning task as a shortest path problem and solve it with an A* algorithm \cite{yuan2011learning}.

Young et al. \cite{young2009using} use \acp{BN} to anonymize a data set so it can be disclosed to the public. They limit their networks to discrete variables and map continuous values (e.g., age) to discrete intervals to facilitate the learning process, which consists of multiple steps:

\begin{enumerate}
    \item The user defines a prior network and conditional probability distributions for each node. Also, an imaginary database is supplied to generate confidence in the prior structure.
    \item The probability distributions of the nodes are updated using the training data.
    \item A greedy search algorithm \cite{bottcher2003deal} starts the search from the prior network, creates all possible networks with one change (edge addition, removal, or reversion), and selects the one with the highest Bayes factor (likelihood of the model according to the data) as the new baseline until no further improvement occurs.
\end{enumerate}

The trained \ac{BN} is now used as an imputation model \cite{graham2009multiply} to generate synthetic data by consecutively drawing random samples from the nodes in the hierarchy.

Suzuki et al. \cite{suzuki2014four} generate fixed-size four-part (alto, tenor, bass, soprano) symbolic harmonies based on the melody of a soprano voice and experiment with the conditioning on chord nodes. The pitches for each voice are classified jointly based on the previous value (Markov property) and the current soprano or chord value. The soprano network without chords produces smoother results.

Zhang et al. \cite{zhang2017privbayes} propose \textit{PrivBayes}, a differentially private\footnote{Differential privacy introduces randomness to the data provision process, resulting in ``plausible deniability of any outcome'' \cite{dwork2014algorithmic}} method to release high-dimensional data sets. First, a \ac{BN} with succinct attribute correlations is created. Then noise is injected into the low-dimensional marginal distributions of the \ac{BN}, and the now noisy approximated data distribution is used to sample a private synthetic data set.

Draghi et al. \cite{draghi2021bayesboost} approach the bias problem in the training data of \acp{BN} by identifying under-represented cases and over-sampling them with synthetic data. They start by training a \ac{BN} on a subset of the original data, modifying it, and generating a new biased data set. Then, a \ac{BN} is trained on this biased data and generates the data set $D_{bias}$ with the same size as the original. Next, a classifier is trained on $D_{bias}$ and tries to predict values from a validation set, which is a subset of the original data and adds samples with an uncertain outcome (low probability) to a data set $D_{unc}$. Finally, the \textit{BayesBoost} is performed by generating $m$ similar samples for each sample in $D_{unc}$ with a \ac{BN} trained on $D_{bias}$ and adding the results to $D_{bias}$, resulting in a new less biased dataset.

\othertab{\acp{BN}}{
    \cite{chen2016synthesizing} & Training a \ac{BN} on motion capture data to produce realistic human 3D poses that are rendered with different textures (e.g., clothes) to produce training data for human 3D pose estimation. & 2016 \\\hline 
    \cite{ping2017datasynthesizer} & DataSynthesizer: Creating a \acl{BN} from data with a \textit{DataDescriber}, injecting noise into the distributions and sampling from it with the \textit{DataGenerator}. The \textit{ModelInspector} compares the properties of the synthetic data to the real one. & 2017 \\\hline 
    \cite{tucker2020generating} & Modeling heterogeneous (continuous and discrete variables) medical patient data as a \ac{BN} to be able to incorporate expert knowledge and generate private data sets. They experiment with three ways to deal with missing values: Deleting the entire entry, adding ``miss states/nodes'' to the \ac{BN}, and using the FCI algorithm \cite{spirtes2000causation} to infer the missing variables. Using probabilistic graphical modeling, the model produces high-fidelity results with a low risk of patient re-identification (cloning of training data). & 2020 \\\hline 
    \cite{draghi2021bayesboost} & Exploring the data bias problem and under-representation in underlying ground truth samples. Specifically, it is an important problem in medical data, where synthetic data generation is used to mask sensitive patient data. The authors propose an approach to identifying under-sampled data and improving data synthesis to correct this problem. & 2021 \\
}

\subsection{Genetic Algorithms}\label{sec:genetic_algorithms}

\acfp{GA} are \ac{ML} algorithms that mimic the natural selection process over time. The population consists of a crowd of individuals at each time step or \textit{generation}. To create the population for the next generation, three actions are performed:

\begin{description}
    \item[Selection] Selection of suitable candidates from the population, using a \textit{fitness function} to eliminate worse candidates and increase the chance of survival of better ones so they can pass on their good properties. Individuals can be selected multiple times to maintain population size.
    \item[Crossover] Information exchange between candidates.
    \item[Mutation] Perturb the candidates' information by randomly changing some properties, usually according to some distribution.
\end{description}

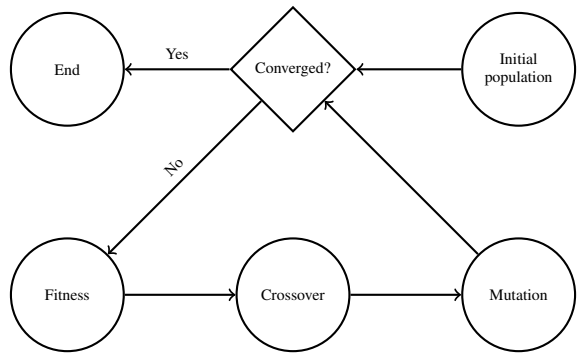
\begin{figure}[ht]
    \centering
    \begin{tikzpicture}[node distance={3cm}, minimum size=15mm, font=\tiny, thick, action/.style = {draw, circle}, decision/.style = {draw, diamond}]
        \node[action] (1) {End};
        \node[decision] (2) [right of=1] {Converged?};
        \node[action] (3) [right of=2, align=center] {Initial\\population};
        \node[action] (4) [below of=1] {Fitness};
        \node[action] (5) [below of=2] {Crossover};
        \node[action] (6) [below of=3] {Mutation};
        \draw[->] (3) -- (2);
        \draw[->] (6) -- (2);
        \draw[->] (2) -- node[midway, sloped, yshift=0.2cm] {Yes} (1);
        \draw[->] (2) -- node[midway, sloped, yshift=0.2cm] {No} (4);
        \draw[->] (4) -- (5);
        \draw[->] (5) -- (6);
    \end{tikzpicture}
    \caption{Process of a \ac{GA}. (Adapted from \cite{you2016automatic} and modified)}
    \label{fig:genetic_algorithm}
\end{figure}

This generation process (illustrated in \autoref{fig:genetic_algorithm}) continues until the system converges (i.e., all candidates are identical) or a user-defined criterion is met. The speed of the \ac{GA} model is measured by the number of generations needed to meet the requirements. \cite{chen2017genetic}

Liu et Ting \cite{liu2012polyphonic} use a \ac{GA} for polyphonic accompaniment generation given a dominant music melody. They remove the impractical need for a human evaluation criterion of previous approaches by building a fitness function based on music theory.

You et Liu \cite{you2016automatic} propose a \ac{GA} that finds similar and suitable chord variations for a given target melody and some example chord progressions from other songs provided as MIDI files. The initial population consists of the exemplar chord patterns with keys shifted to match the key of the target melody. The fitness function and crossover process further incorporate music theory to ensure that the evolving chord patterns harmonize with the melody and conform to basic rules.

Chen et al. \cite{chen2017genetic} implement a \ac{GA} for categorical tabular data with non-hierarchical variables. They start by independently computing the univariate distributions of all columns of the original data; then, they sample a user-defined amount of synthetic tables from these distributions, which are the population of the first generation. The \ac{GA} process then optimizes the statistics of the data sets iteratively until the desired similarity to the original data is reached. The computational workload of table \acp{GA} is significantly higher, and they are more error-prone than \acp{GA} for string data due to the increase in variables and their relationships.

\subsection{Boltzmann Machines}\label{sec:boltzmann_machines}

\begin{figure} [ht]
    \centering
    \begin{subfigure}[t]{0.3\textwidth}
        \centering
        \includegraphics{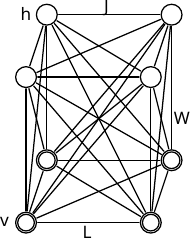}
        \caption{Architecture of a General Boltzmann Machine.}
    \end{subfigure}
    \hfill
    \begin{subfigure}[t]{0.3\textwidth}
        \centering
        \includegraphics{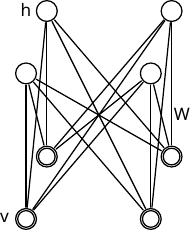}
        \caption{Architecture of a Restricted Boltzmann Machine.}
    \end{subfigure}
    \hfill
    \begin{subfigure}[t]{0.3\textwidth}
        \centering
        \includegraphics{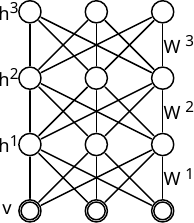}
        \caption{Architecture of a Deep Boltzmann Machine.}
    \end{subfigure}
    \caption{Architectures of different kinds of Boltzmann machines. (Source: \cite{salakhutdinov2009dbm})}
    \label{fig:boltzmann_architectures}
\end{figure}

A Boltzmann machine is an undirected network consisting of binary visible nodes $\textbf{v} \in \{0,1\}^{D}$ and hidden nodes $\textbf{h} \in \{0,1\}^{P}$. The model parameters $\theta = \{\textbf{W},\textbf{L},\textbf{J}\}$ are the visible-to-hidden, visible-to-visible and hidden-to-hidden symmetric interaction terms (matrices) of the graph (see \autoref{fig:boltzmann_architectures}). The model parameters $\theta$ of a Boltzmann machine are trained using gradient ascent. They assign probabilities to the visible units, where the training data is put in, based on the states of the hidden units. \cite{salakhutdinov2009dbm}

\subsubsection{Restricted Boltzmann Machines}\label{sec:rbm}

\acp{RBM} \cite{smolensky1986rbm} are a subset of general Boltzmann machines, where only visible-to-hidden ($\textbf{W}$) connections are allowed, so both $\textbf{J}$ and $\textbf{L}$ are set to zero. These models have the advantage that inference is exact, and learning is significantly more efficient \cite{salakhutdinov2009dbm}. The method used to train a \ac{RBM} is unsupervised learning, so the data is unlabeled. Each training sample is provided as input $\textbf{v}$ and $\theta$ is modified to increase the likelihood function $p(\textbf{v},\theta)$ using, for example, a gradient method or Contrastive Divergence \cite{hua2015deep}.

\begin{figure} [ht]
    \centering
    \includegraphics[width=0.5\textwidth]{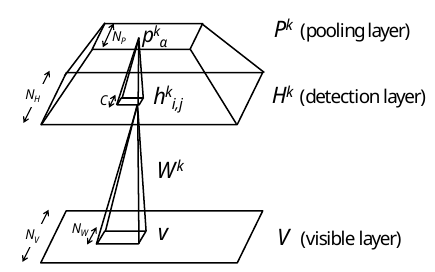}
    \caption{The convolutional \ac{RBM} architecture with only one of the $K$ hidden and max-pooling layers shown for readability. (Source: \cite{lee2009convolutional})}
    \label{fig:crbm}
\end{figure}

Lee et al. \cite{lee2009convolutional} extend the \ac{RBM} with convolution filters to process two-dimensional high-resolution images. The visible input units of size $N_{V}\times N_{V}$ are processed by $K$ filters with size $N_{W}\times N_{W}$ to produce $K$ hidden layers with size $N_{H}\times N_{H}$. Each hidden layer is then partitioned into $C\times C$ blocks that are each connected to exactly one binary unit in the \textit{max-pooling layer} $P$ to shrink the representation. The architecture is depicted in \autoref{fig:crbm}.

Lattner et al. \cite{lattner2018imposing} propose a multi-component model to generate music with consistent local and global structural properties. First, a convolutional \ac{RBM} \cite{lee2009convolutional} learns the local structure of musical pieces based on training data. Then, \textit{constrained sampling} is applied to a randomly initialized ``piano roll'' music representation matrix $\mathbf{v}\in[0,1]^{T\times P}$ consisting of probabilities of active pitches $1<p<P$ over time steps $1<t<T$: The sampling process first applies 20 gradient descent steps to $\mathbf{v}$ with a loss function involving self-similarity, tonality and meter constraints regarding a template piece $\mathbf{x}\in[0,1]^{T\times P}$ before performing alternating steps of Gibbs sampling\footnote{Gibbs sampling is a \ac{MCMC} algorithm used for approximate statistical inference that iteratively samples from the conditional probability distribution of each variable in a multivariate distribution while holding the others fixed.} with the convolutional \ac{RBM} and one step of gradient descent with lower learning rate to $\mathbf{v}$. This process is repeated until the solution no longer improves on the \ac{RBM} and constraints jointly.

\subsubsection{Deep Belief Networks}\label{sec:deep_belief_network}

\begin{figure} [ht]
    \centering
    \begin{subfigure}[t]{0.3\textwidth}
        \centering
        \includegraphics{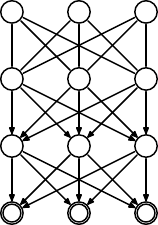}
        \caption{Architecture of a Deep Belief Network.}
    \end{subfigure}
    \hspace{1cm}
    \begin{subfigure}[t]{0.3\textwidth}
        \centering
        \includegraphics{figures/related_work/boltzmann_deep.pdf}
        \caption{Architecture of a Deep Boltzmann Machine.}
    \end{subfigure}
    \caption{Comparison of the architectures of a \ac{DBN} and a \ac{DBM}. (Source: \cite{salakhutdinov2009dbm})}
    \label{fig:dbn_dbm_comparison}
\end{figure}

A \acf{DBN}, similar to its successor \ac{DBM} (see \autoref{sec:dbm}), is a combination of multiple \acp{RBM}, where the hidden layer of one \ac{RBM} becomes the input (visible layer) of the higher-level \ac{RBM}, but only the top two hidden layers are undirected and form an \textit{associative memory} while the lower layers form a \ac{DAG} (see \autoref{fig:dbn_dbm_comparison}). Like the \ac{RBM}, the \ac{DBN} can be used to learn high-level representations of unlabeled data and convert representations back to visible data (e.g., images), but it can also be extended to supervised learning tasks by appending the label as an input to the visible layer. \cite{hinton2006fast}

The layer-wise greedy learning process starts by training the lowest-level \ac{RBM} with the visible and first hidden layer normally. Then, the output of the first hidden layer becomes the input (visible layer) of the next \ac{RBM}, and now the second model is trained. This continues until a sufficient number of layers are reached. The second step introduces a fine-tuning algorithm with a bottom-up and a top-down pass. Now, the ``recognition'' weights $\textbf{W}_{x}^{\top}$ and ``generative'' weights $\textbf{W}_{x}$ are decoupled and modified independently. During the bottom-up pass, fixed recognition weights are used to stochastically determine all hidden values and update the generative weights on the directed connections according to a likelihood metric. The top-down pass starts with a state of the top-level associative memory and uses fixed generative weights to determine the visible layers on which the recognition weights are updated similarly. \cite{hinton2006fast}

Hinton et al. \cite{hinton2006fast} use alternating Gibbs sampling~\cite{tosh2016mixing} in the \ac{DBN}'s associative memory ``until the Markov chain converges to the equilibrium distribution''. Then, images of digits are generated in the \ac{DBN} that was trained on the MNIST dataset by drawing a sample from this distribution and passing it down the generative weights. By fixing the label units, certain digits can be synthetically generated, and with repeated iterations of Gibbs sampling between down-passes, the digits become more realistic.

Lee et al. \cite{lee2009convolutional}, who introduced convolutional \acp{RBM}, stack these on top of one another to create convolutional \acp{DBN}. Different from \cite{hinton2006fast}, they use undirected connections between all layers, like \acp{DBM}. The experimental results with a two-layer convolutional \ac{DBN} on the Kyoto natural image data set show that the first layer learns edge filters and the second filters for contours, corners, angles, and surface boundaries. The hierarchical representations obtained from these layers improve classification results on the Caltech-101 object and MNIST digit classification tasks.

Bickerman et al. \cite{bickerman2010learning} apply \acp{DBN} to create jazz melodies in an unsupervised manner. They divide each beat into 12 slots, each consisting of 30 bits (12 chords, 18 melodies) that encode the note pitch, octave, and length. A two-layer \ac{DBN} can produce short stylistically plausible jazz samples based on a random sequence input but cannot capture regularities in music. Other problems are the large training time and complicated sampling procedure of \acp{DBN}.

Sun \cite{sun2015deephear} employs two \acp{DBN} (one flipped) with pair-wise pre-trained layers as an autoencoder to generate random piano roll bars of music from binary piano roll matrices (rows represent note pitches, columns a 16th note of playing time) of complete or incomplete music. On average, the network copies 56,9\% of the notes during the reconstruction, which still results in noticeably different works being created.

\subsubsection{Temporal Restricted Boltzmann Machines and Related Models}

\begin{figure} [ht]
    \centering
    \includegraphics[width=0.5\textwidth]{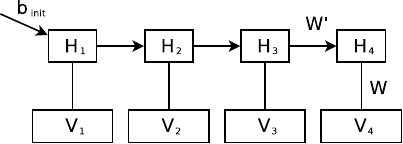}
    \caption{Architecture of a \ac{TRBM}. (Source: \cite{sutskever2008recurrent})}
    \label{fig:trbm}
\end{figure}

The \acf{TRBM} \cite{sutskever2007learning} is a sequence of \acp{RBM} where the biases of one \ac{RBM} depend on the hidden state of the previous \ac{RBM} (see \autoref{fig:trbm}). Training the model works similar to a normal \ac{RBM}, but on sequences instead of fixed-size samples. A significant problem of \acp{TRBM} is that for computing probabilities during inference, the evaluation of all possible states (partition functions) of two \acp{RBM} is required, making a heuristic approach necessary \cite{sutskever2008recurrent}.

\acfp{RTRBM} \cite{sutskever2008recurrent} are slightly modified \acp{TRBM} which offer exact inference and feasible gradient computation. \acp{RTRBM} can be expressed as a \ac{RNN} with the same parameters as the \ac{RTRBM} and its log-likelihood as the cost function, so gradients can be computed using the \ac{BPTT} algorithm.

The \acf{SRTRBM} \cite{mittelman2014structured} learns a dependency structure between pairs of visible and hidden units instead of using full connectivity like the \ac{RTRBM}. The model encourages sparse graphs and can reveal the structure of the underlying time-series data.

Sutskever et al. \cite{sutskever2007learning}, the original authors of the \ac{TRBM}, train their model on 10,000 video sequences with 100 frames and achieve good results on future frame prediction and online denoising (removing artifacts). In their later work on \acp{RTRBM} \cite{sutskever2008recurrent}, they improve on the video generation task.

Mittelmann et al. \cite{mittelman2014structured} compare their \ac{SRTRBM} to the previously proposed models by Sutskever et al. \cite{sutskever2007learning,sutskever2008recurrent} by predicting frames on synthetic bouncing ball videos. Further, they use the \ac{SRTRBM} to make predictions on motion capture and weather data. They improve on the performance of the predecessors in all temporal modeling tasks.

\subsubsection{Deep Boltzmann Machines}\label{sec:dbm}

A \ac{DBM} \cite{salakhutdinov2009dbm} is a combination of multiple \acp{RBM} where the hidden layer of one \ac{RBM} becomes the visible layer of the next one, resulting in a model with one visible and a stack of multiple hidden layers (see \autoref{fig:dbn_dbm_comparison}). Using a greedy layer-wise learning approach, such a multi-layer model can efficiently learn high-level representations from unlabeled data.

Salakhutdinov et al. \cite{salakhutdinov2009dbm} use a \ac{DBM} trained on the MNIST dataset to generate synthetic handwritten digits by initializing the model with random binary states and running a Gibbs sampler for 100,000 steps. Further, they demonstrate the creation of greyscale toy images on the NORB \cite{lecun2004learning} dataset and improve the results by increasing the amount of training data with simple pixel translations.

\subsubsection{Gated Boltzmann Machines}

\begin{figure} [ht]
    \centering
    \begin{subfigure}[t]{0.45\textwidth}
        \centering
        \includegraphics[width=.6\textwidth]{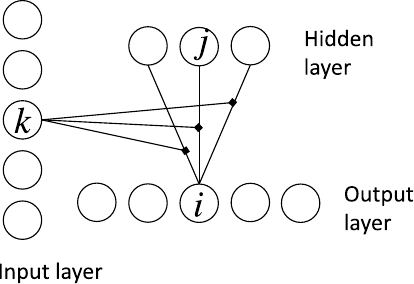}
        \caption{Hidden layer acting as a gate.}
    \end{subfigure}
    \hfill
    \begin{subfigure}[t]{0.45\textwidth}
        \centering
        \includegraphics[width=.6\textwidth]{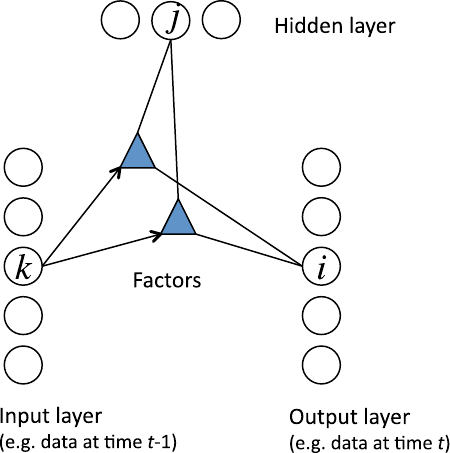}
        \caption{Factorization reducing the number of parameters.}
    \end{subfigure}
    \caption{Gate approaches in a gated Boltzmann machine. (Source: \cite{taylor2011two})}
    \label{fig:gbm_gates}
\end{figure}

A gated Boltzmann machine encodes transformations between two observations using its hidden layer. It can be described as a conditional \ac{RBM} with a visible input $\mathbf{x}$, a hidden transformation representation layer $\mathbf{h}$ that acts as a \textit{gate} (see \autoref{fig:gbm_gates}) and a visible output layer $\mathbf{y}$. Conditional on the input, the inference and learning procedures are tractable \cite{taylor2011two}. The trainable parameters are stored in a three-dimensional parameter ``tensor'' $\mathbf{W}$ and the compatibility between $\mathbf{x}$, $\mathbf{y}$ and $\mathbf{h}$ is computed by an energy function $E(\mathbf{y},\mathbf{h};\mathbf{x})$ \cite{memisevic2007unsupervised}:

\begin{equation}
    E(\mathbf{y},\mathbf{h};\mathbf{x})=-\sum_{ijk}W_{ijk}x_{i}y_{j}h_{k}
\end{equation}

The values for $\mathbf{y}$ and $\mathbf{h}$ can be computed in the same way \cite{memisevic2007unsupervised}:

\begin{equation}\label{eq:gbm_h}
    p(h_{k}\vert\mathbf{x,y})=\frac{1}{1+\exp(-\sum_{ij}W_{ijk}x_{i}y_{j})}
\end{equation}

\begin{equation}\label{eq:gbm_y}
    p(y_{j}\vert\mathbf{x,h})=\frac{1}{1+\exp(-\sum_{ik}W_{ijk}x_{i}h_{k})}.
\end{equation}

The model parameters $\mathbf{W}$ are trained with gradient-based optimization maximizing the average conditional log-likelihood $L=\frac{1}{N}\sum_{\alpha}\log p(\mathbf{y^{\alpha}}\vert\mathbf{x^{\alpha}})$ for training pairs $(\mathbf{x^{\alpha}},\mathbf{y^{\alpha}})$. Because parts of the gradient are intractable, Gibbs sampling is used to approximate partial results with the help of the conditional distributions described in \autoref{eq:gbm_h} and \autoref{eq:gbm_y} in a scheme called contrastive divergence. \cite{memisevic2007unsupervised}

Memisevic et al. \cite{memisevic2007unsupervised} use the gated Boltzmann machine to learn transformation representations on randomly transformed $8\times8$ pixel images and predict the next images on video patches of size $22\times22$ pixels. They notice that the model becomes intractable for larger images due to the cubic parameter space $\mathbf{W}$ and modify it to make it iteratively applicable to smaller image patches. Finally, the model's abilities are demonstrated by analogy making, which means obtaining a transformation from a source image and applying the transformation to a target image using the patch-wise approach.

Memisevic et al. \cite{memisevic2010learning} tackle the problem of the rapidly expanding cubic parameter tensor $\mathbf{W}$ for large inputs by approximating the results using three matrices $w^{x}$, $w^{y}$ and $w^{h}$, so $w_{ijk}=\sum_{f=1}^{F}w_{if}^{x}w_{jf}^{y}w_{kf}^{h}$. Suppose the number of factors $F$ is similar to the number of hidden and visible variables. In that case, the model now only requires $O(N^{2})$ instead of $O(N^{3})$ parameters (see \autoref{fig:gbm_gates} for a comparison). Like \cite{memisevic2007unsupervised}, the model learns filters as transformation representations on larger $40\times40$ pixel images. Also, the analogy experiments are repeated, and motion extraction is successfully performed.

Taylor et al. \cite{taylor2011two} propose three key properties for time series models:

\begin{enumerate}
    \item Distributed (i.e., componential) hidden state instead of sampling from a single category like \acp{HMM} to retain high flexibility and capacity.
    \item Undirected, bipartite graph as the model structure to make inference simple and efficient.
    \item Ability to form deep networks by stacking models and learning one layer at a time to capture more abstract data features.
\end{enumerate}

\begin{figure} [ht]
    \centering
    \begin{subfigure}[t]{0.3\textwidth}
        \centering
        \includegraphics[width=\textwidth]{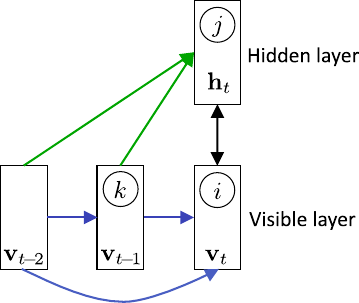}
        \caption{An order $2$ \ac{CRBM}.}
        \label{fig:crbm_temporal}
    \end{subfigure}
    \hfill
    \begin{subfigure}[t]{0.6\textwidth}
        \centering
        \includegraphics[width=\textwidth]{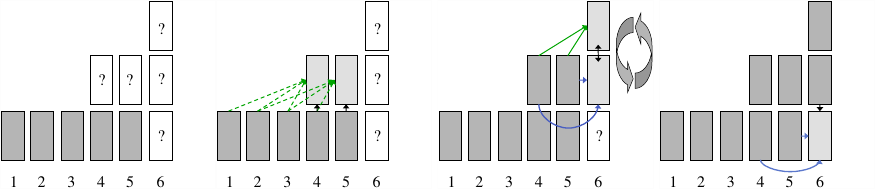}
        \caption{A \ac{CDBN} consisting of an order $2$ and an order $3$ \ac{CRBM} initialized by real data to generate a new sample.}
        \label{fig:cdbn}
    \end{subfigure}
    \caption{Architecture of \ac{CRBM} and the deep version \ac{CDBN}. (Source: \cite{taylor2011two})}
\end{figure}

Guided by these constraints, they introduce the \ac{CRBM}, which takes one or more visible layers from previous time steps as additional inputs to a normal \ac{RBM} (see \autoref{fig:crbm_temporal}), and the \ac{CDBN}, which stacks multiple \acp{CRBM} similar to a \ac{DBN}, to build a generative model for time series initialized by real data (see \autoref{fig:cdbn}). Further, the gated \ac{CRBM} is introduced to enable multiplicative interactions between time steps, allowing the learned transformations to be highly nonlinear. Also, the factorization method from \cite{memisevic2010learning} is reintroduced to reduce parameters, and predefined style labels are added to the gate process. The models are successfully evaluated using the CMU motion data set where motions are continued by the \acp{CRBM} or new motions with mixed or changing styles are created.

\subsection{Autoencoders}\label{sec:autoencoders}

The basic autoencoder is a network that has as input a vector $\textbf{x}\in [0,1]^{d}$ and maps it to a hidden representation $\textbf{y}\in [0,1]^{d'}$ with a function $\textbf{y}=f_{\theta}(\textbf{x})=s(\textbf{Wx}+\textbf{b})$. $\textbf{W}$ is a weight matrix of size $d' \times d$, $\textbf{b}$ a bias vector and together they are the parameters $\theta = \{\textbf{W},\textbf{b}\}$ of $f$. $s(x)=\frac{1}{1+e^{-x}}$ is the sigmoid function. The hidden representation is then mapped back to a vector $\textbf{z}\in [0,1]^{d}$ where $\textbf{z}=g_{\theta'}(\textbf{y})=s(\textbf{W'y}+\textbf{b'})$ with $\theta' = \{\textbf{W'},\textbf{b'}\}$. Optionally, the constraint $\textbf{W'}=\textbf{W}^{\top}$ can be applied. \cite{vincent2008extracting}

During the unsupervised representation learning, the autoencoder adapts its parameters to minimize the \textit{average reconstruction error} for training samples $\textbf{x}^{(i)}$ and corresponding $\textbf{y}^{(i)}$ and $\textbf{z}^{(i)}$:

\begin{equation}
    \theta^{*},\theta'^{*} = \argmin_{\theta,\theta'}\frac{1}{n}\sum_{i=1}^{n}L(\textbf{x}^{(i)},\textbf{z}^{(i)}) = \argmin_{\theta,\theta'}\frac{1}{n}\sum_{i=1}^{n}L(\textbf{x}^{(i)},g_{\theta'}(f_{\theta}(\textbf{x}^{(i)}))).
\end{equation}

The loss function $L$ can be anything compatible with the data, for example, the \textit{squared error} $L(\textbf{x},\textbf{z})=\Vert \textbf{x}-\textbf{z}\Vert^{2}$. Similar to \acp{DBN} and \acp{DBM}, the hidden layer of one autoencoder can become the input layer of another one to learn higher-level representations or pre-train the weights to generate a neural network from them later. \cite{vincent2008extracting}

\acp{RBM} and autoencoders are very similar in structure (\acp{RBM} are undirected, while autoencoders are usually directed graphs), but they differ in training procedure and hidden representation: The autoencoder considers the real-valued mapping from the input as its representation, while the Boltzmann machine samples a binary representation from that real-valued mapping. Autoencoders can also be seen as deterministic feedforward neural networks, while \acp{RBM} can be defined as the generative stochastic variant. \cite{vincent2010stacked,sarroff2014musical}

\othertab{autoencoders}{
    \cite{gregor2014deep} & Deep Autoregressive Network (DARN): A deep autoencoder with a mixture of stochastic and deterministic hidden units that incorporate autoregressive connections in the same layer, so $p(h)=\prod_{j=1}^{n_{h}}p(h_{j}\vert h_{1:j-1})$. Outperforms \ac{RBM} and \ac{NADE} on image log-likelihood. & 2014 \\\hline 
    \cite{sarroff2014musical} & The first evaluation of one and two-layer autoencoders on music audio spectrograms. & 2014 \\\hline 
    \cite{li2015hierarchical} & An hierarchical \acf{LSTM} autoencoder with an attention mechanism that builds and reconstructs embeddings of words, sentences, and paragraphs. The model is capable of coherent multi-sentence generation. & 2015 \\\hline 
    \cite{rasmus2015semi} & Using so-called Ladder Networks, a semi-supervised learning model combining supervised and unsupervised learning in deep neural networks. It enhances learning by adding denoising tasks at each level of the model, fostering more robust feature learning. The model's effectiveness is demonstrated on several benchmark datasets, where it achieves state-of-the-art performance in semi-supervised learning tasks. & 2015 \\\hline
    \cite{maaloe2016auxiliary} & Extending several state-of-the-art network architecture approaches by introducing auxiliary variables to deep generative models, which improve variational distribution approximation. & 2016 \\\hline
    \cite{assouel2018defactor} & DEFactor: Conditional molecule generation with optimal properties using a graph convolutional network \cite{kipf2016semi} as an encoder to produce a latent graph representation, a \ac{LSTM} creates a sequence of node embeddings autoregressively, and a decoder determines the edge and node types using a similarity measure of the node embeddings. Additionally, an existence module \ac{MLP} stops the \ac{LSTM} generator when a non-informative embedding is encountered to generate graphs of arbitrary size. & 2018 \\\hline 
    \cite{ulyanov2018adversarial} & Adversarial Generator-Encoder Network (AGE): A generator creates data from a specified latent distribution $z\sim p(z)$ and an adversarial encoder converts real and generated data to latent vectors. The AGE learns by comparing the distributions of the real and fake latent vectors with $p(z)$. The model is suitable for conditional and unconditional generation, converges quickly, and does not require a discriminator. & 2018 \\\hline
    \cite{tiga-augmentation} & Augmenting time series data through time-warped autoencoders. The authors introduce two techniques - independent and dependent - and showcase their effectiveness in producing synthetic data samples. This method utilizes the characteristics of auto-encoders to create high-quality, realistic time series data. & 2021 \\
}

\subsubsection{Helmholtz Machines}\label{sec:helmholtz_machines}

\begin{figure} [ht]
    \centering
    \includegraphics[width=0.5\textwidth]{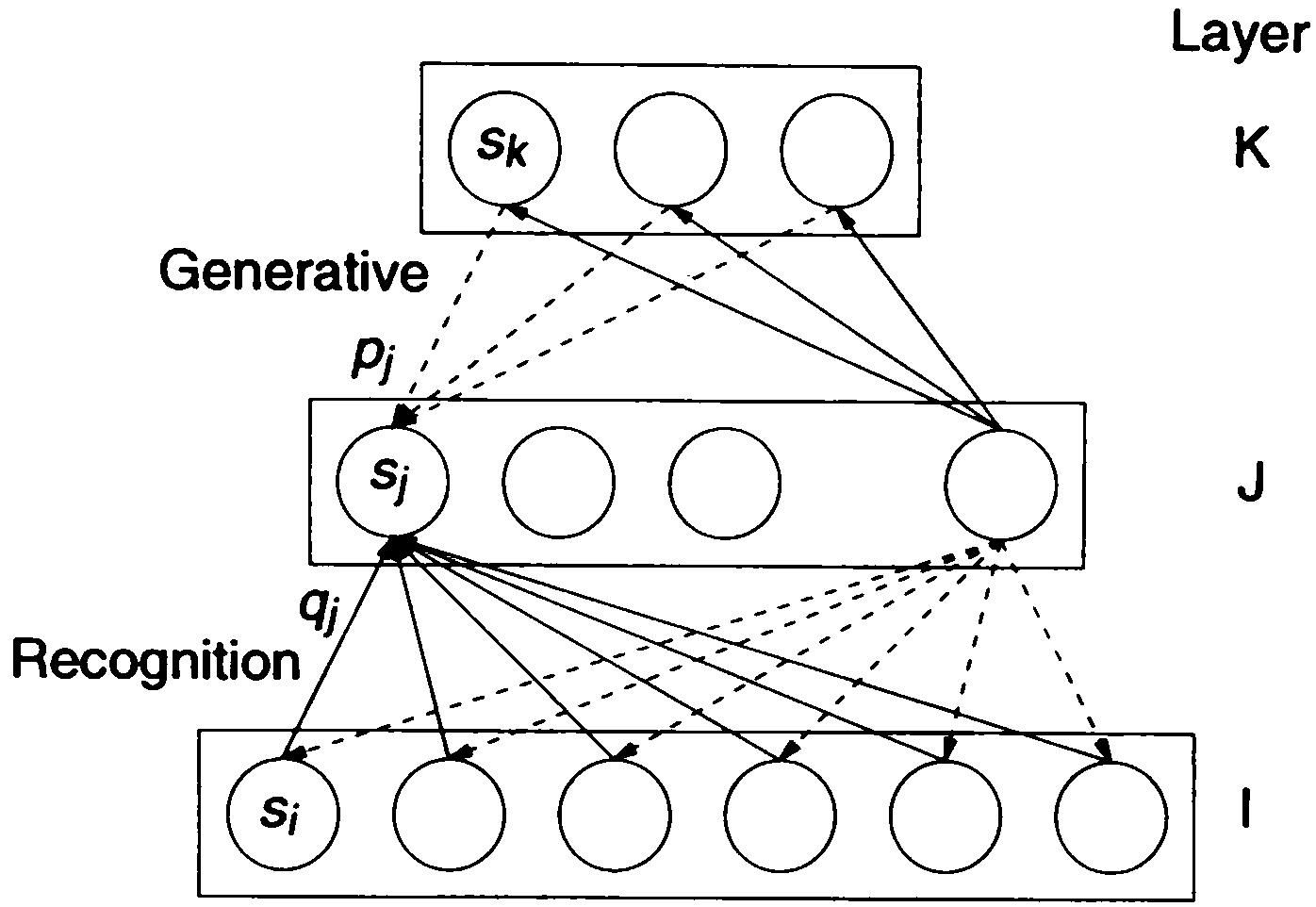}
    \caption{Architecture of a Helmholtz machine with three layers K, L, and I and probabilities $p$ for generation and $q$ for recognition. (Source: \cite{hinton1995wake})}
    \label{fig:helmholtz}
\end{figure}

The Helmholtz machine \cite{dayan1995helmholtz} is a directed deep generative model \cite{li2018exploring} with binary units, biases, and generative and recognition weights for the respective direction (see \autoref{fig:helmholtz}). The weights are trained with the ``wake-sleep'' algorithm \cite{hinton1995wake}: During the ``wake'' (recognition) phase, the input values are propagated bottom-up through the layers to create a representation. The state $s_{v}$ of unit $v$ with bias $b_{v}$ and incoming weights $w_{uv}$ is defined as 
\begin{equation}
    p(s_{v}=1)=\frac{1}{1+\exp(-b_{v}-\sum_{u}s_{u}w_{uv})}.
\end{equation}
Then the generative weights of the hidden states $s^{\alpha}_{i}$ between layers $k$ and $j$ are updated with the delta rule $\Delta w_{kj}=\epsilon s^{\alpha}_{k}(s^{\alpha}_{j}-p^{\alpha}_{j})$ and learning rate $\epsilon$.

The ``sleep'' phase generates the states of the lower layers from the highest layer in a top-down approach. It updates the recognition weights of all states with another delta rule $\Delta w_{jk}=\epsilon s_{j}(s_{k}-q_{k})$ where $q_{k}$ is the probability that unit $k$ is activated by the recognition weights of the states $s_{j}$ of the lower layer.

\subsubsection{Denoising Autoencoder}

\begin{figure} [ht]
    \centering
    \includegraphics[width=0.8\textwidth]{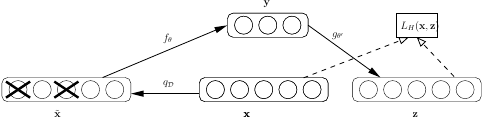}
    \caption{Training process of a \acl{DAE}. (Source: \cite{vincent2008extracting})}
    \label{fig:autoencoder_denoising}
\end{figure}

\acfp{DAE} modify the training process of basic autoencoders by randomly corrupting the input vector $\textbf{x}$ (i.e., setting a fixed proportion of values to zero), resulting in $\tilde{\textbf{x}}$, and then trying to restore it with the transformations $f$ and $g$ (see \autoref{fig:autoencoder_denoising}). The data corruption improves the model's generalization abilities and, therefore, the representations' robustness. Also, the constraint $d'<d$ of the basic autoencoder to prevent overfitting can now be omitted. \cite{vincent2008extracting}

\begin{figure} [ht]
    \centering
    \begin{subfigure}[t]{0.45\textwidth}
        \centering
        \includegraphics{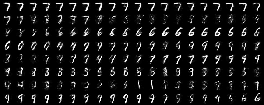}
        \caption{Training with simple corruption.}
    \end{subfigure}
    \hspace{0.5cm}
    \begin{subfigure}[t]{0.45\textwidth}
        \centering
        \includegraphics{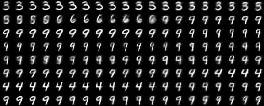}
        \caption{Training with walkback.}
    \end{subfigure}
    \caption{Comparison of the MNIST results of a \ac{DAE} trained normally or with walkback. (Source: \cite{bengio2013generalized})}
    \label{fig:walkback_mnist}
\end{figure}

Bengio et al. \cite{bengio2013generalized} propose a generalized probabilistic interpretation of \acp{DAE}, where an observed random variable $X$ is corrupted using a conditional distribution $\mathcal{C}(\overline{X}\vert X)$ and the \ac{DAE} is trained to estimate the reverse conditional $P_{\theta}(X\vert\overline{X})$, where $\theta$ are the trainable parameters. The sampling process from this \ac{DAE} is realized using a Markov chain where $X_{t}\sim P_{\theta}(X\vert\overline{X}_{t-1})$ and $\overline{X}_{t}\sim\mathcal{C}(\overline{X}\vert X_{t})$, which the authors prove generates the data-generating distribution $P(X)$ with a properly trained model. They introduce \textit{walkback} training, where the default corruption process $\mathcal{C}$ is replaced with a walkback process $\overline{\mathcal{C}}$, which generates one or more boosted ``negative examples'' $\overline{X}^{*}$ for a training sample $X$ by going a random-length walk through the aforementioned Markov chain with the current model parameters. The training with these greatly divergent corruptions, which may not even be represented by the training data or simple corruptions, prevents the \ac{DAE} from deviating too far from the plausible prediction range. The experiments on the MNIST dataset (see \autoref{fig:walkback_mnist}) show that the results of the Markov chain of the walkback-trained model look more natural than the ones obtained from the model trained with a simple corruption process.

\subsubsection{Contractive Autoencoder}

The \acf{CAE} uses the constraint $\textbf{W'}=\textbf{W}^{\top}$ and improves the robustness of the hidden representation $\textbf{y}$ by penalizing sensitivity to the input $ \textbf{x}$ with the Frobenius norm of the Jacobian matrix $J_{f}(\textbf{x})$, which is the sum of squares of all partial derivatives of the extracted features $\textbf{y}=f(\textbf{x})$ concerning the input $\textbf{x}$ \cite{rifai2011contractive,rifai2012generative}:

\begin{equation}
    J_{f}(\textbf{x})=\frac{\partial f(\textbf{x})}{\partial \textbf{x}}
\end{equation}

\begin{equation}
    \Vert J_{f}(\textbf{x})\Vert^{2}_{F}=\sum_{ij}(\frac{\partial f_{j}(\textbf{x})}{\partial\textbf{x}_{i}})^{2}.
\end{equation}

In the case of the sigmoid activation function being used, this results in the following:

\begin{equation}
    \Vert J_{f}(\textbf{x})\Vert^{2}_{F}=\sum_{i=1}^{d'}(\textbf{y}_{i}(1-\textbf{y}_{i}))^{2}\sum_{j=1}^{d}\textbf{W}_{ij}^{2}.
\end{equation}

This objective function is minimized on the data $D_{n}$, with hyperparameter $\lambda\geq0$, parameters $\theta=\{\textbf{W},\textbf{b},\textbf{b'}\}$ and cross-entropy loss $L(\textbf{x},\textbf{z})$:

\begin{equation}
    \mathcal{J}_{CAE}(\theta)=\sum_{\textbf{x}\in D_{n}}(L(\textbf{x},g(f(\textbf{x})))+\lambda\Vert J_{f}(\textbf{x})\Vert^{2}_{F})
\end{equation}

\begin{equation}
    L(\textbf{x},\textbf{z})=-\sum_{i=1}^{d}\textbf{x}_{i}\log(\textbf{z}_{i})+(1-\textbf{x}_{i})\log(1-\textbf{z}_{i})
\end{equation}

Rifai et al. \cite{rifai2012generative} propose an algorithm (see Algorithm \ref{alg:cae_sampling_rifai}) to generate samples from a pre-trained \ac{CAE} that provides an ergodic Harris chain with a stationary distribution $\pi$ under the condition that $J_{t}J_{t}^{\top}$ is full rank.

\begin{algorithm}
    \caption{Sampling algorithm for a \ac{CAE}. (Source: \cite{rifai2012generative})}\label{alg:cae_sampling_rifai}
    \begin{algorithmic}
        \REQUIRE $f$, $g$, step size $\sigma$ and chain length $T$
        \ENSURE Sequence $(x_{1},y_{1})$,$(x_{2},y_{2})$,...,$(x_{T},y_{T})$
        \STATE Initialize $x_{0}$ arbitrarily and $y_{0}=f(x_{0})$.
        \FOR{$t=0$ \TO $T$}
            \STATE Let Jacobian $J_{t}=\frac{\partial f(x_{t})}{\partial x_{t}}$.
            \STATE Let $\epsilon \sim N(0,\sigma I_{k})$ isotropic Gaussian noise.
            \STATE Let perturbation $\Delta y=J_{t}J_{t}^{\top}\epsilon$.
            \STATE Let $x_{t}=g(y_{t-1}+\Delta y)$ and $y_{t}=f(x_{t})$.
        \ENDFOR
    \end{algorithmic}
\end{algorithm}

They test their technique on a \ac{CAE} with two-layer stacks and compare it against a 2-layer \ac{DBN}, resulting in slightly better/worse performance on the \ac{TFD}/MNIST dataset regarding the log-likelihood of the generated samples but significantly reduced sensitivity to deformations of MNIST digits.

Bengio et al. \cite{bengio2013better} hypothesize that sampling from higher-level representations improves the quality (log-likelihood) and class variation of obtained samples. They prove their claims by sampling from high-level representations generated by a Markov chain process from various depths of multi-layer \acp{CAE} and \acp{DBN} and measuring the log-likelihood. Further, they test the mixing of representations of digits at various depths with linear interpolation, which also gives more plausible samples at higher levels.

\subsubsection{Generative Stochastic Network}

\begin{figure} [ht]
    \centering
    \includegraphics[width=0.5\textwidth]{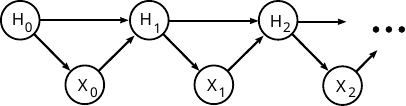}
    \caption{The Markov chain structure of a \ac{GSN}. (Source: \cite{bengio2014deep})}
    \label{fig:gsn_markov_chain}
\end{figure}

A \acf{GSN} \cite{bengio2014deep} is a generalized framework of a deep \ac{DAE} that has the structure of a Markov chain and can be trained with back-propagated gradients and without layer-wise pre-training. The transition operator $P(x_{t},h_{t}\vert x_{t-1},h_{t-1})$ is responsible for generating the next visible state $X_{t}$ and hidden state $H_{t}$ of the Markov chain (see \autoref{fig:gsn_markov_chain}).

To enable the back-propagation of the reconstruction log-likelihood $\log P(X_{1}=x_{0}\vert H_{1})$ into all the parameters of the encoding function $f_{\theta_{1}}$ and reconstruction function $g_{\theta_{2}}$, a deterministic function is used to define $H_{t+1}=f_{\theta_{1}}(X_{t},Z_{t},H_{t})$ with $Z_{t}$ being an independent noise source so $X_{t}$ cannot be exactly recovered from $H_{t+1}$. This resembles the masking of values in the input layer of a \ac{DAE}.

\begin{figure} [ht]
    \centering
    \includegraphics[width=0.8\textwidth]{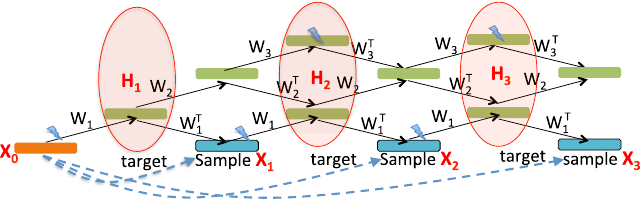}
    \caption{A \ac{GSN} inspired by the Gibbs sampling process of a \ac{DBM}. The lightning symbols indicate the corruption of the samples with salt-and-pepper noise. (Source: \cite{bengio2014deep})}
    \label{fig:gsn_dbm_example}
\end{figure}

Bengio et al. \cite{bengio2014deep} use the \ac{GSN} framework to adapt the Gibbs sampling process of a \ac{DBM} (see \autoref{fig:gsn_dbm_example}), but with the ability to use the \ac{GSN}'s backpropagation at each layer. The chain starts with a training sample $X=x_{0}$ and encodes and reconstructs intermediate samples $x_{i}$ several times. The training of the model is realized using the sum of all log-likelihoods to the target $X=x_{0}$, inspired by the \textit{walkback} objective (see \cite{bengio2013generalized}).

\subsubsection{Variational Autoencoder}\label{sec:vae}

\acfp{VAE} is an autoencoder with encoder $\mathcal{E}(.)$ and decoder $\mathcal{D}(.)$ whose hidden layer is represented as a Gaussian distribution $\mathcal{N}(\mu,\sigma^{2})$ with mean vector $\mu$ and standard deviation vector $\sigma$ that are obtained from the encoder $\mu,\sigma=\mathcal{E}(x)$, where $x\sim\mathbf{X}$. The decoder samples from this distribution and reconstructs the data $\hat{x}=\mathbb{E}_{z\sim\mathcal{N}(\mu,\sigma)}[\mathcal{D}(z)]$. Additionally, the aggregated distribution of $z$ over all data $\mathbf{X}$ is constrained to be $\mathcal{N}(0,\mathbf{I})$, allowing random vectors to be sampled from $\mathcal{N}(0,\mathbf{I})$ to be used for data generation with the decoder. \cite{xu2020synthesizing}

The \ac{VAE} is trained using the \ac{ELBO} loss

\begin{equation}
    \mathcal{L}=\mathbb{E}_{x\sim\mathbf{X}}[\mathbb{E}_{z\sim\mathcal{N}(\mu,\sigma\mathbf{I})}[\Vert\mathcal{D}(z)-x\Vert^{2}_{2}]+\mathbb{KL}(\mathcal{N}(\mu,\sigma\mathbf{I})\Vert\mathcal{N}(0,\mathbf{I}))],
\end{equation}

where the first term encourages the autoencoding part while the second term with the Kullback-Leibler (KL) divergence $\mathbb{KL}(p\Vert q)$ measures the difference between the probability distributions $p$ and $q$. \cite{xu2020synthesizing}

Kingma et al. \cite{kingma2013auto} applied the \ac{VAE} as a generative model with \ac{MLP} encoder and decoder to the MNIST and Frey Face datasets. They achieve faster convergence and a higher marginal likelihood than their reference, the wake-sleep algorithm, described in \autoref{sec:helmholtz_machines}.

Gregor et al. \cite{gregor2015draw} introduce the \textit{Deep Recurrent Attentive Writer} (DRAW) for image generation. The authors follow the human drawing process, where rough outlines are iteratively refined until a realistic picture is generated. Similar to a \ac{VAE}, DRAW consists of an encoder that learns a representation $\mathbf{z}\sim p(\mathbf{z})$ of the input and a decoder that reconstructs the input from $\mathbf{z}$, but both models are \acp{LSTM} that only handle regions of the full input defined by an attention mechanism at each time step. For evaluation, the authors use MNIST, the \ac{SVHN} data set, and CIFAR-10, and they achieve highly realistic results on the first two while overfitting the last data set due to only 50,000 training samples.

Rezende et al. \cite{rezende2016one} propose the class of \textit{sequential generative models}, which generate $T$ groups of $k$ latent variables sequentially instead of generating $K=kT$ latent variables at once and also sequentially reconstruct data to a modifiable canvas while incorporating inference and writing attention mechanisms. At the core of the model are one or more \acp{RNN} (both \acp{LSTM} and \acp{GRU}) and attention-based neural networks like spatial transformers \cite{jaderberg2015spatial} to write the \ac{RNN} output to the canvas. The model performs well on unconditional sampling and \textit{one-shot learning} tasks, where a concept is only encountered once and compelling variations of the concept should be generated, or multiple concepts (e.g., letters of an alphabet) are provided during the sequential inference process, and a plausible new character is generated.

Higgins et al. \cite{higgins2016beta} propose $\beta$-\ac{VAE}, a reformulation of the original unsupervised \ac{VAE} ($\beta=1$) with an additional hyperparameter $\beta$ that encourages the model to learn better-disentangled representations of data, meaning that single hidden units encode single generative factors while being invariant to changes in others (e.g., skin color in face images), by enforcing more independence and less covariance of the latent variables and their distributions. $\beta$-\ac{VAE} outperforms the previous unsupervised state-of-the-art model InfoGAN \cite{chen2016infogan} and semi-supervised DC-IGN \cite{kulkarni2015deep} in terms of disentanglement of factors qualitatively and quantitatively.

Tomczak et al. \cite{tomczak2018vae} implement the \ac{VAE}'s prior as a mixture distribution (e.g., a \ac{GMM}) that can be trained with pseudo inputs and propose multiple layers of variables. They evaluate their model on multiple MNIST data sets, OMNIGLOT, Caltech 101 Silhouette, Frey Face, and Histopathology patches, resulting in state-of-the-art log-likelihood compared to a normal \ac{VAE} and avoidance of its local optima problem.

\othertab{\acp{VAE}}{
    \cite{fabius2014variational} & Variational Recurrent Autoencoder (VRAE): A \ac{VAE} with \ac{RNN} encoder and decoder for modeling sequential data. The model is used for MIDI music generation and creates ``medleys'' of the training data. & 2014 \\\hline 
    \cite{im2015denoising} & Denoising Variational Autoencoder (DVAE): Combination of the noise injection at the input like a \ac{DAE} and the noise injection at the hidden layer of a \ac{VAE} improves average log-likelihood results on MNIST and Frey Face data sets. & 2015 \\\hline 
    \cite{burda2015importance} & Importance-weighted autoencoder (IWAE): A \ac{VAE} with a tighter log-likelihood lower bound on $\log p(\mathbf{x})$ based on importance weighting using multiple samples $q(\mathbf{h}_{i}\vert\mathbf{x})$, that learns richer representations than \acp{VAE}, outperforming them on MNIST density estimation. & 2015 \\\hline 
    \cite{bowman2015generating} & A \ac{VAE} with single-layer \ac{LSTM} encoder and decoder for sentence generation. Interpolation between latent vectors of two sentences provides interesting results. & 2015 \\\hline 
    \cite{kulkarni2015deep} & Deep Convolutional Inverse Graphics Network (DC-IGN): A \ac{VAE} built with a convolutional encoder and deconvolutional decoder. The model is further encouraged to assign certain transformations (e.g., lighting, rotation) in images to disentangled neuron groups unsupervised by training with mini-batches of transformed images. & 2015 \\\hline 
    \cite{huang2015efficient} & Composited spatially transformed \ac{VAE} (CST-VAE): Layer-wise sequential image generation using pose and content encoder-decoder pairs on the partial results and a spatial transformer network \cite{jaderberg2015spatial} to output the next image layer front to back. & 2015 \\\hline 
    \cite{larsen2016autoencoding} & \ac{VAE} training with an additional similarity loss obtained from a jointly trained \ac{GAN} discriminator for higher image quality and better representations. & 2016 \\\hline 
    \cite{yan2016attribute2image} & Attribute2Image: Using a \ac{VAE} to learn disentangled latent representations of attributes (e.g., color, gender, viewpoint) from images and generate new images conditioned on attributes. & 2016 \\\hline 
    \cite{gomez2016automatic} & A \ac{VAE} with recurrent encoder and decoder (similar to \cite{bowman2015generating}) is used to generate SMILES \cite{weininger1988smiles} text encodings of new valid molecules with desirable properties by using gradient-based optimization in the latent space. & 2016 \\\hline 
    \cite{chen2016variational} & Proposal of the variational lossy autoencoder (VLAE), which allows the user to control what the latent variable can contain by limiting the receptive field of the autoregressive encoder and decoder. Combined with an autoregressive model modeling $p(z)$, state-of-the-art results are achieved on MNIST, OMNIGLOT, and Caltech-101 density estimation (competitive on CIFAR-10). & 2016 \\\hline 
    \cite{gulrajani2016pixelvae} & PixelVAE: Combination of a \ac{VAE} with a PixelCNN-based \cite{oord2016pixel} autoregressive decoder that iteratively refines the image result. It performs comparably to PixelCNN with fewer autoregressive layers and a smaller latent variable than a normal \ac{VAE}. & 2016 \\\hline 
    \cite{kusner2017grammar} & GrammarVAE: To better generate discrete data, it is converted to a parse tree using a context-free grammar, and the rules used by the tree are then one-hot encoded in order, formatted as a matric and mapped to a latent space with a \acf{CNN}. A \ac{RNN} decodes from this latent space back to valid rules. The model generates molecule structures and arithmetic expressions, outperforming text-based representations. & 2017 \\\hline 
    \cite{roberts2017hierarchical} & Recurrent hierarchical \ac{VAE} with BiLSTM encoder and 3-layer \ac{LSTM} decoder for the creative reconstruction of short musical sequences with random sampling or interpolation in the latent space. & 2017 \\\hline 
    \cite{yang2017improved} & A \ac{VAE} with \ac{LSTM} encoder and \ac{CNN} decoder with dilated convolutions for text modeling and generation. Changes in the dilation configuration give control over the context size from previous words, and the convolutional decoder is less prone to ignore encoder information because its contextual capacity is lower than an \ac{LSTM}'s. & 2017 \\\hline 
    \cite{oord2017neural} & Vector-Quantised \ac{VAE} (VQ-VAE): The encoder \ac{CNN} outputs $z$ are discrete to enforce more efficient usage of the latent space and prevent ``posterior collapse'', which is often caused by decoders ignoring $z$. The model is combined with an autoregressive decoder (PixelCNN for images, WaveNet for audio) and provides similar results to continuous \acp{VAE}. & 2017 \\\hline 
    \cite{semeniuta2017hybrid} & Character-level text generation \ac{VAE} with convolutional encoder combined with a deconvolutional decoder with recurrent output layer. The \acp{CNN} make \ac{VAE} training easier and an additional cost function encourages reliance of the decoder on the latent vector. Experiments on Twitter tweet generation show more diverse and coherent samples than a \ac{LSTM}-based \ac{VAE}. & 2017 \\\hline 
    \cite{ha2017neural} & Sketch-\ac{RNN}: Sketch-conditional and unconditional stroke-based sketch generation with a sequence-to-sequence \ac{VAE} with bidirectional \ac{RNN} encoder and \ac{RNN} decoder. Possible applications also include latent space interpolation and sketch completion. & 2017 \\\hline 
    \cite{tikhonov2017music} & First application of a recurrent \ac{VAE} \cite{bowman2015generating} to music generation, providing a good balance between local and global structures. & 2017 \\\hline 
    \cite{engel2017latent} & Training a \ac{GAN} to generate and modify the latent code $z$ of an unconditional \ac{VAE} with $z'=G(z,y_{attr})$ to satisfy specific properties enforced by $D(z',y_{attr})$. Allows zero-shot conditional generation and identity-preserving transformation (e.g., same face with different hair color) of data from an unconditional \ac{VAE} model. & 2017 \\\hline 
    \cite{jorgensen2018machine} & Application of the grammar \ac{VAE} \cite{kusner2017grammar} to molecule generation with desired properties by randomly sampling $10^{6}$ samples from $p(\mathbf{z})$ and iteratively encoding and decoding them with the autoencoder many times before filtering the results with neural network regression functions to get the best samples for the desired property. & 2018 \\\hline 
    \cite{simonovsky2018graphvae} & GraphVAE (GVAE): Generation of probabilistic fully connected graphs from which can be sampled. The convolutional encoder gets as input a graph $G=(A,E,F)$ and graph label vector $\mathbf{y}$, where $A$ is the adjacency matrix, $E$ the edge attribute tensor, and $F$ a node attribute matrix, and computes a latent representation $\mathbf{z}$. The deterministic decoder \ac{MLP} takes $\mathbf{z}$ and $\mathbf{y}$ and outputs $\tilde{G}=(\tilde{A},\tilde{E},\tilde{F})$, containing the independent node and edge, edge class and node class \textit{probabilities} respectively for graphs with a fixed maximum number of $k$ nodes. The model is demonstrated on molecule generation tasks and is only suitable for slightly larger graphs than the provided input. & 2018 \\\hline 
    \cite{roberts2018hierarchical} & MusicVAE: Recurrent \ac{VAE} with two-layer bidirectional \ac{LSTM} encoder and hierarchical \ac{RNN} decoder, which consists of a two-layer unidirectional \ac{LSTM} conductor that creates subsequence embeddings from the latent vector, and a two-layer \ac{LSTM} decoder \ac{RNN} that produces the final subsequence output inside these separate embeddings. The model achieves promising results on MIDI music reconstruction tasks. & 2018 \\\hline
    \cite{jin2018junction} & Junction tree \ac{VAE} (JT-VAE): Encoding and decoding molecules using two representations: A fine-grained graph connectivity representation obtained from a message-passing network \cite{gilmer2017neural} and a representation of a junction tree (also from a message passing network), that models a molecule as a composition of valid subgraph components and avoids the node-by-node generation of invalid intermediaries. & 2018 \\\hline 
    \cite{liu2018constrained} & Constrained Graph \ac{VAE} (CGVAE): Using gated graph neural networks as encoder and decoder for molecular graph generation. The encoder maps a graph with a maximum of $N$ nodes to $N$ latent codes $\mathbf{z}_{v}$ conforming to a Gaussian distribution. The decoder initializes a graph with $N$ nodes and a ``stop node'' from the latent codes. Then, a loop starts where new edges are added and labeled, and node representations are updated with messages from neighbors (see~\cite{gilmer2017neural}) until an edge to the stop node is created. Correct atom valency is always enforced to guarantee valid molecules. & 2018 \\\hline 
    \cite{dai2018syntax} & Syntax-directed \ac{VAE} (SD-VAE): Applying syntax and semantic constraints to the decoder of a grammar-based \ac{VAE} similar to GrammarVAE \cite{kusner2017grammar} to enforce syntactically valid and semantically reasonable reconstruction and optimization of molecule structures and program code. & 2018 \\\hline 
    \cite{chen2018differentially} & Differentially private autoencoder-based generative model (DP-AuGM) \& variational autoencoder-based model (DP-VaeGM): Autoencoders are trained using stochastic gradient descent with clipped gradients and noise injection \cite{abadi2016deep}. For DP-AuGM, only the encoder makes confidential data ``private''. The \ac{VAE} version trains one model for each class, samples $z$ from each model, and merges the decoded samples, which is less stable than the former approach. & 2018 \\\hline 
    \cite{huang2018introvae} &  IntroVAE: Integrating the concepts of \ac{VAE} and \ac{GAN} into a single model that is both a generator and a discriminator. This model self-evaluates the quality of generated images and improves itself accordingly, offering stable training and high-resolution image synthesis. The approach combines the advantages of \ac{VAE} and \ac{GAN} without needing extra discriminators, simplifying the architecture and improving training efficiency. & 2018 \\\hline
    \cite{shen2019towards} & A multilevel \ac{VAE} architecture for generating coherent and long text sequences. The encoder consists of a lower and higher-level \ac{CNN} producing separate latent representations where the lower latent vector is additionally conditioned on the upper. The lower representation is then fed to a hierarchical \ac{LSTM} decoder network with a sentence-level and word-level \ac{LSTM}. The model performs well on conditional (title-to-paragraph) and unconditional text generation. & 2019 \\\hline 
    \cite{wang2019topic} & Topic-guided \ac{VAE} (TGVAE): Modeling the latent space of a \ac{VAE} using a \ac{GMM} prior distribution parametrized by a neural topic module that is powered by the bag-of-words representation of the text. A \ac{GRU} encoder and decoder processed the text inputs to outputs. The model is used for text generation and summarization. & 2019 \\\hline 
    \cite{bresson2019two} & A \ac{VAE} for molecule generation that uses a graph-convolutional encoder, a \ac{MLP} to create a ``bag-of-atoms'' (counts for certain atoms in the target reconstruction), and a graph-convolutional decoder that takes the latent representation and the atom bag to create an edge probability matrix from which a beam search generates a discrete output. & 2019 \\\hline 
    \cite{samanta2020nevae} & NeVAE: Generating molecular graphs with variable size from a convolution-style \ac{VAE} with variable-length latent representations (one for each node). In addition to the atoms' types and their bonds, the model also predicts their positions. The decoder can be further optimized with a gradient-based algorithm to maximize specific properties of the generated molecules. & 2020 \\\hline 
    \cite{guo2020interpretable} & Node-Edge Disentangled \ac{VAE} (NED-VAE): Using three convolutional sub-encoders (node, edge, and graph) and two deconvolutional sub-decoders (node and edge), both with access to the graph representation, to reconstruct the node and edge attributes from which a graph can be recreated. The model also enforces the disentanglement of latent factors of nodes, edges, and joint patterns. & 2020 \\\hline 
    \cite{xu2020synthesizing} & Tabular \ac{VAE} (TVAE): Tabular data generation with a \ac{VAE} using probability distributions to encode discrete and continuous values. & 2020 \\\hline 
    \cite{nazabal2020handling} & HI-VAE: A \ac{VAE} that can handle incomplete (missing values at random) and heterogeneous (mixed continuous and discrete) data by learning the influence of input variables on the latent code individually, using special distributions for discrete variables, and only using observed variables for recognition (i.e., replace missing with zero). & 2020 \\\hline 
    \cite{flam2020graph} & Message Passing Graph \ac{VAE} (MPGVAE): Building on top of the GraphVAE \cite{simonovsky2018graphvae}, the authors use message passing neural networks \cite{gilmer2017neural} for both encoder and decoder, where edge and node representations are alternatingly updated multiple times based on messages from their neighbors. & 2020 \\\hline 
    \cite{xu2020synthesizing} & Besides a conditional \ac{GAN}, the authors introduce a tabular \ac{VAE} for generating synthetic tabular data using . This method addresses challenges in modeling tabular data, which often contains a mix of discrete and continuous columns and may exhibit imbalances and non-Gaussian distributions. It employs mode-specific normalization and reversible data transformations to generate synthetic data effectively. & 2020 \\\hline
    \cite{podda2020deep} & Treating the molecule as a sequence of valid SMILES-encoded \cite{weininger1988smiles} fragments/components, this approach uses \acp{GRU} to encode a molecule to a latent vector with Gaussian distribution and decode it back. The whole training process incorporates \textit{low-frequency masking}, which masks rarely encountered fragments with a mask token that is replaced during sampling by any of the suitable masked fragments with uniform probability to improve uniqueness. & 2020 \\ 
}

\subsubsection{Deep Latent Gaussian Models}

A \acf{DLGM} is a deep, directed generative model powered by Gaussian latent variables. A recognition model with $L$ layers encodes the training observations to provide the layer-wise parameters for the Gaussian distributions of the generation model. The generative process starts at the top latent layer ($L$) and draws mutually independent Gaussian variables $\mathbf{\mathcal{E}}_{l}\sim\mathcal{N}(\mathbf{\mathcal{E}}_{l}\vert\mathbf{0},\mathbf{I})$. Each layer $\mathbf{h}_{l}=T_{l}(\mathbf{h}_{l+1})+\mathbf{G}_{l}\mathbf{\mathcal{E}}_{l}$ below the top layer $\mathbf{h}_{L}=\mathbf{G}_{L}\mathbf{\mathcal{E}}_{L}$ depends on the layer above, where $T_{l}$ are \ac{MLP} transformations and $\mathbf{G}_{l}$ are matrices. The visible data $\mathbf{v}=\pi(\mathbf{v}\vert T_{0}(\mathbf{h}_{1}))$ is generated from a distribution $\pi$. The model is trained using stochastic backpropagation, i.e., by computing gradients through random variables. \cite{rezende2014stochastic}

The original \ac{DLGM} authors Rezende et al. \cite{rezende2014stochastic} evaluate the generative abilities of a three-layer \ac{DLGM} on MNIST, CIFAR-10, the Frey faces data set, and NORB. They also propose an imputation use case, where the Gaussian model fills in missing data in the \ac{SVHN}, Frey faces, and MNIST data sets. Their experiments achieve realistic-looking results comparable to other contemporary approaches such as the \ac{RBM} and \ac{DBN}.

\subsubsection{Gated Autoencoders}\label{sec:gae}

A \acf{GAE} is a conditional bi-linear model that learns to represent a linear transformation encoded as \textit{mapping units} $\mathbf{m}$ between two observations $\mathbf{x}^{(1)}$ and $\mathbf{x}^{(2)}$ using parameter matrices $\mathbf{U}$, $\mathbf{V}$ and $\mathbf{W}$ with

\begin{equation}
    \mathbf{m}=\sigma(\mathbf{W}(\mathbf{Ux}^{(1)}\cdot \mathbf{Vx}^{(2)})).
\end{equation}

The mappings, since the \ac{GAE} is a symmentric model, can be used to reconstruct $\mathbf{x}^{(1)}$ or $\mathbf{x}^{(2)}$ respectively based on the other one \cite{michalski2014modeling}:

\begin{equation}
    \tilde{\mathbf{x}}^{(2)}=\mathbf{V}^{T}(\mathbf{Ux}^{(1)}\cdot\mathbf{W}^{T}\mathbf{m})
\end{equation}

\begin{equation}
    \tilde{\mathbf{x}}^{(1)}=\mathbf{U}^{T}(\mathbf{Vx}^{(2)}\cdot\mathbf{W}^{T}\mathbf{m}).
\end{equation}

Training works similarly to a \ac{DAE}: Input pairs are independently corrupted and concatenated. Then backpropagation and gradient-based optimization are used to minimize the loss function, for example, the symmetric reconstruction error \cite{michalski2014modeling,memisevic2011gradient}:

\begin{equation}
    \mathcal{L}=\Vert\mathbf{x}^{(1)}-\tilde{\mathbf{x}}^{(1)}\Vert^{2}+\Vert\mathbf{x}^{(2)}-\tilde{\mathbf{x}}^{(2)}\Vert^{2}
\end{equation}

\begin{figure} [ht]
    \centering
    \begin{subfigure}[t]{0.5\textwidth}
        \centering
        \includegraphics[width=0.9\textwidth]{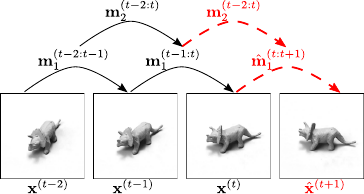}
    \end{subfigure}
    \begin{subfigure}[t]{0.4\textwidth}
        \centering
        \includegraphics[width=0.9\textwidth]{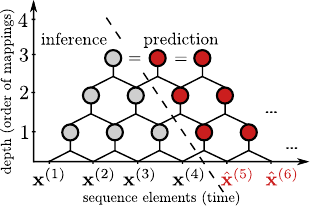}
    \end{subfigure}
    \caption{\textbf{Left:} A 2-layer pyramid model is used to predict the next transformation $\mathbf{\hat{m}}_{1}^{(t:t+1)}=\mathbf{V}_{2}^{T}(\mathbf{U}_{2}\mathbf{m}_{1}^{(t-1:t)}\cdot\mathbf{W}_{2}^{T}\mathbf{m}_{2}^{(t-2:t)})$ and the resulting observation $\mathbf{\hat{x}}^{(t+1)}=\mathbf{V}_{1}^{T}(\mathbf{U}_{1}\mathbf{x}^{(t)}\cdot\mathbf{W}_{1}^{T}\mathbf{\hat{m}}_{1}^{(t:t+1)})$ with $\mathbf{U}$, $\mathbf{V}$ and $\mathbf{W}$ being the filter matrices learned by the respective autoencoders. \textbf{Right:} Multi-step prediction with constant top-layer transformation in a 3-layer pyramid. (Source: \cite{michalski2014modeling})}
    \label{fig:pgp_michalski}
\end{figure}

Michalski et al. \cite{michalski2014modeling} model a time series as a sequence of transformations applied to its elements. During the \textit{predictive training}, a pyramid of stacked \acp{GAE} is used to learn basic and higher-order representations of transformations between observation pairs and predict future observations recurrently with a constant highest-order transformation (see \autoref{fig:pgp_michalski}). The autoencoders are initialized by $k$ \textit{seed frames} corresponding to the $k$ layers of the pyramid and optimized using backpropagation through time. The model is compared against a \ac{RNN} and a \ac{RBM} in the prediction of chirps (sinusoidal waves with changing frequencies) and video frames of bouncing balls and objects of the NORB data set, outperforming them in terms of mean squared error.

\subsubsection{Masked Autoencoders}

\begin{figure} [ht]
    \centering
    \includegraphics[width=0.6\textwidth]{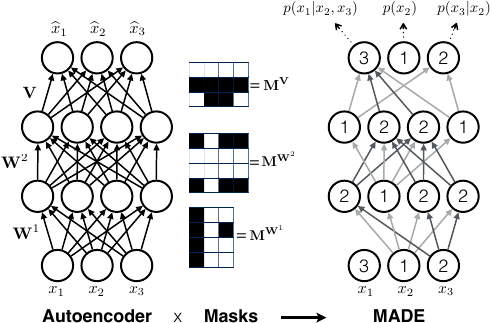}
    \caption{A deep masked autoencoder architecture. (Source: \cite{germain2015made})}
    \label{fig:masked_autoencoder}
\end{figure}

Masked autoencoders set weights in their input-to-hidden or hidden-to-output weight matrices to zero, meaning there is no computational path between certain input and output units, which is necessary, for example, for autoregressive tasks, where future inputs must not be seen. The masking approach also directly applies to deep architectures (see \autoref{fig:masked_autoencoder}). \cite{germain2015made}

Germain et al. \cite{germain2015made} propose the masked autoencoder for distribution estimation (MADE), which computes the joint probability distribution $p(\mathbf{x})$ of data $\mathbf{x}$ autoregressively as $p(\mathbf{x})=\prod_{d=1}^{D}p(x_{d}\vert\mathbf{x}_{<d})$ and therefore needs to mask the future inputs $x_{d},...,x_{D}$ for the respective steps. The model can also be used for sampling according to the calculated probabilities demonstrated by generating binary MNIST images.

\subsection{Neural Autoregressive Distribution Estimators}\label{sec:nade}

The \acf{NADE} is inspired by the \ac{RBM}, whose joint probability estimation of an observation is intractable and can also be interpreted as an autoencoder that assigns probabilities to binary observations. \ac{NADE} models a $D$-dimensional observation vector $\mathbf{x}$ as a product of the one-dimensional conditional distributions $p(\mathbf{x})=\prod_{i=1}^{D}p(x_{i}\vert\mathbf{x}_{<i})$. The probability of $x_{i}$ is conditioned on all previously seen observations $\mathbf{x}_{<i}$, so the ordering of the variables matters. \cite{larochelle2011neural}

Each probability output $\hat{x}_{i}=p(x_{i}=1\vert\mathbf{x}_{<i})$ depends on a $H$-dimensional hidden vector $\mathbf{h}_{i}$ that is computed recursively by

\begin{equation}
    p(x_{i}=1\vert\mathbf{x}_{<i})=\sigma(\mathbf{V}_{.,i}\mathbf{h}_{i}+b_{i})
\end{equation}

\begin{equation}
    \mathbf{h}_{i}=\sigma(\mathbf{W}_{.,<i}\mathbf{x}_{<i}+\mathbf{c}) \mbox{~and~} h_{1}=\sigma(\mathbf{c})
\end{equation}

with $\sigma$ being the sigmoid function and $\mathbf{V}\in\mathbb{R}^{D\times H}$, $\mathbf{b}\in\mathbb{R}^{D}$, $\mathbf{W}\in\mathbb{R}^{H\times D}$, and $\mathbf{c}\in\mathbb{R}^{H}$ being the parameters of the \ac{NADE} model. This corresponds with each $\hat{x}_{i}$ being computed by a neural network and all neural networks having tied weights for each observation. \cite{larochelle2011neural}

\begin{figure} [ht]
    \centering
    \includegraphics[width=0.2\textwidth]{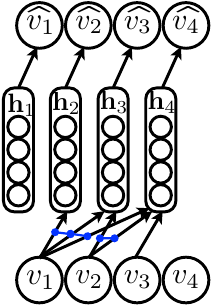}
    \caption{\ac{NADE} architecture (blue lines are tied weights) with $v$ instead of $x$ notation. (Source: \cite{larochelle2011neural})}
    \label{fig:NADE}
\end{figure}

\ac{NADE} is trained using gradient descent on the \ac{NLL} given a training set $\mathbf{X}$ with size $T$ \cite{larochelle2011neural}:

\begin{equation}
    \frac{1}{T}\sum_{t=1}^{T}-\log p(\mathbf{x}_{t})=\frac{1}{T}\sum_{t=1}^{T}\sum_{i=1}^{D}-\log p(x_{i}\vert\mathbf{x}_{<i}) \mbox{~for~} \mathbf{x}_{i}\in\mathbf{X}
\end{equation}

Uria et al. \cite{uria2013rnade} propose real-valued \ac{NADE} (RNADE), where real output values are computed using a \ac{GMM}, so $p(x_{i}\vert\mathbf{x}_{<i})=p_{GMM}(x_{i}\vert\mathbf{\theta_{i}})$ with $\mathbf{\theta_{i}}$ being the parameters of the \ac{GMM}.

In \cite{uria2014deep}, Uria et al. introduce an efficient procedure to train \ac{NADE} and RNADE models for each possible variable ordering simultaneously by using shared weights and stochastic gradient descent to optimize the mean cost over all orderings. After that, the most suitable variable ordering for the data can be determined in constant time. They also introduce a deep \ac{NADE} with multiple hidden layers that is efficient to train and often achieves better log-likelihood results on the test set than single-layer models.

Raiko et al. \cite{raiko2014iterative} propose NADE-$k$, which computed the density of an output as the $k$-th recurrent pass-through of the input $\mathbf{v}^{\langle1\rangle}$ through the neural network hidden layer, so $p(x_{i}=1\vert\mathbf{x}_{\mbox{obs}})=v_{i}^{\langle k\rangle}$. The model outperforms previous \ac{NADE} approaches, \acp{RBM} and \acp{DBN} in density modeling, also on masked inputs, and can generate binary MNIST digits and Caltech-101 silhouettes.

In \cite{uria2016neural}, Uria et al. propose ConvNADE, which replaces the fully connected hidden layers with convolutional layers, allowing exploitation of the spatial structure, for example, of 2D images. They also combine the approach with the DeepNADE \cite{uria2014deep} architecture, which uses masking to improve the results of image modeling tasks.

\subsection{Sparse Coding}

\begin{figure} [ht]
    \centering
    \includegraphics[width=0.8\textwidth]{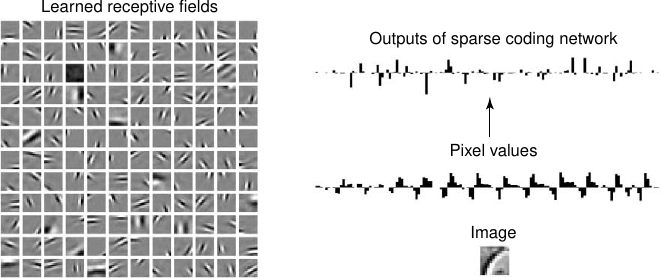}
    \caption{Example of an application of sparse coding to generate images. (Source: \cite{olshausen2004sparse})}
    \label{fig:sparse_coding}
\end{figure}

Sparse coding is usually an optimization problem, where data is reconstructed by a weighted linear combination of as few as possible basis vectors (see \autoref{fig:sparse_coding} for an example application). The reconstruction and sparsity costs of the linear combination representing the data have to be minimized. \cite{tonolini2020variational}

Wang et al. \cite{wang2015deep} propose using sparse representations of image patches for super-resolution. They assume that a low-resolution image patch and its closely related high-resolution pendant share the same sparse code $\alpha_{lowres}=\alpha_{highres}=\alpha$ given properly defined reconstruction dictionaries $D_{lowres}$ and $D_{highres}$. They apply feedforward neural networks to compute approximate sparse codes.

Tonolini et al. \cite{tonolini2020variational} propose Variational Sparse Coding, which incorporates sparse coding at the inputs of a \ac{VAE} recognition model to improve feature disentanglement in the latent code. The model is evaluated on the Fashion MNIST, celebA (celebrity faces), and UCI HAR (accelerometer and gyroscope time-series data of human activities) data sets to investigate the disentanglement of features in the latent space and provides promising results and visuals.

\subsection{Recurrent Neural Networks}

\acfp{RNN} are a superset of feedforward neural networks which include recurrent edges that incorporate hidden states of previous, and in some cases subsequent time steps. This enables the model to process sequential data of arbitrary length one at a time while maintaining a \textit{memory} of the past. In the context of \ac{SDG}, \acp{RNN} are especially useful for speech synthesis, music generation, or time series prediction. \cite{lipton2015critical}

\begin{figure} [ht]
    \centering
    \begin{subfigure}[t]{0.265\textwidth}
        \centering
        \includegraphics[width=0.9\textwidth]{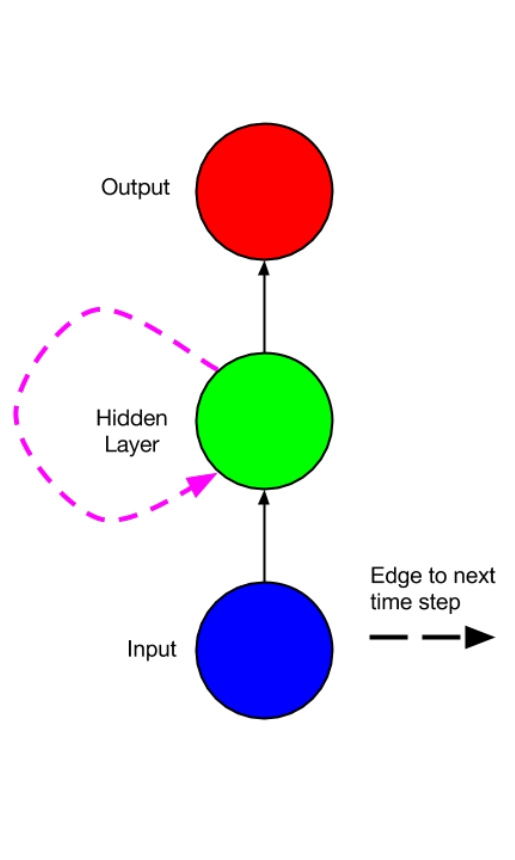}
    \end{subfigure}
    \begin{subfigure}[t]{0.6\textwidth}
        \centering
        \includegraphics[width=0.9\textwidth]{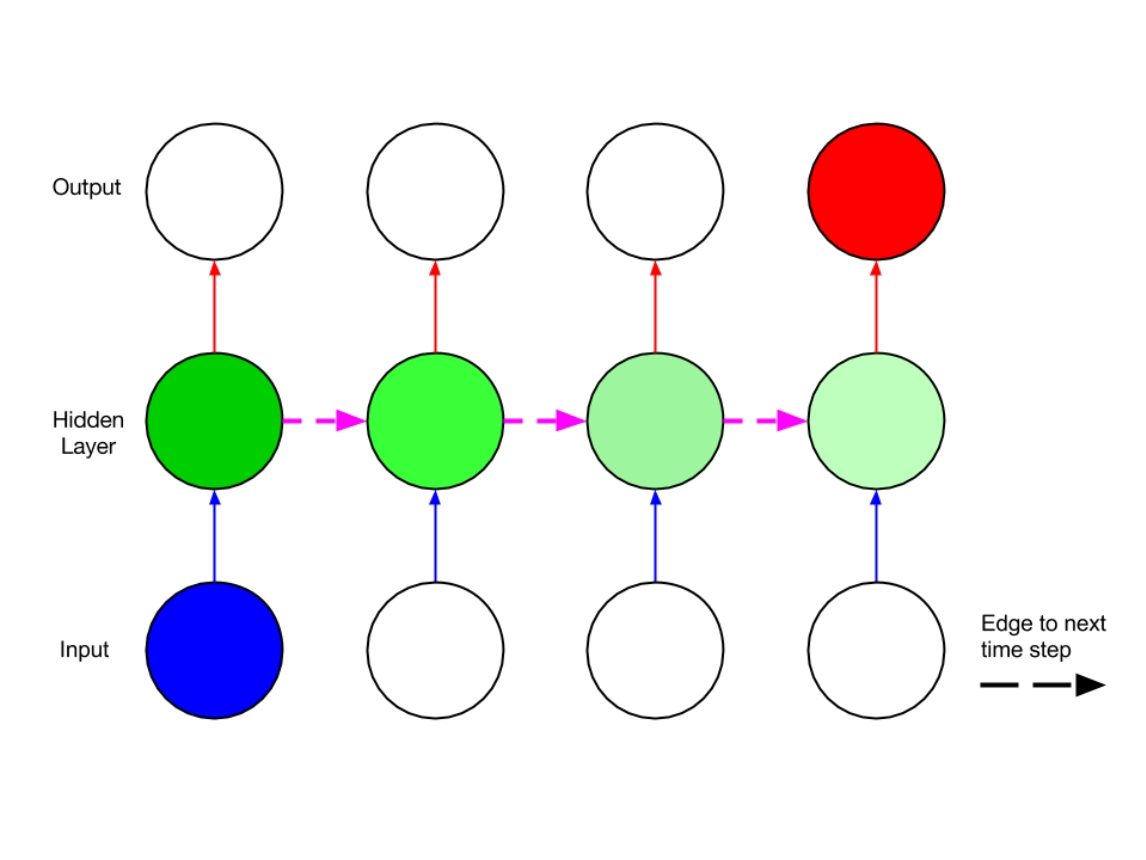}
    \end{subfigure}
    \caption{\textbf{Left:} A simple \ac{RNN} network. \textbf{Right:} Depiction of the vanishing gradient problem, where through weights less than one, the influence of the first input will diminish over time. (Source: \cite{lipton2015critical})}
    \label{fig:rnn_example}
\end{figure}

In a simple \ac{RNN} (see \autoref{fig:rnn_example}, left) at time step $t$, the current hidden state $\mathbf{h}^{(t)}$ depends on the current input example $\mathbf{x}^{(t)}$ and the previous state $\mathbf{h}^{(t-1)}$, resulting in 
\begin{equation}            \mathbf{h}^{(t)}=\sigma(W_{hx}\mathbf{x^{(t)}}+W_{hh}\mathbf{h}^{(t-1)}+\mathbf{b}_{h}) 
\end{equation}
with the sigmoid activation function $\sigma$ and trainable weight matrices $W_{hx}$, $W_{hh}$ and bias $\mathbf{b}_{h}$. The predicted output $\mathbf{\hat{y}}^{(t)}$ is then computed from $\mathbf{h}^{(t)}$, so 
\begin{equation}
\mathbf{\hat{y}}^{(t)}=\softmax(W_{yh}\mathbf{h}^{(t)}+\mathbf{b}_{y})
\end{equation}

with bias $\mathbf{b}_{y}$ and trainable weight $W_{yh}$. The weights are usually trained using \ac{BPTT}, which can suffer from the \textit{vanishing gradient problem} (see \autoref{fig:rnn_example}, right), which \acp{LSTM}, a special kind of \ac{RNN}, and \acp{GRU}, a simplified version of \acp{LSTM} \cite{yu2016video}, aim to solve. \cite{lipton2015critical}

Boulanger-Lewandowski et al. \cite{boulanger2012modeling} generalize \acp{RTRBM} and introduce the \textit{RNN-RBM}, which combines an \ac{RNN} with distinct hidden units with an \ac{RBM}, whose hidden units are related to the \ac{RNN}'s ones and visible units influence the next hidden state of the \ac{RNN}. The \ac{RBM}'s ability to generate complex distributions for each time step allows the authors to model and generate polyphonic music in a binary matrix piano-roll representation but not observe long-term musical structure.

Graves et al. \cite{graves2013generating} propose a deep \ac{RNN} with $N$ stacked recurrently connected \ac{LSTM} hidden layers that, at each step, compute the prediction probability for the next word. The network always starts with a null vector as the first input, so all data is generated without prior information. First, they evaluate their model on one-hot-encoded discrete text data and then on online handwriting data, a sequence of pen tip locations, to generate random handwritten character sequences. Next, they combine the handwriting approach with a target character sequence to generate handwriting for a given text. This works by providing a weighted window on the target text at each \ac{RNN} time step to the hidden layers. The approach is capable of producing realistic results.

Ranzato et al. \cite{ranzato2014video} introduce \textit{rCNN}, an unsupervised recurrent convolutional neural network, to predict the next frame of a video. The model splits the video frame into $8\times8$ pixel patches and feeds a $9\times9$ patch of their quantized values into a \ac{RNN} to create an embedding, which is then processed by two convolutional layers to predict the central patch. The patch-wise convolutional processing allows the rCNN to process videos of arbitrary frame size. The authors evaluate their model on the UCF-101 sport clip data set, resulting in better performance than n-grams and the neural net language model \cite{bengio2000neural}.

Vinyals et al. \cite{vinyals2015show} generate captions for images using an end-to-end encoder-decoder architecture, fully trainable with stochastic gradient descent. The Neural Image Caption (NIC) model encodes images using the last hidden layer of a \ac{CNN} pre-trained for image classification and feeding it to the \ac{LSTM} decoder, which aims to maximize the likelihood of the sentence being a correct transcription of the image. The model achieves state-of-the-art BLEU, METEOR, and CIDER scores for the time (2015) on the described image data sets Pascal VOC 2008, Flickr8k, Flickr30k, MSCOCO, and SBU.

Donahue et al. \cite{donahue2015long} propose the \ac{LRCN} for image interpretation tasks. Images are processed by a single \ac{CNN} to extract visual features, and a \ac{LSTM} encoder creates a total representation from these features. A \ac{LSTM} takes the image representation and the previous word as inputs and generates a description word by word. The authors extract entities and their relations from videos for video description, and the \ac{LSTM} decodes the entity collection into a meaningful sentence.

Srivastava et al. \cite{srivastava2015unsupervised} use \acp{LSTM} to learn representations of video sequences and decode them to predict future frames or reconstruct the input sequence. They also propose a composite model performing both tasks simultaneously to overcome their shortcomings. The training is conducted using backpropagation, and the models are evaluated on the UFC-101, HMDB-51, Sports-1M, and moving MNIST digits data sets. The future prediction results are quite blurry, but the representations work well for action recognition when fed into a classifier.

Mansimov et al. \cite{mansimov2015generating} present \textit{AlignDRAW}, which extends the DRAW model \cite{gregor2015draw} to generate images given a caption. The caption is defined as a sequence of words and is encoded using a bidirectional \ac{RNN}, which consists of a forward and backward \ac{LSTM} whose representations at each time step are concatenated. The generative \ac{RNN} works similarly to DRAW by including the respective caption representation at each time step and iteratively improving the quality of the existing image. The model also generates plausible results for previously unseen types of captions.

Jaques et al. \cite{jaques2016tuning} adopt \acp{LSTM} to music generation by training them to predict the next note on a large training corpus. They then use \ac{RL} to optimize their \textit{Note-RNN} by combining a reward function based on rules of musical theory with the output of another fixed copy Note-RNN. They achieve more coherent results that comply with musical theory, avoiding the sometimes occurring randomness of \acp{RNN}. They also discuss the application of \ac{RL} to other domains such as text generation, which could be used to enforce correct grammar.

\begin{figure} [ht]
    \centering
    \includegraphics[width=0.3\textwidth]{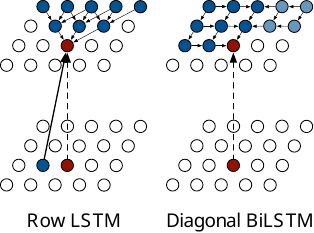}
    \caption{The row \ac{LSTM} (left) and diagonal BiLSTM (right) PixelRNNs. (Source: \cite{oord2016pixel})}
    \label{fig:pixelrnn}
\end{figure}

Van den Oord et al. \cite{oord2016pixel} advance two-dimensional \acp{RNN} to sequentially predict pixels (more specifically, their discrete RGB channel values) along the two spatial dimensions with the help of learned probability distributions and a softmax function. They propose two deep (up to 12 layers) \ac{LSTM} architectures (see \autoref{fig:pixelrnn}), also called \textit{PixelRNNs}: The row \ac{LSTM}, which processes the image row by row from top to bottom, and the diagonal BiLSTM, which uses two \acp{LSTM} starting at each top corner and crossing the image diagonally to capture more context. The diagonal BiLSTM outperforms the row \ac{LSTM} and other models like PixelCNN, DRAW, and \acp{DLGM} in image density estimation tasks.

Waite et al. \cite{waite2016generating} propose \textit{Lookback} and \textit{Attention} \ac{RNN}, which try to improve the long-term structure modeling capabilities of recurrent networks for music generation. The basic \ac{LSTM} takes the one-hot encoded vector of the previous melody event as input. Lookback \ac{RNN} takes two more previous event vectors, the current position in the music measure (e.g., $\frac{4}{4}$), and whether the last event repeats the event of the previous two events. The \ac{LSTM} now has to label new vectors as ``repeat-1-bar-ago'', ``repeat-2-bars-ago'', or a new melody event. The Attention \ac{RNN} uses an attention mechanism to look at the weighted sum of the previous $n$ outputs to generate the current step result. The model can be trained on MIDI files to create similar melodies.

Hadjeres et al. \cite{hadjeres2017interactive} introduce the \textit{Anticipation-RNN}, which enables the interactive generation of music by allowing user-defined positional constraints. Because the incorporation of future constraints in a sequential probabilistic model would be computationally expensive, a backward \textit{Constraint-RNN} going from constraint $N$ to $1$ is proposed, whose step-wise outputs are combined with the input state of the forward \textit{TokenRNN}, which generates the music tokens from $1$ to $N$. The method is general and can be applied to other \ac{RNN}-based approaches.

Oore et al. \cite{oore2020time} train a \ac{LSTM} on the Piano-e-Competition MIDI data set to generate natural-sounding piano performances in the MIDI event space with varying dynamics and velocity. The \ac{LSTM} inputs are one-hot encodings over the MIDI event vocabulary (various NOTE-ON, NOTE-OFF, TIME-SHIFT, and VELOCITY events), and the model computes a softmax probability distribution for the output event conditioned on the input events from which it samples. The model can generate music from an initial starting sequence or from scratch (an empty sequence) that comes close to human improvisation. It does not just copy the training samples but can not decide on a coherent play style. The authors prove that powerful generative models for music are possible without defining rules or heuristics at all and without observing long-term relationships in the data.

\othertab{\acp{RNN}}{
    \cite{bengio2012advances} & RNN-NADE: Combining \acp{RNN} with a \ac{NADE} \cite{larochelle2011neural} to output multivariate predictions (probabilities) for music data. & 2012 \\\hline 
    \cite{mikolov2012subword} & Experiments with subword-level language models, resulting in smaller models than character or word-based approaches with similar performance and no out-of-vocabulary predictions. & 2012 \\\hline 
    \cite{coca2013computer} & A reference melody and chaotic units are given to a \ac{LSTM} as input to generate new melodies with a predefined \textit{melodiouness}. & 2013 \\\hline 
    \cite{bayer2013fast} & Application of fast dropout to \acp{RNN}, which drops each incoming unit of a neuron with a certain probability, resulting in a respective zero value in the weighted sum for the neuron activation. Experiments on polyphonic music generation indicate that shallow \acp{RNN} perform better with dropout through better generalization. & 2013 \\\hline 
    \cite{pascanu2013construct} & Experiments with multiple different deep \acp{RNN} on language modeling and polyphonic music generation tasks. They implement deep state transition and output functions and improve over conventional shallow models, except ones with fast dropout \cite{bayer2013fast} or on character-wise generation, where the subword \ac{RNN} \cite{mikolov2012subword} performs better. & 2013 \\\hline 
    \cite{venugopalan2014translating} & Video description by extracting features from each frame with a \ac{CNN}, applying mean pooling across all frame embeddings and feeding the result to a \ac{LSTM} which generates the description. & 2014 \\\hline 
    \cite{zhang2014chinese} & Chinese poem generation from user-supplied keywords. Chinese symbols are one-hot encoded. A convolutional sentence model creates line embeddings fed to a recurrent context model that forwards a context vector to the recurrent generation model that creates word after word for the current line. & 2014 \\\hline 
    \cite{bayer2014learning} & Stochastic Recurrent Networks (STORNs): Two \acp{RNN}, the recognition (encoder) model $q(z_{t}\vert\mathbf{x}_{1:t-1})$ from which $\mathbf{z}_{t}$ is sampled, and the generating model $p(x_{t}\vert\mathbf{z}_{1:t})$, from which $\mathbf{x}_{t}$ is obtained, form a network easily trainable with stochastic gradient descent that can model multiple variables at each time step. Additionally, the latent representations $\mathbf{z}_{t}$ are conditioned on a prior $p(\mathbf{z})$ like a \ac{VAE}. The model is evaluated on polyphonic music generation and motion capture data continuation, where it outperforms previous \ac{RNN} approaches except RNN-NADE \cite{bengio2012advances}. & 2014 \\\hline 
    \cite{koutnik2014clockwork} & Clockwork RNN: Partition of hidden layers into separate modules $i$ with arbitrary periods $T_{i}$ that are active at time step $t$ only if $t\mod T_{i}=0$. This reduces the number of parameters and increases the performance on sequence reconstruction tasks. & 2014 \\\hline 
    \cite{liu2014bach} & \ac{LSTM} training with resilient propagation \cite{riedmiller1994rprop} instead of \ac{BPTT} for improved music composition represented as ``binary'' piano key presses. & 2014 \\\hline 
    \cite{goel2014polyphonic} & \ac{RNN}-\ac{DBN}: Very similar to the RNN-RBM \cite{boulanger2012modeling}, this model combines \acp{RNN} and \acp{DBN}, which are \acp{RBM} with multiple stacked hidden layers. It is also used for polyphonic music generation and improves upon the RNN-RBM results. & 2014 \\\hline 
    \cite{mao2014deep} & Multimodal \ac{RNN} (m-RNN): A \ac{RNN} that models the next word probability distribution based on the previous word and the encoding of an image provided by a \ac{CNN}. & 2014 \\\hline 
    \cite{theis2015generative} & Recurrent image density estimator (RIDE): Combining spatial \acp{LSTM} \cite{graves2008offline} that compute the hidden state $\mathbf{h}_{ij}$ of the next pixel $x_{ij}$ based on the two axis-wise preceding states $c_{i,j-1}$ and $c_{i-1,j}$, and mixtures of conditional Gaussian scale mixture \cite{theis2012mixtures} that predict the state of the next pixel $p(x_{ij}\vert\mathbf{h}_{ij})$. & 2015 \\\hline 
    \cite{berglund2015bidirectional} & Bidirectional \acp{RNN} as gap fillers in high-dimensional categorical and binary time series data (e.g., music), outperforming unidirectional \acp{RNN} and being applicable in more scenarios. & 2015 \\\hline 
    \cite{chung2015recurrent} & Variational \ac{RNN} (VRNN): A recurrent \ac{VAE} for high-dimensional sequence generation using the hidden state $h_{t}$ of a \ac{RNN} as the parameter for the distribution of the latent random variable $z$ of the \ac{VAE} at each step. The model outperforms simpler configurations in unconditional natural speech and handwriting generation. & 2015 \\\hline 
    \cite{bengio2015scheduled} & Scheduled sampling: To bridge the gap between training, where usually the ground-truth previous token $x_{t-1}$ is taken for the next prediction, and inference (generation of new sequences) distributions, where the model uses its previous prediction $\hat{x}_{t-1}$, it is randomly decided during training for each token prediction, whether $x_{t-1}$ or $\hat{x}_{t-1}$ is taken. Improved results on image captioning (\ac{CNN}encoder) and speech recognition compared to a baseline \ac{LSTM} without scheduled sampling. & 2015 \\\hline 
    \cite{chen2015mind} & Mind's Eye: Learning bi-directional mappings between visual features of images (obtained from a VGG \cite{simonyan2014very}) and their text descriptions with \acp{RNN}. The model can be used to generate in both directions by first training a \ac{RNN} to generate the text from the features and then a second \ac{RNN} on top to reconstruct the visual features from the text. & 2015 \\\hline 
    \cite{karpathy2015deep} & Alignment of visual (object detector \ac{CNN}) and language (bidirectional \ac{LSTM}) representations of image regions for full-frame and region-level image description with \ac{RNN} decoding. & 2015 \\\hline 
    \cite{vohra2015modeling} & DBN-LSTM: Improved version of the RNN-DBN \cite{goel2014polyphonic}, where the \ac{RNN} is replaced with a \ac{LSTM} for better performance in polyphonic music generation. & 2015 \\\hline 
    \cite{lyu2015modelling} & \ac{LSTM}-\ac{RTRBM}: Replacement of some hidden units of a \ac{RTRBM} with \ac{LSTM} ones, which increases performance and learning speed. & 2015 \\\hline 
    \cite{xu2015show} & Image captioning with convolutional feature extraction and an attention-based \ac{LSTM} that generates the caption word-by-word. & 2015 \\\hline 
    \cite{venugopalan2015sequence} & Video description with \ac{CNN}-encoded video frames as input for a two-layer \ac{LSTM} that starts outputting words after the whole video sequence has been processed by the first \ac{LSTM} layer. & 2015 \\\hline 
    \cite{yao2015describing} & Video description with a 3D \ac{CNN} encoder (width $\times$ height $\times$ timesteps of video) and \ac{LSTM} decoder with temporal attention mechanism. & 2015 \\\hline 
    \cite{fraccaro2016sequential} & A stochastic recurrent neural network (SRNN) with separate stochastic and deterministic layers propagates uncertainty in latent space through the network. The hidden representation $\mathbf{z}_{t}$ depends on a prior $p(\mathbf{z_{z}\vert\mathbf{z}_{t-1}})$, similar to a \ac{VAE}, parameterized by a neural network. Achieves state-of-the-art performance on speech and polyphonic music modeling. & 2016 \\\hline 
    \cite{colombo2016algorithmic} & Single-note melody generation and continuation with a multi-layer \ac{GRU} trained on sequences of corresponding pitch and duration one-hot vectors. & 2016 \\\hline 
    \cite{sun2016composing} & Improved \ac{LSTM} training for music composition. Pitches and durations of notes are encoded together in one one-hot vector. Training is split into two parts, where first, the model is trained with real data, and new compositions are generated. These creations are filtered by the \textit{grammar argumented method}, which only allows samples complying with defined musical rules to remain. These are then appended to the training data, and the model is retrained, resulting in the actual generative model. & 2016 \\\hline 
    \cite{bahdanau2016actor} & Training of a sequence prediction \ac{RNN} \textit{actor} with a second \textit{critic} network that has access to the ground-truth data and computes the expected task-specific score. The \ac{RL}-inspired actor-critic training approach fits training data faster than maximum-likelihood learning. It provides more coherent and accurate results on text prediction tasks (i.e., spelling correction and machine translation). & 2016 \\\hline 
    \cite{yang2016review} & Review Network: An encoder-decoder framework with reviewers in between that provide a discriminative loss. Additionally, an attention mechanism between the encoder and the review networks and the review networks and the decoder is implemented. The encoder can be a \ac{CNN} or \ac{RNN}, while the decoder is a \ac{LSTM}. The model is used to caption images and generate comments for Java source code. & 2016 \\\hline 
    \cite{you2016image} & Image captioning based on \ac{CNN} encoder, visual attribute prediction (i.e., what happens in the picture), a \ac{LSTM} for state progression, and two different attention mechanisms for input and output that determine the next word from the vocabulary. & 2016 \\\hline 
    \cite{yu2016video} & Video description in one or multiple sentences using a hierarchical \acp{RNN} with a sentence and paragraph generator. The network is based on \acp{GRU} and uses temporal and spatial attention mechanisms. Input features of the video are obtained from a pre-trained convolutional extractor \cite{simonyan2014very}. & 2016 \\\hline 
    \cite{choi2016text} & Two text-based \acp{LSTM} (word-RNN and char-RNN) learn chord progressions from text representations of music for fully automatic music composition. & 2016 \\\hline 
    \cite{chu2016song} & Multi-track pop music generation with a hierarchical \ac{LSTM} where each layer is responsible for predicting a certain aspect of the song at a time step: The bottom layer predicts the pressed key, the next layer the duration of the press, the third the chord and the fourth and last layer the drum beat. & 2016 \\\hline 
    \cite{sturm2016music} & Folk-RNN: Large-scale generation of Celtic folk music transcriptions with \acp{LSTM} trained with a vocabulary of tokens on single transcriptions. Evaluation is performed on the large-scale distributions of real and generated data and the single transcription level. The \acp{LSTM} automatically learn to conform to structural constraints of folk music. & 2016 \\\hline 
    \cite{liang2016bachbot} & BachBot: A three-layer stacked \ac{LSTM} with optimized parameters generates Bach chorales or transfers their style to other melodies. It is trained with \textit{teacher forcing} (always continue prediction with the correct previous token) on frame-based representations of the polyphonic piano roll data. & 2016 \\\hline 
    \cite{huang2016deep} & Two-layer \ac{LSTM} for token-level music generation from a short seed sequence. Each MIDI message or existing note combination from a piano roll representation is treated as a separate token. The results are comparable to RNN-NADE \cite{bengio2012advances}. & 2016 \\\hline 
    \cite{mehri2016samplernn} & SampleRNN: An unsupervised and unconditional end-to-end model for raw audio waveform synthesis with hierarchical \acp{RNN} that cover different temporal ranges. & 2016 \\\hline 
    \cite{serban2016building} & Application of a (bidirectional) hierarchical recurrent encoder-decoder (HRED) to dialogue generation by utilizing question-answer pairs and pre-trained word embeddings. The model consists of a \ac{RNN} encoder, whose final representation is fed into a context \ac{RNN}, which maps the representation into the dialogue context. The context state is then fed to each step of the decoder \ac{RNN}. & 2016 \\\hline 
    \cite{lamb2016professor} & The ``Professor Forcing'' algorithm is designed to align the behaviors of recurrent neural networks during training and sampling phases, addressing a common issue in \ac{RNN} training. This method applies adversarial domain adaptation, where the network is trained to make its behavior indistinguishable between these two phases under the scrutiny of a discriminator. This approach helps in producing more coherent and structured outputs. The algorithm serves as a regularization technique, enhancing the network's ability to generalize from its training data, leading to improvements in performance on various tasks including language modeling and image synthesis. & 2016 \\\hline
    \cite{sotelo2017char2wav} & Char2Wav: End-to-end speech synthesis from text. A bidirectional \ac{RNN} encodes text, and a \ac{RNN} with attention decodes the representation at different time steps to produce features for a vocoder. The vocoder is a conditional SampleRNN \cite{mehri2016samplernn} that produces raw waveform output from the vocoder features. & 2017 \\\hline 
    \cite{zilly2017recurrent} & Recurrent Highway Network: \acp{LSTM} with multiple highway layers \cite{srivastava2015training} that allow deep step-to-step transition functions that are easily trainable. The model outperforms previous \acp{RNN} in character prediction tasks on Wikipedia texts. & 2017 \\\hline 
    \cite{colombo2017deep} & Deep Artificial Composer (DAC): Extension of \cite{colombo2016algorithmic} that uses \acp{LSTM} (duration and pitch \ac{RNN}) to generate note transitions and is trained on a corpus with two different musical styles. It also emphasizes a new novelty measure (fraction of transitions found in a defined corpus) that is used to improve the creativity of the model. & 2017 \\\hline 
    \cite{liu2017attention} & Improvement of image captioning with attention \cite{xu2015show} by supervised training of the attention mechanism with ground truth text entity-image region mappings obtained from humans. & 2017 \\\hline 
    \cite{gan2017semantic} & Semantic Composition Network (SCN): Image captioning in multiple parts: A \ac{CNN} extracts a feature vector, and a \ac{MLP} computes probabilities of tags based on the most used words in the training descriptions. A \ac{LSTM} then generates the description based on the feature vector and the tag probabilities. & 2017 \\\hline 
    \cite{hadjeres2017deepbach} & DeepBach: Generation or reharmonization of Bach chorales/MIDI data using deep \acp{LSTM} working in opposite directions and neural networks that merge the \ac{RNN} results. Sampling from this dependency network is performed using pseudo-Gibbs sampling. & 2017 \\\hline 
    \cite{makris2017combining} & Conditional drum rhythm generation with a combination of a two-layer stacked \ac{LSTM} that learns drum sequences and a feedforward fully connected layer that processes metrical rhythm and a bass sequence as constraints. Their predictions are merged to predict the next drum event. & 2017 \\\hline 
    \cite{jaques2017sequence} & Sequence Tutor: An improved fine-tuning approach for \acp{RNN} that first trains an \ac{RNN} on data with maximum-likelihood estimation and uses its output as a policy (originally proposed in \cite{jaques2016tuning}) for a second \ac{RNN} trained with \ac{RL} for a specific domain. The effectiveness is demonstrated on music melody and molecule generation (i.e., SMILES \cite{weininger1988smiles} strings). & 2017 \\\hline 
    \cite{lim2017chord} & Chord progression generation from monophonic melodies with Bi\acp{LSTM}. & 2017 \\\hline 
    \cite{serban2017hierarchical} & Latent variable hierarchical recurrent encoder-decoder (VHRED): A stack of encoder, context, and decoder \acp{RNN} combined with a stochastic latent variable conditioned on all previously observed tokens that can capture the dependencies of sub-sequences of sequential data. The stochastic variable allows for diverse and coherent dialogues to be generated. & 2017 \\\hline 
    \cite{hutchings2017using} & Harmonic Improviser: A \ac{LSTM} harmony agent trained on Jazz chord progressions and a rule-based melody agent manipulating provided melodies take turns improvising music in real-time and are rewarded for harmonic consistency and melodic flow. & 2017 \\\hline 
    \cite{simon2017performance-rnn} & Performance RNN: MIDI music generation with a \ac{LSTM} showing timing and dynamics, but lacking long-term coherence. & 2017 \\\hline 
    \cite{johnson2017generating} & TP-LSTM-NADE \& BALSTM (biaxial \acp{LSTM}): Modification of an RNN-NADE \cite{bengio2012advances} to model relative differences between nodes for transposition-invariant polyphonic music generation. & 2017 \\\hline 
    \cite{manzelli2018conditioning} & Combination of a biaxial \ac{LSTM} \cite{johnson2017generating} for symbolic music generation and a conditional WaveNet-based audio generator for waveform music generation. & 2018 \\\hline 
    \cite{li2018learning} & \acfp{GNN}: Computing and iteratively updating node and graph embeddings using fully connected neural networks, \acp{GRU} for nodes and \acp{LSTM} for edge modifications. \acp{MLP} are used to sequentially compute probabilities of adding new nodes (and their type) and edges based on these representations. The model can work conditionally and unconditionally and is demonstrated on molecule generation. & 2018 \\\hline 
    \cite{kalchbrenner2018efficient} & WaveRNN: A sparse single-layer \ac{RNN} mainly for text-to-speech synthesis where the majority of weights are pruned and subscaling is employed to fold long sequences into a batch (matrix) of short ones, which allows generating multiple samples (i.e., over multiple rows) at one step. This allows the model to produce results in real time on a mobile CPU. & 2018 \\\hline 
    \cite{santoro2018relational} & Relational \ac{RNN}: A \ac{LSTM} with a multi-slot memory matrix instead of the hidden state vector. Input is concatenated to the matrix as a new row, and multi-head dot product attention \cite{vaswani2017attention} is applied to generate the next memory state. The model achieves state-of-the-art results on language modeling tasks (next-word probability). & 2018 \\\hline 
    \cite{mao2018deepj} & DeepJ: A model built upon the biaxial \ac{LSTM} \cite{johnson2017generating} structure and incorporate dynamics (i.e., relative note volume) in note embeddings and global style and context (e.g., genre) conditioning in the network. They train three outputs simultaneously (play and replay probability and dynamics) for the predicted notes. & 2018 \\\hline 
    \cite{you2018graphrnn} & GraphRNN: Generate large variable-length graphs without node ordering by treating the problem as a sequence of node and edge additions. At each step $i$, the graph-level \ac{GRU} updates the graph state $h_{i}$ and adds a new node. The edge-level \ac{RNN} then creates the adjacency vector for the new node to all old nodes or the end-of-sequence token from $h_{i}$. The model is trained on data obtained using breadth-first-search through any graph permutation with a random starting point. The graphs are evaluated by computing a \acf{MMD} score between the degree and clustering coefficient distributions and orbit count statistics between sets of graphs. & 2018 \\\hline 
    \cite{shen2018natural} & Tacotron 2: A recurrent sequence-to-sequence text-to-speech network consisting of multiple \acp{LSTM} and \acp{CNN} that maps character embeddings to simplified \textit{mel} spectrograms that are then converted to waveform audio with WaveNet \cite{oord2016wavenet}. & 2018 \\\hline 
    \cite{liao2019efficient} & Graph Recurrent Attention Network (GRAN): Improvement upon GraphRNN \cite{you2018graphrnn} using a \ac{GNN} \cite{li2018learning} with attention at each step to generate a block of new nodes and edges based on the already existing graph. Further, they propose training on adjacency matrices conforming to families of node orderings (e.g., nodes sorted by node degree, breadth/depth-first-search ordering from largest degree node, original data order) to improve model understanding. & 2019 \\\hline 
    \cite{bacciu2019graph} & Generating undirected, fully connected graphs without self-loops as an ordered edge sequence using \acp{GRU}. The first \ac{GRU} predicts the sequence of the first nodes of the edge pairs. The second \ac{GRU} outputs the probabilities for the second node of the edge. The graph nodes are assigned a fixed order in advance and the \acp{RNN} are trained to maximize the node probabilities observed in the training data. This simple model performs similarly to the GraphRNN \cite{you2018graphrnn}. & 2019 \\\hline 
    \cite{popova2019molecularrnn} & MolecularRNN: Extension of GraphRNN \cite{you2018graphrnn} that uses a \textit{NodeRNN} to compute the next atom type and an \textit{EdgeRNN} to compute the bond types to the previous atoms. The model enforces valid per-atom valency. It is first trained to reconstruct the training data and then fine-tuned using a \ac{RL} critic that rewards molecules with certain properties. & 2019 \\\hline 
    \cite{khodayar2019deep} & Deep Graph Distribution Learning (DeepGDL): Decomposing a graph into densely connected components with sparse connections between these communities. A \ac{GRU} is then trained to learn the distribution of nodes and edges in these communities based on earlier node and edge observations. Synthetic communities are sampled from these predicted distributions and probabilistically connected to produce new large graphs (synthetic power grids) with similar properties to the real data. & 2019 \\\hline 
    \cite{hernandez2020novel} & Generation of biomedical signals (electrocardiogram, etc.) for patients or specific events using bidirectional \acp{RNN} that are trained with real patient data. In the first optional stage, noise is injected into the real signals, and then the signal is segmented according to certain events or classes. Finally, the BiRNN generates new similar data based on the input, and a statistical stage evaluates the data quality. & 2020 \\\hline 
    \cite{goyal2020graphgen} & GraphGen: A domain-agnostic and scalable labeled graph generation method using a \ac{LSTM} to sequentially append tuples containing the sequence and types of nodes and edges in a depth-first search order. Other models like GraphRNN \cite{you2018graphrnn} are outperformed in almost every evaluated criterion. & 2020 \\\hline 
    \cite{privato2022creative} & Scramble: Music generation with pitch transitions generated by a Markov chain and a \ac{LSTM} that imposes a learned style (velocity, rhythm, and beats per minute) on the pitch sequence, which the user can also tweak. & 2022 \\ 
}

\subsection{Convolutional Neural Networks}

\begin{figure} [ht]
    \centering
    \begin{subfigure}[t]{\textwidth}
        \includegraphics[width=\textwidth]{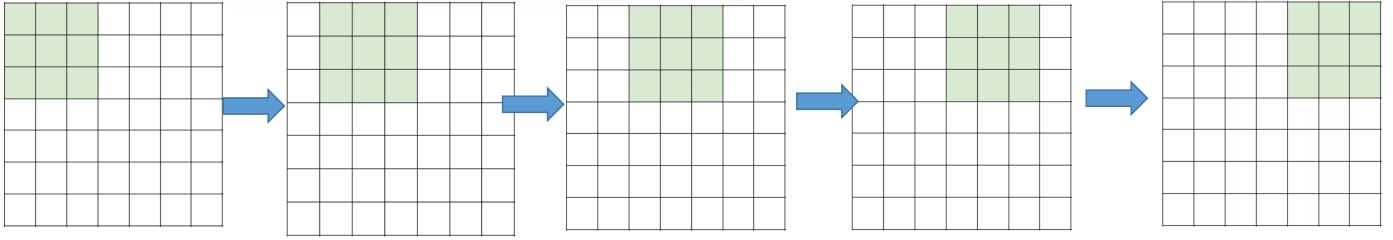}
        \caption{Application of kernels or pooling functions to an input matrix with stride $1$. (Source: \cite{albawi2017understanding})}
        \label{fig:cnn_stride}
    \end{subfigure}

    \vspace{2em}

    \begin{subfigure}[t]{\textwidth}
        \includegraphics[width=\textwidth]{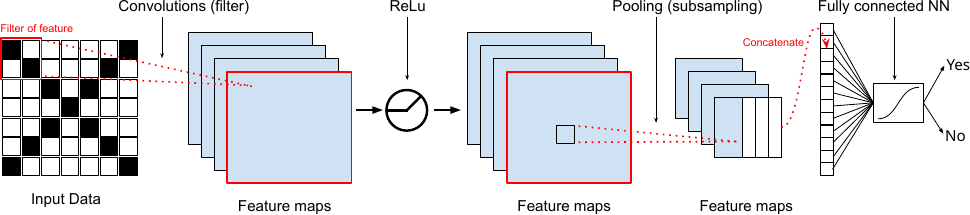}
        \caption{Example \ac{CNN} architecture for an image classification task. (Source: \cite{eigenschink2021deep})}
        \label{fig:cnn_architecture}
    \end{subfigure}
    \caption{Illustrations of the structure of a \ac{CNN}.}
    \label{fig:cnn_example}
\end{figure}

\acfp{CNN} are artificial neural networks based on matrix operations that can handle data of large sizes like images or speech with significantly fewer parameters to train than a fully connected neural network. The model can contain different types of layers \cite{albawi2017understanding}:

\begin{description}
    \item[Convolution] A \textit{kernel} or \textit{filter}, which is a lower-size matrix consisting of trainable weights, is applied to parts of the input matrix to extract local features, for example, to detect edges in an image. The kernel is moved over the input left-to-right, and top-to-bottom by a \textit{stride}, which defines the number of units shifted at each step, to compute the output value at this kernel position through matrix multiplication of the weights with the respective values in the input at the kernel's position (see \autoref{fig:cnn_stride}).
    \item[Padding] To prevent loss of information at the border of the input or to preserve the input size, a zero-padding can be added around the matrix.
    \item[Non-linearity] After a convolution layer, a non-linear function is used to modify or cut off the output. Typically, the \ac{ReLU} function $ReLU(x)=max(0,x)$ is used.
    \item[Pooling] A pooling layer downsizes its input to reduce complexity for later model layers. Popular implementations are max-pooling or average-pooling, which return a region's maximum value or average, respectively. They are applied in the same manner as a kernel.
    \item[Fully connected layer] Finally, a computationally expensive fully connected layer is applied to the significantly smaller input to, for example, classify an image or perform another task. An illustration of such a \ac{CNN} architecture is provided in \autoref{fig:cnn_architecture}.
\end{description}

Lotter et al. \cite{lotter2015unsupervised} predict future frames of image sequences by first learning a representation of each single input image with a \ac{CNN}. Then, a \ac{LSTM} processes the images sequentially in order before the final output is forwarded to a deconvolutional (i.e., reversed) \ac{CNN}, which produces an image from the \ac{RNN} prediction. They train the model using mean squared error but also experiment with adversarial loss \cite{goodfellow2014generative} realized by a similar \ac{CNN}-\ac{LSTM} discriminator whose final prediction together with an encoding of the generator's result image or the real next frame is passed to a \ac{MLP}. The model can predict movements and rotations in videos well, which supports the idea that prediction may enable the development of transformation-tolerant object representations.

Bruna et al. \cite{bruna2015super} propose \acp{CNN} for high-dimensional structured prediction problems such as image super-resolution. They model the conditional distribution $p(y\vert x)$ as a Gibbs density $p(y\vert x)\varpropto\exp(-\Vert\Phi(x)-\Psi(y)\Vert^{2})$, where $\Phi:\mathbb{R}^{N}\rightarrow\mathbb{R}^{P}$ and $\Psi:\mathbb{R}^{M}\rightarrow\mathbb{R}^{P}$ are highly-informative non-linear mappings (sufficient statistics) obtained from deep \acp{CNN} that ``minimize the uncertainty of $y$ given $x$''. The model can provide solutions with spatial coherence. Still, the inference is computationally costly compared to \acp{GAN}, and a trade-off between sharpness and stability must always be made.

\begin{figure} [ht]
    \centering
    \includegraphics[width=0.4\textwidth]{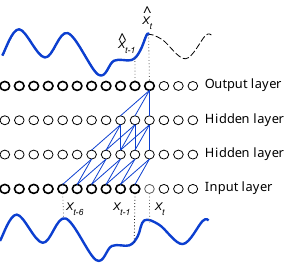}
    \caption{Auto-regressive 1D signal modeling with a masked \ac{CNN}. (Source: \cite{reed2016generating})}
    \label{fig:masked_cnn}
\end{figure}

Van den Oord et al. \cite{oord2016pixel} propose the \textit{PixelCNN}, which consists of multiple resolution-preserving convolutional layers with masks (see \autoref{fig:masked_cnn}) to ignore future pixels. Like the PixelRNNs, a softmax function is used to compute the discrete RGB values of pixels sequentially for image generation and completion tasks. The advantage of the PixelCNN over the PixelRNN is the possibility of parallelization during training and evaluation of test images, but the performance is worse. The model has been successfully adapted to video continuation as a decoder \cite{kalchbrenner2017video}.

\begin{figure} [ht]
    \centering
    \begin{subfigure}[t]{.49\textwidth}
        \includegraphics[width=\textwidth]{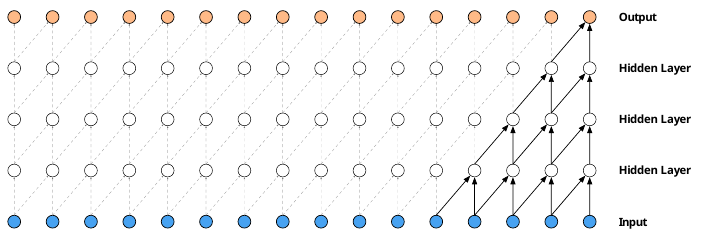}
        \caption{Causal convolution: A masked convolution where only previous timesteps are considered.}
    \end{subfigure}
    \hfill
    \begin{subfigure}[t]{.49\textwidth}
        \includegraphics[width=\textwidth]{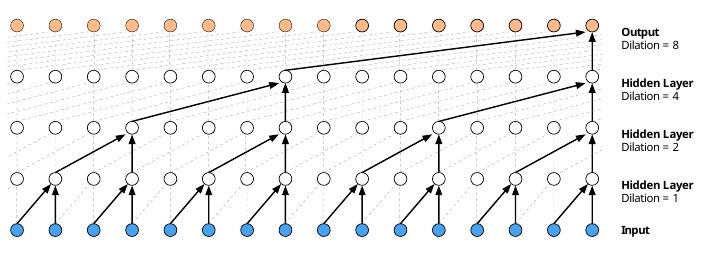}
        \caption{Dilated causal convolution: Only timesteps with a certain step between them are considered.}
    \end{subfigure}
    \caption{Illustrations of a causal and dilated convolution. (Source: \cite{oord2016wavenet})}
    \label{fig:wavenet_causal_dilated_convolution}
\end{figure}

Van den Oord et al. \cite{oord2016wavenet} introduce \textit{WaveNet}, which operates on the raw audio waveform and is similar to PixelCNN \cite{oord2016pixel} in that it models the conditional probability distribution $p(x_{t}\vert x_{1},...,x_{t-1})$ with a stack of convolutional layers. It uses dilated causal convolutions (see \autoref{fig:wavenet_causal_dilated_convolution}) to preserve ordering and process an area larger than the convolution length by skipping values by a step. The model can be trained in parallel but generates new audio sequentially. The model can also easily be transformed to incorporate additional input (e.g., text and speaker identity for text-to-speech) by appending it to the conditional distribution, resulting in a conditional WaveNet. WaveNet achieves state-of-the-art results in text-to-speech tasks and allows conditioning on different speakers. Further, the model can generate novel and realistic musical fragments.

Gatys et al. \cite{gatys2016image} propose to use a deep \ac{CNN} \cite{simonyan2014very} to separate style (texture) and content (object recognition) representations of an image for style transfer. The style transfer works by extracting multiple layers of style and content representations from style and content reference images, respectively. Then, a white noise image $\vec{x}$ is initialized and iteratively optimized using gradient descent with respect to the pixel values based on the combined style and content losses, which are the sum of squared errors between the respective representations (style image and $\vec{x}$, $\vec{x}$ and content image).

Kim et al. \cite{kim2016deeply} use a recursive \ac{CNN} for image super-resolution. First, an embedding network, similar to a \ac{MLP}, encodes an image as a set of feature maps. Then the recursive \ac{CNN} applies the same convolution followed by a \ac{ReLU} to the embedding to increase the observed range before a reconstruction network, also similar to a \ac{MLP}, creates the final output. To solve the problems of vanishing/exploding gradients and finding the optimal amount of recursions during training, each recursive layer is reconstructed, and a weighted average of all predictions produces the final output (\textit{recursive supervision}). A skip connection between the input image and the reconstruction network is established to improve results further. The method outperforms SRCNN \cite{dong2014learning,dong2015image} and provides clearer images.

Salimans et al. \cite{salimans2017pixelcnn} accelerate PixelCNN training and generate state-of-the-art results on class-conditional and unconditional image generation task CIFAR-10 with their improved implementation called \textit{PixelCNN++}. They compute the RGB values of pixels assuming linear dependence and using continuous distributions that are rounded instead of 256-way softmax to reduce training cost. Further, they introduce downsampling and shortcut connections between layers to better capture the input structure and use dropout regularization to prevent overfitting.

\othertab{\acp{CNN}}{
    \cite{dong2014learning} & SRCNN: A deep \ac{CNN} learns an end-to-end mapping from low to high-resolution images for super-resolution. The first layer extracts feature maps from overlapping patches, the second maps these feature maps to higher-resolution maps, and the third combines the predictions in a neighboring area for the final result. & 2014 \\\hline      \cite{mordvintsev2015deepdream,mordvintsev2015inceptionism} & DeepDream: An image classification \ac{CNN} that is reversed to (iteratively) amplify high-level features in random noise or real images, often resulting in dream-like over-interpretations. This approach allows a user to visually inspect what a model has learned about objects or concepts and can be used to create art-like images. & 2015 \\\hline 
    \cite{dong2015image} & Improvement of SRCNN \cite{dong2014learning} for simultaneous 3-channel color handling. Further, different model architectures (larger filters and more layers) and parameters are explored. & 2015 \\\hline 
    \cite{oord2016conditional} & Gated PixelCNN: Improved image quality over the original PixelCNN by utilizing gated convolutional layers, matching the PixelRNN. The model is suitable for class-conditional image generation and as a powerful decoder for an autoencoder. & 2016 \\\hline 
    \cite{reed2016generating} & Proposal of the gated conditional PixelCNN with text (\ac{GRU} encoder), segmentation map, and keypoint (i.e., annotated locations of certain human/bird body parts in the image) conditioning for image generation. & 2016 \\\hline 
    \cite{johnson2016perceptual} & Image super-resolution and style transfer like \cite{gatys2016image}, but three orders of magnitude faster with qualitatively similar results by using a perceptual loss obtained from a pre-trained VGG network \cite{simonyan2014very} instead of a per-pixel loss. & 2016 \\\hline 
    \cite{shi2016real} & ESPCN: Image and video super-resolution with sub-pixel \ac{CNN} learning upscaling filters from low-resolution feature maps into high-resolution output. & 2016 \\\hline 
    \cite{ping2017deep} & Deep Voice 3: A fully-convolutional encoder-decoder model with position-augmented attention mechanism for text-to-speech synthesis. The model can produce different parameters for various waveform synthesis models (e.g., WaveNet \cite{oord2016wavenet}) and incorporate a speaker representation to capture different speech styles. The model achieves state-of-the-art quality in human mean opinion score evaluations. & 2017 \\\hline 
    \cite{chen2017pixelsnail} & PixelSNAIL: Application of SNAIL \cite{mishra2017simple}, a general purpose autoregressive meta-learning model using causal convolutions and self-attention to maximize its context size, to sequential image generation, resulting in state-of-the-art likelihood, but slow density estimation performance. & 2017 \\\hline 
    \cite{menick2018generating} & Subscale Pixel Network (SPN): Images of size $N\times N$ are split into slices of size $\frac{N}{S}\times\frac{N}{S}$ that are interleaved. The network consists of a convolutional encoder that embeds previously processed slices and a convolutional decoder with masked convolution and self-attention that predicts the next slice given the embedding. The model is especially suitable for upscaling and generates coherent and exact samples. & 2018 \\\hline 
    \cite{wang2019end} & Joint training of an ensemble of shallow and deep \acp{CNN} to generate super-resolution images end-to-end. Optimization during training is alleviated by letting the shallow \ac{CNN} restore the main structure of the image and the deep \ac{CNN} fill in the details. The deep \ac{CNN} extracts features, upscales them to the target factor, and uses multi-scale reconstruction to capture the context better and produce the output pixels. & 2019 \\ 
}

\subsection{Transformers}\label{sec:transformers}

\begin{figure}[htb!]
    \centering
    \begin{minipage}{0.5\linewidth}
        \begin{subfigure}{\linewidth}
            \centering
            \includegraphics[width=0.18\linewidth]{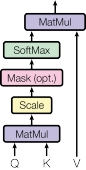}
            \caption{The scaled dot-product attention maps weights to values of key-value pairs $(K,V)$ according to the correspondence between a query $Q$ and the key.}
            \label{fig:transformer_attention}
        \end{subfigure}
        \begin{subfigure}{\textwidth}
            \centering
            \includegraphics[width=0.33\linewidth]{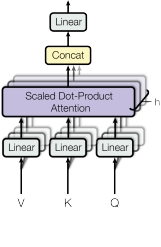}
            \caption{Multi-head attention uses multiple attention layers and concatenates the results.}
            \label{fig:transformer_attention_multi}
        \end{subfigure}
    \end{minipage}
    \hfil
    \begin{minipage}{0.45\linewidth}
        \begin{subfigure}{\linewidth}
            \centering
            \includegraphics[width=0.75\linewidth]{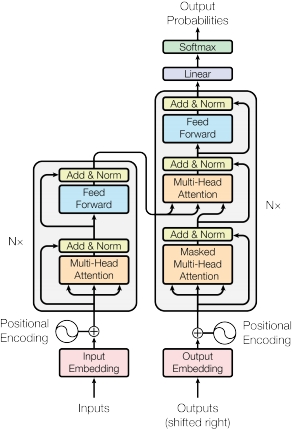}
            \caption{Encoder-decoder architecture of the transformer. Both parts consist of $N=6$ layers.}
            \label{fig:transformer}
        \end{subfigure}
    \end{minipage}
    \caption{Structure of the transformer itself and essential parts. (Source: \cite{vaswani2017attention})}
\end{figure}

Transformers (see \autoref{fig:transformer}) are sequence-to-sequence transduction models with an encoder-decoder structure. They advance previous recurrent and convolutional encoder-decoder architectures because they allow for more parallelization and modeling of dependencies with arbitrary distance in the input and output sequences with a constant number of sequential operations. For that, the transformer utilizes a multi-headed \textit{self-attention} mechanism (see \autoref{fig:transformer_attention} and \autoref{fig:transformer_attention_multi}) instead of recurrent layers. Originally, the transformer was used for language translation tasks, where the original sentence was first encoded, and the decoder used attention to the encoder embeddings and the already generated output in the target language to compute the probabilities of the next token. \cite{vaswani2017attention}

Liu et al. \cite{liu2018generating} generate English Wikipedia articles by providing the target article title and summarizing multiple non-Wikipedia reference documents. The training data consists of Wikipedia articles, their citations, and the top 10 web search results for each article section title. The paragraphs of all reference documents are ranked by importance according to the target article title. Then, the first $L$ tokens of these ranked paragraphs are used as the input for the generative model. The generative model is a transformer without an encoder that concatenates the input sequence and the desired output sequence into a single vector for the decoder input and is trained to predict the next token based on the previous ones. The authors also modify the attention mechanism by splitting the tokens into blocks on which attention is independently applied and performing convolution on the key-value pairs to reduce memory requirements on long input sequences. The results show that the model can split articles into reasonable sections and fill in factual information from many different references.

\begin{figure} [ht]
    \centering
    \includegraphics[width=0.5\textwidth]{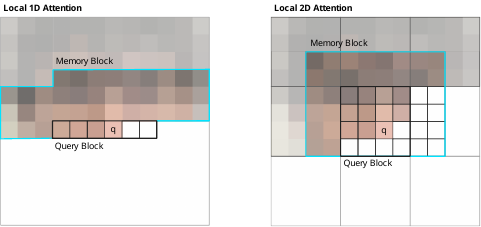}
    \caption{Illustration of the 1D and 2D local attention mechanisms of the \textit{Image Transformer}. The 2D attention performs slightly better in terms of perceptual image quality evaluated by humans. (Source: \cite{parmar2018image})}
    \label{fig:image_transformer}
\end{figure}

Parmar et al. \cite{parmar2018image} propose the \textit{Image Transformer} conditioned on a few class embeddings (decoder only) or low-resolution pictures (encoder-decoder architecture) to generate high-resolution images. The generation process is formulated as a sequence modeling problem, where the RGB channel values of the next pixel are predicted based on the other pixels' values in the local neighborhood to allow for larger image sizes. The authors propose two different attention mechanisms, 1D and 2D local attention (see \autoref{fig:image_transformer}). Experiments on CIFAR-10, ImageNet, and celebA data sets show that the transformer architecture outperforms previous state-of-the-art architectures like \acp{RNN}, \acp{CNN}, and \acp{GAN} in terms of processible image size and quality.

Huang et al. \cite{huang2018music} use a transformer to generate symbolic music represented as a sequence of discrete tokens. Since pieces often repeat and modify previous motifs or sections, relations between such sections are explicitly modeled. Therefore, the authors adopt a relation-aware self-attention \cite{shaw2018self} that creates relative position embeddings and optimizes the memory consumption to allow long sequences to be autoregressively modeled. They achieve the best \ac{NLL} scores and most win in a human ``musicality'' comparison test when compared against the PerformanceRNN and LookBackRNN \cite{waite2016generating} \ac{LSTM} models and a baseline transformer when continuing a music sequence they were initialized on.

Child et al. \cite{child2019generating} introduce the \textit{sparse transformer}, which uses sparse factorizations of the attention matrix to reduce the time and memory requirements from $O(n^{2})$ to $O(n\sqrt{n})$ for the sequence length $n$ without performance loss. This works by splitting the attention operation into multiple faster operations that only access a subset of all previous positions and combining their results to approximate the full attention. The model can generate large unconditional sequence samples in various domains such as natural images (CIFAR-10, ImageNet64) and raw audio data of classical music and achieves state-of-the-art results in density modeling tasks compared to contemporary models. 

\begin{figure} [ht]
    \centering
    \includegraphics[width=0.9\textwidth]{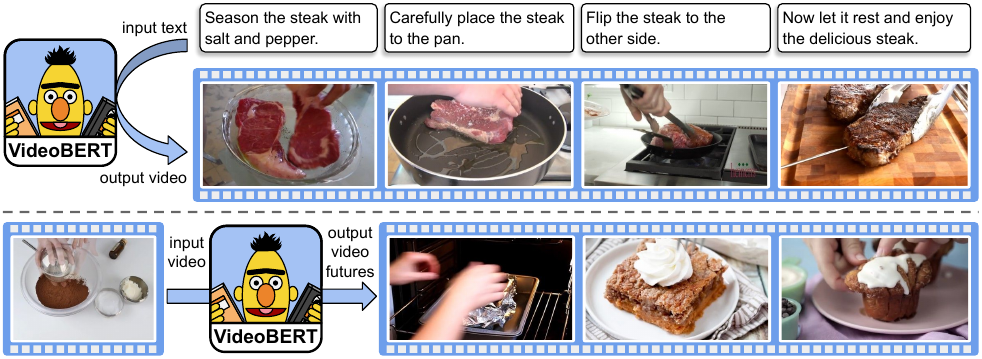}
    \caption{Text-to-video token generation and future token prediction with VideoBERT. The images depicted from the training data have the most similar token representation to the prediction. (Source: \cite{sun2019videobert})}
    \label{fig:videobert}
\end{figure}

Sun et al. \cite{sun2019videobert} build \textit{VideoBERT}, a joint visual-linguistic model in the style of BERT \cite{devlin2018bert}, which uses masked language model and next sentence prediction tasks to train the language understanding of a transformer. VideoBERT pairs representations of videos, consisting of spatiotemporal features extracted with pre-trained video classification models and their text description, obtained through an automatic speech recognition system, and trains the transformer on filling in masked tokens in both data types (representations of frames, not raw image data) or deciding whether the text matches the video features. The trained model is then used for action classification, video captioning, future video token forecasting, and prediction of video tokens for text descriptions (see \autoref{fig:videobert}), achieving coherent results and state-of-the-art captions.

Liu et al. \cite{liu2019auto} propose a \textit{graph transformer} that replaces the edge-output network of the GraphRNN \cite{you2018graphrnn} with a transformer decoder with self-attention layers and attention layers referring to the hidden graph state of the node \ac{RNN}. The results are competitive or better than GraphRNN on a variety of metrics.

\subsection{Generative Adversarial Networks}\label{sec:GANs}

\acfp{GAN} are frameworks consisting of a generator $G$ creating synthetic data from random noise and a discriminator $D$ determining whether a provided sample came from $G$ or the training data. The authors describe their system as a ``minimax two-player game'' \cite{goodfellow2014generative}, where the generator tries to deceive the discriminator. \cite{feng2020generative}

\begin{figure} [ht]
    \centering
    \includegraphics[width=0.8\textwidth]{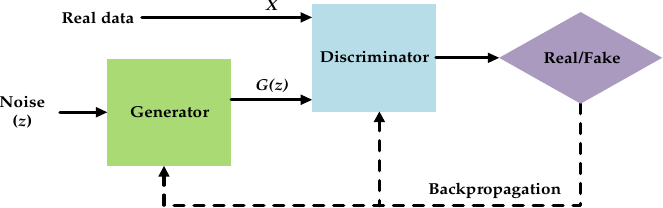}
    \caption{The original \ac{GAN} implementation. (Source: \cite{feng2020generative})}
    \label{fig:gan_original}
\end{figure}

As seen in \autoref{fig:gan_original}, both $D$ and $G$, originally implemented as \acp{MLP} to provide a network model with non-linear mapping, are learning from the results of $D$. $G$ learns the probability distribution of the real data $p_{data}(x)$ and $D$ the distribution of the random noise $p_{z}(z)$. The goal of the optimization process in Algorithm \ref{alg:gan_optimization} is reaching the \textit{Nash equilibrium} between $D$ and $G$, so both distributions become indistinguishable and the probability of $D$ classifying a sample as either fake or real approaches 50\%. \cite{feng2020generative}

In practice, minimizing the cost of the discriminator and generator jointly is difficult and often leads to instability because minimizing one cost function often means increasing the other. A \ac{GAN} may fail to converge. Further, \acp{GAN} often lack diversity due to the \textit{mode collapse problem} induced by the effort of the generator to deceive the discriminator, not represent a realistic data distribution. This often leads to only certain ``easy'' data types being generated and repeated. \cite{hong2019generative}

\begin{algorithm}
    \caption{Training algorithm for \acp{GAN}. $\theta_{d}$ and $\theta_{g}$ are the parameters of the respective \acp{MLP} $D$ and $G$ that are updated. In the original experiments, $k=1$ is used, but a higher $k$ is recommended to keep $D$ close to the optimal solution and prevent $G$ from overfitting. (Source: \cite{goodfellow2014generative})}\label{alg:gan_optimization}
    \begin{algorithmic}
        \FOR{number of training iterations}
            \FOR{$k$ steps}
                \STATE Sample minibatch of $m$ noise samples $\{z^{(1)},...,z^{(m)}\}$ from $p_{z}(z)$.
                \STATE Sample minibatch of $m$ examples $\{x^{(1)},...,x^{(m)}\}$ from $p_{data}(x)$.
                \STATE Update the discriminator by ascending its stochastic gradient: \center{$\nabla_{\theta_{d}}\frac{1}{m}\sum_{i=1}^{m}[\log D(x^{(i)})+\log(1-D(G(z^{(i)})))]$.}
            \ENDFOR
            \STATE Sample minibatch of $m$ noise samples $\{z^{(1)},...,z^{(m)}\}$ from $p_{z}(z)$.
            \STATE Update the generator by descending its stochastic gradient: \center{$\nabla_{\theta_{g}}\frac{1}{m}\sum_{i=1}^{m}\log(1-D(G(z^{(i)})))$.}
        \ENDFOR
        \STATE The gradient-based updates can use any standard gradient-based learning rule. We used momentum in our experiments.
    \end{algorithmic}
\end{algorithm}

The authors of the original \ac{GAN} framework, Goodfellow et al. \cite{goodfellow2014generative}, train a setup of two \acp{MLP} on three datasets: MNIST, \ac{TFD} and CIFAR-10. They achieve first and second-place results on the MNIST and \ac{TFD} datasets, respectively, when compared against a \ac{DBN}, a deep \ac{GSN}, and a stacked \ac{CAE} in terms of log-likelihood estimates \cite{breuleux2011quickly}.

\begin{figure} [ht]
    \centering
    \includegraphics[width=0.8\textwidth]{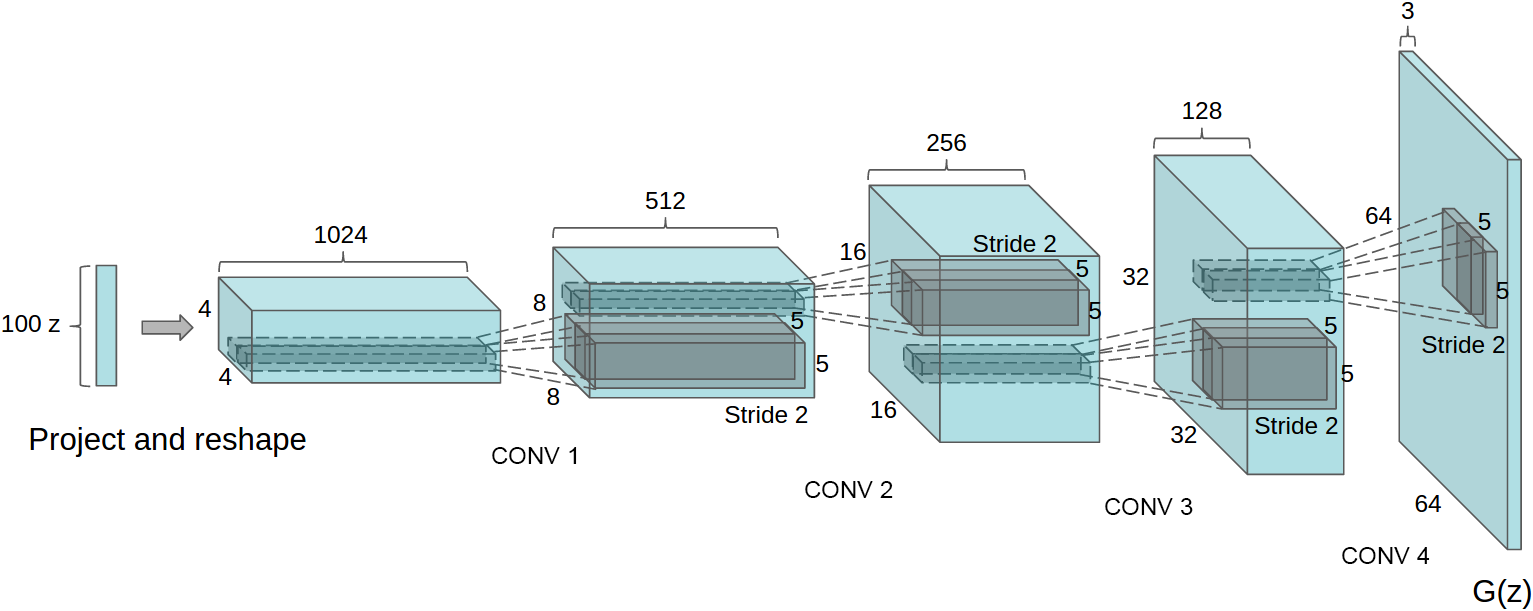}
    \caption{The \ac{DCGAN} generator architecture used to generate bedroom scene images trained on the LSUN data set. (Source: \cite{radford2015unsupervised})}
    \label{fig:dcgan_lsun}
\end{figure}

Radford et al. \cite{radford2015unsupervised} develop the \ac{DCGAN} architecture, which aims to adopt \acp{CNN} to unsupervised learning tasks by using the \ac{GAN} framework. \acp{DCGAN} replace the pooling functions of \acp{CNN} with stridden convolutions for the discriminator and fractional-stridden convolutions for the generator and remove fully connected layers entirely. Further, the neural networks now utilize \ac{ReLU} activation functions and batch normalization for all parts. An example of a \ac{DCGAN} architecture for image modeling is depicted in \autoref{fig:dcgan_lsun}. Frid-Adar et al. \cite{frid2018synthetic} apply the \ac{DCGAN} to generate labeled liver lesion images. 

Metz et al. \cite{metz2016unrolled} unroll the parameters of \ac{GAN} discriminators $\theta_{D}^{0}=\theta_{D}$ for $K$ future steps 
\begin{equation}
\theta_{D}^{k+1}=\theta_{D}^{k}+\eta^{k}\frac{df(\theta_{G},\theta_{D}^{k})}{d\theta_{D}^{k}}
\end{equation} and update the generator parameters
\begin{equation}
    \theta_{G}\leftarrow\theta_{G}-\eta\frac{df(\theta_{G},\theta_{D}^{K})}{d\theta_{G}}
\end{equation} 
with learning rate $\eta$ and objective function $f$, accordingly to stabilize the training of the generator at the expense of increased computational cost during training. By letting the generator ``see into the future'', the next discriminator step becomes less effective, which balances the two models better and improves convergence.

Elgammal et al. \cite{elgammal2017can} train a \ac{DCGAN} \cite{radford2015unsupervised} to produce art images. They achieve this by changing the training objectives of both the generator and discriminator to encourage learning about styles and art separately. The \ac{CAN} generator aims to deviate as much as possible from learned styles while keeping the learned art aspects. The discriminator, on the other hand, decides whether an input image is art or not and classifies the art style. During training, the discriminator is trained with art images with style labels to refine the decisions and style classifications. The generator receives the art/not art decision and the style ambiguity of the discriminator as a loss. An evaluation with human subjects indicates that generated art is indistinguishable from real art.

Donahue et al. \cite{donahue2018adversarial} introduce \textit{WaveGAN}, which synthesizes one-second clips of raw-waveform audio unsupervised. WaveGAN can generate words, bird chirping, and instruments by capturing periodic patterns in the sampled waveforms with convolutions. The approach is based on \ac{DCGAN} \cite{radford2015unsupervised} but has a modified one-dimensional convolution kernel with a higher stride. Further, \textit{phase shuffle} is used to shift the activations of each layer's activations by a random integer $\in[-n,n]$ to prevent the discriminator from learning trivial policies. They compare WaveGAN to another of their models based on \ac{DCGAN}, \textit{SpecGAN}, which works on spectrograms, and find that SpecGAN has a higher \ac{IS} and label accuracy through humans, but WaveGAN produces samples of higher sound quality.

\othertab{\acp{GAN}}{
    \cite{durugkar2016generative} & Generative Multi-Adversarial Networks (GMANs): The first introduction of \ac{DCGAN} \cite{radford2015unsupervised} with multiple discriminators demonstrated on image generation tasks, resulting in higher-quality images and robustness to mode collapse. & 2016 \\\hline 
    \cite{che2016mode} & Mode regularized \acp{GAN} for stable training and reduced risk of mode collapse: Train an encoder $E(x):X\rightarrow Z$ together with generator $G(z):Z\rightarrow X$ and add a similarity loss $\mbox{distance}(x,G(E(x)))$ for stable training gradients. Further, a mode regularizer objective $D_{1}(G(E(x)))$ on the training data $x$ is employed to force the generator to cover the whole data space. After training with $x$, a second discriminator $D_{2}$ discriminates $G(z)$ and $G(E(x))$ to bring both distributions to the same manifold efficiently. & 2016 \\\hline 
    \cite{wu2016learning} & 3D-GAN: A \ac{CNN} generator creates objects in 3D voxel space from a random vector, and the discriminator, which is a reversed generator, judges whether the objects are real. The model learns mappings between low-dimensional latent vectors and 3D objects, which can be used, apart from generative purposes, for object recognition and description. They further introduce 3D-VAE-GAN, which consists of a convolutional image encoder and allows one-shot image-conditional 3D model generation. & 2016 \\\hline 
    \cite{zhang2016generating} & TextGAN: Text generation with a \ac{LSTM} generator and a \ac{CNN} discriminator/encoder. Instead of the standard \ac{GAN} objective, the generator is trained to minimize the covariance matrices of the real and synthetic sentence feature vectors obtained from the \ac{CNN} encoder. The discriminator is trained with the standard adversarial loss, latent code reconstruction loss, and generator loss. The generator is pre-trained in an autoencoder \ac{LSTM} setting, while the discriminator is pre-trained to classify sentences with swapped words from true sentences. In \cite{zhang2017adversarial}, they go into further detail and explore a \ac{MMD}-based feature matching loss instead of covariance. & 2016 \\\hline 
    \cite{liu2016coupled} & Coupled GAN (CoGAN): Multiple generator-discriminator pairs with shared weights in the higher abstraction layers allow learning of image relations without tuples of corresponding images. The model learns to generate corresponding images for different domains from the same noise vector $\mathbf{z}$, making the model suitable for unsupervised domain adaptation and image transformation (e.g., learning the relationship between depth and color in RGBD images). & 2016 \\\hline 
    \cite{zhao2016energy} & Energy-based \ac{GAN} (EBGAN): Defining the discriminator as an energy function \cite{lecun2006tutorial} that assigns low energies to data points near the data manifold and vice versa. The discriminator is implemented as an autoencoder, and the reconstruction error is the energy function. During training, the EBGAN is more stable than regular \acp{GAN} and generates high-resolution images. & 2016 \\\hline 
    \cite{mogren2016c} & C-RNN-GAN: A model with deep \ac{LSTM} generator and discriminator to create continuous sequential data with one or more values at each step. The adversarially trained model generates polyphonic music with a sense of timing and variation but is distinguishable from real music. & 2016 \\\hline 
    \cite{brock2016neural} & Introspective adversarial network (IAN): A hybrid of a \ac{VAE} for representation learning and reconstruction and a \ac{GAN} to improve \ac{VAE} performance. The network combines \ac{GAN} discriminator and \ac{VAE} encoder and \ac{GAN} generator and \ac{VAE} decoder. Used for realistic photo editing/interpolation. & 2016 \\\hline 
    \cite{chen2016infogan} & InfoGAN: An unsupervised \ac{GAN} that learns disentangled representations with a mutual information objective on latent variable subsets. The model can separate writing styles from digits on MNIST and hairstyles, emotions, and eyeglasses from faces on the CelebA data set. & 2016 \\\hline 
    \cite{nowozin2016f} & f-GAN: Training of generative neural samplers (probabilistic feedforward neural networks), which efficiently convert a random input vector to a sample, using an auxiliary discriminative neural network or other $f$-divergences (e.g., Kullback-Leibler, Jensen-Shannon). Results, also with DCGAN \cite{radford2015unsupervised}, show that the discriminator approach does not necessarily perform better. & 2016 \\\hline 
    \cite{im2016generating} & Generative Recurrent Adversarial Network (GRAN): A recurrent attention-based generator/decoder applies sequential changes to a canvas based on the previous hidden state and random noise at each time step, while an attention-based encoder/discriminator provides a feature-based loss. An evaluation method involving a ``battle'' between separately trained generators and discriminators is proposed, where GRAN outperforms DRAW \cite{gregor2015draw} and the denoising \ac{VAE} \cite{im2015denoising}. & 2016 \\\hline 
    \cite{salimans2016improved} & Offering a suite of methodologies for improved training of \acp{GAN}. These techniques, such as feature matching, minibatch discrimination, and historical averaging, address the challenges of \ac{GAN} training, particularly in achieving convergence. The authors further contributed to stabilizing \ac{GAN} training and proposed a novel evaluation metric now widely used for GAN performance evaluation, the \ac{IS}, to assess sample quality. & 2016 \\\hline
    \cite{arjovsky2017wasserstein} & Wasserstein GAN (WGAN): Replacement of the discriminator with a \textit{critic} that approximates the Earth-Mover distance (Wasserstein-1) between the real data distribution and the generator data distribution. Since the Wasserstein distance is continuous and differentiable under most circumstances, the generator can be trained with gradient descent, solving the training instability and mode collapse problems of \acp{GAN} and providing meaningful learning curves, as demonstrated on image generation tasks. In \cite{gulrajani2017improved}, the training is improved with a gradient penalty with respect to the critic input. & 2017 \\\hline 
    \cite{berthelot2017began} & Boundary Equilibrium \ac{GAN} (BEGAN): Similar to EBGAN \cite{zhao2016energy}, an autoencoder is used as a discriminator, but instead of using the reconstruction loss of samples directly, the Wasserstein distance between error distributions of real and generated samples is computed. Further, the equilibrium condition between the expected errors of generated and real samples is relaxed with a hyperparameter $\gamma\in[0,1]$ that influences the diversity of the generated images. BEGAN outperforms previous models in image quality at higher resolutions. & 2017 \\\hline 
    \cite{mao2017least} & Least Squares GAN (LSGAN): Using the least squares loss for the discriminator instead of the sigmoid cross-entropy loss, resulting in higher quality images generated and a more stable training procedure compared to regular \acp{GAN}. The new loss penalizes correct decisions far beyond the decision boundary (i.e., too much certainty by the discriminator) to prevent vanishing gradients and move the generator towards the decision boundary, which converges to the real data manifold. & 2017 \\\hline 
    \cite{lin2017adversarial} & RankGAN: Instead of binary judgments of individual data samples by the discriminator, generator outputs are mixed with real data and ranked by the discriminator according to a reference. The more detailed feedback allows the generator to learn better what makes data realistic. The model generates natural language sentences but could be extended for image generation and captioning. & 2017 \\\hline 
    \cite{grnarova2017online} & Chekhov GAN: Treating \ac{GAN} training as a zero-sum game that can be solved by finding a mixed strategy. Using ideas from online learning, where a player aims to minimize a sequentially-revealed cumulative loss function, a no-regret strategy for both generator and discriminator that incorporates the history of the model's actions is employed. The model improves stability and mode collapse problems and is guaranteed to converge to equilibrium for specific \ac{GAN} architectures. The model is evaluated on image generation and density estimation tasks. & 2017 \\\hline 
    \cite{kim2017adversarially} & Adversarially Regularized Autoencoder (ARAE): A framework that trains a Wasserstein GAN \cite{arjovsky2017wasserstein} to produce latent codes of a simultaneously trained autoencoder to create a \ac{GAN} for discrete data (e.g., images, text). & 2017 \\\hline 
    \cite{mroueh2017mcgan} & Mean and covariance feature matching GAN (McGAN): Training \acp{GAN} by matching statistics (e.g., embedded mean or covariance of features) of real and fake data instead of classifying individual samples. The approach is adapted to DCGAN \cite{radford2015unsupervised} and used to generate images. & 2017 \\\hline 
    \cite{mroueh2017fisher} & Fisher GAN: Instead of penalizing the gradients of the Wasserstein GAN discriminator \cite{gulrajani2017improved}, a constraint is imposed on its second-order moments, inspired by the \textit{Fisher Discriminant Analysis} method. The approach is applied to a DCGAN \cite{radford2015unsupervised} architecture, outperforming other models in unconditional image generation. & 2017 \\\hline 
    \cite{rosca2017variational} & $\alpha$-GAN: Combining an autoencoder with a \ac{GAN}, where the \ac{GAN} generator is trained to reconstruct latent representations of an encoder and the latent space is encouraged to conform to a Gaussian distribution. Further, a discriminator is used to evaluate the realism of the generated or reconstructed samples. The combination of autoencoders and \acp{GAN} (DCGAN \cite{radford2015unsupervised} in particular) solves blurriness and mode collapse issues for image generation. & 2017 \\\hline 
    \cite{yu2017seqgan} & SeqGAN: Application of the \ac{GAN} architecture to the generation of sequences of discrete tokens. Since the default \ac{GAN} generator generates continuous values and receives instant feedback from the discriminator, and the discriminator can only evaluate complete sequences, the authors model the generator as a \ac{LSTM} and \ac{RL} agent with a stochastic policy whose intermediate actions are rewarded after the complete sequence has been judged by the \ac{CNN} discriminator. Lee et al.~\cite{lee2017seqgan} applied SeqGAN to polyphonic music generation using efficient representations of polyphonic MIDI files. & 2017 \\\hline 
    \cite{saito2017temporal} & Temporal \ac{GAN} (TGAN): The video generator consists of a temporal generator \ac{CNN} that generates $T$ latent variables $z_{1}^{t}$ from initial noise $z_{0}$ and an image generator that generates $T$ video frames from $z_{0}$ and the respective $z_{1}^{t}$. The discriminator receives all frames and evaluates, with multiple convolutional layers, whether the video is real or generated. Image generator and discriminator are very similar to \ac{DCGAN} \cite{radford2015unsupervised}, and training also incorporates the WGAN \cite{arjovsky2017wasserstein} objective. & 2017 \\\hline 
    \cite{srivastava2017veegan} & VEEGAN: Using a reconstructor network $F(x)$ to map the true data distribution $p(x)$ back to the Gaussian distribution $p(z)$ of the generator $G(z)$'s latent code. The Kullback-Leibler divergence serves as an expressive loss function between distributions $F(G(p(z)))$ and $p(z)$ because they should be identical. This autoencoder-styled structure aims to solve the mode collapse problem of \acp{GAN} by checking if this whole distribution is covered. The model is demonstrated on density estimation and image generation tasks, where it is less prone to mode collapse issues than other \ac{GAN} approaches. & 2017 \\\hline 
    \cite{hjelm2017boundary} & Boundary-seeking GAN (BGAN): Using the estimated difference measure of a discriminator as a differentiable policy gradient for the generator to allow \acp{GAN} to generate discrete data. The approach is also suitable for continuous data and is demonstrated in natural language and image generation. & 2017 \\\hline 
    \cite{choi2017generating} & medGAN: Generation of synthetic \acfp{EHR} for privacy-preserving data sharing for medical research. An autoencoder provides discrete real data $\mathbf{x}$ or discrete data from noise $\mathbf{z}$ to a discriminator $D$ in the form $Dec(Enc(\mathbf{x}))$ or $Dec(G(\mathbf{z}))$ respectively, where generator $G$ and $D$ are feedforward neural networks. & 2017 \\\hline 
    \cite{yang2017midinet} & MidiNet: A generator \ac{CNN} creates symbolic MIDI melodies from random noise and prior knowledge (e.g., chord progression, previous bars, priming melody) from a conditioner \ac{CNN} with \textit{transposed convolutions} \cite{dumoulin2016guide} and a discriminator \ac{CNN} predicts whether an input score is real or fake. & 2017 \\\hline 
    \cite{karras2017progressive} & Progressive GAN: Accelerating and stabilizing \ac{GAN} training by progressively adding convolutional layers to the generator and discriminator, increasing the resolution for image generation iteratively. & 2017 \\\hline 
    \cite{kodali2017train} & DRAGAN: Treating \ac{GAN} training as a zero-sum game solved using a regret minimization technique. Mode collapse is interpreted as a problem caused by local equilibria in the training ``game''. The model is stable and outperforms Wasserstein GAN \cite{arjovsky2017wasserstein,gulrajani2017improved} significantly. & 2017 \\\hline 
    \cite{xu2017gang} & Gang of \acp{GAN} (GoGAN): Improvement over Wasserstein \ac{GAN} \cite{arjovsky2017wasserstein} by using a margin-based discriminator loss that enforces a certain distance $\epsilon$ between discriminator scores of fake and real samples and progressively training multiple \acp{GAN} that are evaluated against each other with a maximum margin ranking loss that ensures that later \acp{GAN} perform significantly better than earlier ones. The model is used for image completion. & 2017 \\\hline 
    \cite{guo2017long} & LeakGAN: The \ac{CNN} discriminator leaks its high-level abstracted features of the currently evaluated sentence to the \ac{LSTM} generator that takes the additional input as guidance for next word generation for a text. & 2017 \\\hline 
    \cite{dumoulin2017adversarially} & Combining a classic \ac{GAN} architecture with an inference network. This allows the discriminator to distinguish between joint latent/data-space samples from the generative network and joint samples from the inference network (which receives a data sample as input and outputs synthetic data). The setup enhances the performance of state-of-the-art tasks. & 2017 \\\hline
    \cite{yang2017lr} &  LR-GAN: Generating images by creating backgrounds and foregrounds separately and then stitching them together. It employs a recursive approach where each layer (background or foreground) is generated step-by-step, each with its own shape and pose. This method allows for a contextually relevant composition of images, leading to more natural and realistic image generation. & 2017 \\\hline
    \cite{warde-farley2017improving} & Enhancing \acp{GAN} using denoising feature matching. This technique guides the generator towards more probable configurations of abstract discriminator features, generating more object-like samples. The approach uses a denoising auto-encoder to estimate and track the distribution of these features derived from real data. This is combined with the original \ac{GAN} loss, and the augmented training procedure is shown to improve its stability. & 2017 \\\hline
    \cite{gulrajani2017improved} & Applying Wasserstein \ac{GAN} for data generation, using a gradient penalty to enforce a Lipschitz constraint on the discriminator. This addresses the training instability associated with weight clipping in Wasserstein \ac{GAN}. The method enables stable training across various \ac{GAN} architectures, significantly reducing the need for hyperparameter tuning. & 2017 \\\hline
    \cite{heusel2017gans} & In contrast to \ac{GAN} training training, the authors update the discriminator and generator at different rates. This method facilitates more stable convergence and enables the networks to reach a local Nash equilibrium effectively. Additionally, the introduction of the \acf{FID} provides a more reliable and sensitive metric for \ac{GAN} performance evaluation when compared to the \ac{IS}. The effectiveness was validated using several \ac{GAN} architectures, improving stability and image quality in the generated samples. & 2017 \\\hline
    \cite{tulyakov2018mocogan} & Motion and content decomposed GAN (MoCoGAN): Video generation using a fixed content vector and a sequence of motion vectors that are converted to a state using a \ac{RNN}. An image generator \ac{CNN} then creates a frame from both vectors, and \ac{CNN}-based video and image discriminators provide feedback. & 2018 \\\hline 
    \cite{dong2018musegan} & MuseGAN: A WGAN \cite{arjovsky2017wasserstein} utilizing deep \acp{CNN} that generate multi-track music sequences with piano-roll representations. It combines a jamming model that improvises one track with a composer model that creates multiple accompanying tracks at once, so multiple hybrid generators with inter-track random vector $\mathbf{z}$ and intra-track random vectors $\mathbf{z}_{i}$ are paired with one discriminator. & 2018 \\\hline 
    \cite{jaiswal2018capsulegan} & CapsuleGAN: Replaces the \ac{CNN} discriminator in a \ac{DCGAN} \cite{radford2015unsupervised} with a capsule network (CapsNet) \cite{sabour2017dynamic} classifier to improve accuracy and generator performance. & 2018 \\\hline 
    \cite{cao2018molgan} & MolGAN: Training a \ac{MLP} generator to produce adjacency and node annotation matrices for small molecules with a graph-convolutional discriminator and reward network. The reward network, like in \ac{RL}, learns to score non-differentiable metrics (e.g., solubility in water) with the help of external software and to guide the molecule generation towards a specific target. & 2018 \\\hline 
    \cite{bojchevski2018netgan} & NetGAN: Generating graphs as a set of random walks (node sequences) with a \ac{LSTM} generator outputting vertex after vertex of the sequence and a discriminator \ac{LSTM} that judges whether the sequence belongs to the real graph or is fake. Training samples (random walks) are obtained from a real graph, and synthetic graphs are obtained by merging the random walks. & 2018 \\\hline 
    \cite{park2018data} & table-GAN: Application of the \ac{DCGAN} \cite{radford2015unsupervised} architecture to the generation of private and useful tabular data. Original table entries are fed as square zero-padded matrices to the \ac{GAN} for training, with a neural network classifier $C$ besides the generator and discriminator that enforces data consistency learned from the original table. & 2018 \\\hline 
    \cite{patel2018correlated} & corrGAN: Generation of correlated discrete data (e.g., table entries, binary images) with an autoencoder that learns mappings between discrete input/output and continuous latent space while considering the correlations between subsets of the discrete variables. The decoder then serves as the output layer of a \ac{GAN} generator modeling the continuous latent space. & 2018 \\\hline 
    \cite{xu2018synthesizing} & Tabular GAN (TGAN): A \ac{LSTM} generates discrete and continuous values encoded with probability distributions column by column for each entry, and a \ac{MLP} discriminator scores the likelihood and diversity of the data. & 2018 \\\hline 
    \cite{wei2018improving} & Consistency Term (CT) GAN: Improved training of improved Wasserstein \acp{GAN} \cite{arjovsky2017wasserstein,gulrajani2017improved} by additionally training the discriminator on once and twice perturbed real data and evaluating the responses. Combined with other small optimizations, this approach achieves state-of-the-art generative results and is better suitable for semi-supervised learning than other \acp{GAN}. & 2018 \\\hline 
    \cite{bora2018ambientgan} & AmbientGAN: Training a \ac{GAN} on lossy measurements by passing the generator output through a simulated random measurement function that corrupts it. The discriminator then tries to distinguish the lossy real and fake measurements. The model can successfully inpaint lossy images. & 2018 \\\hline 
    \cite{anand2018generative} & Protein structure generation and completion by encoding them as pair-wise distances between $\alpha$-carbons in a matrix using a DCGAN \cite{radford2015unsupervised} and recovering the 3D structure using the ``alternating directional method of multipliers'' (ADMM) algorithm. & 2018 \\\hline 
    \cite{ouyang2018non} & Human trajectory generation as a sequence of stays with a location $(x,y)$, a start time $t$, and a duration $d$. A \ac{GAN} made of \acp{CNN} creates and evaluates ``maps'' with the aforementioned times as coordinate values. & 2018 \\\hline 
    \cite{hoang2018mgan} & MGAN: Tackling the mode collapse in \acp{GAN} by using multiple generators. This method employs a mixture of generators with a shared classifier and discriminator, aiming to produce diverse outputs that cover different data modes. The authors demonstrate that their approach effectively minimizes the Jensen-Shannon Divergence between the mixed generator distributions and the real data distribution. Empirical results showed the model's ability to generate diverse and recognizable images, indicating a significant improvement over single-generator \ac{GAN}. & 2018 \\\hline
    \cite{nie2018relgan} & RelGAN: Text generation with a configurable trade-off between sample quality and diversity. The recurrent generator incorporates relational memory \cite{santoro2018relational} (multiple memory slots with self-attention \cite{vaswani2017attention})  to model long-range dependencies, and the \ac{CNN} discriminator creates multiple representations for each sentence to evaluate the sentence from different aspects and provide better feedback. & 2018 \\\hline 
    \cite{tran2018dist-gan} & Dist-GAN: A \ac{GAN} enhanced with distance constraints. It addresses two key issues in \ac{GAN} training: gradient vanishing and mode collapse. The novel approach includes integrating an autoencoder with the generator and implementing two distance constraints, one in the latent space and another based on discriminator scores. This stabilizes the \ac{GAN} training while retaining competitive scores in classification tasks. & 2018 \\\hline
    \cite{miyato2018spectral} & Introducing spectral normalization, a technique for stabilizing \ac{GAN} training. It normalizes the spectral norm of the weight matrices in the discriminator, using a Lipschitz constant as the only hyper-parameter to be tuned. This method is simple, computationally efficient, and stabilizes \ac{GAN} performance, particularly in image generation tasks on datasets. & 2018 \\\hline
    \cite{he2018bayesian} & Bayesian modeling is employed to tackle mode collapse in \acp{GAN}, a well-documented and actively researched problem. The authors propose learning the distributions of generators, as it mirrors a common approach of training a \ac{GAN} with multiple generators to alleviate the mode collapse. & 2018 \\\hline
    \cite{genevay2018learning} &  Using the Sinkhorn divergence to train generative models based on regularized optimal transport with an entropy penalty. This method addresses the computational and statistical challenges of using optimal transport metrics in training generative models. The Sinkhorn divergence interpolates between Wasserstein and \ac{MMD} losses, leveraging the geometrical properties of the optimum transport and the favorable high-dimensional sample complexity of \ac{MMD}. & 2018 \\\hline
    \cite{karras2019style} & StyleGAN: A style-transfer related progressive \ac{GAN} \cite{karras2017progressive} capable of unsupervised learning of disentangled high-level attributes of images in latent space and generation of realistic images from that latent space and random noise. The model can also interpolate (style mixing) in the latent space. & 2019 \\\hline 
    \cite{jordon2018pate} & PATE-GAN: Differentially private data generation with the original \ac{GAN} utilizing the private aggregation of teacher ensembles (PATE) \cite{papernot2018scalable} framework which allows to bound the influence of individual samples on the generator. A set of \textit{teacher} discriminators is trained on equal amounts of generator output and disjoint training data. A \textit{student} discriminator is trained on the output labels of the teachers, and the generator is trained on the student's output. & 2019 \\\hline 
    \cite{chen2019adversarial} & Improved text generation with \acp{GAN} by segmenting sentences into sub-sequences and providing simultaneous discriminator feedback for all sub-sequences and the entire sentence instead of the sentence alone. Applying this approach to previous state-of-the-art \acp{GAN} \cite{yu2017seqgan,guo2017long,nie2018relgan} significantly improves their results. & 2019 \\\hline 
    \cite{ghelfi2019adversarial} & SemGAN: Generation of pixel-level semantic images from latent vectors with prior distribution to speed up convergence. Class probabilities for each pixel are forwarded as a softmax distribution to the discriminator, allowing for detailed feedback. The SemGAN produces significantly cleaner results than a default GAN working with RGB semantic mappings. Using the image-to-image \ac{GAN} from \cite{isola2017image}, the creation of natural images from these artificial semantic images is demonstrated. & 2019 \\\hline 
    \cite{baowaly2019synthesizing} & Medical Wasserstein GAN (medWGAN) \& Medical Boundary-seeking GAN (medBGAN): Modified versions of medGAN \cite{choi2017generating} with better performance. medBGAN outperforms all other approaches on synthetic \ac{EHR} generation. & 2019 \\\hline 
    \cite{sedigh2019generating} & Using a default \ac{GAN} to generate images of skin cancer, a domain for which only a few labeled samples are available. The synthetic data is used to boost the performance of a cancer detection \ac{CNN}. & 2019 \\\hline 
    \cite{gong2019autogan}  & AutoGAN: An automated approach of solving neural architecture search specifically tailored to \ac{GAN} architectures. To do so, an additional \ac{RNN} controller is incorporated for the search process, while the \ac{IS} is used as the success reward of the reinforcement. & 2019 \\\hline
    \cite{doppelganger} & DoppelGANger: A generative model for creating synthetic time series data tackling several issues about \acp{GAN} and data privacy. This model effectively captures and generates complex correlations between time series and its attributes while addressing challenges such as mode collapse and variable data lengths. & 2019 \\\hline
    \cite{time-gan} & TimeGAN: Generating realistic time-series data by combining supervised and adversarial training, addressing the challenge of preserving temporal dynamics in generated sequences. It features an embedding network for dimensional reduction, enhancing the generative model's learning of temporal relationships. & 2019 \\\hline
    \cite{karras2020analyzing} & StyleGAN2: Improvement of StyleGAN \cite{karras2019style} by simplifying and restructuring the generator architecture. They further employ lazy regularization and remove the progressive growing in favor of skip connections between up- and downsampling layers. They also developed a method to project images back to latent space where style could be changed/interpolated. & 2020 \\\hline 
    \cite{yale2020generation} & HealthGAN: A feedforward neural network based \ac{GAN} that combines ideas from medGAN \cite{choi2017generating} with the Wasserstein GAN gradient penalty \cite{gulrajani2017improved} and data transformations to enable the use of continuous and categorical data. The model is purposefully small to prevent memorization and used to generate private \ac{EHR} records. & 2020 \\\hline 
    \cite{alam2020synthetic} & Using a progressively growing \ac{GAN} (similar to \cite{karras2017progressive}) with Wasserstein gradient penalty loss \cite{gulrajani2017improved} to generate magnetic resonance imaging of brains that can be used for Attention Deficit Hyperactivity Disorder prediction. & 2020 \\\hline 
    \cite{islam2020gan} & Using \ac{DCGAN} \cite{radford2015unsupervised} to generate positron emission tomography brain images for three stages of Alzheimer's disease to build an automated diagnosis model. & 2020 \\\hline 
    \cite{hazra2020synsiggan} & SynSigGAN: Generation of fixed-length labeled biomedical signals (electrocardiogram and other time series data) using a bidirectional grid \ac{LSTM} and a \ac{CNN} discriminator with a small amount of training data. The data can be used for automatical medical diagnosis or training medical students and achieves state-of-the-art performance when used as training data for a classifier. & 2020 \\\hline 
    \cite{gao2020adversarial} & By analyzing the loss of the generator and the discriminator during the training process, overarching calculations such as \ac{IS} or \ac{FID} can be eliminated. This reduces the computing time greatly and optimizes the network at the same time, especially compared to either a manual \ac{GAN} configuration or similar automated approaches based on the \ac{IS} oder \ac{FID}. & 2020 \\\hline
    \cite{cot-gan} & COT-GAN: A generative model for sequential data employing causal optimal transport for training implicit generative models, integrating classic optimal transport methods with a temporal causality constraint. The COT-GAN framework is adept at generating low- and high-dimensional time series data. & 2020 \\\hline
    \cite{zhao2020differentiable} & Using differentiable augmentations applied to real and fake samples during training, this approach effectively prevents overfitting in \ac{GAN} and reduces the training size. The method shows notable improvements across various \ac{GAN} architectures and achieves state-of-the-art results while notably only using 20\% of the training data. & 2020 \\\hline
    \cite{dewi2021synthetic} & Using \ac{DCGAN} \cite{radford2015unsupervised} to boost the amount of training data for traffic sign recognition, resulting in increased accuracy and reduced detection time. & 2021 \\\hline 
    \cite{imtiaz2021synthetic} & Combining the boundary-seeking GAN \cite{hjelm2017boundary} with \textit{noise addition} directly on the data to generate differentially private smart health care data sets of populations for medical research. Training data is obtained from Fitbit smartwatches. & 2021 \\\hline 
    \cite{gans-for-ecg} & Introducing a method for generating synthetic \acfp{ECG} using \acp{GAN} to address privacy concerns in medical data sharing. The authors presented two \ac{GAN} models, WaveGAN and Pulse2Pulse, trained on real normal \acp{ECG} to produce synthetic, plausible \acp{ECG}. The presented approach allows for generating an arbitrary amount of normally very sensitive patient data and open-sources the over 100,000 normal \acp{ECG}. & 2021 \\\hline
    \cite{tts-gan} & TTS-GAN: Generating synthetic time-series data using transformer architecture. It employs transformer encoders in both generator and discriminator networks, overcoming limitations of \ac{RNN}-based \ac{GAN} in handling long sequences. It showcases improved performance in generating realistic sequences across multiple datasets. & 2022 \\
}

\subsubsection{Conditional GANs}

\begin{figure} [ht]
    \centering
    \includegraphics[width=0.5\textwidth]{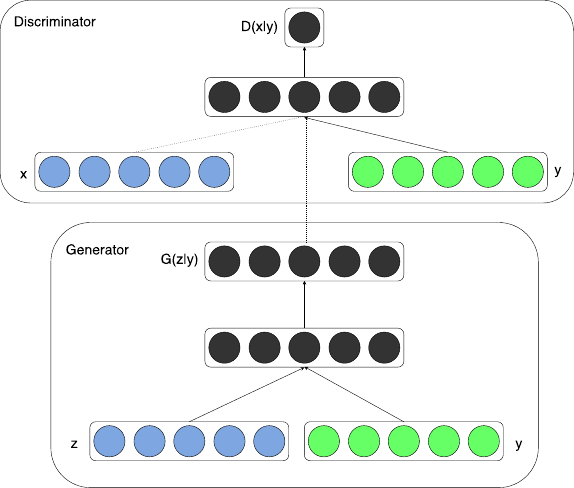}
    \caption{A conditional \ac{GAN} conditioned on additional input $\mathbf{y}$. (Source: \cite{mirza2014conditional})}
    \label{fig:conditional_gan}
\end{figure}

Normal \acp{GAN} have no control over the type of data the generator outputs. Conditional \acp{GAN} solve this problem by conditioning generator and discriminator on additional information $\mathbf{y}$, which could be class labels, for example, that are provided at the input layer (see \autoref{fig:conditional_gan}). The objective function of the \ac{GAN} is modified as follows \cite{mirza2014conditional}:

\begin{equation}
    \min_{G}\max_{D}V(D,G)=\mathbb{E}_{\mathbf{x}\sim p_{data}(\mathbf{x})}[\log D(\mathbf{x}\vert\mathbf{y})]+\mathbb{E}_{\mathbf{z}\sim p_{z}(\mathbf{z})}[\log(1-D(G(\mathbf{z}\vert\mathbf{y})))].
\end{equation}

Mathieu et al. \cite{mathieu2015deep} adopt the \ac{GAN} architecture to predict the next video sequence frames and introduce a loss function that improves the sharpness of the image predictions. The generator and discriminator use multi-scale networks, which are \acp{CNN} that up- or down-scale the resolution of a prediction multiple times, respectively, until the target image size or a single scalar output is reached. The model is trained by providing a sequence of video frames to both models and the real or generated additional frames to the discriminator. The model parameters are updated using stochastic gradient descent. They achieve state-of-the-art sharpness and similarity on a collection of sports clips from the Sports1m and UCF-101 data sets.

Zhu et al. \cite{zhu2016generative} propose using the \ac{DCGAN} \cite{radford2015unsupervised} architecture to learn the approximate natural image manifold by training the generator to produce realistic images given a 100-dimensional random vector $z$. Then, real images are projected to such a latent representation $z_{0}$, and manipulating operations (coloring, sketching, or warping executed with brush tools) are applied to that vector, resulting in $z_{i}$. The changes in the artificially generated images $G(z_{i})$ are then sequentially transferred to the original photo using optical flow methods to maintain image quality. Besides interactive image manipulation, the \textit{iGAN} model can also generate objects from drawn sketches.

Reed et al. \cite{reed2016generative} generate images from textual descriptions (sentences) using a \ac{DCGAN} \cite{radford2015unsupervised} architecture. First, a text embedding encoder model (hybrid character-level convolutional recurrent neural network) is pre-trained by comparing its embeddings to the ones encoded by a corresponding image encoder (deep \ac{CNN}). The \ac{GAN} consists of a deep convolutional generator that takes as input the text embedding and random noise and a deep convolutional discriminator that decides whether the image is real or fake based on the text embedding and a provided image. They also test a modified discriminator with additional real training samples with mismatched text to condition the model on matching texts in addition to image realism. A further addition to the generator training objective is the interpolation of text representations to ensure that gaps in the training data also correspond to the data manifold:

\begin{equation}
    \mathbb{E}_{t_{1},t{2}\sim p_{data}}[\log(1-D(G(z,\beta t_{1}+(1-\beta) t_{2})))]
\end{equation}

with noise sample $z$, text embeddings $t_{1}$ and $t_{2}$ and $\beta=0.5$, which works well as long as the discriminator can recognize the matching image and text pairs. Further, the author developed a style encoder, which, in combination with the trained generator, allows them to combine the style of one image and a description to generate a same-style image with properties matching the text (e.g., an image of an eagle with the description of a red bird results in a red eagle).

In \cite{reed2016learning}, Reed et al. introduce the Generative Adversarial What-Where Network (GAWWN) as an extension of \cite{reed2016generative}, that allows specification of locations (bounding-boxes) of objects and their parts (keypoints) for text-to-image synthesis. They present two models, the bounding-box-conditional and the keypoint-conditional text-to-image model, and apply a \ac{GAN} architecture to generate missing keypoints given some user-defined keypoints and the text description or text alone, which they demonstrate also works.

Isola et al. \cite{isola2017image} perform image-to-image translation using a conditional \ac{DCGAN} \cite{radford2015unsupervised} with a generator $G: \{x,z\}\rightarrow y$ and a discriminator $D: \{x,y\}\rightarrow [\mbox{real},\mbox{fake}]$, where $x$ is the ``concept'' of an image (e.g., an edge drawing or blueprint), $y$ is the real or generated image and $z$ is random noise, which is later replaced by dropout on several layers because $G$ tends to ignore $z$. The generator is a deep \ac{CNN} which first down- and then up-samples $x$ to obtain essential features and create a fitting image in the target ``style''. The discriminator \textit{PatchGAN} is a convolutional model that classifies if each $N \times N$ patch of an image is real or fake and averages all outputs. The loss function combines the discriminator output and the L1 loss (mean absolute error) of the generated image and the ground truth. The model is then trained using alternating gradient descent steps on $D$ and $G$.

\begin{figure} [ht]
    \centering
    \begin{subfigure}{.22\textwidth}
        \includegraphics[width=\textwidth]{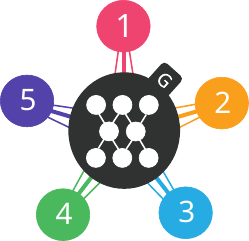}
        \caption{Overall architecture of StarGAN with one generator and multiple discriminators.}
        \label{fig:stargan_arch}
    \end{subfigure}
    \hfill
    \begin{subfigure}{.67\textwidth}
        \includegraphics[width=\textwidth]{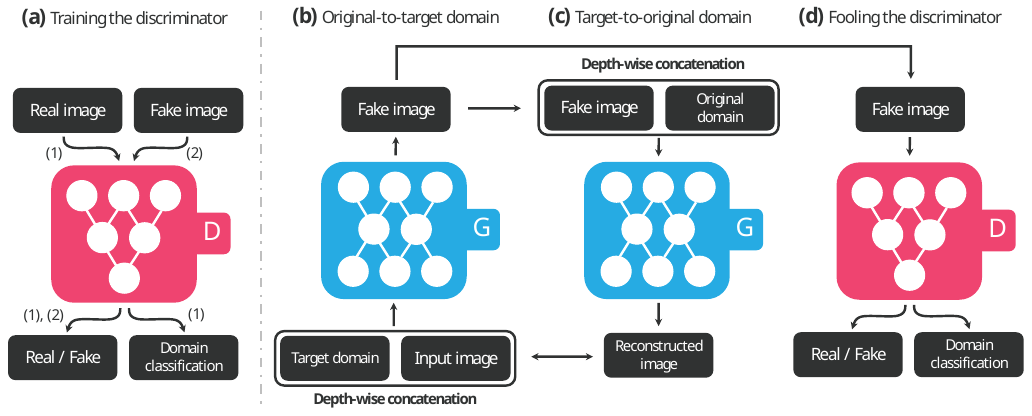}
        \caption{\textbf{Right:} Discriminator training. \textbf{Left:} Generator training with reconstruction loss \textbf{(c)} and adversarial loss by the discriminator \textbf{(d)}.}
        \label{fig:stargan_training}
    \end{subfigure}
    \caption{Illustrations of StarGAN. (Source: \cite{choi2017stargan})}
    \label{fig:stargan}
\end{figure}

Choi et al. \cite{choi2017stargan} introduce \textit{StarGAN} based on \cite{zhu2017unpaired} for multi-domain image-to-image translation using one generator and multiple discriminators, one for every pair of image domains (see \autoref{fig:stargan_arch}). This allows StarGAN to be trained on multiple data sets with different labels simultaneously and combine the information learned into one generative model.

\begin{figure} [ht]
    \centering
    \begin{subfigure}{.6\textwidth}
        \includegraphics[width=\textwidth]{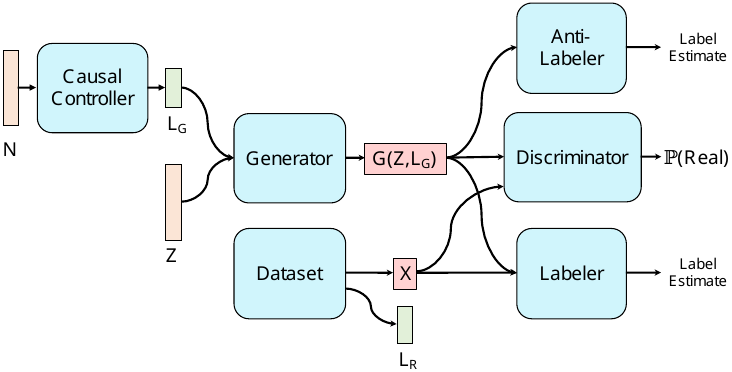}
        \caption{Overall architecture of CausalGAN.}
        \label{fig:causalGAN_overall}
    \end{subfigure}
    \hfill
    \begin{subfigure}{.35\textwidth}
        \includegraphics[width=\textwidth]{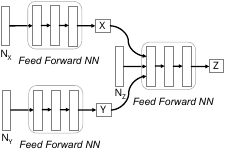}
        \caption{The \ac{BN}-like generator of CausalGAN.}
        \label{fig:causalGAN_generator}
    \end{subfigure}
    \caption{Illustrations of CausalGAN. (Source: \cite{kocaoglu2017causalgan})}
    \label{fig:causalGAN}
\end{figure}

Kocaoglu et al. \cite{kocaoglu2017causalgan} propose \textit{CausalGAN} (\autoref{fig:causalGAN_overall}), whose generator is structured according to a given causal graph, similar to a \ac{BN} (see \autoref{fig:causalGAN_generator}). Given binary labels and some real observations for the discriminator, the model can be conditioned to generate data, for example, face images, according to the labels (e.g., sex, hair color). First, the \textit{causal controller}, implemented as a Wasserstein GAN \cite{arjovsky2017wasserstein}, generates the labels. Then, the generator produces data according to the labels and noise. Finally, the generator competes with three adversaries to produce realistic samples (discriminator) with correct labels (labeler) while avoiding easy-to-label unrealistic image distributions (anti-labeler).

Zhang et al. \cite{zhang2019self} develop the self-attention \ac{GAN} (SAGAN) to combine the computational and statistical efficiency of a convolutional \ac{GAN} for class-conditional image generation with the long-range dependency modeling capability of self-attention mechanisms (see \autoref{sec:transformers}). SAGAN significantly outperforms previous state-of-the-art models \cite{odena2017conditional} by increasing the \ac{IS} from 36.8 to 52.52 and decreasing the \ac{FID} from 27.62 to 18.65.

Krishna et al. \cite{krishna2019medical} propose a conditional \ac{GAN} to merge the style of one \ac{CT} image with the content of another to solve the problem of scarcity and privacy issues in clinical training data. They introduce a convolutional encoder-decoder generator trained to map the style of a \ac{CT} image to the segmentation map of other images per organ influenced by random noise. The training requires only a small data set and combines style and content loss to guide the generator and the convolutional discriminator.

Alonso et al. \cite{alonso2019adversarial} extend a \ac{GAN} to generate images of handwritten words. They use a 4-layer bidirectional \ac{LSTM} to create an embedding of the target character sequence fed to the generator. Further, an auxiliary text recognition network consisting of a \ac{CNN} encoder and \ac{LSTM} decoder evaluates the generated image in addition to the discriminator to encourage faithful recreation of the target word. The resulting \ac{GAN} produces realistic images of French and Arabic words that improve text recognition results of a neural network when used as additional training data.

Brock et al. \cite{brock2019large} build upon SAGAN \cite{zhang2019self}, increase the batch size by factor 8 and the width of each layer by 50\%, doubling the number of parameters in the generator and the discriminator. Their \textit{BigGAN} achieves new state-of-the-art results in class-conditional image synthesis with high resolutions up to $512\times 512$ and a controllable trade-off between detail and variety.

Lu{\v{c}}i{\'c} et al. \cite{luvcic2019high} aim to train the BigGAN \cite{brock2019large} with unlabeled data for conditional image generation. They experiment with unsupervised clustering and linear classifiers (pre-trained or co-trained with the discriminator) on top of representations of the images to substitute large parts of the labels (up to 90\%), achieving similar or better inception and \ac{FID} scores compared to the normal BigGAN.

Feng et al. \cite{feng2020generative} introduce CA-GAN, a symmetric and convolutional \ac{GAN} that implements collaborative learning between the generator and discriminator to provide real sample information to the generator and an attention mechanism for the generator to remove spurious features. The generator and discriminator interact at multiple convolution steps with the attention mechanism. CA-GAN is used to generate label-conditional high-quality \acp{HSI}, which are pictures with multiple layers of features per pixel, with the help of a limited sample set and classify them.

\othertab{conditional \acp{GAN}}{
    \cite{mirza2014conditional} & First introduction of conditional \acp{GAN} on MNIST digit and image tags generation. & 2014 \\\hline 
    \cite{gauthier2014conditional} & Conditional face image generation with the conditional \ac{GAN} \cite{mirza2014conditional} architecture with deconvolutional generator and convolutional discriminator. The conditional vector allows age specification and other attributes found in the training data. To avoid the reproduction of the training data by the generator, the conditional vectors are not directly used but are sampled from a kernel density estimate of the training data. & 2014 \\\hline 
    \cite{springenberg2015unsupervised} & Utilizing an objective function balancing mutual information between observed examples and predicted class distributions against the robustness to an \ac{GAN}. This approach extends the regularized information maximization to robust classification against an optimal adversary. The study includes empirical evaluations of synthetic data and image classification tasks, demonstrating the robustness of learned classifiers and the fidelity of samples generated by the adversarial generator. & 2015 \\\hline
    \cite{perarnau2016invertible} & Invertible conditional GAN (IcGAN): Using two encoders to create an inverse mapping from images to a latent representation $z$ and conditional attributes, allowing modifications of images with a conditional \ac{DCGAN} \cite{radford2015unsupervised} whose generator also functions as the decoder. The model is evaluated with different configurations, enabling realistic and complex image modifications. & 2016 \\\hline 
    \cite{karacan2016learning} & Attribute-Layout Conditioned GAN (AL-CGAN): Generating images of outdoor scenes from semantic layouts and scene attributes (e.g., weather) with a \ac{DCGAN} \cite{radford2015unsupervised}. & 2016 \\\hline 
    \cite{vondrick2016generating} & GAN for video (VGAN): A spatio-temporal \ac{DCGAN} \cite{radford2015unsupervised} architecture for unconditional and conditional video generation (e.g., future prediction from static images) with moving objects and a static camera. Generates one-second-long videos that resemble real videos with separate foreground and background generators. & 2016 \\\hline 
    \cite{shrivastava2017learning} & SimGAN: Use a \ac{GAN} to refine synthetic data from a simulator and make it realistic while preserving the simulator's annotations. An additional loss is employed to minimize the applied per-pixel difference of the \ac{CNN} refiner. & 2017 \\\hline 
    \cite{ehsani2017segan} & SeGAN: Segmentation and painting of occluded parts of an image. A segmentation \ac{CNN} creates the complete segment mask from a partial mask obtained by another pre-trained model \cite{he2016deep} and the RGB image. A conditional \ac{GAN} (similar to \cite{isola2017image}) then paints the invisible parts of the object based on the masks and the image. & 2017 \\\hline 
    \cite{bousmalis2017unsupervised} & Training a \ac{GAN} generator to learn a pixel-level image transformation from one domain to another (e.g., depth for an RGB image) using convolutions and fully connected layers. The discriminator is structured similarly and outputs the probability that an image was sampled from the target domain. For some tasks, the generator is further conditioned using a content similarity loss that penalizes large differences between the input and output of the generator. & 2017 \\\hline 
    \cite{xue2017segan} & SegAN: Training a \ac{CNN} to produce segmentation masks for medical images (\ac{CT} images) and an adversarial critic network compares the generated mask to the ground truth by evaluating features extracted with a \ac{CNN}. & 2017 \\\hline 
    \cite{yi2017dualgan} & DualGAN: Training two convolutional \acp{GAN} \cite{isola2017image} to translate images between two domains in both directions from unpaired data. Also, since both directions' models are available, a reconstruction loss is employed to improve training. Instead of a dedicated noise vector $z$, dropout is used in multiple layers. & 2017 \\\hline 
    \cite{volkhonskiy2017steganographic} & Steganographic \ac{GAN}: Using a \ac{DCGAN} \cite{radford2015unsupervised} generator to encode encrypted messages in natural-looking images, a discriminator to ensure data realism, and a steganalyzer network to retrieve the message using a shared key. & 2017 \\\hline 
    \cite{ledig2017photo} & SRGAN: Generator and discriminator consisting of multiple convolutional layers generate super-resolution (factor $4\times$) images. The \ac{GAN} is additionally trained on a perceptual loss obtained from feature maps of a VGG network \cite{simonyan2014very}, a deep \ac{CNN} for image classification. & 2017 \\\hline 
    \cite{hsu2017voice} & Variational autoencoding Wasserstein GAN (VAW-GAN): Voice conversion with a conditional \ac{VAE} encoding phonetic content from speech parameters (spectral frames) and decoding it depending on speaker identity. The \ac{VAE} decoder is then treated as the generator of a Wasserstein GAN \cite{arjovsky2017wasserstein} and trained to generate clearer speech. & 2017 \\\hline 
    \cite{li2017triple} & Triple-GAN: Jointly train a generator, discriminator, and classifier for semi-supervised learning, resulting in state-of-the-art classification results and smooth class-conditional generation and interpolation in the latent space. & 2017 \\\hline 
    \cite{esteban2017real} & Recurrent \ac{GAN} (RGAN) and Recurrent Conditional \ac{GAN} (RCGAN) for real-valued (medical) time series generation. The conditional version has additional inputs besides noise and data for the generator and discriminator \acp{LSTM}, respectively. The models are evaluated using an evaluation scheme, trained on synthetic data, and evaluated on real data. & 2017 \\\hline 
    \cite{ghosh2017multi} & Multi-Agent Diverse GAN (MAD-GAN): Using multiple generators and one discriminator that also has to identify the generator that created the sample to generate images with different classes in a conditional and unconditional setting. The discriminator pushes certain generators to different classes (modes) and improves the overall performance. & 2017 \\\hline 
    \cite{gorijala2017image} & Variational InfoGAN (ViGAN): Combining \acp{VAE} with InfoGAN \cite{chen2016infogan}, using the \ac{VAE} encoder to generate a latent representation $z$ of an observation that combined with a set of interpretable variables $c$ allows to generate/decode modified images $\tilde{x}\sim P(x\vert z,c)$. These images are then evaluated by a discriminator and a recognizer that reconstructs $c$. & 2017 \\\hline 
    \cite{odena2017conditional} & Auxiliary Classifier GAN (AC-GAN): A class-conditional deconvolutional generator produces images, and the convolutional discriminator outputs the real/fake probabilities and a probability distribution over all classes for input images. The objective function encourages both models to maximize the class likelihood, while the generator aims to minimize the correct real/fake judgments that the discriminator tries to maximize. The model works better on higher-resolution images and achieves a state-of-the-art \ac{IS} on CIFAR-10. & 2017 \\\hline 
    \cite{dai2017scan} & Structure Correcting Adversarial Network (SCAN): A \ac{CNN} generator learns segmentation masks of chest x-rays, and a critic network (\ac{CNN} and fully connected network for classification) learns to discriminate between real and generated masks for an image. To mitigate mode collapse, the generator is pre-trained with pixel-wise loss. & 2017 \\\hline 
    \cite{dong2017semantic} & Image manipulation with text descriptions using a generator that consists of a \ac{CNN} image encoder, a pre-trained \ac{RNN} text encoder, a residual transformation unit that jointly encodes text and image embeddings further and a decoder that reconstructs the image with upscaling layers. A \ac{CNN} discriminator evaluates the probabilities of the generated image and the text description matching. & 2017 \\\hline 
    \cite{zhu2017unpaired} & CycleGAN: Unpaired image-to-image translation (i.e., two unrelated sets of images with no relation information provided) by learning mapping $G:X\rightarrow Y$ and inverse mapping $F:Y\rightarrow X$ and adding a cycle consistency loss $F(G(X))\approx X$ next to the adversarial loss $G(X)\approx Y$. The model is implemented as \acp{CNN} and used for style transfer, object transfiguration, season transfer, and photo enhancement. & 2017 \\\hline 
    \cite{zhu2017toward} & BicycleGAN: Image translation (e.g., night-to-day image conversion) with invertible latent codes to prevent many-to-one mappings (i.e., mode collapse). For that purpose, a generator/decoder is conditioned on an input image and a latent vector to produce a suitable output image, and a \ac{VAE} encoder is trained to map the output back to the same latent code and a predefined distribution. & 2017 \\\hline 
    \cite{liu2017unsupervised} & Unsupervised image-to-image translation with coupled \acp{GAN} utilizing a \textit{shared latent space assumption}. Two encoders $E_{1}$ and $E_{2}$ translate images from two domains to the same latent space $z$. The generators $G_{1}$ and $G_{2}$ serve as decoders of a \ac{VAE} and recreate images from $z$ suitable for the domain. Their performance is evaluated by discriminators $D_{1}$ and $D_{2}$ respectively. The high-level connection weights between the encoders and the generators are tied to account for the shared latent space assumption. & 2017 \\\hline 
    \cite{lu2017conditional} & Conditional CycleGAN: Extending CycleGAN \cite{zhu2017unpaired} with an additional attribute condition vector (e.g., hair color, gender, smiling) for face image super-resolution and attribute-conditional translation. & 2017 \\\hline 
    \cite{grinblat2017class} & Showcasing that \acp{GAN} can produce higher quality samples by using a conditional setup, more precisely, when class labels are provided. The approach proposed augments class labels by clustering the representation space learned by the model itself. The method is, however, based on the more computationally costly Wasserstein \ac{GAN} approach. & 2017 \\\hline
    \cite{zhang2018synthetic} & Using the \textit{pix2pix} \cite{isola2017image} approach to generate infrared images and videos with tracking labels for entities from labeled RGB video frames. & 2018 \\\hline 
    \cite{wang2018perceptual} & Perceptual Adversarial Network (PAN): An image-to-image transformation \ac{CNN} $T$ learns mappings between domains and a discriminator \ac{CNN} $D$ tries to find discrepancies between transformed images and ground truth. & 2018 \\\hline 
    \cite{nam2018text} & Text-adaptive \ac{GAN} (TAGAN): A \ac{GAN} that allows the modification of images using natural language descriptions while preserving text-irrelevant features of the original image, for example, the form of an object. The generator is an encoder-decoder architecture derived from \cite{dong2017semantic} that combines the image embedding with the text representation obtained from a bidirectional \ac{GRU} working on pre-trained fastText \cite{bojanowski2017enriching} word vectors. The text-adaptive discriminator classifies each attribute independently using word-level discriminators. & 2018 \\\hline 
    \cite{wang2018video} & vid2vid: Video-to-video synthesis from segmentation masks or other sources to photorealistic output with \acp{GAN} and a spatiotemporal adversarial objective. A recursively applied feedforward neural network generates the next frame based on the current source frame and the past source and generated frames. A conditional image discriminator compares the current source and the generated image. Additionally, a conditional video discriminator estimates the optical flow of the generated sequence to ensure plausible temporal dynamics. & 2018 \\\hline 
    \cite{trigueros2018generating} & Training a progressive GAN \cite{karras2017progressive} to produce face images conditioned on the identity embedding of a person. The identity is built with discrete labels and converted to a continuous representation that follows a simple distribution. The model can be used for conditional and unconditional image generation and improves and alleviates the training of face recognition models by boosting the training data size with interpolated image reconstructions. & 2018 \\\hline 
    \cite{gecer2018semi} & Produce identity-preserving photorealistic face images from renderings of a 3D morphable model showing different poses, expressions, and illuminations with a conditional \ac{GAN}. The \ac{GAN} uses semi-supervised learning, so it trains with a few pairs of real and rendered face images and has a large amount of unpaired data available. & 2018 \\\hline 
    \cite{guo2018deep} & Graph-translation \ac{GAN} (GT-GAN): A \ac{GAN} consisting of a (de)convolutional graph translator with an encoder-decoder structure pitted against a conditional graph discriminator that takes the translated/real graph pairs as input and uses the same \ac{CNN} as the encoder to classify whether they are real or fake. & 2018 \\\hline 
    \cite{fedus2018maskgan} & MaskGAN: Training a \ac{GAN} for text generation. The generator consists of a \ac{LSTM} encoder that generates a context representation of text with masked tokens and a \ac{LSTM} decoder that takes the context and the masked text to autoregressively fill in the gaps and reconstruct the sequence probabilistically. The discriminator is a \ac{LSTM} similar to the decoder and outputs the probability of a token in the sequence to be real. On top of the discriminator, a critic network computes a reward function based on the predicted token likelihoods used to train the generator. The model is also used for unconditional text generation by masking the entire input. & 2018 \\\hline 
    \cite{shin2018medical} & Using the \textit{pix2pix} \cite{isola2017image} \ac{GAN} approach to generate images and segmentation maps of brains and tumors. & 2018 \\\hline 
    \cite{triastcyn2018generating} & Making the improved Wasserstein \ac{GAN} with gradient penalty \cite{gulrajani2017improved} differentially private by adding Gaussian noise to the activations of the second-to-last layer of the discriminator. The model is used to produce label-conditioned images. & 2018 \\\hline 
    \cite{engel2019gansynth} & GANsynth: High-fidelity and locally-coherent audio (music notes) synthesis using high-resolution frequency spectrograms, several orders of magnitude faster than autoregressive models. The generator is additionally conditioned on a one-hot pitch label that the discriminator tries to predict with an auxiliary classifier. & 2019 \\\hline 
    \cite{fan2019labeled} & Labeled-Graph \ac{GAN} (LGGAN): An improved Wasserstein \ac{GAN} \cite{wei2018improving} approach that uses a \ac{MLP} generator to produce adjacency and one-hot node label matrices for graphs and a graph convolutional network \cite{kipf2016semi} with residual connections \cite{he2016deep} as the discriminator. Both conditional and unconditional configurations are presented. & 2019 \\\hline 
    \cite{torkzadehmahani2019dp} & DP-CGAN: Differentially private data generation using a conditional \ac{GAN} that injects Gaussian noise and clips the discriminator gradients to limit the amount of information from the training data transferred to the generator. The model is trained until the privacy budget monitored by a Rényi differential privacy accountant \cite{mironov2017renyi} is spent. The model improves on differentially private results on MNIST over a simpler baseline conditional \ac{GAN} model. & 2019 \\\hline 
    \cite{zhou2019misc} & Misc-GAN: Translating graphs from a source to a target domain (e.g., for anonymization) by extracting coarse (partial) structures from real graphs using clustering methods, permutating them, generating new coarse graphs from these templates, and combining the generated coarse graphs to produce a full synthetic graph. & 2019 \\\hline 
    \cite{lin2019cocogan} & COCO-GAN: A conditional \ac{GAN} only trained using so-called micropatches of image datasets. The generator and discriminator learn only parts of the image via conditional formatting; however, they can generate full images during the inference phase.  & 2019 \\\hline
    \cite{xu2020synthesizing} & Conditional Tabular GAN (CTGAN): A \ac{GAN} to effectively generate tabular data with mixed discrete and continuous columns. The generator is conditioned on a one-hot vector determining a randomly selected category and samples a row (each column independently) from the learned marginal distributions. The critic then scores the generated sample against a random sample from the training data matching the same criterion. & 2020 \\\hline 
    \cite{armanious2020medgan} & MedGAN: Image-to-image translation in the medical domain using a \textit{CasNet} generator that progressively refines the image via encoder-decoder pairs. This feature-extracting discriminator evaluates the presence of desired modalities and non-adversarial losses that evaluate the generated image's style (e.g., structure and texture application). & 2020 \\\hline 
    \cite{maziarka2020mol} & Mol-CycleGAN: Using CycleGAN \cite{zhu2017unpaired} to enable the bidirectional translation of molecule embeddings between molecules with and without an optimized property. The GAN operates in the latent space of a junction tree \ac{VAE} \cite{jin2018junction}, simplifying training because the similarity measure used for this molecule representation is differentiable. & 2020 \\\hline 
    \cite{rashidian2020smooth} & SMOOTH-GAN: \ac{EHR} generation conditioned on diagnosis codes (binary vector of present diseases) built upon the conditional GAN \cite{mirza2014conditional} with the Wasserstein GAN gradient penalty loss \cite{arjovsky2017wasserstein,gulrajani2017improved}. & 2020 \\\hline 
    \cite{sarp2021wg2an} & WG$^{2}$GAN: Image-to-image translation to generate wounds from segmentation maps with a conditional \ac{GAN}. & 2021 \\\hline 
    \cite{amin2021quantum} & Using a modified \ac{DCGAN} \cite{radford2015unsupervised} to generate lung \ac{CT} images for COVID-19 classification. & 2021 \\ 
}

\subsubsection{Deep/Stacked GANs}

Deep \acp{GAN} contain multiple layers of generators and discriminators or share partial results (e.g., images at different resolutions and the respective discriminative feedback) at specific steps with each other. This allows the discriminator(s) to evaluate the final product, assess higher-order features, and provide more detailed feedback to the generator(s), often resulting in better results.

Denton et al. \cite{denton2015deep} introduce LAPGAN, a Laplacian pyramid framework having conditional \acp{CNN} trained with \acp{GAN} at each layer to generate coarse-to-fine images. For this purpose, the generator upsamples the image from the previous stage to double the width and height and then applies a convolution based on the upscaled image and random noise. During the training process, a high-resolution image $I$ is downscaled, blurred, and upscaled again. This modified version $l$ is then used to either generate a real high-pass image $h=I-l$ or synthetic sample $G(z,l)$ that is given to the discriminator together with $l$ to determine whether it is real or generated. Log-likelihood evaluation with Parzen window estimates and human evaluation show that LAPGAN produces more realistic results than a standard \ac{GAN}.

Zhang et al. \cite{zhang2017stackgan} propose \textit{StackGAN}, where the text-to-image task is divided into two stages. The first stage \ac{GAN} draws the shape and color of the object described by the text. In stage two, a second \ac{GAN} refines the sketch given the text description to produce photo-realistic images. StackGAN outperforms previous state-of-the-art conditional \ac{GAN} models GAN-INT-CLS \cite{reed2016generative} and GAWWN \cite{reed2016learning} in terms of resolution ($256\times256$ pixels) and realism.

\othertab{deep/stacked \acp{GAN}}{
    \cite{huang2017stacked} & Stacked \acp{GAN}: A stack of symmetric encoder-decoder layers with layer-wise representation pair discriminators and a Q-Net that evaluates the diversity of the output of the generator/decoder at each layer. Each generator layer has its independent noise input and previous layer input, which allows conditioning on the output of the encoder. The resulting label-conditioned images are of higher quality than compared shallow \acp{GAN}. & 2017 \\\hline 
    \cite{guibas2017synthetic} & A two-stage \ac{GAN} approach for retina images. In the first stage \ac{GAN}, vessel tree segmentation masks are generated with the help of a small human data set. In stage two, a conditional \ac{DCGAN} \cite{radford2015unsupervised} learns to create the corresponding retina fundus image for the vessel tree. & 2017 \\\hline 
    \cite{trans-gan} & TransGAN: A deep \ac{GAN} consisting of a purely transformer-based architecture, excluding conventional convolutional layers. TransGAN consists of a memory-friendly transformer-based generator that progressively increases feature resolution and a multi-scale discriminator that captures both semantic contexts and low-level textures. The study includes a new grid self-attention module for high-resolution image generation and a unique training procedure to address the instability issues associated with TransGAN. & 2021 \\\hline
}

\subsubsection{Bidirectional GANs}\label{sec:bi_gans}

\begin{figure} [ht]
    \centering
    \includegraphics[width=0.6\textwidth]{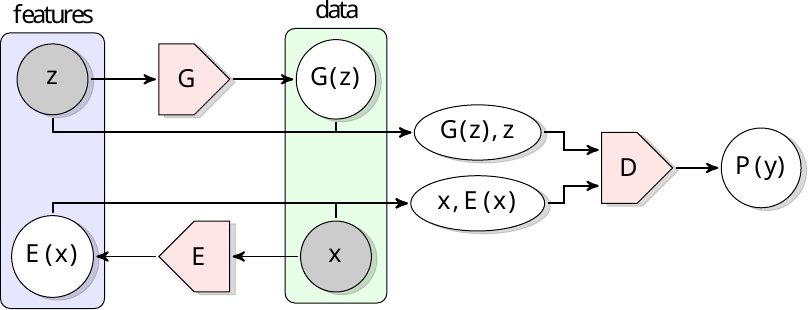}
    \caption{The \ac{BiGAN} architecture. (Source: \cite{donahue2016adversarial})}
    \label{fig:bigan}
\end{figure}

A \ac{BiGAN} \cite{donahue2016adversarial}, also introduced as \ac{ALI} \cite{dumoulin2016adversarially}, lets the discriminator evaluate data-representation pairs $(x,z)$ of either a generator $x=G(z)$ creating data from a representation or an encoder $z=E(x)$ inversely generating representations for data and decide, whether $x$ is real or fake (see \autoref{fig:bigan}). The result is a model that learns meaningful data representations for detection or classification tasks despite the encoder and generator being unable to communicate. It can also be used similarly to an autoencoder $\tilde{x}=G(E(x))$ for generative tasks, for example, image generation. \cite{donahue2016adversarial}

Donahue et al. \cite{donahue2019large} propose \textit{BigBiGAN}, which combines the BigGAN \cite{brock2019large} with the encoder of a BiGAN. They further modify the BigGAN discriminator to compute a score for $x$ and $z$, respectively, and a joint score. BigBiGAN achieves state-of-the-art results in unsupervised representation learning and unconditional image generation tasks, outperforming BiGAN and BigGAN, respectively.

\subsubsection{Adversarial Autoencoders}

\begin{figure} [ht]
    \centering
    \includegraphics[width=0.6\textwidth]{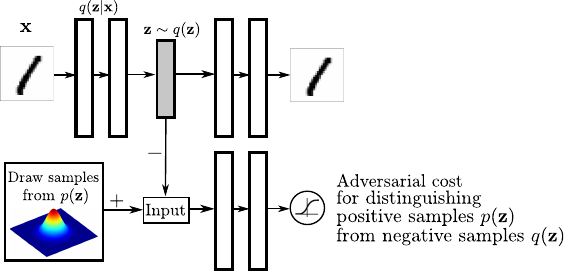}
    \caption{Architecture of an \acl{AAE}. (Source: \cite{makhzani2015adversarial})}
    \label{fig:aae}
\end{figure}

\begin{figure} [ht]
    \centering
    \includegraphics[width=0.8\textwidth]{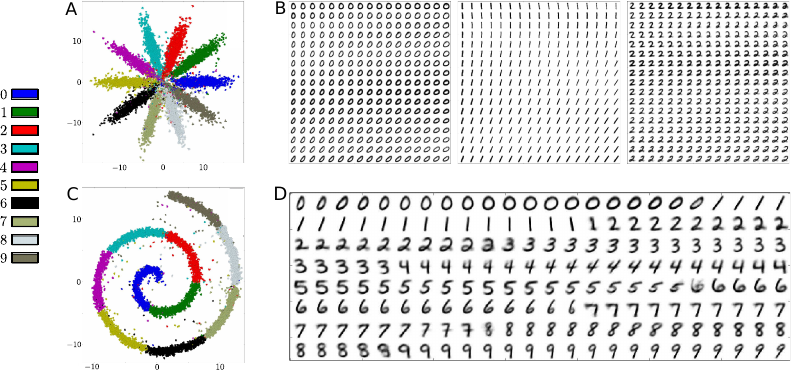}
    \caption{By forcing the latent space distribution of an \acl{AAE} into specific regions, for example, to separate different numbers of MNIST, the quality and interpretability of representations can be greatly improved. (Source: \cite{makhzani2015adversarial})}
    \label{fig:aae_clustering}
\end{figure}

The \acf{AAE} is a probabilistic autoencoder whose latent representations $\mathbf{z}\sim q(\mathbf{z})$ of the encoded data are forwarded to a discriminator who tries to distinguish the samples $q(\mathbf{z})$ from ones of a prior user-defined distribution $p(\mathbf{z})$ and provides an adversarial loss, as seen in \autoref{fig:aae}. By using the variational inference technique similar to a \ac{VAE} (see \autoref{sec:vae}), the autoencoder learns to map data to and generate meaningful samples from the whole space defined by the prior $p(\mathbf{z})$, which is configurable by the user and allows learning of powerful representations (see \autoref{fig:aae_clustering} for example) for classification, disentangling of style and content, unsupervised clustering or data visualization. \cite{makhzani2015adversarial}

Makhzani et al. \cite{makhzani2015adversarial} demonstrate the generative capabilities of the \ac{AAE} on MNIST, \ac{SVHN} and \ac{TFD}, achieving state-of-the-art log-likelihood on the test data compared to the standard \ac{GAN} \cite{goodfellow2014generative}, \ac{GMMN} \cite{li2015generative}, \ac{DBN} \cite{hinton2006fast} and \ac{GSN} \cite{bengio2014deep} and allowing the combination of disentangled styles (writing style or font) and contents (numbers) on MNIST and \ac{TFD}.

Tolstikhin et al. \cite{tolstikhin2017wasserstein} propose the \acf{WAE} as a generalization of the \ac{AAE}, inspired by the Wasserstein GAN \cite{arjovsky2017wasserstein} adversary that uses the probability distribution discrepancy between real and generated data as the adversarial loss. Two approaches to compute the discrepancy between $q$ and $p$ are presented: The \ac{WAE}-GAN incorporates an adversary that approximates the Jensen-Shannon (JS) divergence as loss and is trained together with the autoencoder, while the WAE-MMD used an adversary-free \ac{MMD}-based loss, similar to \acp{GMMN} (see \autoref{sec:gmmns}). WAE-GAN is less stable but generates higher quality samples than WAE-MMD and a normal \ac{VAE}.

\othertab{\acp{AAE}}{
    \cite{kadurin2017cornucopia} & An \ac{AAE} is used to generate suitable molecule fingerprints for cancer treatment (trained on 6252 samples), and these probabilistic fingerprints are used to search for similar real molecules in the PubChem database with 72 million entries. & 2017 \\\hline 
    \cite{kadurin2017drugan} & druGAN: An improved \ac{AAE} trained on larger data sets than Kadurin et al.~\cite{kadurin2017cornucopia} that demonstrates superior reconstruction results over a comparable \ac{VAE} with the same sampling variability (i.e., coverage), which they identify is a trade-off necessary for generative autoencoders. & 2017 \\\hline 
    \cite{costa2017end} & End-to-end retinal image generation with an \ac{AAE} generating the vessel network and a conditional \ac{GAN} creating the corresponding fundus image. The \ac{GAN} discriminator then evaluates the vessel-fundus pairs in terms of realism. & 2017 \\\hline 
    \cite{ae-breathing-gen} & An \ac{AAE} with an additional classification step for biomedical time series generation. This allows the authors to generate labeled time series data using a semi-supervised approach. A dimensionality reduction is further introduced, transforming three-dimensional biomedical breathing data into one-dimensional time series and back. & 2021 \\
}

\subsection{Generative Moment Matching Networks}\label{sec:gmmns}

\acfp{GMMN} use a neural network to learn deterministic mappings from an easy-to-sample data distribution to the real data distribution, similar to a \ac{GAN} generator. The model starts with a top hidden layer $\mathbf{h}\in\mathbb{R}^{H}$ whose elements are usually independently sampled from a uniform distribution so that $p(\mathbf{h})=\prod_{j=1}^{H}U(h_{j})$. Then, $\mathbf{h}$ is passed through multiple neural network layers until the final output is returned as a data sample. \cite{li2015generative}

The \ac{GMMN} is trained, unlike a \ac{GAN} generator, minimizing the \ac{MMD} criterion instead of adversarial training with a discriminator. The \ac{MMD} criterion is defined as

\begin{equation}
    \mathcal{L}_{MMD^{2}}=\frac{1}{N^{2}}\sum_{i=1}^{N}\sum_{i'=1}^{N}k(x_{i},x_{i'})-\frac{2}{NM}\sum_{i=1}^{N}\sum_{j=1}^{M}k(x_{i},y_{j})+\frac{1}{M^{2}}\sum_{j=1}^{M}\sum_{j'=1}^{M}k(y_{j},k_{j'})
\end{equation}

and computes, whether the generating distributions for two sets of samples $X=\{x_{i}\}_{i=1}^{N}$ and $Y=\{y_{j}\}_{j=1}^{M}$ are the same using a Gaussian kernel $k(x,x')=\exp(-\frac{1}{2\sigma}\vert x-x'\vert^{2})$ with $\sigma$ being the bandwidth parameter. The gradient is differentiable; that is, the gradient can be backpropagated to update the parameters of the neural network. The larger the dimensionality of the data, the more training data is required to estimate the data distribution. \cite{li2015generative}

Li et al. \cite{li2015generative} use a \ac{GMMN} with four intermediate non-linear \ac{ReLU} layers and a logistic sigmoid output layer to model the MNIST and \ac{TFD} data distribution. Further, they improve their generative model by training their \ac{GMMN} on top of the latent representation of an autoencoder to reduce the data dimensionality and, consequently, the amount of training data needed.

Li et al. \cite{li2017mmd} propose \textit{MMD GAN}, which replaces the Gaussian kernel of a \ac{GMMN} with an adversarial kernel learning technique to reduce the amount of required training data and improve the model accuracy and efficiency. The Gaussian kernel is modified with injective functions $f_{\phi}$ with optimizable parameters $\phi$, resulting in the new kernel function $\tilde{k}(x,x')=\exp(-\Vert f_{\phi}(x)-f_{\phi}(x')\Vert^{2})$. The \ac{GMMN} model (generator) parameters $\theta$ and \ac{MMD} (discriminator) kernel parameters $\phi$ are then adversarially optimized in a minimax game $\min_{\theta}\max_{\phi}\mathcal{L}_{\phi}(X,Y_{\theta})$, like in a \ac{GAN}, minimizing the generator loss for the worst possible discriminator result.

\othertab{\acp{GMMN}}{
    \cite{dziugaite2015training} & MMD nets: Proposal of training/initializing generative neural networks with the inexpensive \ac{MMD} statistic instead of a \ac{GAN} discriminator at the same time as Li et al.~\cite{li2015generative}. & 2015 \\\hline 
    \cite{ren2016conditional} & Conditional \ac{GMMN}: Feeding a sample drawn from a simple distribution and the conditional variables to the network that generates the target sample. A conditional \ac{MMD} criterion is developed to learn the parameters. The model is used for conditional generation of MNIST and face images, combined with an autoencoder like Li et al.~\cite{li2015generative}. & 2016 \\\hline 
    \cite{takamichi2017sampling} & Speech synthesis by sampling speech parameters from a \ac{GMMN}-learned distribution to induce variation in the generated speech and make it more natural. & 2017 \\\hline 
    \cite{wang2018improving} & Introducing a repulsive loss function to address the limitations of existing \ac{MMD} loss functions that may hinder learning fine details in data. The repulsive loss function enhances learning by emphasizing differences among real data samples. Additionally, the study proposes a bounded Gaussian kernel to stabilize \ac{MMD}-\ac{GAN} training. The methods are applied to unsupervised image generation tasks on datasets, showing significant improvements over the traditional \ac{MMD} loss without additional computational costs. The paper also explores regularization techniques for \ac{MMD} and the discriminator, contributing to the stability and effectiveness of the training process. & 2018 \\\hline
    \cite{liao2022scenario} & Generation of high-dimensional load curves (cooling, heating, power) of integrated energy systems with a \ac{CNN} generator and transformation of real and generated samples to latent space with an autoencoder where the distributions are compared using an \ac{MMD} loss. & 2022 \\ 
}

\subsection{Plug \& Play Generative Networks}

\acp{PPGN} consist of a pre-trained generator network $G$ with latent variable space, usually obtained from a \ac{GAN}, and a replaceable (``plug and play'') pre-trained condition network $C$, which can be a classifier or image captioning network for example and ``tells the generator what to draw''. The class probabilities of the classifier are used to perform gradient ascent on the latent space of the generator, iteratively improving the generated results. $G$ and $C$ can even be trained on different data sets or domains, allowing for at least a limited form of domain transfer. \cite{nguyen2016synthesizing,nguyen2017plug}

\begin{figure} [ht]
    \centering
    \includegraphics[width=0.8\textwidth]{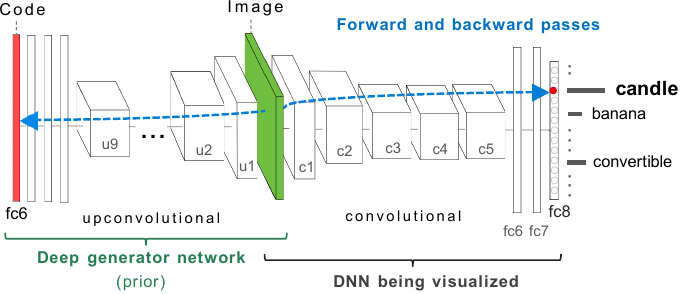}
    \caption{Example of the first \ac{PPGN}, the DGN-AM. (Source: \cite{nguyen2016synthesizing})}
    \label{fig:ppgn_dgn_am}
\end{figure}

Nguyen et al. \cite{nguyen2016synthesizing} introduce the deep generator network with activation maximization, short DGN-AM (see \autoref{fig:ppgn_dgn_am}), which is the origin of \acp{PPGN}. It aims to synthesize data from a pre-trained generator, which maximizes the activation of a specific neuron of a classifier. The architecture produces images of state-of-the-art quality but little diversity \cite{nguyen2017plug}.

In another work \cite{nguyen2017plug}, Nguyen et al. improve upon the lack of diversity of DGN-AM by learning a prior distribution for the latent variable of the generator using a \ac{DAE}, which serves as an additional optimization criterion. They further define the generalized class of \acp{PPGN} and experiment with other building blocks like image captioning networks for the condition network. They achieve higher sample quality and diversity than DGN-AM.

\subsection{Copulas}

A copula $C$ models and decomposes the joint probability distribution of a continuous random vector $\mathbf{X}$ into a product of the marginal distributions and a representation of the marginal variables' dependence structure. \cite{tagasovska2019copulas}

The creation of a copula model relies heavily on estimation: First, the marginal cumulative distribution functions $F_{i}$ for individual random variables $\mathbf{X}_{i}$ are approximated with the help of training data and converted to a uniform distribution using probability integral transform. Then, the variables' dependencies and correlations can be estimated in various ways, for example, nonparametrically via kernel estimation \cite{kulkarni2018generative,tagasovska2019copulas} or decomposition into trees of pair-copulas called \textit{vines} \cite{tagasovska2019copulas}.

Li et al. \cite{li2014differentially} introduce \textit{DPCopula}, a technique to synthesize differentially private multi-dimensional data with copula functions efficiently. They propose two metrics, maximum likelihood estimation, and Kendall's $\tau$ correlation, to estimate the parameters of the Gaussian copula function. Both methods are analyzed in terms of privacy guarantees and computational complexity on census data sets, outperforming previous state-of-the-art approaches like \ac{PSD} and P-HP (lossy compression), especially on high-dimensional data sets.

Kulkarni et al. \cite{kulkarni2018generative} use vine copulas to iteratively generate mobility trajectories, which are defined as a temporally ordered sequence $T=\langle(l_{1},t_{1}),...,(l_{n},t_{n})\rangle$ with locations $l_{i}$ (cell IDs) and timestamps $t_{i}$. They find that the copulas model observed statistical and semantic/geographic similarities particularly well at a fraction of the computational cost of neural network approaches while also incorporating long-range dependencies. Less accurate models like \acp{RNN} and \acp{GAN} perform better in their privacy tests.

Tagaskova et al. \cite{tagasovska2019copulas} propose \ac{VCAE}, which utilize a \ac{CNN} autoencoder based on \ac{DCGAN} \cite{radford2015unsupervised} to extract lower-dimensional representations from data and fit a nonparametric vine copula to learn the representations' distribution. By sampling from the trained copula, the decoder of the autoencoder can be used as a generative model. The \ac{VCAE} is tested on three real-world image data sets in terms of \ac{MMD} score and \ac{C2ST} accuracy: MNIST, \ac{SVHN}, and CelebA. \ac{VCAE} performs similar to \acs{DCGAN} and outperforms the \acs{VAE}.

\subsection{Normalizing Flow Models}\label{sec:normalizing_flows}

Normalizing flows are deterministic invertible transformations $f:\mathcal{E}\rightarrow\mathcal{Z}$ with parameters $\theta$ between a base distribution $\mathcal{E}$ (e.g., Gaussian distribution) and observational space $\mathcal{Z}$. The transformation can be used to calculate the exact density (probability) of an observation by using $f^{-1}:\mathcal{Z}\rightarrow\mathcal{E}$ or sample new observations by sampling from the simpler distribution $\epsilon\in\mathcal{E}$ and transforming it to observation space $z\in\mathcal{Z}$. \cite{shi2020graphaf}

\acfp{AF} are a variant of normalizing flows that models an observation $z\in\mathbb{R}^{D}$ as

\begin{equation}\label{eq:af_prob}
    p(z_{d}\vert z_{1:d-1})=\mathcal{N}(z_{d}\vert\mu_{d},(\alpha_{d})^{2})\mbox{, with }\mu_{d}=g_{\mu}(z_{1:d-1};\theta), \alpha_{d}=g_{\alpha}(z_{1:d-1};\theta),
\end{equation}

where $g_{\mu}$ and $g_{\alpha}$ are unconstrained positive scalar functions, usually implemented as neural networks, that compute the mean and deviation for a normal distribution $\mathcal{N}$. The transformations are defined as

\begin{equation}\label{eq:af_transformation}
    f_{\theta}(\epsilon_{d})=z_{d}=\mu_{d}+\alpha_{d}\cdot\epsilon_{d}\mbox{;   }f^{-1}_{\theta}(z_{d})=\epsilon_{d}=\frac{z_{d}-\mu_{d}}{\alpha_{d}}.
\end{equation}

To sample $z$ from an \ac{AF}, first $\epsilon\in\mathbb{R}^{D}$ is sampled, then $z_{1}$ is computed using \autoref{eq:af_transformation}. Finally, each subsequent $z_{d}$ can be computed using \autoref{eq:af_prob}. \cite{shi2020graphaf}

Shi et al. \cite{shi2020graphaf} propose \textit{GraphAF} for graph-based molecule generation with parallel training, which uses \acp{AF} to generate graphs sequentially. At each step, the type of the next node (made continuous with a dequantization technique) is predicted before all edges to existing nodes are determined. The node and edge distribution parameters respectively $\mu^{X}_{i}$ and $\alpha^{X}_{i}$ and $\mu^{A}_{i}$ and $\alpha^{A}_{i}$ are computed with \acp{MLP} that are conditioned on the node embeddings $H_{i}$ that are obtained from a relational graph convolutional network \cite{schlichtkrull2018modeling}. The authors additionally employ \textit{valency checking} \cite{popova2019molecularrnn} to reject invalid edges and implement a \ac{RL} approach for goal-directed molecule generation.

\othertab{normalizing flow models}{
    \cite{dinh2014nice} & Non-linear independent components estimation (NICE): Learning of a non-linear deterministic and easily invertible transformation (deep neural network) $f$ that maps an input $x$ to a hidden representation $h=f(x)$ and back, so $x=f^{-1}(h)$. The training criterion is the exact log-likelihood. Since the transformations are deterministic, noise is injected at $h$ for generation. Results in image generation achieve high log-likelihood but often do not look realistic. & 2014 \\\hline 
    \cite{dinh2016density} & Real-valued non-volume preserving (real NVP) transformations: Efficient invertible mapping of images to latent variables. & 2016 \\\hline 
    \cite{kingma2018glow} & Glow: Using an invertible $1\times 1$ convolution to generate realistic and large images. & 2018 \\\hline 
    \cite{chen2018neural} & Neural \acf{ODE}: A continuous-time mapping $z(t)$ from latent variables to data defined by \acp{ODE}, also called continuous normal flow \cite{grathwohl2018ffjord}. The model is used to accurately model and extrapolate time series or replace discrete hidden layers of a neural network. & 2018 \\\hline 
    \cite{grathwohl2018ffjord} & FFJORD: Combining the continuous normal flow procedure from \cite{chen2018neural} with an estimator of the log density for training instead of using maximum likelihood. This significantly reduces the computational cost and allows unrestricted architectures. The model outperforms previous flow-based methods on density estimation and also image generation. & 2018 \\\hline 
    \cite{madhawa2019graphnvp} & GraphNVP: The first flow-based invertible graph generation model that can handle fixed-size node type assignments and adjacency matrices together. The model maps the node feature and annotation matrices to latent representations, which can also be randomly sampled for generative purposes. They also train a simple linear regressor on the latent space for property-targeted molecule generation. & 2019 \\\hline
    \cite{fourier-flows} & Fourier Flows: This approach operates in the frequency domain, using discrete Fourier transformation to handle variable-length time-series with varying sampling rates and leveraging the more computationally efficient convolutions in the frequency domain. Fourier Flows applies data-dependent spectral filters to the frequency-transformed data, enabling efficient Jacobian determinants and inverse mapping computation. This method shows competitive performance compared to state-of-the-art models.  & 2021 \\
}

\subsection{Reinforcement Learning}\label{sec:reinforcement_learning}

\begin{figure} [ht]
    \centering
    \includegraphics[width=0.4\textwidth]{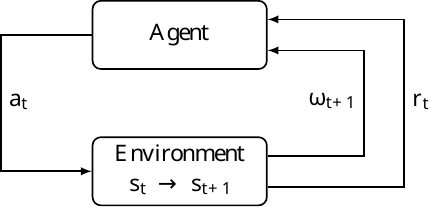}
    \caption{Agent-environment interaction in \ac{RL}. (Source: \cite{franccois2018introduction})}
    \label{fig:reinforcement_learning_agent}
\end{figure}

In \acf{RL}, an agent starts in a state $s_{0}\in\mathcal{S}$ within its environment and obtains an initial observation $\omega_{0}\in\Omega$. The agent then has to decide on an action $a_{t}\in\mathcal{A}$ at each time step $t$. After performing the action, the agent receives a reward $r_{t}\in\mathcal{R}$, the state transitions to $s_{t+1}\in\mathcal{S}$ and the agent gets a new observation $\omega_{t+1}\in\Omega$, as seen in \autoref{fig:reinforcement_learning_agent}. The goal of \ac{RL} is to learn and optimize a policy $\pi$ so that the actions taken maximize the cumulative reward. Therefore, the agent uses a value function $V$ to predict the reward for an action. The policy can be \textit{deterministic}, so it can be defined as $\pi(s):\mathcal{S}\rightarrow\mathcal{A}$, or \textit{stochastic}, assigning a probability $\pi(s,a):\mathcal{S}\times\mathcal{A}\rightarrow[0,1]$ to each action. A significant advantage of \ac{RL} is that the agent does not need complete knowledge or control of the environment, which often makes it more computationally efficient than classic supervised and unsupervised \ac{ML} methods. \cite{franccois2018introduction} 

Oh et al. \cite{oh2015action} predict the next frames of Atari video games conditioned on previous frames and player actions. Input frames are encoded with a \ac{CNN} (and optionally a \ac{RNN}) to extract spatio-temporal features and then combined with a 1-hot encoded action variable in a transformation layer to obtain a high-level prediction of the next frame. A \ac{CNN} takes this prediction and uses upsampling to generate a full-size frame. The model is trained on emulator recordings with the corresponding user inputs using stochastic gradient descent with backpropagation through time. The trained model can generate future frames for arbitrary input sequences. The evaluation shows realistic 100-step future frames for a variety of Atari games.

Jia et al. \cite{jia2019paintbot} train a painting agent's policy with \ac{RL} to brush strokes step-by-step on a canvas guided by a reference image. The painting process of \textit{Paintbot} consists of multiple steps:

\begin{enumerate}
    \item A random stroke starting point $p$ on the canvas is selected.
    \item Image patches centered around $p$ from the reference image and the canvas act as the observation $o$.
    \item While the predicted reward $V_{\pi}(o)$ is positive, actions (strokes) are performed, consisting of continuous values for angle, length, color, and width, the canvas is updated, and $p$ and $o$ are both updated to the new position.
\end{enumerate}

\begin{figure} [ht]
    \centering
    \begin{subfigure}[t]{0.45\textwidth}
        \centering
        \includegraphics[width=0.9\textwidth]{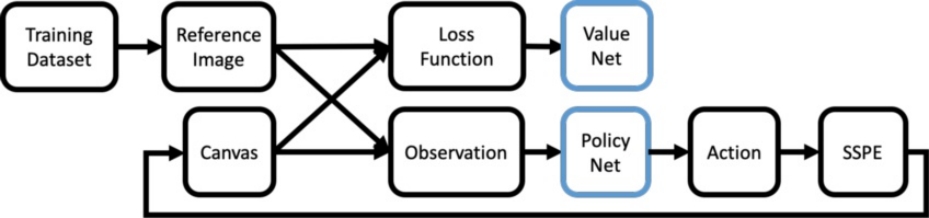}
        \caption{During training.}
    \end{subfigure}
    \begin{subfigure}[t]{0.4\textwidth}
        \centering
        \includegraphics[width=0.9\textwidth]{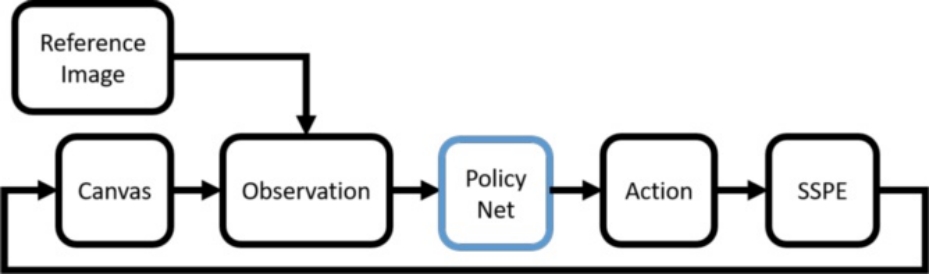}
        \caption{After training.}
    \end{subfigure}
    \caption{The Paintbot painting process during and after training. (Source: \cite{jia2019paintbot})}
    \label{fig:paintbot_process}
\end{figure}

This process is repeated until a specified similarity threshold between the reference image and painting is reached. The training process consists of an additional loss function used to train the deep neural network for reward prediction $V$ (see \autoref{fig:paintbot_process}).

Krishna et al. \cite{krishna2021image} use a \ac{RL} approach to synthesize full-resolution \ac{CT} images. They create anatomically correct semantic masks and use their existing conditional \ac{GAN} style transfer network \cite{krishna2019medical} to fill the generated masked areas with correct textures. Semantic masks are represented as vectors via b-splines and principal component analysis for which the agent learns a policy. An image classifier \ac{CNN} is used as the reward predictor and trained with human feedback on the agent's generated masks through an interface.

\othertab{\ac{RL}}{
    \cite{li2016deep} & Dialogue text generation with two agents, which are encoder-decoder \acp{LSTM}. Agents are rewarded for displaying three conversational properties: Informativity, coherence, and ease of answering. & 2016 \\\hline 
    \cite{li2017adversarial} & Extending the idea from \cite{li2016deep} for dialogue generation, but using an adversarial discriminator to distinguish between human-generated and synthetic dialogues and using the probabilities put out by the discriminator as the reward function. & 2017 \\\hline 
    \cite{guimaraes2017objective} & ORGAN: Combination of a SeqGAN \cite{yu2017seqgan} with \ac{RL} where the \ac{GAN} generator is trained with a tunable reward function consisting of the discriminator classification result, a repetition penalty, and domain-specific objective functions. The network is evaluated on SMILES \cite{weininger1988smiles} molecule and musical melody generation. & 2017 \\\hline 
    \cite{olivecrona2017molecular} & Generation of SMILES \cite{weininger1988smiles} representations of molecules with specific properties using a prior \ac{RNN} trained on a SMILES database and a user-defined scoring function as rewards for an \ac{RL} agent \ac{RNN} initialized by the prior \ac{RNN}. & 2017 \\\hline 
    \cite{shi2018toward} & Application of inverse \ac{RL} to text generation, where a reward function is alternatingly learned on training data and an agent learns an optimal policy to maximize the total reward. In the implementation, a reward approximator \ac{MLP} aims to maximize the log-likelihood of the training set samples, and the text generator \ac{LSTM} is trained with a policy gradient \cite{williams1992simple} technique based on the reward and entropy regularization to encourage more diverse results. & 2018 \\\hline 
    \cite{you2018graph} & Goal-directed molecular graph generation with a graph convolutional policy network (GCPN): Molecule generation as a Markov decision process where the next graph state only depends on the previous one. At each step, a new subgraph, in this case, predefined as all single-node graphs of all allowed atom types, is connected to an existing node, or two existing nodes are connected by the GCPN. The GCPN is rewarded by a sum of domain-specific and adversarial rewards to ensure the molecule's utility, realism, and validity. & 2018 \\\hline 
    \cite{kumar2019polyphonic} & Amadeus: Train a \ac{LSTM} on a representation of multiple monophonic note streams that provide a polyphonic piece of music to simplify the learning process. Then \ac{RL} is applied to select high-level \ac{LSTM} configurations that produce the desired outputs instead of modifying weights or outputs. & 2019 \\ \hline
    \cite{ts-gen-contrastive-imitation} & Addressing the compounding errors in sequential generation by combining contrastive imitation and an energy model. The model aims to capture both the step-wise transitions and the overall trajectory distribution, balancing the local and global properties of time-series data. & 2021 \\
}

\subsection{Diffusion Models}\label{sec:diffusion_models}

\begin{figure} [ht]
    \centering
    \includegraphics[width=0.8\textwidth]{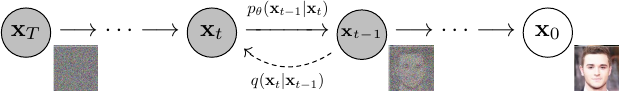}
    \caption{Diffusion forward and reverse processes. (Source: \cite{ho2020denoising})}
    \label{fig:diffusion_process}
\end{figure}

Diffusion models are Markov chains that iteratively add Gaussian noise to data $\mathbf{x}_{0}$ in a \textit{forward process} over $T$ steps and also learn the \textit{reverse process} that iteratively maps the noise input back to the data distribution (see \autoref{fig:diffusion_process}). \cite{ho2020denoising}

The forward process is defined as

\begin{equation}
    q(\mathbf{x}_{1:T}\vert\mathbf{x}_{0})=\prod_{t=1}^{T}q(\mathbf{x}_{t}\vert\mathbf{x}_{t-1})\mbox{,   }q(\mathbf{x}_{t}\vert\mathbf{x}_{t-1})=\mathcal{N}(\mathbf{x}_{t};\sqrt{1-\beta_{t}}\mathbf{x}_{t-1},\beta_{t}\mathbf{I}),
\end{equation}

where the variance schedule $\beta_{1},...,\beta_{T}$ can be constant or learned hyperparameters. The reverse process is defined as

\begin{equation}
    p_{\theta}(\mathbf{x}_{0:T})=p(\mathbf{x}_{T})\prod_{t=1}^{T}p_{\theta}(\mathbf{x}_{t-1}\vert\mathbf{x}_{t})\mbox{,   }p_{\theta}(\mathbf{x}_{t-1}\vert\mathbf{x}_{t})=\mathcal{N}(\mathbf{x}_{t-1};\mathbf{\mu}_{\theta}(\mathbf{x}_{t},t),\mathbf{\Sigma}_{\theta}(\mathbf{x}_{t},t)),
\end{equation}

where $\mathbf{\mu}_{\theta}(\mathbf{x}_{t},t)$ and $\mathbf{\Sigma}_{\theta}(\mathbf{x}_{t},t)$ are functions that provide the mean and covariance for the Gaussian and are defined using \acp{MLP}. \cite{dickstein2015deep,ho2020denoising}

The model is trained by maximizing the variational lower bound on the \ac{NLL} (like a \ac{VAE})

\begin{equation}
    \mathbb{E}[-\log p_{\theta}(\mathbf{x}_{0})]\leq\mathbb{E}_{q}[-\log\frac{p_{\theta}(\mathbf{x}_{0:T})}{q(\mathbf{x}_{1:T}\vert\mathbf{x}_{0})}]=L,
\end{equation}

which is done by optimizing the aforementioned \acp{MLP}. \cite{dickstein2015deep,ho2020denoising}

Sohl-Dickstein et al. \cite{dickstein2015deep} provide the first implementation of diffusion models and apply it to the generation and inpainting of images and binary sequences, performing worse than \acp{GAN}, but better than \acp{DBN}, \acp{GSN} and \acp{CAE} in terms of log-likelihood.

Ho et al. \cite{ho2020denoising} propose the \ac{DDPM}, which simplifies the training process of \cite{dickstein2015deep} by replacing $\mathbf{\Sigma}_{\theta}(\mathbf{x}_{t},t)$ with untrained time-dependent constants $\sigma_{t}^{2}\mathbf{I}$ in the reverse process and training an estimator $\mathbf{\epsilon}_{\theta}(\mathbf{x}_{t},t)$ to predict the noise $\mathbf{\epsilon}$ added to $\mathbf{x}_{t}$ instead of predicting the mean $\mathbf{\mu}_{\theta}(\mathbf{x}_{t},t)$. The variational lower bound is simplified to optimize the difference between actual and predicted error. For the estimator $\mathbf{\epsilon}_{\theta}$, PixelCNN++ \cite{salimans2017pixelcnn} is adopted with self-attention at the small feature maps.

Nichol et al. \cite{nichol2021improved} introduce the improved \ac{DDPM} for better log-likelihood results. It reintroduces $\mathbf{\Sigma}_{\theta}(\mathbf{x}_{t},t)$ as a trainable model and replaces the linear noise schedule for $\beta_{t}$ with a cosine schedule, which spreads the noise addition in the forward process more evenly. They also reduce the number of diffusion steps to improve sampling speed with very little quality loss by scaling the schedule parameters, and increasing the model size also increases performance.

Dhariwal et al. \cite{dhariwal2021diffusion} propose the ablated diffusion model (ADM) with classifier guidance (ADM-G), which improves upon \cite{nichol2021improved} by using more attention at different scales, class conditioning, a deeper model architecture, and a classifier to guide the generation process more precisely.

Ramesh et al. \cite{ramesh2022hierarchical} add text embeddings produced by a decoder-only transformer to the existing timestep embedding of a diffusion model to produce text-conditional images. Images can be manipulated based on text by reconstructing images from $\mathbf{x}_{T}$ with the diffusion model conditioned on new/interpolated text embeddings.

\subsection{Virtual Environments}\label{sec:virtual_environments}

Virtual environments are computer-simulated ``graphic and real-like models of real-life objects'' \cite{korakakis2018short} that are simple to annotate by nature since object locations, classes, and other properties are apparent in the simulation software. More realistic depictions of virtual objects have become possible because of the increasing computational processing power of \acp{GPU} in recent years. They allowed virtual environments to be adopted for various tasks, such as autonomous driving or gesture recognition. Based on \cite{korakakis2018short}, we derive three categories for virtual environment usages:

\begin{description}
    \item[Graphic Models] In this simple scenario, single 3D \ac{CAD} models or compositions are used to alleviate \ac{ML} model training for, e.g., gesture recognition and object recognition from different viewpoints or building 3D models from images.
    \item[Virtual Worlds] These models are computer-simulated and emulate a complex real-world environment. They are populated with many objects that can sometimes move or interact with the world or each other. They are especially popular for generating annotated training data for autonomous driving systems.
    \item[Interactive Environments] These interactive virtual worlds, usually video games or simulators with an inherent goal and well-defined rules, are especially suitable for the training and benchmarking/competition of \ac{RL} agents because they allow user inputs and directly present a result. The advantages are their high availability for various scenarios (e.g., driving simulations, open-world, and strategy games) and their often simple adaptability to research tasks through mod support and editor software.
\end{description}

For optimal results with rendered synthetic data for computer vision tasks, Mayer et al. \cite{mayer2018makes} present several findings: Multistage training on different data sets works better than mixing or training on one data set alone, complex and more realistic lighting does not necessarily help, and incorporating flaws of a real camera during training improves model performance. 

\subsubsection{Graphic Models}\label{sec:graphic_models}

Butler et al. \cite{butler2012naturalistic} introduce \textit{MPI-Sintel}, an optimized version of the open-source 3D animated short film Sintel \cite{roosendaal2010sintel} for optical flow evaluation. As ground truth, motion vectors for each pixel are computed using a modified version of Blender's \cite{blender} motion blur pipeline. The advantages over other data sets for optical flow evaluation are the long sequences and motions and the ability to render the film with various effects like motion blur, specular reflections, and atmospheric effects enabled or disabled.

Handa et al. \cite{handa2014benchmark} create the first RGB-D (RGB plus depth) image collection with camera trajectory and full 3D scene ground truth attached to each frame. They provide raytraced renderings of two rooms, the office room and the living room, via the POVRay \cite{povray} raytracing software. They also separately apply artificial noise to the RGB and depth values to make the images more realistic. The data set is then used to benchmark algorithms for visual odometry, 3D reconstruction, and \ac{SLAM}.

Su et al. \cite{su2015render} utilize 3D models rendered on top of real background images to train \acp{CNN} for viewpoint estimation. The models are rotated and inserted at different positions in the picture for that purpose. Their ``render for CNN'' approach achieves state-of-the-art performance on a benchmark data set at negligible human cost using existing 3D model repositories.

Peng et al. \cite{peng2015learning} train deep \ac{CNN} object detectors with synthetic images rendered from non-photorealistic 3D \ac{CAD} models that are freely available on the Internet to detect novel object categories not available in the real training data. They evaluate the models on real images, showing better detection performance than models trained on real data from a different domain.

Handa et al. \cite{handa2016scenenet} introduce a framework to randomly generate realistic and automatically annotated 3D indoor environments using 3D objects from public databases. The \textit{SceneNet} uses a hierarchical scene generator that learns relationships (co-occurrence frequency) between objects from prior indoor scene data sets. In another work~\cite{handa2016understanding}, they demonstrate their model's utility by using its renderings from random perspectives with added noise to train a deep model for depth-based semantic per-pixel segmentation.

\othertab{graphical models}{
    \cite{peris2012towards} & Creation of the Tsukuba CG Stereo data set, which is a collection of photo-realistic renderings of the \textit{head and lamp} scene under varying conditions and camera perspectives for stereo matching task training and evaluation. & 2012 \\\hline  
    \cite{molina2014natural} & Kinematic hand model with random initial global rotation is used to render hand gesture sequences for human-computer interfaces. & 2014 \\\hline 
    \cite{sun2014virtual} & Training of object detectors with 3D model renderings combined with domain adaptation using discriminative decorrelation performs comparably to models trained on real data (ImageNet). & 2014 \\\hline 
    \cite{rematas2014image} & (Re-)Synthesis of natural images from different viewpoints aided by structural information from 3D models of the same object class aligned to an image. & 2014 \\\hline 
    \cite{papon2015semantic} & Simultaneous class, pose, and location prediction of possible objects using a deep \ac{CNN} trained on RGB-D renderings of randomly generated 3D rooms built with intersection detection and plausibility checks (e.g., sofas are usually near walls). & 2015 \\\hline 
    \cite{fischer2015flownet} & A collection of \textit{Flying Chairs} image sequences combining natural background images with renderings of 3D chair models is used to train the \textit{FlowNet} \ac{CNN} for optical flow estimation. & 2015 \\\hline 
    \cite{kortylewski2018training} & Using a 3D morphable face model to create synthetic training data for face recognition systems reduces the amount of real data needed significantly and improves performance. & 2018 \\\hline 
    \cite{philipp2021synthetic} & Rendering video sequences of neurosurgical instrument movements from a microscope's perspective in Blender \cite{blender} for optical flow estimation benchmarks in this domain. & 2021 \\\hline 
    \cite{boikov2021synthetic} & Using Blender \cite{blender} to render steel pieces with defects (e.g., cracks in the surface texture) and masks for a steel defect detection task. & 2021 \\ 
}

\subsubsection{Virtual Worlds}\label{sec:virtual_worlds}

\begin{figure} [ht]
    \centering
    \includegraphics[width=\textwidth]{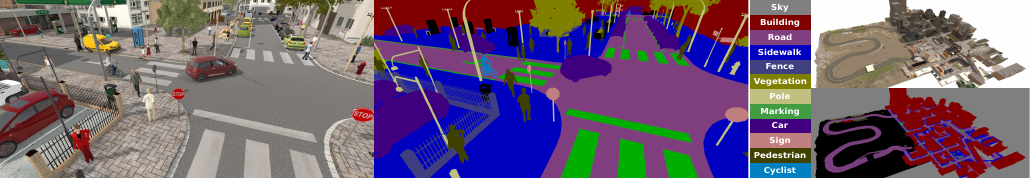}
    \caption{The SYNTHIA data set: Image rendered from the virtual world (left), ground truth segmentation map (middle), and city overview (right). (Source: \cite{ros2016synthia})}
    \label{fig:synthia}
\end{figure}

Haltakov et al. \cite{haltakov2013framework} propose a framework built on top of the open-source driving simulator \textit{VDrift} \cite{vdrift}. The modified software allows them to render realistic synthetic images with pixel-wise object annotations, depth, and optical flow maps (movements) in various scenarios, perspectives, and driving styles. The framework is then applied to create a large image training set for a multi-class image segmentation task.

Richter et al. \cite{richter2016playing} utilize the \textit{detouring} technique to generate semantic label maps for images from closed-source modern computer games. They evaluate the communication between the game and graphics hardware using a graphics API wrapper library, hashing rendering resources like textures or geometry and creating persistent object signatures to which labels for urban scene understanding are applied. The game training data, obtained from \textit{Grand Theft Auto V}, boosts the accuracy of segmentation models and reduces the need for expensive labeled real-world images.

Johnson-Robertson et al. \cite{johnson2016driving} use open-source plugins and \ac{GPU} buffer data to extract annotated and realistic images from \textit{Grand Theft Auto V} for vehicle detection tasks. They achieve state-of-the-art performance on real data with a model trained only on simulated images.

Ros et al. \cite{ros2016synthia} introduce the \textit{SYNTHIA} data set (see \autoref{fig:synthia}), which provides a semantically segmented collection of images of urban scenes obtained from the Unity game engine \cite{unity}. Since SYNTHIA is intended for autonomous driving tasks, it not only contains images from various perspectives in the virtual world but also four video sequences of a virtual car driving through the urban simulated landscape, one for each season. The virtual car consists of two multi-cameras 0.8 meters apart, each consisting of four monocular cameras with depth sensors, a common center, and 90-degree rotation between them. In combination with real data, SYNTHIA can significantly improve segmentation model accuracy.

\othertab{virtual worlds}{
    \cite{shafaei2016play} & Assessment of the effectiveness of models trained with computer games on real-world data in image segmentation and depth estimation tasks. The results show similar or better performance than models trained on real data. & 2016 \\\hline 
    \cite{gaidon2016virtual} & Virtual KITTI: A labeled video data set containing computer-rendered sequences of driving through realistic virtual worlds obtained from the Unity \cite{unity} game engine. The virtual worlds are created with positional information from the original KITTI \cite{geiger2012we} data set and human optimization. & 2016 \\\hline 
    \cite{richter2017playing} & Visual perception benchmark (VIPER): More than 250,000 high-resolution video frames with annotations, for instance, segmentation, optical flow, object detection, tracking, visual odometry, and object-level 3D scene layout tasks obtained by moving through the world of the video game \textit{Grand Theft Auto V}. & 2017 \\\hline 
    \cite{kar2019meta} & Meta-Sim: Using probabilistic scene grammars to create and render valid virtual environments. Performance improvements of the method are demonstrated by training a task network on the synthetic data and comparing it to a model trained on real data. & 2019 \\ 
}

\subsubsection{Interactive Environments}\label{sec:video_games}

Bellemare et al. \cite{bellemare2012arcade} build the Arcade Learning Environment (ALE) on top of an Atari 2600 emulator to evaluate \ac{AI} techniques like \ac{RL} and planning algorithms on arbitrary Atari games, which are usually split into a training and testing set. ALE provides an interface to the game controls, screen information, RAM, and registers for an \ac{AI} agent to control or read. The reward function for an agent is defined per game based on the score difference between frames.

Kempka et al. \cite{kempka2016vizdoom} build upon the idea of \cite{bellemare2012arcade}, where 2D Atari 2600 games are used as an evaluation platform for \ac{RL} agents, and propose \textit{VizDoom}, a more realistic 3D game based on the first-person shooter Doom, as a research platform for visual reinforcement learning. They train competent bots using convolutional deep neural networks and \ac{RL} on various tasks and scenarios.

Sadeghi et al. \cite{sadeghi2016cad2rl} make deep reinforcement learning applicable to safety-critical domains such as autonomous flights by training in virtual environments built entirely with \ac{CAD} models. The CAD$^{2}$RL method trains a \ac{RL} agent (deep \ac{CNN}) on RGB images of a monocular camera mounted to a drone in the virtual environment to output velocity commands that avoid collisions. The authors find that by using highly randomized rendering settings, the agent's policy can be trained to generalize well to real-world applications, which they demonstrate by letting the trained agent fly a real drone through indoor environments.

Vinyals et al. \cite{vinyals2017starcraft} introduce the \textit{StarCraft II Learning Environment} (SC2LE), which combines a Python-based interface for the game engine with specifications for possible observations, actions, and rewards. As a complex multi-agent problem with incomplete information, long-term strategies, and large action space, StarCraft provides a difficult class of problems to evaluate \ac{RL} models on. The authors test agents on various mini-games, resulting in agent behavior similar to a novice player and on the main game, where the agents cannot progress noticeably.

\othertab{interactive environments}{
    \cite{beattie2016deepmind} & \textit{DeepMind Lab}: First-person 3D game framework based on the \textit{Quake III} game engine for easy task and \ac{AI} design. & 2016 \\\hline 
    \cite{synnaeve2016torchcraft} & \textit{TorchCraft}: Providing an interface between the \textit{Torch} \ac{ML} framework and real-time strategy game ``StarCraft: Brood War''. & 2016 \\\hline 
    \cite{johnson2016malmo} & Project Malmo: An \ac{AI} experimentation platform for complex navigation, survival, collaboration, and problem-solving tasks built on top of Minecraft, a popular game mimicking the real world as a collection of blocks and friendly and hostile entities (e.g., animals, zombies). & 2016 \\ 
}

%% file: classification.tex
\section{Classification of Generative Models}\label{ch:classification}

In this section, we classify the models we presented in \autoref{ch:models} by criteria presented in \autoref{sec:classification_criteria}. In \autoref{sec:trend_analysis}, we perform a trend analysis before finally presenting a guideline for model selection in \autoref{sec:guideline}.

\subsection{Criteria for Classification}\label{sec:classification_criteria}

\begin{figure} [ht]
    \centering
    \begin{subfigure}{0.55\textwidth}
        \centering
        \includegraphics[width=0.45\textwidth]{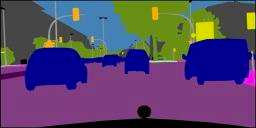}
        \includegraphics[width=0.45\textwidth]{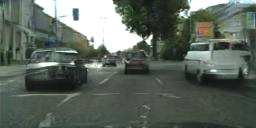}
        \caption{Image-to-image translation (Source: \cite{isola2017image})}
        \label{fig:i_to_i_translation}
    \end{subfigure}
    \hfill
    \begin{subfigure}{0.44\textwidth}
        \centering
        \includegraphics[width=0.9\textwidth]{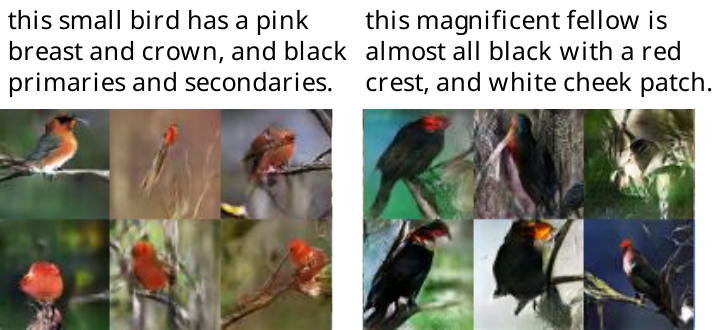}
        \caption{Text-to-image translation (Source: \cite{goodfellow2016nips})}
        \label{fig:t_to_i_translation}
    \end{subfigure}
    \caption{Examples of \acp{GAN} being applied to translative tasks.}
    \label{fig:translation_example}
\end{figure}

For each model presented in this work, we collect certain information we find to be comparable and useful for our trend analysis and guidelines:

\begin{description}
    \item[Metadata] We collect name, release year, model (sub-)category (according to \autoref{ch:models}) and citations (according to Google scholar). Optional entries are predecessors (other models that are the foundation for this model) and combinations, which describe the categories a proposed model combines (e.g., a \ac{GAN} can combine a \ac{CNN} generator with a \ac{RNN} discriminator).
    \item[Data Structure] Determines if the size/amount of samples of the data generated by the model is limited, for example, an image with a static resolution or a fixed-length video, or (theoretically) infinite, which is often required for processing arbitrarily-sized sequences.
    \item[Data Type] Type of the data generated by the model, for example, natural language text, time series, music, or images.
    \item[Sampling Requirements] The data generation process of a model can be unconditional or conditional, which means an input is required based on which synthetic data is generated, or both (see \autoref{fig:translation_example}). If an input is accepted, we also capture the types of input that the model accepts.
    \item[Sampling Process] Describes whether a sample is generated in ``one go'' or iteratively refined by the model.
    \item[Training Process] Describes how the generative model is trained and is described by two aspects, inspired by \cite{wang2019generative}:
        \begin{description}
            \item[Loss Type] The loss function(s) that is/are used to optimize the model.
            \item[Optimization] Additional penalization of the model or modification of the loss function to improve results.
        \end{description}
    \item[Data Sets] The data sets used to train and evaluate the models.
    \item[Model Performance] Comparing different models is complicated because no commonly used performance measure exists, and many proposed measures only work for specific data types or domains (e.g., music can not be evaluated in the same way as natural language). To work around this issue, we collect the \textit{performance predecessors}, that is, the list of outperformed models, from the evaluation section of the respective paper, if available.
    \item[Privacy] Shows whether the generated data is considered differentially private or private by another criterion defined by the respective authors.
\end{description}

\subsection{Data Evaluation and Trend Analysis}\label{sec:trend_analysis}

This section presents the findings from our survey of existing \ac{SDG} literature. In the following subsections that build on one another, we investigate the criteria proposed in \autoref{sec:classification_criteria} before concluding.

\subsection{Metadata}\label{sec:class_metadata}

\begin{figure}[ht]
    \begin{center}
        \input{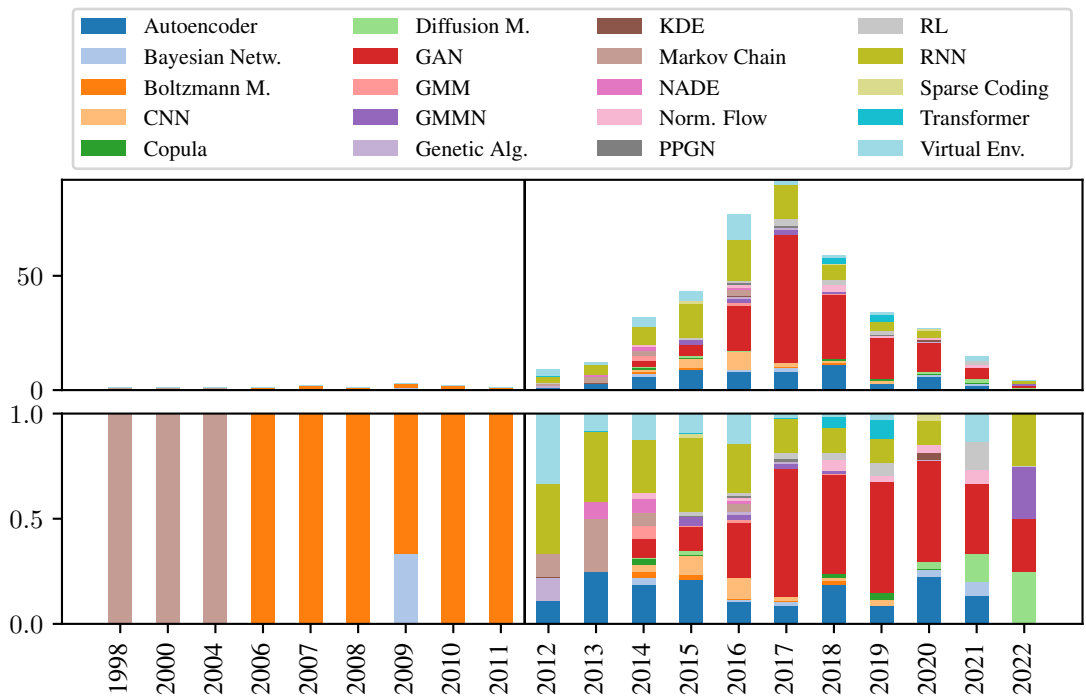}
    \end{center}
    \caption{Total (top) and normalized (bottom) amount of models we present in our work grouped by year and model category.}
    \label{fig:class_categories_by_year}
\end{figure}

First, we look at the metadata from our \works{} models. In \autoref{fig:class_categories_by_year}, we show the number of papers grouped by model category we evaluated for each year. We focus primarily on models proposed in the last ten years, so the data starting in 2012 is primarily relevant. \acp{GAN}, \acp{RNN}, autoencoders (especially \acp{VAE}), virtual environments, and \acp{CNN} experience high usage throughout the years in the literature we evaluated, with \acp{GAN} quickly surpassing the other approaches since their first proposal in 2014 \cite{goodfellow2014generative}. The usage of Markov chain models and Boltzmann machines declined over the years, while \ac{RL} and diffusion models slightly gained popularity.

\begin{figure}[ht!]
    \begin{center}
        \input{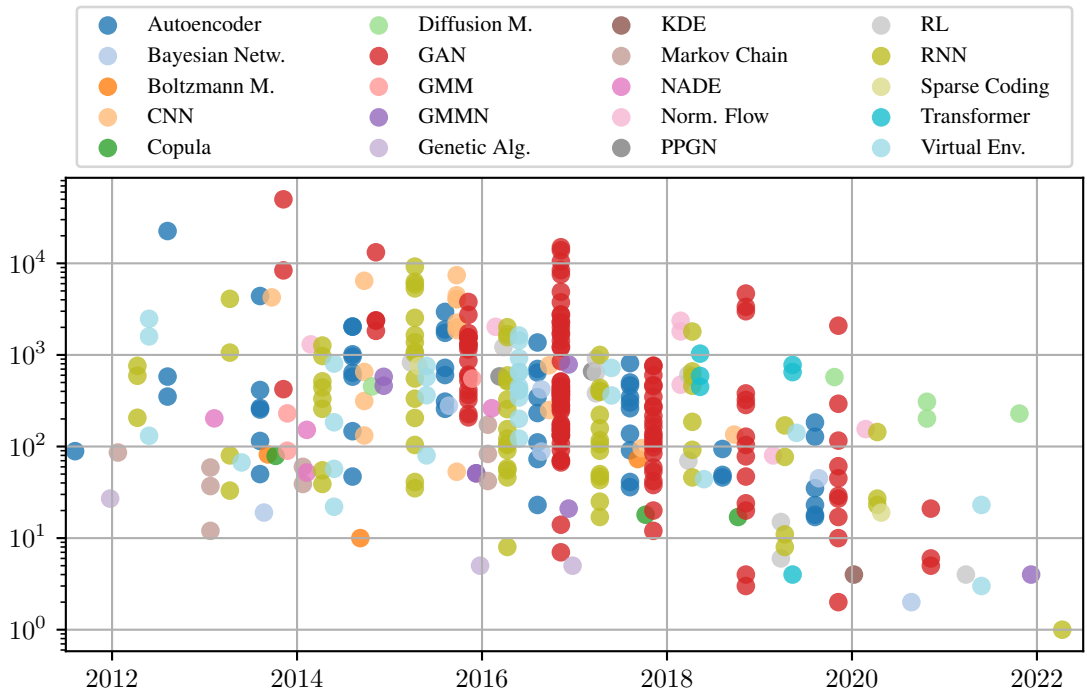}
    \end{center}
    \caption{Google Scholar citations for different model categories over the last ten years on a logarithmic scale. For better readability, we apply a $[-0.4,0.4]$ offset to the points' year value in the order of their legend appearance (top to bottom, left to right). The numbers were last obtained \textcolor{red}{Date}.}
    \label{fig:class_citations}
\end{figure}

Next, in \autoref{fig:class_citations}, we evaluate the amount of Google Scholar citations per model category. Compared to other models covered by our work, \acp{GAN} have received the highest amount of citations over the last ten years. In addition, \acp{RNN}, \acp{CNN}, and autoencoders often receive comparable attention.

\begin{figure} [htp]
    \centering
    \begin{subfigure}[c]{0.55\textwidth}
        \centering
        \scalebox{0.6}{
            \input{figures/classification/hierarchy_predecessors_gan.tex}
        }
        \caption{\acp{GAN}.}
    \end{subfigure}
    \begin{subfigure}[c]{0.39\textwidth}
        \centering
        \scalebox{0.6}{
            \input{figures/classification/hierarchy_predecessors_rnn.tex}
        }
        \caption{\acp{RNN}.}
    \end{subfigure}
    \begin{subfigure}[c]{0.49\textwidth}
        \centering
        \scalebox{0.6}{
            \input{figures/classification/hierarchy_predecessors_diff.tex}
        }
        \caption{Diffusion models.}
    \end{subfigure}
    \begin{subfigure}[c]{0.49\textwidth}
        \centering
        \scalebox{0.6}{
            \input{figures/classification/hierarchy_predecessors_ae.tex}
        }
        \caption{Autoencoders.}
    \end{subfigure}
    \caption{Inheritance graphs of the model types for which we acquired a significant amount of data in the \textit{predecessors} section of the metadata. We omit models without documented edges, and some years are spread across multiple rows to accommodate width limitations.}
    \label{fig:hierarchy_predecessors}
\end{figure}
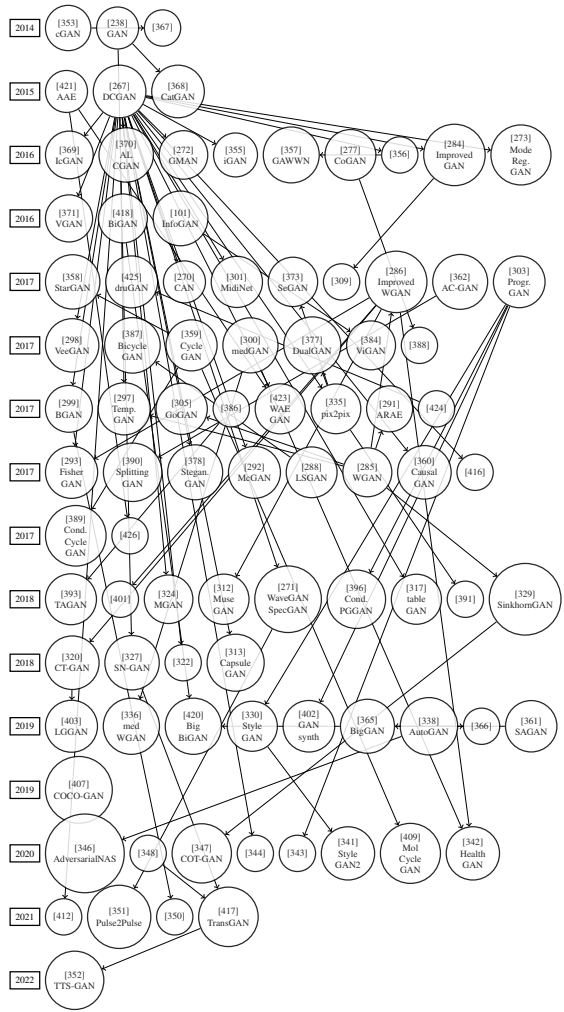
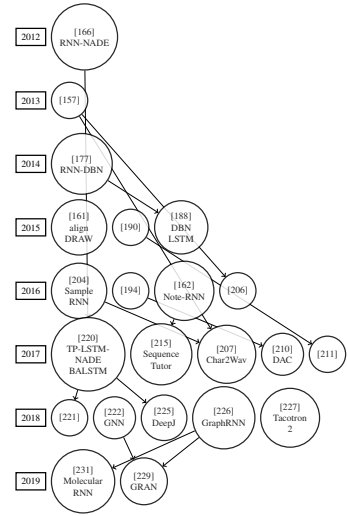
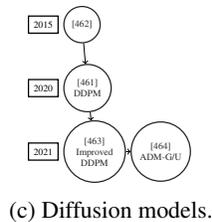
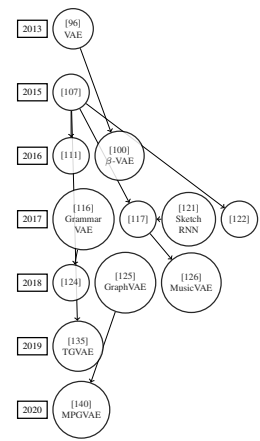

In \autoref{fig:composite_model_flow_graph}, we evaluate which architectures and concepts our models borrow from other model types. The models often used as submodels according to \autoref{fig:composite_model_flow_graph_total} are \acp{CNN}, followed by \acp{RNN} and autoencoders. In \autoref{fig:composite_model_flow_graph_normalized}, we also show that some apparent assumptions are confirmed:

\begin{itemize}
    \item \acp{GAN} and the very similar \acp{GMMN} often use autoencoders, especially their decoders, as generative networks.
    \item 70\% of our \ac{RL} models use \acp{RNN}, which are suitable for guidance by reward functions due to their sequential data generation process. \ac{RL} is also applied to \ac{GAN} and \ac{CNN} training.
    \item Autoencoders often incorporate \acp{RNN} and \acp{CNN} as their encoders and decoders.
    \item Diffusion models, \acp{GAN} and normalizing flow models heavily utilize \acp{CNN}. This is because many \ac{GAN} models are based on the \ac{CNN}-based \ac{DCGAN} \cite{radford2015unsupervised}, and diffusion models are also mostly related to the \ac{CNN}-based DDPM \cite{ho2020denoising}, which is illustrated in \autoref{fig:hierarchy_predecessors}.
\end{itemize}

\begin{figure} [ht]
    \begin{subfigure}{\textwidth}
        \centering
        \input{figures/classification/composite_flow_graph_total.tex}
        \caption{Total amount bottom models borrowed from upper ones.}
        \label{fig:composite_model_flow_graph_total}
    \end{subfigure}
    \begin{subfigure}{\textwidth}
        \centering
        \input{figures/classification/composite_flow_graph_normalized.tex}
        \caption{Normalized (in \%, rounded down) by the number of models of bottom model categories that use the upper models.}
        \label{fig:composite_model_flow_graph_normalized}
    \end{subfigure}
    \caption{Weighted dependency graph for model compositions. The bottom models rely on submodels (e.g., encoders and decoders) from the upper categories. The edges' size and labels denote the amount or percentage of models that use the respective submodel. We omitted edges with less than three total usages for readability.}
    \label{fig:composite_model_flow_graph}
\end{figure}
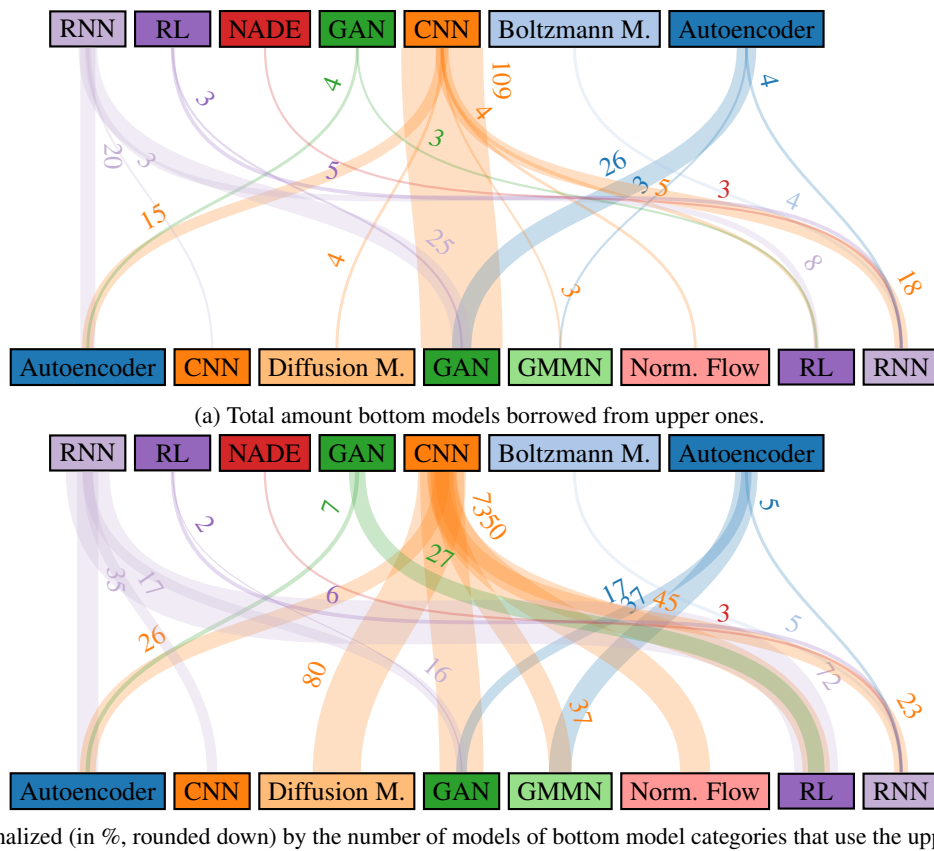

Finally, we discuss the \textit{predecessors} class of our metadata. In \autoref{fig:hierarchy_predecessors}, we build a graph of all predecessor connections for notable model categories. We observe that especially \acp{GAN} and \acp{RNN} are strongly interconnected. For autoencoders, we only found strong relations between \ac{VAE} models. Four of the five diffusion models we presented significantly depend on each other, while the fifth model is extended for text-conditional tasks and, therefore, considerably changed.

\subsection{Data Structure}\label{sec:data_structure}

\begin{figure}[ht]
    \begin{center}
        \input{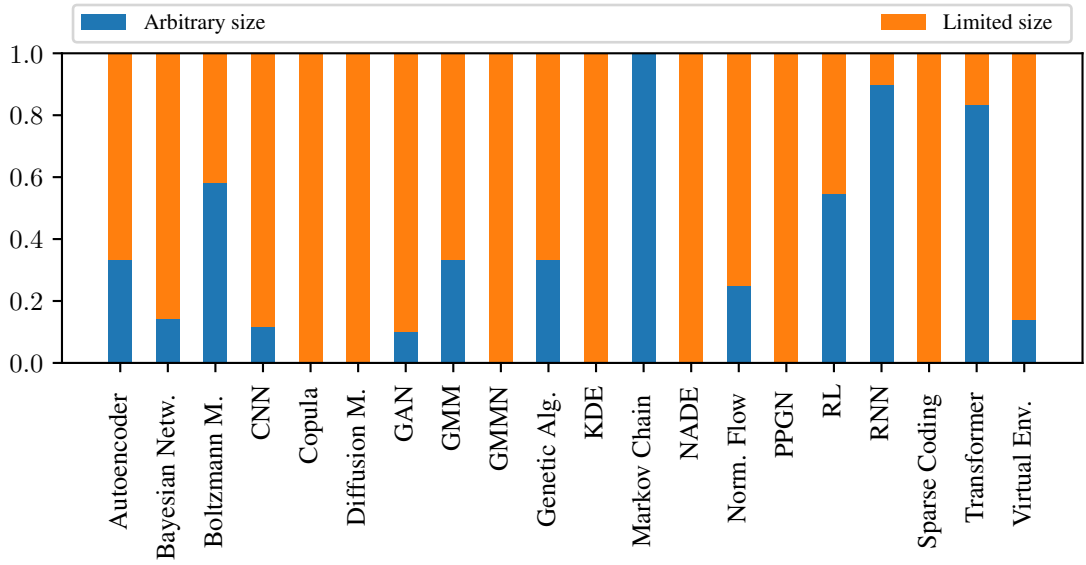}
    \end{center}
    \caption{Fraction of data structures different types of models put out.}
    \label{fig:data_structure}
\end{figure}

In \autoref{fig:data_structure}, we discuss the data modeling capabilities of different model types. As expected, sequence-based models such as Markov chains, \acp{RNN}, and transformers are mainly applied to generate arbitrarily-sized data. Autoencoders and \acp{GAN} are based initially on neural network architectures with fixed input and output sizes but can process sequences by adopting \acp{RNN}. These correlations are easy to observe by comparing the \ac{RNN} connections in \autoref{fig:composite_model_flow_graph_normalized} to our findings in this section. \ac{RL}, which also heavily relies on \acp{RNN}, works well for unlimited data lengths as well, also due to the reward function being a powerful tool to learn and improve during the generation process. Modern Boltzmann machines like the \ac{TRBM} are also strongly related to \acp{RNN}, making them noticeable in our data structure evaluation.

\subsection{Data Types}\label{sec:data_types}

We identify several data types in our research, of which multiple are often used by a single model. We categorize them as follows:

\begin{description}
    \item[Audio] Represents raw waveform audio. Due to the high sampling rate required to produce natural-sounding audio, this data type usually requires complex models and thorough training to produce good results. We further differentiate between music and speech generation due to their difference in complexity (e.g., note and instrument vs. text, speaker, and character transitions).
    \item[Image] Describes classic two-dimensional bitmaps, usually with RGB or grayscale pixel values to which we refer as natural images. Binary images are simplified versions where pixels can either be on or off (black and white). Segmentation masks describe images where pixels are of a particular class instead of color. Images with more information encapsulate other pixel values besides color, for example, depth in RGB-D images or \acp{HSI}.
    \item[Text] This class describes a sequence of characters. We differentiate between natural language, as spoken by humans, and text representations that encode other data types, for example, SMILES strings \cite{weininger1988smiles} for molecules.
    \item[Time Series] Sequences of one (univariate) or more (multivariate) variable values that have to be determined for each step. The more potentially interdependent variables have to be specified, the more complex the task becomes. We additionally define symbolic music as an additional category because of the considerable interest in the topic and the complex constraints provided by music theory that must be considered. Depending on the task definition, symbolic music can be regarded as a univariate or multivariate time series.
    \item[Graphs and Molecules] Graphs are collections of nodes with connections (edges) between them. Nodes and edges can have additional properties and values assigned to them. They are usually represented by adjacency matrices or time series of node and edge creations. Many presented works consider molecules as a subset of graphs, where atoms are the nodes connected in a specific way, and they are often represented as SMILES \cite{weininger1988smiles} text representations in addition to the representation above types.
    \item[Tabular Data] We consider tabular data to be tuples, sequences of tuples, or matrices containing categorical and numerical values. Besides generating tables, it is often used to provide additional information to another data type. Virtual environments use it to provide, for example, emulator or game information.
    \item[Video] A sequence of images as defined above.
\end{description}

\begin{figure}[ht!]
    \begin{center}
        \input{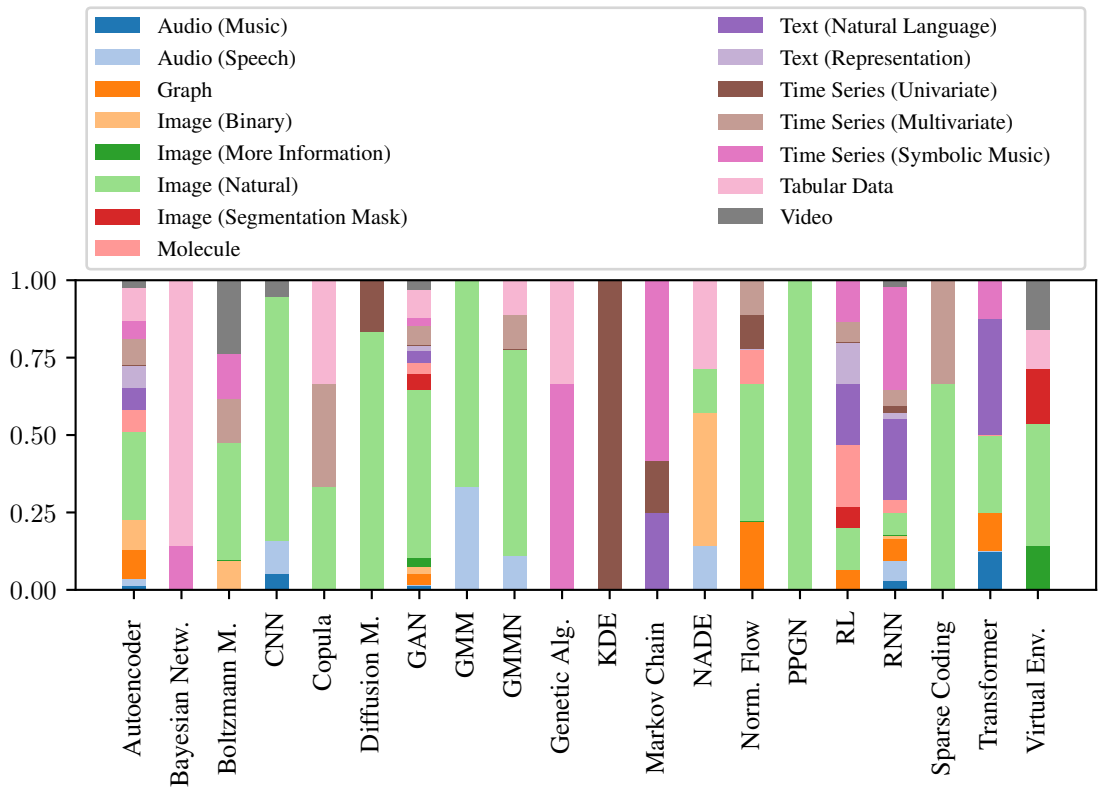}
    \end{center}
    \caption{Fraction of data type usages per model category.}
    \label{fig:data_type_v_bar}
\end{figure}

\autoref{fig:data_type_v_bar} shows that natural image generation is the data type of most overall importance. \acp{RNN} are predominantly applied to sequential arbitrary-length domains like natural language text and symbolic music. Due to their flexible architecture, autoencoders and \acp{GAN} are applied to almost all data types.

\begin{figure}[ht!]
    \begin{center}
        \import{figures/classification/}{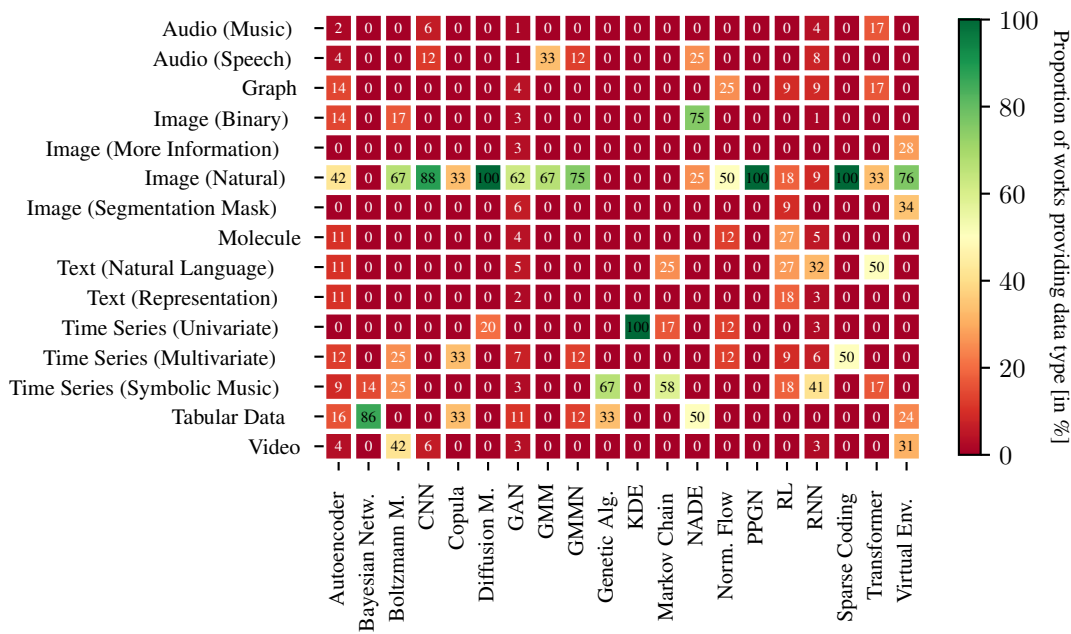}
    \end{center}
    \caption{Heatmap showcasing the data output provided by different types of models.}
    \label{fig:data_type_heatmap}
\end{figure}

\autoref{fig:data_type_heatmap} illustrates that some model types are limited to specific data types: Boltzmann machines are exclusively applied to time series, video, and image data. \acp{CNN} mainly focus on images and other high-dimensional formats like waveform audio. Virtual environments are exclusively applied to visual data and provide additional information. \ac{RL} puts itself forward for domains suited for sequential generation.

\subsection{Sampling Requirements}\label{sec:sampling_requirements}

\begin{figure}[ht!]
    \begin{center}
        \input{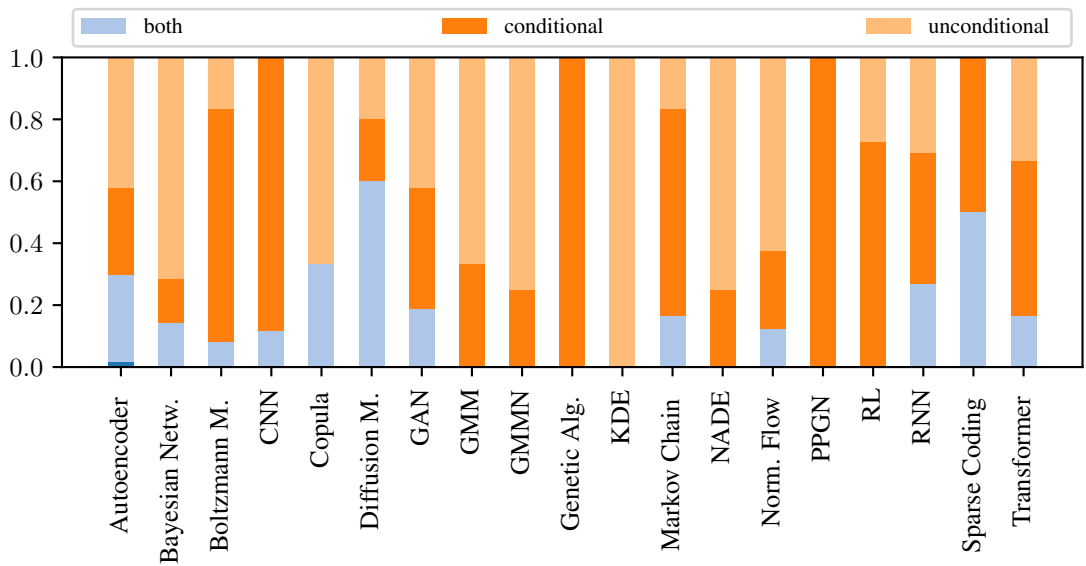}
    \end{center}
    \caption{Fraction of models conditioned on input data per model category.}
    \label{fig:sampling_requirements_multi_v_bar}
\end{figure}

In \autoref{fig:sampling_requirements_multi_v_bar}, we compare the models' reliance on additional information. Genetic algorithms, \acp{PPGN}, sparse coding models, and \acp{CNN} that we covered can always be conditioned on additional information to guide the generation process. Boltzmann machines, \acp{RNN}, transformers, \ac{RL}, Markov chains, diffusion models, \acp{GAN}, and autoencoders are also very flexible and often support conditional generation. The other model types are mainly used for unconditional generation; they only require random noise as input or during the generation process.

\begin{figure}[ht!]
    \begin{center}
        \import{figures/classification/}{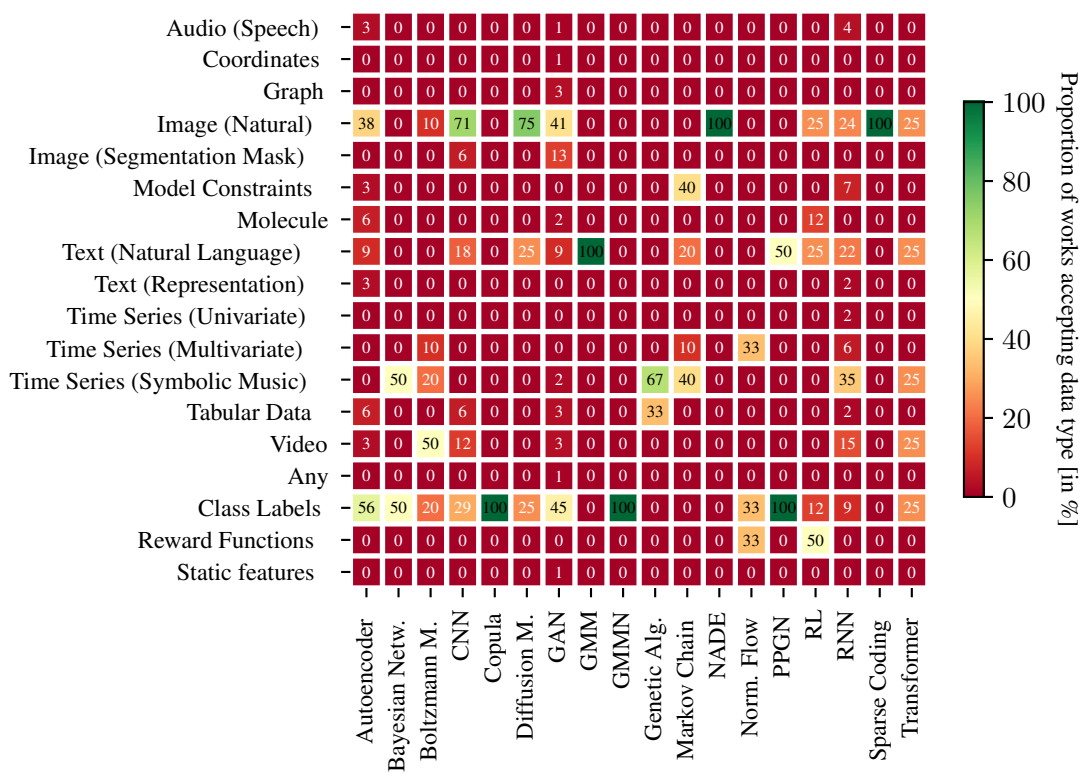}
    \end{center}
    \caption{Heatmap showcasing the data types accepted by different types of models.}
    \label{fig:sampling_requirements_heatmap}
\end{figure}

In \autoref{fig:sampling_requirements_heatmap}, we provide a more detailed overview of the reliance of specific model types on certain conditioning input types. We extend the data types proposed in \autoref{sec:data_types} with three new entries:

\begin{description}
    \item[Model Constraints] Constraints imposed on the model by the user, for example, static generation rules from music theory or positional constraints that require a specific step in a sequence to have a particular value.
    \item[Class Labels] A one-hot vector or embedding that specifies certain aspects the generated data should have. This could be hair or skin color for human face image generation.
    \item[Reward Functions] User-defined functions that provide a specific value to be optimized by the model in addition to its default targets. For example, a generated molecule should have a particular chemical property.
\end{description}

The most often used sampling requirements across all model types are images and class labels, followed by natural language text and music. Due to their simple structure, Markov chains are suitable for applying model constraints. In contrast, \ac{RL} models are predestined for using reward functions due to their flexible policy learning architecture. In the presented literature, we find segmentation masks to be exclusively used by \acp{CNN} and \acp{GAN}, which also often use \acp{CNN} as generators (see \autoref{sec:class_metadata}).

\subsection{Sampling Process}\label{sec:sampling_process}

\begin{figure}[ht]
    \begin{center}
        \input{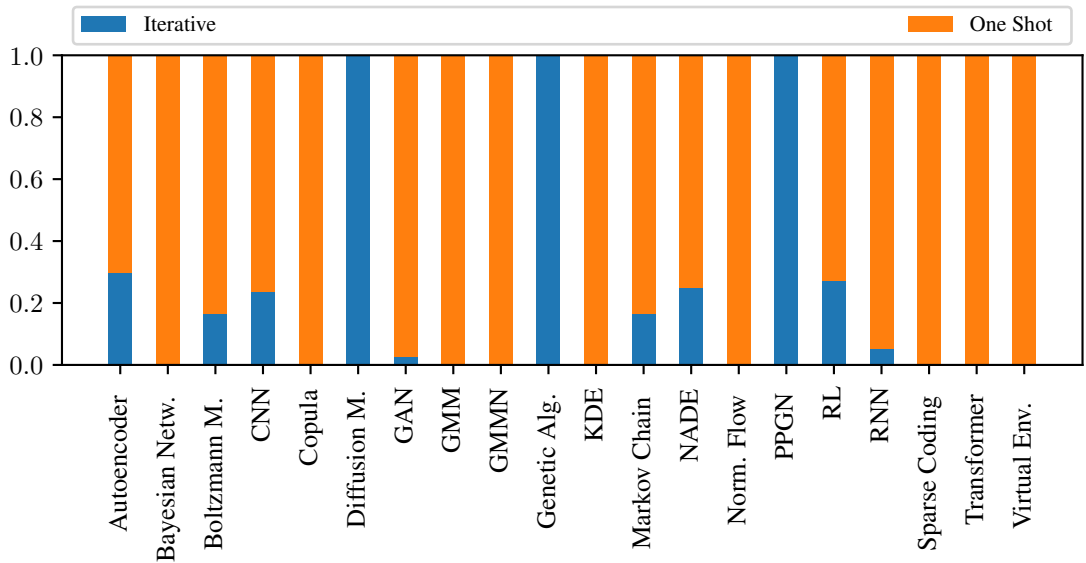}
    \end{center}
    \caption{Fraction of models utilizing a specific sampling process.}
    \label{fig:sampling_process_multi_v_bar}
\end{figure}

We found two types of sampling processes to be relevant: Determining the values of a sample or sequence of samples ``in one go'' (one shot) and iteratively refining the sample values a specific number of steps or until a criterion is reached. We present our findings in \autoref{fig:sampling_process_multi_v_bar}, where we show that most models use one-shot sampling. \acp{PPGN}, genetic algorithms and diffusion models always iteratively sample data because of their architecture. Other models like autoencoders, \acp{CNN}, Markov chains (especially \acp{HMM}), \acp{NADE}, \ac{RL}, and \acp{RNN} are also suitable to be applied multiple times or as deep architectures consisting of multiple submodels to the data to refine the results, but less often used in that way.

\subsection{Training Process}\label{sec:training_process}

\begin{figure}[ht]
    \begin{center}
        \import{figures/classification/}{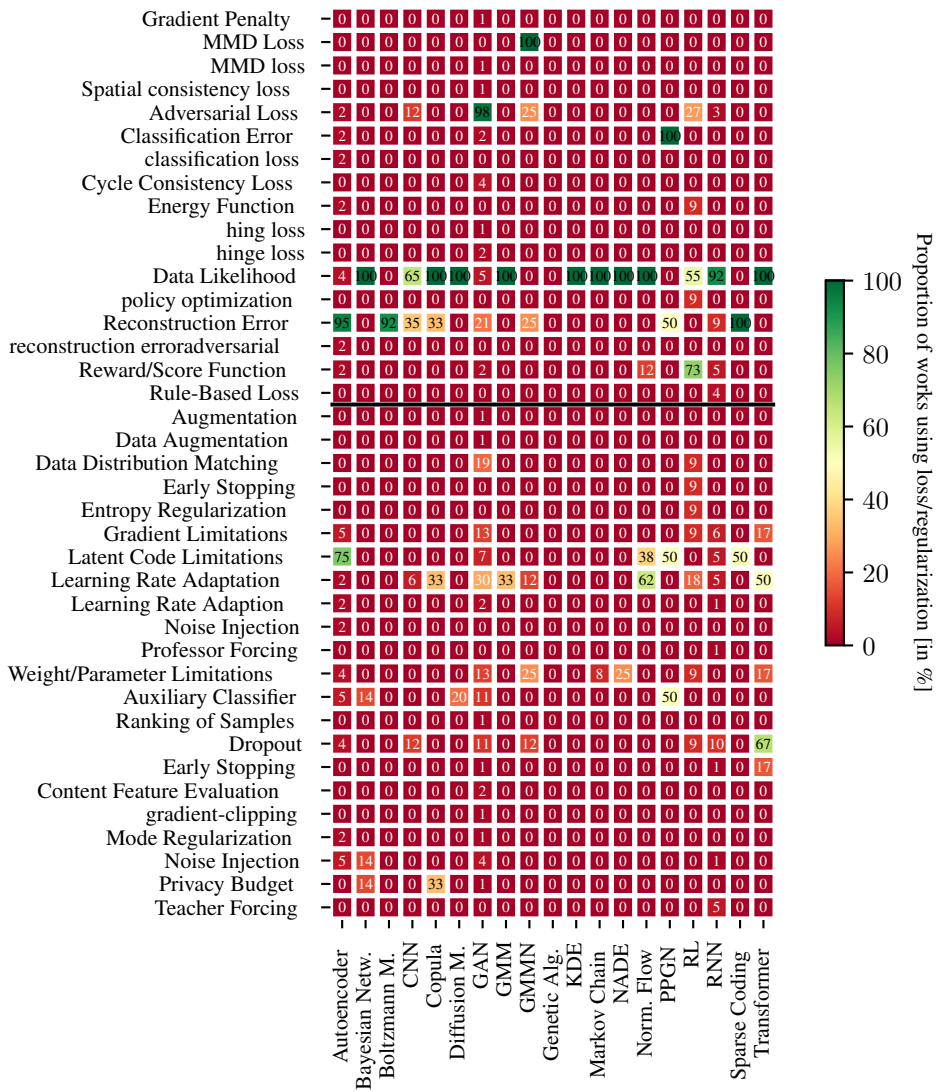}
    \end{center}
    \caption{Loss (above) and additional optimization (below black line) heatmap for different model types.}
    \label{fig:loss_and_regularization_heatmap}
\end{figure}

In \autoref{fig:loss_and_regularization_heatmap}, we provide a combined overview of different losses and optimization techniques used to train and improve generative models:

\begin{description}
    \item[\ac{MMD} Loss/Data Distribution Matching] Comparison of overall data statistics between sets of real and generated samples. \acp{GMMN} use the \ac{MMD} metric as their main training objective. Many \acp{GAN} use the Wasserstein distance in addition to the adversarial loss to prevent mode collapse.
    \item[Adversarial Loss/Classification Error/Auxiliary Classifier] Describes the competition of a generative model against a classifier model that judges whether its input is real or fake (adversarial loss, the foundation of \acp{GAN}) or the probabilities that the input belongs to a specific class. Classifiers are often used besides other loss types (auxiliary) or in \acp{PPGN} as the primary training objective, which can be considered a more powerful \ac{GAN} discriminator.
    \item[Data Likelihood] A simplistic evaluation approach of the model performance is often used by older models, letting the model assign a generation probability to real test data.
    \item[Reconstruction Error/Cycle Consistency Loss] Reconstruction error is the foundation of autoencoders, Boltzmann machines, and the more general sparse coding model that takes data as input, converts it to a small representation, and then aims to reconstruct the data as accurately as possible. The accuracy is measured by the error metric, which is often a mean absolute or mean squared error. Cycle consistency loss, introduced by CycleGAN \cite{zhu2017unpaired}, allows a \ac{GAN} to train based on reconstruction loss by training conditional generators and discriminators for both directions in an unpaired image translation task.
    \item[Reward/Score Function and Content Feature Evaluation] Evaluation of specific aspects of the generated data for model training and guidance. This is especially relevant for policy-learning models like \ac{RL}.
    \item[Rule-Based Loss] Hard constraints imposed on the generated data and model, implemented by humans. It is used to force \acp{RNN} to comply with the basic rules of music theory.
    \item[Limitations and Adaptation] The training process of many models is significantly modified. Model training with gradient descent is often improved and accelerated by using a dynamically adapted learning rate or restricting the gradient itself (e.g., gradient clipping, normalization, penalization). A model's network weights or parameters can also be heavily restricted by normalization, freezing, decay, clipping, or connection of some of them (i.e., weight sharing). Models that work with latent codes representing data (mostly autoencoders) also often impose distribution constraints on them, usually to simplify the sampling of new data.
    \item[Ranking of Samples] A ranking of fake samples among real ones provides more detailed feedback to the generative model.
    \item[Dropout] Set a random fraction of neurons of a neural network to zero at each training step to prevent neurons from learning the same features.
    \item[Early Stopping] Stop optimization on the training data when model performance stops increasing on the evaluation data to prevent overfitting.
    \item[Mode Regularization] Offering an incentive to the model or penalizing it for its coverage of data classes or the data distribution. Makes models more resilient to imbalanced data sets or mode collapse in the case of \acp{GAN}.
    \item[Noise Injection] Adding noise to the data or at specific layers in a neural network. Forces the model to generalize more.
    \item[Privacy Budget] Used to prevent differentially-private models from disclosing too much information from the original data.
    \item[Teacher Forcing] Applied to \acp{RNN} to avoid error propagation by feeding the correct token of the real data instead of the faulty predicted token to the next recurrent step.
\end{description}

\subsection{Data Sets}\label{sec:data_sets}

\begin{figure}[ht]
    \begin{center}
        \input{figures/classification/data_set_counts.pgf}
    \end{center}
    \caption{Amount of models (excluding virtual environments) using at least one public, only private, and only private data sets while needing to enforce privacy constraints. Additionally, we cover the most popular data sets with more than ten occurrences.}
    \label{fig:data_set_counts}
\end{figure}
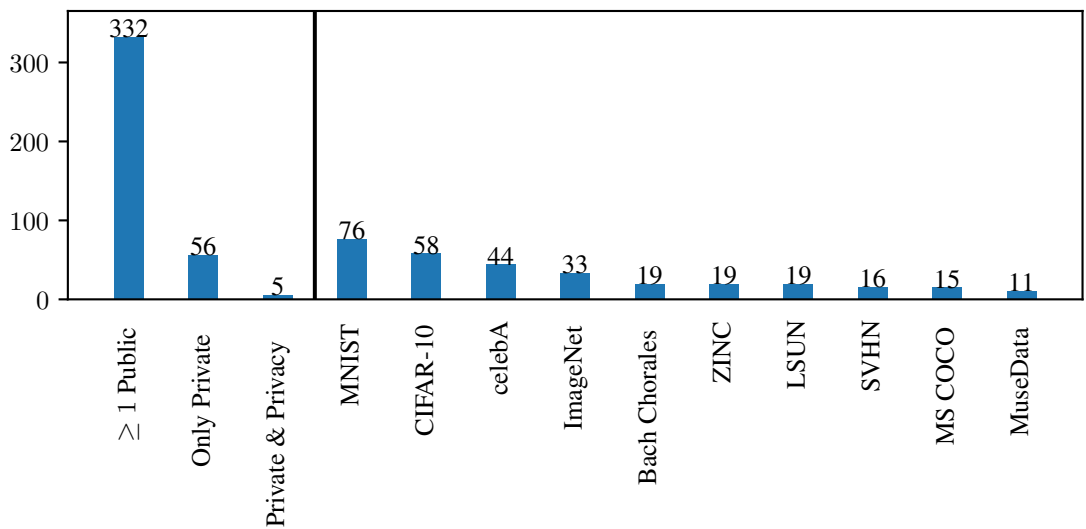

Common data sets are essential for model training and evaluation. They allow researchers to compare their models against others in a meaningful manner. These data sets must be commonly available to the research community for that purpose. In \autoref{fig:data_set_counts}, we investigate if the generative models presented in this survey use public or private data sets. We find that 355 out of 388 models (excluding virtual environments) use at least one publicly available data set for their evaluation, while 33 do not disclose their data, of which only five models are used for privacy preservation, where restrictions often apply (e.g., healthcare). Additionally, 47 models utilize both private and public data sets for their evaluations. Each paper presented uses 1.74 data sets on average.

We further present the most often used public data sets used in our survey: MNIST \cite{lecun1998gradient}, CIFAR-10 \cite{krizhevsky2009learning}, celebA \cite{liu2015faceattributes}, ImageNet \cite{deng2009imagenet}, LSUN (Bedrooms) \cite{yu2015lsun}, MS COCO \cite{lin2014microsoft} and \ac{SVHN} \cite{netzer2011reading} are collections of labeled or unlabeled images, which also are the most popular application area of \ac{SDG} as we identified in \autoref{sec:data_types}. Other domains where commonly used data sets exist are music with the Bach Chorales \cite{bach_chorales} and MuseData\footnote{\url{www.musedata.org}} data sets, and graphs/molecules, where ZINC \cite{sterling2015zinc} is popular.

\subsection{Model Performance}\label{sec:model_performance}

\begin{figure}
    \begin{subfigure}{\textwidth}
        \centering
        \scalebox{0.55}{
            \input{figures/classification/graph_performance_predecessors.tex}
        }
    \end{subfigure}

    \vspace{3mm}

    \begin{subfigure}{\textwidth}
        \begin{center}
            \import{figures/classification/}{color_bar_performance.pgf}
        \end{center}
    \end{subfigure}
    \caption{The relationships between models and their performance predecessors. The fill color indicates the model category as in \autoref{fig:class_categories_by_year} while the border color shows how often other models have outperformed a model.}
    \label{fig:graph_performance_predecessors}
\end{figure}

In \autoref{fig:graph_performance_predecessors}, we visualize the models and their relationships to other approaches that they claim to outperform in their respective evaluations. We can see that almost all arrows run from the top to the bottom, meaning that newer models tend to outperform older ones. Most models only evaluate against a small number of other works (on average 0.65 of the presented approaches), leading to a small in-degree and usually also out-degree. Some approaches like \ac{DCGAN} \cite{radford2015unsupervised} and \cite{gomez2016automatic} are popular to compare against, as shown by their large out-degree. \acp{GAN} are overall most often compared against (83 times), followed by autoencoders (62 times), \acp{RNN} (44 times), \acp{CNN} (25 times), normalizing flow models (12 times), Boltzmann machines (7 times), \ac{RL} approaches (7 times), diffusion models (5 times), \acp{NADE} (3 times), transformers (3 times), \acp{GMMN} (2 times), and \acp{PPGN} (1 time).

\acp{GAN} and transformers tend to outperform other model types like \acp{CNN} and autoencoders. In graph/molecule generation domain, \acp{RNN} also hold up well. Since the resurgence of diffusion models in 2020 \cite{ho2020denoising}, they also outperform \acp{GAN} in unconditional and text-conditional image generation.

We also investigate the evaluation metrics used by the presented models: The overall most commonly used metric is the \acf{NLL}, which describes the probabilities assigned to the observed ground truth by the model. As Borji \cite{borji2022pros} points out, a low \ac{NLL} score does not necessarily result in high data quality, and the metric is difficult to compute for high-dimensional data. Other common metrics for image data and \acp{GAN} in particular are \acf{IS} and \acf{FID}, which use a pre-trained image classifier to compare the real and fake data distributions \cite{borji2022pros}. We also often encounter evaluations by humans using a \ac{MOS} on a scale from one to five or one to ten to rate several features of the produced data. A comprehensive evaluation based on these metrics is not possible, due to most works using different metrics specialized for their task, and even if the same metric is used, the data sets (see \autoref{sec:data_sets}) are often different.

\subsection{Privacy}

\begin{figure}[ht!]
    \begin{center}
        \input{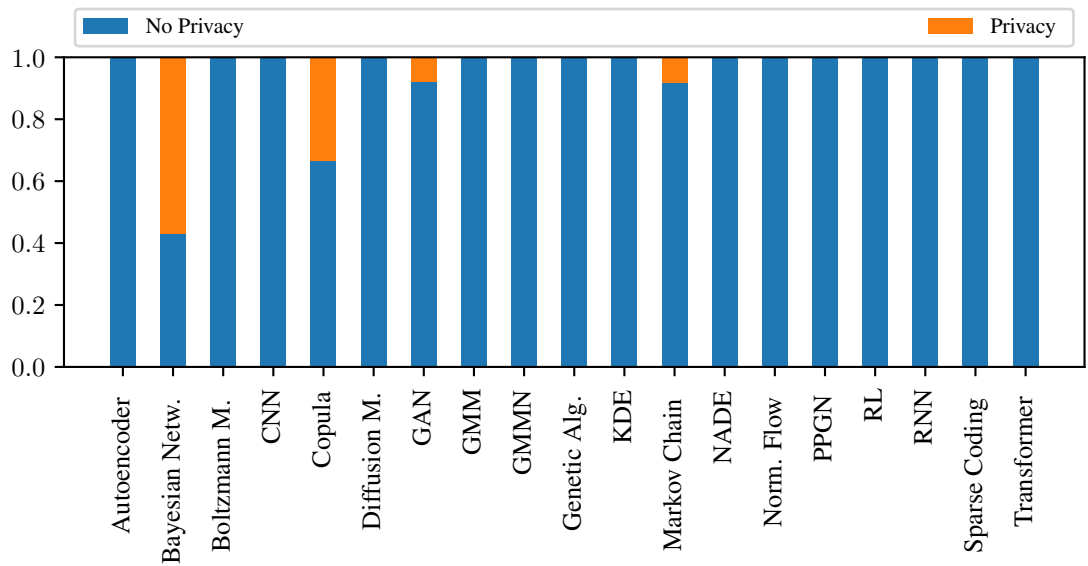}
    \end{center}
    \caption{Amount of models with privacy considerations per model category absolute (top) and normalized (bottom).}
    \label{fig:privacy_multi_v_bar}
\end{figure}

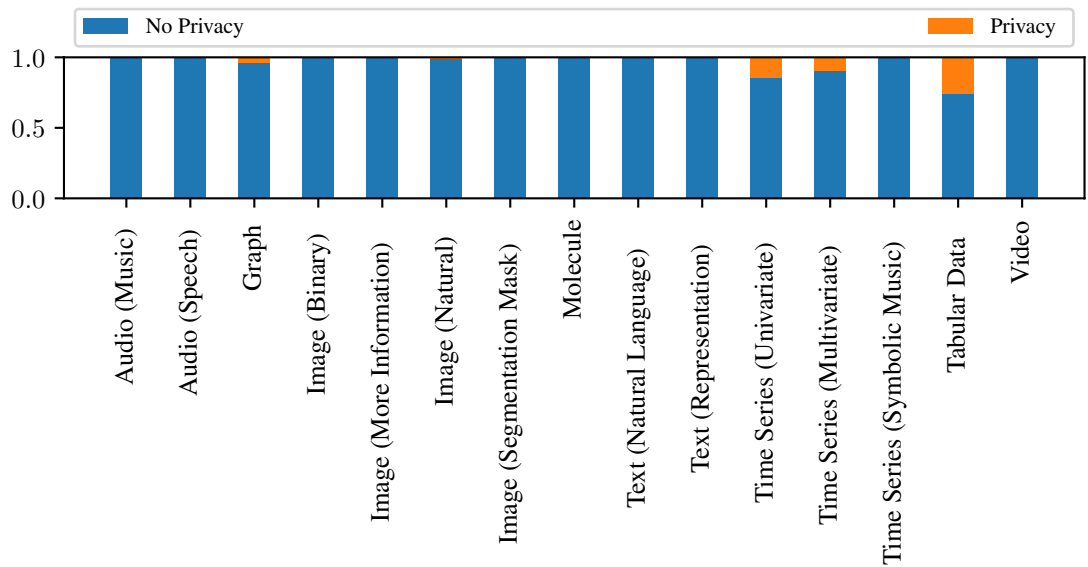
\begin{figure}[ht!]
    \begin{center}
        \input{figures/classification/privacy_output_data_multi_v_bar.pgf}
    \end{center}
    \caption{Privacy considerations per output data type absolute (top) and normalized (bottom).}
    \label{fig:privacy_output_data_multi_v_bar}
\end{figure}

Many modern applications of \ac{ML} are in areas such as the health care sector, where sensitive data of real persons has to be processed. This leads to the problem that large amounts of data for training and evaluation of models are required but cannot be disclosed. \ac{SDG} can be a sensible solution to this problem \cite{xu2020synthesizing}, but proving that no sensitive information is leaked by the \ac{SDG} model requires special techniques such as differential privacy~\cite{dwork2014algorithmic}.

In \autoref{fig:privacy_multi_v_bar}, we look at our surveyed models grouped by category and evaluate whether they provide a privacy guarantee. Usually, simplistic models with limited learning capabilities, such as \acp{BN}, Markov chains, and copulas, are used to generate private data. These models have the advantage that they can be inspected and modified by humans and with simple distribution modifications or noise injection. Thus, it becomes difficult to determine and extract information from real or personal data. More complex neural network models are seldom used, which is likely because they lack the advantages above and learn lots of details about their training data. That is, to make models more privacy-preserving, the realism and, ultimately, the utility of the data is reduced \cite{park2018data}. \acp{GAN}, however, are an exception: They have the advantage that their generator never sees the real training data, and it is demonstrated that with thoughtful design of the loss function and the model architecture, privacy guarantees for these models can be given \cite{park2018data}. 

In \autoref{fig:privacy_output_data_multi_v_bar}, we evaluate the types of private data provided by our presented models. The most used data type is tabular data, which encapsulates \acp{EHR} and most other personal information often encountered in the real world. The only other data types we encountered were graphs and time series, which can be used to store mobility trajectories of persons or other medical data (e.g., an electrocardiogram). We did not encounter private \ac{SDG} of higher-dimensional data like audio, images, video, or text. 

\subsection{Summary}\label{sec:summary}
In the earlier sections, we took a look at various aspects of \ac{SDG} models that we now summarize: The most popular model type is the \ac{GAN}, which is flexible and, by design, can create large amounts of data because it does not directly train on the training data. \acp{RNN} and \acp{CNN} are used as standalone models but also serve as building blocks for \acp{GAN}, \ac{RL} approaches, diffusion models, and autoencoders. They are suitable for generating sequences of samples and high-dimensional data (e.g., audio, images) respectively.

We find visual data generation, that is, images and videos, to be by far the most essential use case for \ac{SDG}. Virtual environments are exclusively applied to this domain. At the same time, \acp{GAN} and autoencoders are flexibly employed, and \acp{RNN} are preferably used for sequential data such as time series and text. Most models sample the data ``in one go''. In contrast, iterative sample refinement was less popular despite achieving similar results until recently, when diffusion models quickly became competitive against \acp{GAN} for image generation.

Our performance evaluation found that newer models usually outperform older ones. Especially \acp{GAN}, transformers, diffusion models, and \acp{RNN}, sometimes combined with \ac{RL}, often come out on top. But we also encountered two significant problems: First, no common standardized evaluation metric for \ac{SDG} models exists. \ac{FID}, \ac{IS}, \ac{NLL}, and human evaluation are frequently used but are not suited for all tasks. The second problem is the evaluation data, which is also not standardized. We identified several common data sets for different domains, but direct comparison of models was often impossible due to different metric and data set combinations. Another aspect of model performance that was seldom mentioned is the computational complexity of models, meaning the training and sampling resources and time and the amount of training data required to achieve the quality of the presented results. A more systematic evaluation approach could solve the comparability problem: To measure the quality of generated images, for example, a model author could compare against a fixed set of other popular models (e.g., \ac{DCGAN} \cite{radford2015unsupervised}), on a larger set of common data sets (e.g., celebA, MNIST, CIFAR-10) using predetermined metrics (e.g., \ac{FID}, \ac{IS}). This common foundation would allow for a more precise comparison of approaches.

According to our findings and also the research of others \cite{dankar2021fake}, privacy-preserving data generation is still in the early stages of development. It is limited to low-dimensional data like table entries or time series and is usually performed with simple models observable by humans like Markov chains or Bayesian networks. The only more complex model that seems to be suitable for this task is the \ac{GAN}, whose generator never sees the actual data. The main challenge also identified by others \cite{dankar2021fake} is the trade-off between data utility and privacy, which means that the modifications required to make the data more private reduce the representativeness. Another problem is that the more complex and powerful neural-network-based approaches are known \cite{chang2018privacy} to covertly encode individual samples from the training data in their parameters that can be reconstructed.

\subsection{Guideline for Synthetic Data Generation}\label{sec:guideline}

\begin{figure}[ht]
    \centering
    \begin{tikzpicture}[grow=right, treenode/.style = {shape=rectangle, rounded corners, draw, align=center}, font=\small, endnode/.style = {treenode, fill=blue!20, align=left}, sibling distance = 4cm, level distance = 4cm, level 2/.style = {sibling distance = 2cm}, edgenode/.style = {above, sloped, align=center, font=\footnotesize}, edge from parent/.append style={->}]
        \node [treenode] {Privacy?}
            child { node [treenode] {Task Complexity?}
                child { node [endnode] {Bayesian Network\\Markov Chain}
                    edge from parent node [edgenode, xshift=0.2cm] {simple}
                }
                child { node [endnode, yshift=-0.8cm] {GAN}
                    edge from parent node [edgenode] {complex}
                }
                edge from parent node [edgenode] {Yes}
            }
            child { node [treenode] {Data Type?}
                child { node [endnode, xshift=2cm] {Best Quality: Diffusion Models\\Faster: GANs\\Disentangled Features: VAEs\\Accurate Labels: Virtual Environments}
                    edge from parent node [edgenode] {Images}
                }
                child { node (seqnode) [endnode, xshift=2.5cm] {Very Simple: Markov chains\\Simple: RNNs\\With Guidance: RNNs \& RL\\Complex: Transformers}
                    edge from parent node [edgenode] {Text, Music,\\Time Series, Video}
                }
                child { node (structnode) [treenode] {Data Structure?}
                    child { node [endnode, xshift=0.5cm] {VAEs, CNNs}
                        edge from parent node [edgenode] {Fixed Size}
                    }
                    edge from parent node [edgenode] {Graphs,\\Molecules}
                }
                (structnode) edge [->] node [edgenode, xshift=0.28cm] {Sequ.} (seqnode)
                edge from parent node [edgenode] {No}
            };
    \end{tikzpicture}
    \caption{A simple decision tree for model selection. (Sequ. = Sequential)}
    \label{fig:guideline_decision_tree}
\end{figure}
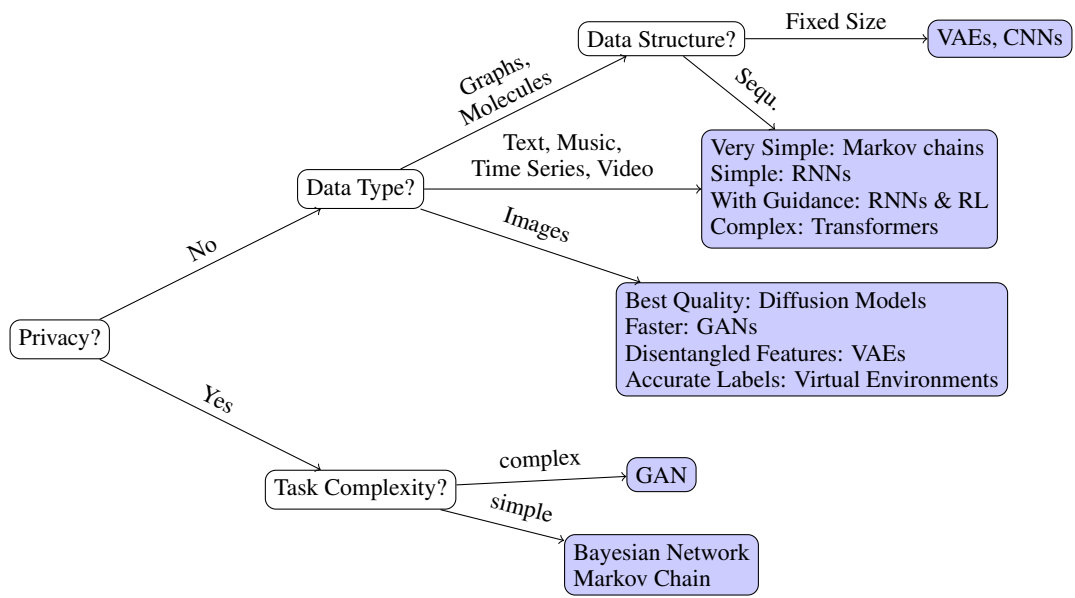

After classifying a large amount of \ac{SDG} models and presenting our findings in written and visual form, we finally provide a guideline for model selection for various use cases. We explain our suggestions in written form and finally illustrate them as a decision tree in \autoref{fig:guideline_decision_tree}:

\begin{itemize}
    \item Use recent and up-to-date models. More powerful models are usually more expensive to train.
    \item For image generation and related purposes, diffusion models and \acp{GAN} are the best options quality-wise. \acp{GAN} are better for situations where less training data is available. Autoencoders are a less powerful but also a less resource-demanding alternative.
    \item Generation of sequential data like symbolic music, time series, most graph/molecule representations, and text is a suitable task for transformers and \acp{RNN}. Markov chains are a less powerful but simple and efficient alternative that humans can inspect. 
    \item Autoencoders, especially \acp{VAE}, are good unsupervised feature learners that can disentangle and recombine features from training data to produce new data with desired properties. 
    \item \acp{CNN} are a common building block of models to make them suitable to process high-dimensional data like large-resolution images and audio waveforms, whose sizes otherwise significantly slow down the training process or increase model size beyond a processable point. 
    \item For the guidance of \ac{SDG} models towards producing data with specific properties (e.g., molecular properties/validity), two approaches have proven to be simple yet powerful: \ac{RL}, especially in combination with \acp{RNN}, allows users to define rewards for \ac{SDG} models to produce the desired output. Models like \acp{PPGN} pair a trained \ac{GAN} generator with a classifier to fine-tune for a specific task. Such approaches can greatly reduce the required training data. 
    \item For private data generation, Markov chains or Bayesian networks with modified probabilities or noise injection in the sampling process can be used. \acp{GAN} are a modern and more complex solution to private data generation when applied correctly \cite{park2018data}. 
    \item When visual data with highly accurate labels and high configurability is required, virtual environments are a good option, but they often require a considerable amount of human interaction to build and configure. 
\end{itemize}

%% file: figures/classification/composite_flow_graph_total.tex
\begin{tikzpicture}[node distance={1mm}, minimum width=1cm, font=\normalsize, thick, action/.style = {draw}, auto]
\node[action] (RNNl) [fill={rgb,255:red,197; green,176; blue,213}] {RNN};
\node[action] (RLl) [fill={rgb,255:red,148; green,103; blue,189}, right=of RNNl] {RL};
\node[action] (NADEl) [fill={rgb,255:red,214; green,39; blue,40}, right=of RLl] {NADE};
\node[action] (GANl) [fill={rgb,255:red,44; green,160; blue,44}, right=of NADEl] {GAN};
\node[action] (CNNl) [fill={rgb,255:red,255; green,127; blue,14}, right=of GANl] {CNN};
\node[action] (Boltzml) [fill={rgb,255:red,174; green,199; blue,232}, right=of CNNl] {Boltzmann M.};
\node[action] (AEl) [fill={rgb,255:red,31; green,119; blue,180}, right=of Boltzml] {Autoencoder};
\node[action] (AEr) [fill={rgb,255:red,31; green,119; blue,180}, below=4cm of RNNl] {Autoencoder};
\node[action] (CNNr) [fill={rgb,255:red,255; green,127; blue,14}, right=of AEr] {CNN};
\node[action] (Diffr) [fill={rgb,255:red,255; green,187; blue,120}, right=of CNNr] {Diffusion M.};
\node[action] (GANr) [fill={rgb,255:red,44; green,160; blue,44}, right=of Diffr] {GAN};
\node[action] (GMMNr) [fill={rgb,255:red,152; green,223; blue,138}, right=of GANr] {GMMN};
\node[action] (NFMr) [fill={rgb,255:red,255; green,152; blue,150}, right=of GMMNr] {Norm. Flow};
\node[action] (RLr) [fill={rgb,255:red,148; green,103; blue,189}, right=of NFMr] {RL};
\node[action] (RNNr) [fill={rgb,255:red,197; green,176; blue,213}, right=of RLr] {RNN};
\draw (RNNl) edge[color={rgb,255:red,197; green,176; blue,213}, out=270, in=90, line width=2.0mm, opacity=.25] node[pos=0.35, sloped, above, opacity=1] {20} (AEr);
\draw (CNNl) edge[color={rgb,255:red,255; green,127; blue,14}, out=270, in=90, line width=1.5mm, opacity=.25] node[pos=0.70, sloped, above, opacity=1] {15} (AEr);
\draw (GANl) edge[color={rgb,255:red,44; green,160; blue,44}, out=270, in=90, line width=0.4mm, opacity=.25] node[pos=0.10, sloped, above, opacity=1] {4} (AEr);
\draw (RNNl) edge[color={rgb,255:red,197; green,176; blue,213}, out=270, in=90, line width=0.3mm, opacity=.25] node[pos=0.40, sloped, above, opacity=1] {3} (CNNr);
\draw (CNNl) edge[color={rgb,255:red,255; green,127; blue,14}, out=270, in=90, line width=0.4mm, opacity=.25] node[pos=0.75, sloped, above, opacity=1] {4} (Diffr);
\draw (CNNl) edge[color={rgb,255:red,255; green,127; blue,14}, out=270, in=90, line width=10.9mm, opacity=.25] node[pos=0.10, sloped, above, opacity=1] {109} (GANr);
\draw (AEl) edge[color={rgb,255:red,31; green,119; blue,180}, out=270, in=90, line width=2.6mm, opacity=.25] node[pos=0.45, sloped, above, opacity=1] {26} (GANr);
\draw (RNNl) edge[color={rgb,255:red,197; green,176; blue,213}, out=270, in=90, line width=2.5mm, opacity=.25] node[pos=0.80, sloped, above, opacity=1] {25} (GANr);
\draw (RLl) edge[color={rgb,255:red,148; green,103; blue,189}, out=270, in=90, line width=0.3mm, opacity=.25] node[pos=0.15, sloped, above, opacity=1] {3} (GANr);
\draw (AEl) edge[color={rgb,255:red,31; green,119; blue,180}, out=270, in=90, line width=0.3mm, opacity=.25] node[pos=0.50, sloped, above, opacity=1] {3} (GMMNr);
\draw (CNNl) edge[color={rgb,255:red,255; green,127; blue,14}, out=270, in=90, line width=0.3mm, opacity=.25] node[pos=0.85, sloped, above, opacity=1] {3} (GMMNr);
\draw (CNNl) edge[color={rgb,255:red,255; green,127; blue,14}, out=270, in=90, line width=0.4mm, opacity=.25] node[pos=0.20, sloped, above, opacity=1] {4} (NFMr);
\draw (CNNl) edge[color={rgb,255:red,255; green,127; blue,14}, out=270, in=90, line width=0.5mm, opacity=.25] node[pos=0.55, sloped, above, opacity=1] {5} (RLr);
\draw (RNNl) edge[color={rgb,255:red,197; green,176; blue,213}, out=270, in=90, line width=0.8mm, opacity=.25] node[pos=0.90, sloped, above, opacity=1] {8} (RLr);
\draw (GANl) edge[color={rgb,255:red,44; green,160; blue,44}, out=270, in=90, line width=0.3mm, opacity=.25] node[pos=0.25, sloped, above, opacity=1] {3} (RLr);
\draw (Boltzml) edge[color={rgb,255:red,174; green,199; blue,232}, out=270, in=90, line width=0.4mm, opacity=.25] node[pos=0.60, sloped, above, opacity=1] {4} (RNNr);
\draw (CNNl) edge[color={rgb,255:red,255; green,127; blue,14}, out=270, in=90, line width=1.8mm, opacity=.25] node[pos=0.90, sloped, above, opacity=1] {18} (RNNr);
\draw (RLl) edge[color={rgb,255:red,148; green,103; blue,189}, out=270, in=90, line width=0.5mm, opacity=.25] node[pos=0.30, sloped, above, opacity=1] {5} (RNNr);
\draw (NADEl) edge[color={rgb,255:red,214; green,39; blue,40}, out=270, in=90, line width=0.3mm, opacity=.25] node[pos=0.65, sloped, above, opacity=1] {3} (RNNr);
\draw (AEl) edge[color={rgb,255:red,31; green,119; blue,180}, out=270, in=90, line width=0.4mm, opacity=.25] node[pos=0.10, sloped, above, opacity=1] {4} (RNNr);
\end{tikzpicture}

%% file: figures/classification/composite_flow_graph_normalized.tex
\begin{tikzpicture}[node distance={1mm}, minimum width=1cm, font=\normalsize, thick, action/.style = {draw}, auto]
\node[action] (RNNl) [fill={rgb,255:red,197; green,176; blue,213}] {RNN};
\node[action] (RLl) [fill={rgb,255:red,148; green,103; blue,189}, right=of RNNl] {RL};
\node[action] (NADEl) [fill={rgb,255:red,214; green,39; blue,40}, right=of RLl] {NADE};
\node[action] (GANl) [fill={rgb,255:red,44; green,160; blue,44}, right=of NADEl] {GAN};
\node[action] (CNNl) [fill={rgb,255:red,255; green,127; blue,14}, right=of GANl] {CNN};
\node[action] (Boltzml) [fill={rgb,255:red,174; green,199; blue,232}, right=of CNNl] {Boltzmann M.};
\node[action] (AEl) [fill={rgb,255:red,31; green,119; blue,180}, right=of Boltzml] {Autoencoder};
\node[action] (AEr) [fill={rgb,255:red,31; green,119; blue,180}, below=4cm of RNNl] {Autoencoder};
\node[action] (CNNr) [fill={rgb,255:red,255; green,127; blue,14}, right=of AEr] {CNN};
\node[action] (Diffr) [fill={rgb,255:red,255; green,187; blue,120}, right=of CNNr] {Diffusion M.};
\node[action] (GANr) [fill={rgb,255:red,44; green,160; blue,44}, right=of Diffr] {GAN};
\node[action] (GMMNr) [fill={rgb,255:red,152; green,223; blue,138}, right=of GANr] {GMMN};
\node[action] (NFMr) [fill={rgb,255:red,255; green,152; blue,150}, right=of GMMNr] {Norm. Flow};
\node[action] (RLr) [fill={rgb,255:red,148; green,103; blue,189}, right=of NFMr] {RL};
\node[action] (RNNr) [fill={rgb,255:red,197; green,176; blue,213}, right=of RLr] {RNN};
\draw (RNNl) edge[color={rgb,255:red,197; green,176; blue,213}, out=270, in=90, line width=2.81mm, opacity=.25] node[pos=0.35, sloped, above, opacity=1] {35} (AEr);
\draw (CNNl) edge[color={rgb,255:red,255; green,127; blue,14}, out=270, in=90, line width=2.11mm, opacity=.25] node[pos=0.70, sloped, above, opacity=1] {26} (AEr);
\draw (GANl) edge[color={rgb,255:red,44; green,160; blue,44}, out=270, in=90, line width=0.56mm, opacity=.25] node[pos=0.10, sloped, above, opacity=1] {7} (AEr);
\draw (RNNl) edge[color={rgb,255:red,197; green,176; blue,213}, out=270, in=90, line width=1.41mm, opacity=.25] node[pos=0.40, sloped, above, opacity=1] {17} (CNNr);
\draw (CNNl) edge[color={rgb,255:red,255; green,127; blue,14}, out=270, in=90, line width=6.40mm, opacity=.25] node[pos=0.75, sloped, above, opacity=1] {80} (Diffr);
\draw (CNNl) edge[color={rgb,255:red,255; green,127; blue,14}, out=270, in=90, line width=5.85mm, opacity=.25] node[pos=0.10, sloped, above, opacity=1] {73} (GANr);
\draw (AEl) edge[color={rgb,255:red,31; green,119; blue,180}, out=270, in=90, line width=1.40mm, opacity=.25] node[pos=0.45, sloped, above, opacity=1] {17} (GANr);
\draw (RNNl) edge[color={rgb,255:red,197; green,176; blue,213}, out=270, in=90, line width=1.34mm, opacity=.25] node[pos=0.80, sloped, above, opacity=1] {16} (GANr);
\draw (RLl) edge[color={rgb,255:red,148; green,103; blue,189}, out=270, in=90, line width=0.16mm, opacity=.25] node[pos=0.15, sloped, above, opacity=1] {2} (GANr);
\draw (AEl) edge[color={rgb,255:red,31; green,119; blue,180}, out=270, in=90, line width=3.00mm, opacity=.25] node[pos=0.50, sloped, above, opacity=1] {37} (GMMNr);
\draw (CNNl) edge[color={rgb,255:red,255; green,127; blue,14}, out=270, in=90, line width=3.00mm, opacity=.25] node[pos=0.85, sloped, above, opacity=1] {37} (GMMNr);
\draw (CNNl) edge[color={rgb,255:red,255; green,127; blue,14}, out=270, in=90, line width=4.00mm, opacity=.25] node[pos=0.20, sloped, above, opacity=1] {50} (NFMr);
\draw (CNNl) edge[color={rgb,255:red,255; green,127; blue,14}, out=270, in=90, line width=3.64mm, opacity=.25] node[pos=0.55, sloped, above, opacity=1] {45} (RLr);
\draw (RNNl) edge[color={rgb,255:red,197; green,176; blue,213}, out=270, in=90, line width=5.82mm, opacity=.25] node[pos=0.90, sloped, above, opacity=1] {72} (RLr);
\draw (GANl) edge[color={rgb,255:red,44; green,160; blue,44}, out=270, in=90, line width=2.18mm, opacity=.25] node[pos=0.25, sloped, above, opacity=1] {27} (RLr);
\draw (Boltzml) edge[color={rgb,255:red,174; green,199; blue,232}, out=270, in=90, line width=0.41mm, opacity=.25] node[pos=0.60, sloped, above, opacity=1] {5} (RNNr);
\draw (CNNl) edge[color={rgb,255:red,255; green,127; blue,14}, out=270, in=90, line width=1.85mm, opacity=.25] node[pos=0.90, sloped, above, opacity=1] {23} (RNNr);
\draw (RLl) edge[color={rgb,255:red,148; green,103; blue,189}, out=270, in=90, line width=0.51mm, opacity=.25] node[pos=0.30, sloped, above, opacity=1] {6} (RNNr);
\draw (NADEl) edge[color={rgb,255:red,214; green,39; blue,40}, out=270, in=90, line width=0.31mm, opacity=.25] node[pos=0.65, sloped, above, opacity=1] {3} (RNNr);
\draw (AEl) edge[color={rgb,255:red,31; green,119; blue,180}, out=270, in=90, line width=0.41mm, opacity=.25] node[pos=0.10, sloped, above, opacity=1] {5} (RNNr);
\end{tikzpicture}

%% file: figures/classification/hierarchy_predecessors_gan.tex
\begin{tikzpicture}[node distance={1mm}, font=\tiny, thick, year/.style = {draw, align=left}, entry/.style = {draw, circle, minimum width=3mm, align=center, fill=white, opacity=.85}, auto]
\node[year] (year20140) {2014};
\node[entry] (mirza2014conditional) [right=of year20140] {\cite{mirza2014conditional}\\cGAN};
\node[entry] (goodfellow2014generative) [right=of mirza2014conditional] {\cite{goodfellow2014generative}\\GAN};
\node[entry] (gauthier2014conditional) [right=of goodfellow2014generative] {\cite{gauthier2014conditional}};
\node[year] (year20150) [below=10mm of year20140] {2015};
\node[entry] (makhzani2015adversarial) [right=of year20150] {\cite{makhzani2015adversarial}\\AAE};
\node[entry] (radford2015unsupervised) [right=of makhzani2015adversarial] {\cite{radford2015unsupervised}\\DCGAN};
\node[entry] (springenberg2015unsupervised) [right=of radford2015unsupervised] {\cite{springenberg2015unsupervised}\\CatGAN};
\node[year] (year20160) [below=10mm of year20150] {2016};
\node[entry] (perarnau2016invertible) [right=of year20160] {\cite{perarnau2016invertible}\\IcGAN};
\node[entry] (karacan2016learning) [right=of perarnau2016invertible] {\cite{karacan2016learning}\\AL\\CGAN};
\node[entry] (durugkar2016generative) [right=of karacan2016learning] {\cite{durugkar2016generative}\\GMAN};
\node[entry] (zhu2016generative) [right=of durugkar2016generative] {\cite{zhu2016generative}\\iGAN};
\node[entry] (reed2016learning) [right=of zhu2016generative] {\cite{reed2016learning}\\GAWWN};
\node[entry] (liu2016coupled) [right=of reed2016learning] {\cite{liu2016coupled}\\CoGAN};
\node[entry] (reed2016generative) [right=of liu2016coupled] {\cite{reed2016generative}};
\node[entry] (salimans2016improved) [right=of reed2016generative] {\cite{salimans2016improved}\\Improved\\GAN};
\node[entry] (che2016mode) [right=of salimans2016improved] {\cite{che2016mode}\\Mode\\Reg.\\GAN};
\node[year] (year20161) [below=10mm of year20160] {2016};
\node[entry] (vondrick2016generating) [right=of year20161] {\cite{vondrick2016generating}\\VGAN};
\node[entry] (donahue2016adversarial) [right=of vondrick2016generating] {\cite{donahue2016adversarial}\\BiGAN};
\node[entry] (chen2016infogan) [right=of donahue2016adversarial] {\cite{chen2016infogan}\\InfoGAN};
\node[year] (year20170) [below=10mm of year20161] {2017};
\node[entry] (choi2017stargan) [right=of year20170] {\cite{choi2017stargan}\\StarGAN};
\node[entry] (kadurin2017drugan) [right=of choi2017stargan] {\cite{kadurin2017drugan}\\druGAN};
\node[entry] (elgammal2017can) [right=of kadurin2017drugan] {\cite{elgammal2017can}\\CAN};
\node[entry] (yang2017midinet) [right=of elgammal2017can] {\cite{yang2017midinet}\\MidiNet};
\node[entry] (ehsani2017segan) [right=of yang2017midinet] {\cite{ehsani2017segan}\\SeGAN};
\node[entry] (warde-farley2017improving) [right=of ehsani2017segan] {\cite{warde-farley2017improving}};
\node[entry] (gulrajani2017improved) [right=of warde-farley2017improving] {\cite{gulrajani2017improved}\\Improved\\WGAN};
\node[entry] (odena2017conditional) [right=of gulrajani2017improved] {\cite{odena2017conditional}\\AC-GAN};
\node[entry] (karras2017progressive) [right=of odena2017conditional] {\cite{karras2017progressive}\\Progr.\\GAN};
\node[year] (year20171) [below=10mm of year20170] {2017};
\node[entry] (srivastava2017veegan) [right=of year20171] {\cite{srivastava2017veegan}\\VeeGAN};
\node[entry] (zhu2017toward) [right=of srivastava2017veegan] {\cite{zhu2017toward}\\Bicycle\\GAN};
\node[entry] (zhu2017unpaired) [right=of zhu2017toward] {\cite{zhu2017unpaired}\\Cycle\\GAN};
\node[entry] (choi2017generating) [right=of zhu2017unpaired] {\cite{choi2017generating}\\medGAN};
\node[entry] (yi2017dualgan) [right=of choi2017generating] {\cite{yi2017dualgan}\\DualGAN};
\node[entry] (gorijala2017image) [right=of yi2017dualgan] {\cite{gorijala2017image}\\ViGAN};
\node[entry] (liu2017unsupervised) [right=of gorijala2017image] {\cite{liu2017unsupervised}};
\node[year] (year20172) [below=10mm of year20171] {2017};
\node[entry] (hjelm2017boundary) [right=of year20172] {\cite{hjelm2017boundary}\\BGAN};
\node[entry] (saito2017temporal) [right=of hjelm2017boundary] {\cite{saito2017temporal}\\Temp.\\GAN};
\node[entry] (xu2017gang) [right=of saito2017temporal] {\cite{xu2017gang}\\GoGAN};
\node[entry] (dong2017semantic) [right=of xu2017gang] {\cite{dong2017semantic}};
\node[entry] (tolstikhin2017wasserstein) [right=of dong2017semantic] {\cite{tolstikhin2017wasserstein}\\WAE\\GAN};
\node[entry] (isola2017image) [right=of tolstikhin2017wasserstein] {\cite{isola2017image}\\pix2pix};
\node[entry] (kim2017adversarially) [right=of isola2017image] {\cite{kim2017adversarially}\\ARAE};
\node[entry] (kadurin2017cornucopia) [right=of kim2017adversarially] {\cite{kadurin2017cornucopia}};
\node[year] (year20173) [below=10mm of year20172] {2017};
\node[entry] (mroueh2017fisher) [right=of year20173] {\cite{mroueh2017fisher}\\Fisher\\GAN};
\node[entry] (grinblat2017class) [right=of mroueh2017fisher] {\cite{grinblat2017class}\\Splitting\\GAN};
\node[entry] (volkhonskiy2017steganographic) [right=of grinblat2017class] {\cite{volkhonskiy2017steganographic}\\Stegan.\\GAN};
\node[entry] (mroueh2017mcgan) [right=of volkhonskiy2017steganographic] {\cite{mroueh2017mcgan}\\McGAN};
\node[entry] (mao2017least) [right=of mroueh2017mcgan] {\cite{mao2017least}\\LSGAN};
\node[entry] (arjovsky2017wasserstein) [right=of mao2017least] {\cite{arjovsky2017wasserstein}\\WGAN};
\node[entry] (kocaoglu2017causalgan) [right=of arjovsky2017wasserstein] {\cite{kocaoglu2017causalgan}\\Causal\\GAN};
\node[entry] (guibas2017synthetic) [right=of kocaoglu2017causalgan] {\cite{guibas2017synthetic}};
\node[year] (year20174) [below=10mm of year20173] {2017};
\node[entry] (lu2017conditional) [right=of year20174] {\cite{lu2017conditional}\\Cond.\\Cycle\\GAN};
\node[entry] (costa2017end) [right=of lu2017conditional] {\cite{costa2017end}};
\node[year] (year20180) [below=10mm of year20174] {2018};
\node[entry] (nam2018text) [right=of year20180] {\cite{nam2018text}\\TAGAN};
\node[entry] (triastcyn2018generating) [right=of nam2018text] {\cite{triastcyn2018generating}};
\node[entry] (hoang2018mgan) [right=of triastcyn2018generating] {\cite{hoang2018mgan}\\MGAN};
\node[entry] (dong2018musegan) [right=of hoang2018mgan] {\cite{dong2018musegan}\\Muse\\GAN};
\node[entry] (donahue2018adversarial) [right=of dong2018musegan] {\cite{donahue2018adversarial}\\WaveGAN\\SpecGAN};
\node[entry] (trigueros2018generating) [right=of donahue2018adversarial] {\cite{trigueros2018generating}\\Cond.\\PGGAN};
\node[entry] (park2018data) [right=of trigueros2018generating] {\cite{park2018data}\\table\\GAN};
\node[entry] (zhang2018synthetic) [right=of park2018data] {\cite{zhang2018synthetic}};
\node[entry] (genevay2018learning) [right=of zhang2018synthetic] {\cite{genevay2018learning}\\SinkhornGAN};
\node[year] (year20181) [below=10mm of year20180] {2018};
\node[entry] (wei2018improving) [right=of year20181] {\cite{wei2018improving}\\CT-GAN};
\node[entry] (miyato2018spectral) [right=of wei2018improving] {\cite{miyato2018spectral}\\SN-GAN};
\node[entry] (anand2018generative) [right=of miyato2018spectral] {\cite{anand2018generative}};
\node[entry] (jaiswal2018capsulegan) [right=of anand2018generative] {\cite{jaiswal2018capsulegan}\\Capsule\\GAN};
\node[year] (year20190) [below=10mm of year20181] {2019};
\node[entry] (fan2019labeled) [right=of year20190] {\cite{fan2019labeled}\\LGGAN};
\node[entry] (baowaly2019synthesizing) [right=of fan2019labeled] {\cite{baowaly2019synthesizing}\\med\\WGAN};
\node[entry] (donahue2019large) [right=of baowaly2019synthesizing] {\cite{donahue2019large}\\Big\\BiGAN};
\node[entry] (karras2019style) [right=of donahue2019large] {\cite{karras2019style}\\Style\\GAN};
\node[entry] (engel2019gansynth) [right=of karras2019style] {\cite{engel2019gansynth}\\GAN\\synth};
\node[entry] (brock2019large) [right=of engel2019gansynth] {\cite{brock2019large}\\BigGAN};
\node[entry] (gong2019autogan) [right=of brock2019large] {\cite{gong2019autogan}\\AutoGAN};
\node[entry] (luvcic2019high) [right=of gong2019autogan] {\cite{luvcic2019high}};
\node[entry] (zhang2019self) [right=of luvcic2019high] {\cite{zhang2019self}\\SAGAN};
\node[year] (year20191) [below=10mm of year20190] {2019};
\node[entry] (lin2019cocogan) [right=of year20191] {\cite{lin2019cocogan}\\COCO-GAN};
\node[year] (year20200) [below=10mm of year20191] {2020};
\node[entry] (gao2020adversarial) [right=of year20200] {\cite{gao2020adversarial}\\AdversarialNAS};
\node[entry] (zhao2020differentiable) [right=of gao2020adversarial] {\cite{zhao2020differentiable}};
\node[entry] (cot-gan) [right=of zhao2020differentiable] {\cite{cot-gan}\\COT-GAN};
\node[entry] (islam2020gan) [right=of cot-gan] {\cite{islam2020gan}};
\node[entry] (alam2020synthetic) [right=of islam2020gan] {\cite{alam2020synthetic}};
\node[entry] (karras2020analyzing) [right=of alam2020synthetic] {\cite{karras2020analyzing}\\Style\\GAN2};
\node[entry] (maziarka2020mol) [right=of karras2020analyzing] {\cite{maziarka2020mol}\\Mol\\Cycle\\GAN};
\node[entry] (yale2020generation) [right=of maziarka2020mol] {\cite{yale2020generation}\\Health\\GAN};
\node[year] (year20210) [below=10mm of year20200] {2021};
\node[entry] (amin2021quantum) [right=of year20210] {\cite{amin2021quantum}};
\node[entry] (gans-for-ecg) [right=of amin2021quantum] {\cite{gans-for-ecg}\\Pulse2Pulse};
\node[entry] (imtiaz2021synthetic) [right=of gans-for-ecg] {\cite{imtiaz2021synthetic}};
\node[entry] (trans-gan) [right=of imtiaz2021synthetic] {\cite{trans-gan}\\TransGAN};
\node[year] (year20220) [below=10mm of year20210] {2022};
\node[entry] (tts-gan) [right=of year20220] {\cite{tts-gan}\\TTS-GAN};
\begin{pgfonlayer}{bg}\draw[->] (radford2015unsupervised) -- (elgammal2017can);
\draw[->] (radford2015unsupervised) -- (donahue2018adversarial);
\draw[->] (radford2015unsupervised) -- (durugkar2016generative);
\draw[->] (radford2015unsupervised) -- (che2016mode);
\draw[->] (arjovsky2017wasserstein) -- (gulrajani2017improved);
\draw[->] (arjovsky2017wasserstein) -- (kim2017adversarially);
\draw[->] (radford2015unsupervised) -- (mroueh2017mcgan);
\draw[->] (gulrajani2017improved) -- (mroueh2017fisher);
\draw[->] (radford2015unsupervised) -- (mroueh2017fisher);
\draw[->] (radford2015unsupervised) -- (saito2017temporal);
\draw[->] (arjovsky2017wasserstein) -- (saito2017temporal);
\draw[->] (radford2015unsupervised) -- (srivastava2017veegan);
\draw[->] (radford2015unsupervised) -- (hjelm2017boundary);
\draw[->] (radford2015unsupervised) -- (yang2017midinet);
\draw[->] (arjovsky2017wasserstein) -- (xu2017gang);
\draw[->] (gulrajani2017improved) -- (dong2018musegan);
\draw[->] (radford2015unsupervised) -- (jaiswal2018capsulegan);
\draw[->] (radford2015unsupervised) -- (park2018data);
\draw[->] (gulrajani2017improved) -- (wei2018improving);
\draw[->] (radford2015unsupervised) -- (anand2018generative);
\draw[->] (karras2017progressive) -- (karras2019style);
\draw[->] (choi2017generating) -- (baowaly2019synthesizing);
\draw[->] (karras2019style) -- (karras2020analyzing);
\draw[->] (choi2017generating) -- (yale2020generation);
\draw[->] (gulrajani2017improved) -- (yale2020generation);
\draw[->] (karras2017progressive) -- (alam2020synthetic);
\draw[->] (hjelm2017boundary) -- (imtiaz2021synthetic);
\draw[->] (radford2015unsupervised) -- (zhu2016generative);
\draw[->] (radford2015unsupervised) -- (reed2016generative);
\draw[->] (reed2016generative) -- (reed2016learning);
\draw[->] (radford2015unsupervised) -- (isola2017image);
\draw[->] (zhu2017unpaired) -- (choi2017stargan);
\draw[->] (radford2015unsupervised) -- (kocaoglu2017causalgan);
\draw[->] (zhang2019self) -- (brock2019large);
\draw[->] (brock2019large) -- (luvcic2019high);
\draw[->] (mirza2014conditional) -- (gauthier2014conditional);
\draw[->] (radford2015unsupervised) -- (perarnau2016invertible);
\draw[->] (radford2015unsupervised) -- (karacan2016learning);
\draw[->] (radford2015unsupervised) -- (vondrick2016generating);
\draw[->] (isola2017image) -- (ehsani2017segan);
\draw[->] (isola2017image) -- (yi2017dualgan);
\draw[->] (radford2015unsupervised) -- (volkhonskiy2017steganographic);
\draw[->] (chen2016infogan) -- (gorijala2017image);
\draw[->] (mao2017least) -- (zhu2017toward);
\draw[->] (liu2016coupled) -- (liu2017unsupervised);
\draw[->] (zhu2017unpaired) -- (lu2017conditional);
\draw[->] (isola2017image) -- (zhang2018synthetic);
\draw[->] (dong2017semantic) -- (nam2018text);
\draw[->] (karras2017progressive) -- (trigueros2018generating);
\draw[->] (gulrajani2017improved) -- (triastcyn2018generating);
\draw[->] (karras2017progressive) -- (engel2019gansynth);
\draw[->] (wei2018improving) -- (fan2019labeled);
\draw[->] (zhu2017unpaired) -- (maziarka2020mol);
\draw[->] (radford2015unsupervised) -- (islam2020gan);
\draw[->] (radford2015unsupervised) -- (amin2021quantum);
\draw[->] (radford2015unsupervised) -- (guibas2017synthetic);
\draw[->] (brock2019large) -- (donahue2019large);
\draw[->] (donahue2016adversarial) -- (donahue2019large);
\draw[->] (makhzani2015adversarial) -- (tolstikhin2017wasserstein);
\draw[->] (radford2015unsupervised) -- (tolstikhin2017wasserstein);
\draw[->] (kadurin2017cornucopia) -- (kadurin2017drugan);
\draw[->] (makhzani2015adversarial) -- (costa2017end);
\draw[->] (trans-gan) -- (tts-gan);
\draw[->] (zhao2020differentiable) -- (trans-gan);
\draw[->] (miyato2018spectral) -- (trans-gan);
\draw[->] (goodfellow2014generative) -- (miyato2018spectral);
\draw[->] (gong2019autogan) -- (gao2020adversarial);
\draw[->] (genevay2018learning) -- (cot-gan);
\draw[->] (donahue2018adversarial) -- (gans-for-ecg);
\draw[->] (arjovsky2017wasserstein) -- (genevay2018learning);
\draw[->] (radford2015unsupervised) -- (hoang2018mgan);
\draw[->] (radford2015unsupervised) -- (salimans2016improved);
\draw[->] (odena2017conditional) -- (grinblat2017class);
\draw[->] (salimans2016improved) -- (warde-farley2017improving);
\draw[->] (goodfellow2014generative) -- (springenberg2015unsupervised);
\end{pgfonlayer}\end{tikzpicture}

%% file: figures/classification/hierarchy_predecessors_rnn.tex
\begin{tikzpicture}[node distance={1mm}, font=\tiny, thick, year/.style = {draw, align=left}, entry/.style = {draw, circle, minimum width=3mm, align=center, fill=white, opacity=.85}, auto]
\node[year] (year20120) {2012};
\node[entry] (bengio2012advances) [right=of year20120] {\cite{bengio2012advances}\\RNN-NADE};
\node[year] (year20130) [below=10mm of year20120] {2013};
\node[entry] (graves2013generating) [right=of year20130] {\cite{graves2013generating}};
\node[year] (year20140) [below=10mm of year20130] {2014};
\node[entry] (goel2014polyphonic) [right=of year20140] {\cite{goel2014polyphonic}\\RNN-DBN};
\node[year] (year20150) [below=10mm of year20140] {2015};
\node[entry] (mansimov2015generating) [right=of year20150] {\cite{mansimov2015generating}\\align\\DRAW};
\node[entry] (xu2015show) [right=of mansimov2015generating] {\cite{xu2015show}};
\node[entry] (vohra2015modeling) [right=of xu2015show] {\cite{vohra2015modeling}\\DBN\\LSTM};
\node[year] (year20160) [below=10mm of year20150] {2016};
\node[entry] (mehri2016samplernn) [right=of year20160] {\cite{mehri2016samplernn}\\Sample\\RNN};
\node[entry] (colombo2016algorithmic) [right=of mehri2016samplernn] {\cite{colombo2016algorithmic}};
\node[entry] (jaques2016tuning) [right=of colombo2016algorithmic] {\cite{jaques2016tuning}\\Note-RNN};
\node[entry] (lamb2016professor) [right=of jaques2016tuning] {\cite{lamb2016professor}};
\node[year] (year20170) [below=10mm of year20160] {2017};
\node[entry] (johnson2017generating) [right=of year20170] {\cite{johnson2017generating}\\TP-LSTM-\\NADE\\BALSTM};
\node[entry] (jaques2017sequence) [right=of johnson2017generating] {\cite{jaques2017sequence}\\Sequence\\Tutor};
\node[entry] (sotelo2017char2wav) [right=of jaques2017sequence] {\cite{sotelo2017char2wav}\\Char2Wav};
\node[entry] (colombo2017deep) [right=of sotelo2017char2wav] {\cite{colombo2017deep}\\DAC};
\node[entry] (liu2017attention) [right=of colombo2017deep] {\cite{liu2017attention}};
\node[year] (year20180) [below=10mm of year20170] {2018};
\node[entry] (manzelli2018conditioning) [right=of year20180] {\cite{manzelli2018conditioning}};
\node[entry] (li2018learning) [right=of manzelli2018conditioning] {\cite{li2018learning}\\GNN};
\node[entry] (mao2018deepj) [right=of li2018learning] {\cite{mao2018deepj}\\DeepJ};
\node[entry] (you2018graphrnn) [right=of mao2018deepj] {\cite{you2018graphrnn}\\GraphRNN};
\node[entry] (shen2018natural) [right=of you2018graphrnn] {\cite{shen2018natural}\\Tacotron\\2};
\node[year] (year20190) [below=10mm of year20180] {2019};
\node[entry] (popova2019molecularrnn) [right=of year20190] {\cite{popova2019molecularrnn}\\Molecular\\RNN};
\node[entry] (liao2019efficient) [right=of popova2019molecularrnn] {\cite{liao2019efficient}\\GRAN};
\begin{pgfonlayer}{bg}\draw[->] (goel2014polyphonic) -- (vohra2015modeling);
\draw[->] (mehri2016samplernn) -- (sotelo2017char2wav);
\draw[->] (graves2013generating) -- (sotelo2017char2wav);
\draw[->] (colombo2016algorithmic) -- (colombo2017deep);
\draw[->] (xu2015show) -- (liu2017attention);
\draw[->] (jaques2016tuning) -- (jaques2017sequence);
\draw[->] (bengio2012advances) -- (johnson2017generating);
\draw[->] (johnson2017generating) -- (manzelli2018conditioning);
\draw[->] (johnson2017generating) -- (mao2018deepj);
\draw[->] (you2018graphrnn) -- (liao2019efficient);
\draw[->] (li2018learning) -- (liao2019efficient);
\draw[->] (you2018graphrnn) -- (popova2019molecularrnn);
\draw[->] (graves2013generating) -- (lamb2016professor);
\end{pgfonlayer}\end{tikzpicture}

%% file: figures/classification/hierarchy_predecessors_diff.tex
\begin{tikzpicture}[node distance={1mm}, font=\tiny, thick, year/.style = {draw, align=left}, entry/.style = {draw, circle, minimum width=3mm, align=center, fill=white, opacity=.85}, auto]
\node[year] (year20150) {2015};
\node[entry] (dickstein2015deep) [right=of year20150] {\cite{dickstein2015deep}};
\node[year] (year20200) [below=10mm of year20150] {2020};
\node[entry] (ho2020denoising) [right=of year20200] {\cite{ho2020denoising}\\DDPM};
\node[year] (year20210) [below=10mm of year20200] {2021};
\node[entry] (nichol2021improved) [right=of year20210] {\cite{nichol2021improved}\\Improved\\DDPM};
\node[entry] (dhariwal2021diffusion) [right=of nichol2021improved] {\cite{dhariwal2021diffusion}\\ADM-G/U};
\begin{pgfonlayer}{bg}\draw[->] (dickstein2015deep) -- (ho2020denoising);
\draw[->] (ho2020denoising) -- (nichol2021improved);
\draw[->] (nichol2021improved) -- (dhariwal2021diffusion);
\end{pgfonlayer}\end{tikzpicture}

%% file: figures/classification/hierarchy_predecessors_ae.tex
\begin{tikzpicture}[node distance={1mm}, font=\tiny, thick, year/.style = {draw, align=left}, entry/.style = {draw, circle, minimum width=3mm, align=center, fill=white, opacity=.85}, auto]
\node[year] (year20130) {2013};
\node[entry] (kingma2013auto) [right=of year20130] {\cite{kingma2013auto}\\VAE};
\node[year] (year20150) [below=10mm of year20130] {2015};
\node[entry] (bowman2015generating) [right=of year20150] {\cite{bowman2015generating}};
\node[year] (year20160) [below=10mm of year20150] {2016};
\node[entry] (gomez2016automatic) [right=of year20160] {\cite{gomez2016automatic}};
\node[entry] (higgins2016beta) [right=of gomez2016automatic] {\cite{higgins2016beta}\\$\beta$-VAE};
\node[year] (year20170) [below=10mm of year20160] {2017};
\node[entry] (kusner2017grammar) [right=of year20170] {\cite{kusner2017grammar}\\Grammar\\VAE};
\node[entry] (roberts2017hierarchical) [right=of kusner2017grammar] {\cite{roberts2017hierarchical}};
\node[entry] (ha2017neural) [right=of roberts2017hierarchical] {\cite{ha2017neural}\\Sketch\\RNN};
\node[entry] (tikhonov2017music) [right=of ha2017neural] {\cite{tikhonov2017music}};
\node[year] (year20180) [below=10mm of year20170] {2018};
\node[entry] (jorgensen2018machine) [right=of year20180] {\cite{jorgensen2018machine}};
\node[entry] (simonovsky2018graphvae) [right=of jorgensen2018machine] {\cite{simonovsky2018graphvae}\\GraphVAE};
\node[entry] (roberts2018hierarchical) [right=of simonovsky2018graphvae] {\cite{roberts2018hierarchical}\\MusicVAE};
\node[year] (year20190) [below=10mm of year20180] {2019};
\node[entry] (wang2019topic) [right=of year20190] {\cite{wang2019topic}\\TGVAE};
\node[year] (year20200) [below=10mm of year20190] {2020};
\node[entry] (flam2020graph) [right=of year20200] {\cite{flam2020graph}\\MPGVAE};
\begin{pgfonlayer}{bg}\draw[->] (kingma2013auto) -- (higgins2016beta);
\draw[->] (bowman2015generating) -- (gomez2016automatic);
\draw[->] (bowman2015generating) -- (roberts2017hierarchical);
\draw[->] (ha2017neural) -- (roberts2017hierarchical);
\draw[->] (bowman2015generating) -- (tikhonov2017music);
\draw[->] (kusner2017grammar) -- (jorgensen2018machine);
\draw[->] (roberts2017hierarchical) -- (roberts2018hierarchical);
\draw[->] (bowman2015generating) -- (wang2019topic);
\draw[->] (simonovsky2018graphvae) -- (flam2020graph);
\end{pgfonlayer}\end{tikzpicture}

%% file: figures/classification/data_set_counts.pgf
\begingroup%
\makeatletter%
\begin{pgfpicture}%
\pgfpathrectangle{\pgfpointorigin}{\pgfqpoint{5.816308in}{2.904096in}}%
\pgfusepath{use as bounding box, clip}%
\begin{pgfscope}%
\pgfsetbuttcap%
\pgfsetmiterjoin%
\definecolor{currentfill}{rgb}{1.000000,1.000000,1.000000}%
\pgfsetfillcolor{currentfill}%
\pgfsetlinewidth{0.000000pt}%
\definecolor{currentstroke}{rgb}{1.000000,1.000000,1.000000}%
\pgfsetstrokecolor{currentstroke}%
\pgfsetdash{}{0pt}%
\pgfpathmoveto{\pgfqpoint{0.000000in}{0.000000in}}%
\pgfpathlineto{\pgfqpoint{5.816308in}{0.000000in}}%
\pgfpathlineto{\pgfqpoint{5.816308in}{2.904096in}}%
\pgfpathlineto{\pgfqpoint{0.000000in}{2.904096in}}%
\pgfpathlineto{\pgfqpoint{0.000000in}{0.000000in}}%
\pgfpathclose%
\pgfusepath{fill}%
\end{pgfscope}%
\begin{pgfscope}%
\pgfsetbuttcap%
\pgfsetmiterjoin%
\definecolor{currentfill}{rgb}{1.000000,1.000000,1.000000}%
\pgfsetfillcolor{currentfill}%
\pgfsetlinewidth{0.000000pt}%
\definecolor{currentstroke}{rgb}{0.000000,0.000000,0.000000}%
\pgfsetstrokecolor{currentstroke}%
\pgfsetstrokeopacity{0.000000}%
\pgfsetdash{}{0pt}%
\pgfpathmoveto{\pgfqpoint{0.405556in}{1.293675in}}%
\pgfpathlineto{\pgfqpoint{5.716308in}{1.293675in}}%
\pgfpathlineto{\pgfqpoint{5.716308in}{2.804096in}}%
\pgfpathlineto{\pgfqpoint{0.405556in}{2.804096in}}%
\pgfpathlineto{\pgfqpoint{0.405556in}{1.293675in}}%
\pgfpathclose%
\pgfusepath{fill}%
\end{pgfscope}%
\begin{pgfscope}%
\pgfpathrectangle{\pgfqpoint{0.405556in}{1.293675in}}{\pgfqpoint{5.310752in}{1.510421in}}%
\pgfusepath{clip}%
\pgfsetbuttcap%
\pgfsetmiterjoin%
\definecolor{currentfill}{rgb}{0.121569,0.466667,0.705882}%
\pgfsetfillcolor{currentfill}%
\pgfsetlinewidth{0.000000pt}%
\definecolor{currentstroke}{rgb}{0.000000,0.000000,0.000000}%
\pgfsetstrokecolor{currentstroke}%
\pgfsetstrokeopacity{0.000000}%
\pgfsetdash{}{0pt}%
\pgfpathmoveto{\pgfqpoint{0.646954in}{1.293675in}}%
\pgfpathlineto{\pgfqpoint{0.802695in}{1.293675in}}%
\pgfpathlineto{\pgfqpoint{0.802695in}{2.666785in}}%
\pgfpathlineto{\pgfqpoint{0.646954in}{2.666785in}}%
\pgfpathlineto{\pgfqpoint{0.646954in}{1.293675in}}%
\pgfpathclose%
\pgfusepath{fill}%
\end{pgfscope}%
\begin{pgfscope}%
\pgfpathrectangle{\pgfqpoint{0.405556in}{1.293675in}}{\pgfqpoint{5.310752in}{1.510421in}}%
\pgfusepath{clip}%
\pgfsetbuttcap%
\pgfsetmiterjoin%
\definecolor{currentfill}{rgb}{0.121569,0.466667,0.705882}%
\pgfsetfillcolor{currentfill}%
\pgfsetlinewidth{0.000000pt}%
\definecolor{currentstroke}{rgb}{0.000000,0.000000,0.000000}%
\pgfsetstrokecolor{currentstroke}%
\pgfsetstrokeopacity{0.000000}%
\pgfsetdash{}{0pt}%
\pgfpathmoveto{\pgfqpoint{1.036305in}{1.293675in}}%
\pgfpathlineto{\pgfqpoint{1.192046in}{1.293675in}}%
\pgfpathlineto{\pgfqpoint{1.192046in}{1.525284in}}%
\pgfpathlineto{\pgfqpoint{1.036305in}{1.525284in}}%
\pgfpathlineto{\pgfqpoint{1.036305in}{1.293675in}}%
\pgfpathclose%
\pgfusepath{fill}%
\end{pgfscope}%
\begin{pgfscope}%
\pgfpathrectangle{\pgfqpoint{0.405556in}{1.293675in}}{\pgfqpoint{5.310752in}{1.510421in}}%
\pgfusepath{clip}%
\pgfsetbuttcap%
\pgfsetmiterjoin%
\definecolor{currentfill}{rgb}{0.121569,0.466667,0.705882}%
\pgfsetfillcolor{currentfill}%
\pgfsetlinewidth{0.000000pt}%
\definecolor{currentstroke}{rgb}{0.000000,0.000000,0.000000}%
\pgfsetstrokecolor{currentstroke}%
\pgfsetstrokeopacity{0.000000}%
\pgfsetdash{}{0pt}%
\pgfpathmoveto{\pgfqpoint{1.425657in}{1.293675in}}%
\pgfpathlineto{\pgfqpoint{1.581397in}{1.293675in}}%
\pgfpathlineto{\pgfqpoint{1.581397in}{1.314355in}}%
\pgfpathlineto{\pgfqpoint{1.425657in}{1.314355in}}%
\pgfpathlineto{\pgfqpoint{1.425657in}{1.293675in}}%
\pgfpathclose%
\pgfusepath{fill}%
\end{pgfscope}%
\begin{pgfscope}%
\pgfpathrectangle{\pgfqpoint{0.405556in}{1.293675in}}{\pgfqpoint{5.310752in}{1.510421in}}%
\pgfusepath{clip}%
\pgfsetbuttcap%
\pgfsetmiterjoin%
\definecolor{currentfill}{rgb}{0.121569,0.466667,0.705882}%
\pgfsetfillcolor{currentfill}%
\pgfsetlinewidth{0.000000pt}%
\definecolor{currentstroke}{rgb}{0.000000,0.000000,0.000000}%
\pgfsetstrokecolor{currentstroke}%
\pgfsetstrokeopacity{0.000000}%
\pgfsetdash{}{0pt}%
\pgfpathmoveto{\pgfqpoint{1.815008in}{1.293675in}}%
\pgfpathlineto{\pgfqpoint{1.970749in}{1.293675in}}%
\pgfpathlineto{\pgfqpoint{1.970749in}{1.608002in}}%
\pgfpathlineto{\pgfqpoint{1.815008in}{1.608002in}}%
\pgfpathlineto{\pgfqpoint{1.815008in}{1.293675in}}%
\pgfpathclose%
\pgfusepath{fill}%
\end{pgfscope}%
\begin{pgfscope}%
\pgfpathrectangle{\pgfqpoint{0.405556in}{1.293675in}}{\pgfqpoint{5.310752in}{1.510421in}}%
\pgfusepath{clip}%
\pgfsetbuttcap%
\pgfsetmiterjoin%
\definecolor{currentfill}{rgb}{0.121569,0.466667,0.705882}%
\pgfsetfillcolor{currentfill}%
\pgfsetlinewidth{0.000000pt}%
\definecolor{currentstroke}{rgb}{0.000000,0.000000,0.000000}%
\pgfsetstrokecolor{currentstroke}%
\pgfsetstrokeopacity{0.000000}%
\pgfsetdash{}{0pt}%
\pgfpathmoveto{\pgfqpoint{2.204359in}{1.293675in}}%
\pgfpathlineto{\pgfqpoint{2.360100in}{1.293675in}}%
\pgfpathlineto{\pgfqpoint{2.360100in}{1.533556in}}%
\pgfpathlineto{\pgfqpoint{2.204359in}{1.533556in}}%
\pgfpathlineto{\pgfqpoint{2.204359in}{1.293675in}}%
\pgfpathclose%
\pgfusepath{fill}%
\end{pgfscope}%
\begin{pgfscope}%
\pgfpathrectangle{\pgfqpoint{0.405556in}{1.293675in}}{\pgfqpoint{5.310752in}{1.510421in}}%
\pgfusepath{clip}%
\pgfsetbuttcap%
\pgfsetmiterjoin%
\definecolor{currentfill}{rgb}{0.121569,0.466667,0.705882}%
\pgfsetfillcolor{currentfill}%
\pgfsetlinewidth{0.000000pt}%
\definecolor{currentstroke}{rgb}{0.000000,0.000000,0.000000}%
\pgfsetstrokecolor{currentstroke}%
\pgfsetstrokeopacity{0.000000}%
\pgfsetdash{}{0pt}%
\pgfpathmoveto{\pgfqpoint{2.593711in}{1.293675in}}%
\pgfpathlineto{\pgfqpoint{2.749451in}{1.293675in}}%
\pgfpathlineto{\pgfqpoint{2.749451in}{1.475654in}}%
\pgfpathlineto{\pgfqpoint{2.593711in}{1.475654in}}%
\pgfpathlineto{\pgfqpoint{2.593711in}{1.293675in}}%
\pgfpathclose%
\pgfusepath{fill}%
\end{pgfscope}%
\begin{pgfscope}%
\pgfpathrectangle{\pgfqpoint{0.405556in}{1.293675in}}{\pgfqpoint{5.310752in}{1.510421in}}%
\pgfusepath{clip}%
\pgfsetbuttcap%
\pgfsetmiterjoin%
\definecolor{currentfill}{rgb}{0.121569,0.466667,0.705882}%
\pgfsetfillcolor{currentfill}%
\pgfsetlinewidth{0.000000pt}%
\definecolor{currentstroke}{rgb}{0.000000,0.000000,0.000000}%
\pgfsetstrokecolor{currentstroke}%
\pgfsetstrokeopacity{0.000000}%
\pgfsetdash{}{0pt}%
\pgfpathmoveto{\pgfqpoint{2.983062in}{1.293675in}}%
\pgfpathlineto{\pgfqpoint{3.138803in}{1.293675in}}%
\pgfpathlineto{\pgfqpoint{3.138803in}{1.430159in}}%
\pgfpathlineto{\pgfqpoint{2.983062in}{1.430159in}}%
\pgfpathlineto{\pgfqpoint{2.983062in}{1.293675in}}%
\pgfpathclose%
\pgfusepath{fill}%
\end{pgfscope}%
\begin{pgfscope}%
\pgfpathrectangle{\pgfqpoint{0.405556in}{1.293675in}}{\pgfqpoint{5.310752in}{1.510421in}}%
\pgfusepath{clip}%
\pgfsetbuttcap%
\pgfsetmiterjoin%
\definecolor{currentfill}{rgb}{0.121569,0.466667,0.705882}%
\pgfsetfillcolor{currentfill}%
\pgfsetlinewidth{0.000000pt}%
\definecolor{currentstroke}{rgb}{0.000000,0.000000,0.000000}%
\pgfsetstrokecolor{currentstroke}%
\pgfsetstrokeopacity{0.000000}%
\pgfsetdash{}{0pt}%
\pgfpathmoveto{\pgfqpoint{3.372413in}{1.293675in}}%
\pgfpathlineto{\pgfqpoint{3.528154in}{1.293675in}}%
\pgfpathlineto{\pgfqpoint{3.528154in}{1.372257in}}%
\pgfpathlineto{\pgfqpoint{3.372413in}{1.372257in}}%
\pgfpathlineto{\pgfqpoint{3.372413in}{1.293675in}}%
\pgfpathclose%
\pgfusepath{fill}%
\end{pgfscope}%
\begin{pgfscope}%
\pgfpathrectangle{\pgfqpoint{0.405556in}{1.293675in}}{\pgfqpoint{5.310752in}{1.510421in}}%
\pgfusepath{clip}%
\pgfsetbuttcap%
\pgfsetmiterjoin%
\definecolor{currentfill}{rgb}{0.121569,0.466667,0.705882}%
\pgfsetfillcolor{currentfill}%
\pgfsetlinewidth{0.000000pt}%
\definecolor{currentstroke}{rgb}{0.000000,0.000000,0.000000}%
\pgfsetstrokecolor{currentstroke}%
\pgfsetstrokeopacity{0.000000}%
\pgfsetdash{}{0pt}%
\pgfpathmoveto{\pgfqpoint{3.761765in}{1.293675in}}%
\pgfpathlineto{\pgfqpoint{3.917505in}{1.293675in}}%
\pgfpathlineto{\pgfqpoint{3.917505in}{1.372257in}}%
\pgfpathlineto{\pgfqpoint{3.761765in}{1.372257in}}%
\pgfpathlineto{\pgfqpoint{3.761765in}{1.293675in}}%
\pgfpathclose%
\pgfusepath{fill}%
\end{pgfscope}%
\begin{pgfscope}%
\pgfpathrectangle{\pgfqpoint{0.405556in}{1.293675in}}{\pgfqpoint{5.310752in}{1.510421in}}%
\pgfusepath{clip}%
\pgfsetbuttcap%
\pgfsetmiterjoin%
\definecolor{currentfill}{rgb}{0.121569,0.466667,0.705882}%
\pgfsetfillcolor{currentfill}%
\pgfsetlinewidth{0.000000pt}%
\definecolor{currentstroke}{rgb}{0.000000,0.000000,0.000000}%
\pgfsetstrokecolor{currentstroke}%
\pgfsetstrokeopacity{0.000000}%
\pgfsetdash{}{0pt}%
\pgfpathmoveto{\pgfqpoint{4.151116in}{1.293675in}}%
\pgfpathlineto{\pgfqpoint{4.306857in}{1.293675in}}%
\pgfpathlineto{\pgfqpoint{4.306857in}{1.372257in}}%
\pgfpathlineto{\pgfqpoint{4.151116in}{1.372257in}}%
\pgfpathlineto{\pgfqpoint{4.151116in}{1.293675in}}%
\pgfpathclose%
\pgfusepath{fill}%
\end{pgfscope}%
\begin{pgfscope}%
\pgfpathrectangle{\pgfqpoint{0.405556in}{1.293675in}}{\pgfqpoint{5.310752in}{1.510421in}}%
\pgfusepath{clip}%
\pgfsetbuttcap%
\pgfsetmiterjoin%
\definecolor{currentfill}{rgb}{0.121569,0.466667,0.705882}%
\pgfsetfillcolor{currentfill}%
\pgfsetlinewidth{0.000000pt}%
\definecolor{currentstroke}{rgb}{0.000000,0.000000,0.000000}%
\pgfsetstrokecolor{currentstroke}%
\pgfsetstrokeopacity{0.000000}%
\pgfsetdash{}{0pt}%
\pgfpathmoveto{\pgfqpoint{4.540467in}{1.293675in}}%
\pgfpathlineto{\pgfqpoint{4.696208in}{1.293675in}}%
\pgfpathlineto{\pgfqpoint{4.696208in}{1.359849in}}%
\pgfpathlineto{\pgfqpoint{4.540467in}{1.359849in}}%
\pgfpathlineto{\pgfqpoint{4.540467in}{1.293675in}}%
\pgfpathclose%
\pgfusepath{fill}%
\end{pgfscope}%
\begin{pgfscope}%
\pgfpathrectangle{\pgfqpoint{0.405556in}{1.293675in}}{\pgfqpoint{5.310752in}{1.510421in}}%
\pgfusepath{clip}%
\pgfsetbuttcap%
\pgfsetmiterjoin%
\definecolor{currentfill}{rgb}{0.121569,0.466667,0.705882}%
\pgfsetfillcolor{currentfill}%
\pgfsetlinewidth{0.000000pt}%
\definecolor{currentstroke}{rgb}{0.000000,0.000000,0.000000}%
\pgfsetstrokecolor{currentstroke}%
\pgfsetstrokeopacity{0.000000}%
\pgfsetdash{}{0pt}%
\pgfpathmoveto{\pgfqpoint{4.929819in}{1.293675in}}%
\pgfpathlineto{\pgfqpoint{5.085559in}{1.293675in}}%
\pgfpathlineto{\pgfqpoint{5.085559in}{1.355714in}}%
\pgfpathlineto{\pgfqpoint{4.929819in}{1.355714in}}%
\pgfpathlineto{\pgfqpoint{4.929819in}{1.293675in}}%
\pgfpathclose%
\pgfusepath{fill}%
\end{pgfscope}%
\begin{pgfscope}%
\pgfpathrectangle{\pgfqpoint{0.405556in}{1.293675in}}{\pgfqpoint{5.310752in}{1.510421in}}%
\pgfusepath{clip}%
\pgfsetbuttcap%
\pgfsetmiterjoin%
\definecolor{currentfill}{rgb}{0.121569,0.466667,0.705882}%
\pgfsetfillcolor{currentfill}%
\pgfsetlinewidth{0.000000pt}%
\definecolor{currentstroke}{rgb}{0.000000,0.000000,0.000000}%
\pgfsetstrokecolor{currentstroke}%
\pgfsetstrokeopacity{0.000000}%
\pgfsetdash{}{0pt}%
\pgfpathmoveto{\pgfqpoint{5.319170in}{1.293675in}}%
\pgfpathlineto{\pgfqpoint{5.474911in}{1.293675in}}%
\pgfpathlineto{\pgfqpoint{5.474911in}{1.339170in}}%
\pgfpathlineto{\pgfqpoint{5.319170in}{1.339170in}}%
\pgfpathlineto{\pgfqpoint{5.319170in}{1.293675in}}%
\pgfpathclose%
\pgfusepath{fill}%
\end{pgfscope}%
\begin{pgfscope}%
\definecolor{textcolor}{rgb}{0.000000,0.000000,0.000000}%
\pgfsetstrokecolor{textcolor}%
\pgfsetfillcolor{textcolor}%
\pgftext[x=0.759547in, y=0.546376in, left, base,rotate=90.000000]{\color{textcolor}\rmfamily\fontsize{10.000000}{12.000000}\selectfont \(\displaystyle \geq1\) Public}%
\end{pgfscope}%
\begin{pgfscope}%
\definecolor{textcolor}{rgb}{0.000000,0.000000,0.000000}%
\pgfsetstrokecolor{textcolor}%
\pgfsetfillcolor{textcolor}%
\pgftext[x=1.148898in, y=0.414815in, left, base,rotate=90.000000]{\color{textcolor}\rmfamily\fontsize{10.000000}{12.000000}\selectfont Only Private}%
\end{pgfscope}%
\begin{pgfscope}%
\definecolor{textcolor}{rgb}{0.000000,0.000000,0.000000}%
\pgfsetstrokecolor{textcolor}%
\pgfsetfillcolor{textcolor}%
\pgftext[x=1.538249in, y=0.100000in, left, base,rotate=90.000000]{\color{textcolor}\rmfamily\fontsize{10.000000}{12.000000}\selectfont Private \& Privacy}%
\end{pgfscope}%
\begin{pgfscope}%
\definecolor{textcolor}{rgb}{0.000000,0.000000,0.000000}%
\pgfsetstrokecolor{textcolor}%
\pgfsetfillcolor{textcolor}%
\pgftext[x=1.927601in, y=0.737347in, left, base,rotate=90.000000]{\color{textcolor}\rmfamily\fontsize{10.000000}{12.000000}\selectfont MNIST}%
\end{pgfscope}%
\begin{pgfscope}%
\definecolor{textcolor}{rgb}{0.000000,0.000000,0.000000}%
\pgfsetstrokecolor{textcolor}%
\pgfsetfillcolor{textcolor}%
\pgftext[x=2.316952in, y=0.579168in, left, base,rotate=90.000000]{\color{textcolor}\rmfamily\fontsize{10.000000}{12.000000}\selectfont CIFAR-10}%
\end{pgfscope}%
\begin{pgfscope}%
\definecolor{textcolor}{rgb}{0.000000,0.000000,0.000000}%
\pgfsetstrokecolor{textcolor}%
\pgfsetfillcolor{textcolor}%
\pgftext[x=2.706303in, y=0.791360in, left, base,rotate=90.000000]{\color{textcolor}\rmfamily\fontsize{10.000000}{12.000000}\selectfont celebA}%
\end{pgfscope}%
\begin{pgfscope}%
\definecolor{textcolor}{rgb}{0.000000,0.000000,0.000000}%
\pgfsetstrokecolor{textcolor}%
\pgfsetfillcolor{textcolor}%
\pgftext[x=3.095655in, y=0.610032in, left, base,rotate=90.000000]{\color{textcolor}\rmfamily\fontsize{10.000000}{12.000000}\selectfont ImageNet}%
\end{pgfscope}%
\begin{pgfscope}%
\definecolor{textcolor}{rgb}{0.000000,0.000000,0.000000}%
\pgfsetstrokecolor{textcolor}%
\pgfsetfillcolor{textcolor}%
\pgftext[x=3.485006in, y=0.321452in, left, base,rotate=90.000000]{\color{textcolor}\rmfamily\fontsize{10.000000}{12.000000}\selectfont Bach Chorales}%
\end{pgfscope}%
\begin{pgfscope}%
\definecolor{textcolor}{rgb}{0.000000,0.000000,0.000000}%
\pgfsetstrokecolor{textcolor}%
\pgfsetfillcolor{textcolor}%
\pgftext[x=3.874357in, y=0.856947in, left, base,rotate=90.000000]{\color{textcolor}\rmfamily\fontsize{10.000000}{12.000000}\selectfont ZINC}%
\end{pgfscope}%
\begin{pgfscope}%
\definecolor{textcolor}{rgb}{0.000000,0.000000,0.000000}%
\pgfsetstrokecolor{textcolor}%
\pgfsetfillcolor{textcolor}%
\pgftext[x=4.263709in, y=0.824153in, left, base,rotate=90.000000]{\color{textcolor}\rmfamily\fontsize{10.000000}{12.000000}\selectfont LSUN}%
\end{pgfscope}%
\begin{pgfscope}%
\definecolor{textcolor}{rgb}{0.000000,0.000000,0.000000}%
\pgfsetstrokecolor{textcolor}%
\pgfsetfillcolor{textcolor}%
\pgftext[x=4.653060in, y=0.806792in, left, base,rotate=90.000000]{\color{textcolor}\rmfamily\fontsize{10.000000}{12.000000}\selectfont SVHN}%
\end{pgfscope}%
\begin{pgfscope}%
\definecolor{textcolor}{rgb}{0.000000,0.000000,0.000000}%
\pgfsetstrokecolor{textcolor}%
\pgfsetfillcolor{textcolor}%
\pgftext[x=5.042411in, y=0.529014in, left, base,rotate=90.000000]{\color{textcolor}\rmfamily\fontsize{10.000000}{12.000000}\selectfont MS COCO}%
\end{pgfscope}%
\begin{pgfscope}%
\definecolor{textcolor}{rgb}{0.000000,0.000000,0.000000}%
\pgfsetstrokecolor{textcolor}%
\pgfsetfillcolor{textcolor}%
\pgftext[x=5.431763in, y=0.576467in, left, base,rotate=90.000000]{\color{textcolor}\rmfamily\fontsize{10.000000}{12.000000}\selectfont MuseData}%
\end{pgfscope}%
\begin{pgfscope}%
\pgfsetbuttcap%
\pgfsetroundjoin%
\definecolor{currentfill}{rgb}{0.000000,0.000000,0.000000}%
\pgfsetfillcolor{currentfill}%
\pgfsetlinewidth{0.803000pt}%
\definecolor{currentstroke}{rgb}{0.000000,0.000000,0.000000}%
\pgfsetstrokecolor{currentstroke}%
\pgfsetdash{}{0pt}%
\pgfsys@defobject{currentmarker}{\pgfqpoint{-0.048611in}{0.000000in}}{\pgfqpoint{-0.000000in}{0.000000in}}{%
\pgfpathmoveto{\pgfqpoint{-0.000000in}{0.000000in}}%
\pgfpathlineto{\pgfqpoint{-0.048611in}{0.000000in}}%
\pgfusepath{stroke,fill}%
}%
\begin{pgfscope}%
\pgfsys@transformshift{0.405556in}{1.293675in}%
\pgfsys@useobject{currentmarker}{}%
\end{pgfscope}%
\end{pgfscope}%
\begin{pgfscope}%
\definecolor{textcolor}{rgb}{0.000000,0.000000,0.000000}%
\pgfsetstrokecolor{textcolor}%
\pgfsetfillcolor{textcolor}%
\pgftext[x=0.238889in, y=1.245450in, left, base]{\color{textcolor}\rmfamily\fontsize{10.000000}{12.000000}\selectfont \(\displaystyle {0}\)}%
\end{pgfscope}%
\begin{pgfscope}%
\pgfsetbuttcap%
\pgfsetroundjoin%
\definecolor{currentfill}{rgb}{0.000000,0.000000,0.000000}%
\pgfsetfillcolor{currentfill}%
\pgfsetlinewidth{0.803000pt}%
\definecolor{currentstroke}{rgb}{0.000000,0.000000,0.000000}%
\pgfsetstrokecolor{currentstroke}%
\pgfsetdash{}{0pt}%
\pgfsys@defobject{currentmarker}{\pgfqpoint{-0.048611in}{0.000000in}}{\pgfqpoint{-0.000000in}{0.000000in}}{%
\pgfpathmoveto{\pgfqpoint{-0.000000in}{0.000000in}}%
\pgfpathlineto{\pgfqpoint{-0.048611in}{0.000000in}}%
\pgfusepath{stroke,fill}%
}%
\begin{pgfscope}%
\pgfsys@transformshift{0.405556in}{1.707263in}%
\pgfsys@useobject{currentmarker}{}%
\end{pgfscope}%
\end{pgfscope}%
\begin{pgfscope}%
\definecolor{textcolor}{rgb}{0.000000,0.000000,0.000000}%
\pgfsetstrokecolor{textcolor}%
\pgfsetfillcolor{textcolor}%
\pgftext[x=0.100000in, y=1.659037in, left, base]{\color{textcolor}\rmfamily\fontsize{10.000000}{12.000000}\selectfont \(\displaystyle {100}\)}%
\end{pgfscope}%
\begin{pgfscope}%
\pgfsetbuttcap%
\pgfsetroundjoin%
\definecolor{currentfill}{rgb}{0.000000,0.000000,0.000000}%
\pgfsetfillcolor{currentfill}%
\pgfsetlinewidth{0.803000pt}%
\definecolor{currentstroke}{rgb}{0.000000,0.000000,0.000000}%
\pgfsetstrokecolor{currentstroke}%
\pgfsetdash{}{0pt}%
\pgfsys@defobject{currentmarker}{\pgfqpoint{-0.048611in}{0.000000in}}{\pgfqpoint{-0.000000in}{0.000000in}}{%
\pgfpathmoveto{\pgfqpoint{-0.000000in}{0.000000in}}%
\pgfpathlineto{\pgfqpoint{-0.048611in}{0.000000in}}%
\pgfusepath{stroke,fill}%
}%
\begin{pgfscope}%
\pgfsys@transformshift{0.405556in}{2.120850in}%
\pgfsys@useobject{currentmarker}{}%
\end{pgfscope}%
\end{pgfscope}%
\begin{pgfscope}%
\definecolor{textcolor}{rgb}{0.000000,0.000000,0.000000}%
\pgfsetstrokecolor{textcolor}%
\pgfsetfillcolor{textcolor}%
\pgftext[x=0.100000in, y=2.072625in, left, base]{\color{textcolor}\rmfamily\fontsize{10.000000}{12.000000}\selectfont \(\displaystyle {200}\)}%
\end{pgfscope}%
\begin{pgfscope}%
\pgfsetbuttcap%
\pgfsetroundjoin%
\definecolor{currentfill}{rgb}{0.000000,0.000000,0.000000}%
\pgfsetfillcolor{currentfill}%
\pgfsetlinewidth{0.803000pt}%
\definecolor{currentstroke}{rgb}{0.000000,0.000000,0.000000}%
\pgfsetstrokecolor{currentstroke}%
\pgfsetdash{}{0pt}%
\pgfsys@defobject{currentmarker}{\pgfqpoint{-0.048611in}{0.000000in}}{\pgfqpoint{-0.000000in}{0.000000in}}{%
\pgfpathmoveto{\pgfqpoint{-0.000000in}{0.000000in}}%
\pgfpathlineto{\pgfqpoint{-0.048611in}{0.000000in}}%
\pgfusepath{stroke,fill}%
}%
\begin{pgfscope}%
\pgfsys@transformshift{0.405556in}{2.534437in}%
\pgfsys@useobject{currentmarker}{}%
\end{pgfscope}%
\end{pgfscope}%
\begin{pgfscope}%
\definecolor{textcolor}{rgb}{0.000000,0.000000,0.000000}%
\pgfsetstrokecolor{textcolor}%
\pgfsetfillcolor{textcolor}%
\pgftext[x=0.100000in, y=2.486212in, left, base]{\color{textcolor}\rmfamily\fontsize{10.000000}{12.000000}\selectfont \(\displaystyle {300}\)}%
\end{pgfscope}%
\begin{pgfscope}%
\pgfpathrectangle{\pgfqpoint{0.405556in}{1.293675in}}{\pgfqpoint{5.310752in}{1.510421in}}%
\pgfusepath{clip}%
\pgfsetbuttcap%
\pgfsetroundjoin%
\pgfsetlinewidth{1.505625pt}%
\definecolor{currentstroke}{rgb}{0.000000,0.000000,0.000000}%
\pgfsetstrokecolor{currentstroke}%
\pgfsetdash{}{0pt}%
\pgfpathmoveto{\pgfqpoint{1.698203in}{1.293675in}}%
\pgfpathlineto{\pgfqpoint{1.698203in}{2.804096in}}%
\pgfusepath{stroke}%
\end{pgfscope}%
\begin{pgfscope}%
\pgfsetrectcap%
\pgfsetmiterjoin%
\pgfsetlinewidth{0.803000pt}%
\definecolor{currentstroke}{rgb}{0.000000,0.000000,0.000000}%
\pgfsetstrokecolor{currentstroke}%
\pgfsetdash{}{0pt}%
\pgfpathmoveto{\pgfqpoint{0.405556in}{1.293675in}}%
\pgfpathlineto{\pgfqpoint{0.405556in}{2.804096in}}%
\pgfusepath{stroke}%
\end{pgfscope}%
\begin{pgfscope}%
\pgfsetrectcap%
\pgfsetmiterjoin%
\pgfsetlinewidth{0.803000pt}%
\definecolor{currentstroke}{rgb}{0.000000,0.000000,0.000000}%
\pgfsetstrokecolor{currentstroke}%
\pgfsetdash{}{0pt}%
\pgfpathmoveto{\pgfqpoint{5.716308in}{1.293675in}}%
\pgfpathlineto{\pgfqpoint{5.716308in}{2.804096in}}%
\pgfusepath{stroke}%
\end{pgfscope}%
\begin{pgfscope}%
\pgfsetrectcap%
\pgfsetmiterjoin%
\pgfsetlinewidth{0.803000pt}%
\definecolor{currentstroke}{rgb}{0.000000,0.000000,0.000000}%
\pgfsetstrokecolor{currentstroke}%
\pgfsetdash{}{0pt}%
\pgfpathmoveto{\pgfqpoint{0.405556in}{1.293675in}}%
\pgfpathlineto{\pgfqpoint{5.716308in}{1.293675in}}%
\pgfusepath{stroke}%
\end{pgfscope}%
\begin{pgfscope}%
\pgfsetrectcap%
\pgfsetmiterjoin%
\pgfsetlinewidth{0.803000pt}%
\definecolor{currentstroke}{rgb}{0.000000,0.000000,0.000000}%
\pgfsetstrokecolor{currentstroke}%
\pgfsetdash{}{0pt}%
\pgfpathmoveto{\pgfqpoint{0.405556in}{2.804096in}}%
\pgfpathlineto{\pgfqpoint{5.716308in}{2.804096in}}%
\pgfusepath{stroke}%
\end{pgfscope}%
\begin{pgfscope}%
\definecolor{textcolor}{rgb}{0.000000,0.000000,0.000000}%
\pgfsetstrokecolor{textcolor}%
\pgfsetfillcolor{textcolor}%
\pgftext[x=0.724824in,y=2.666785in,,base]{\color{textcolor}\rmfamily\fontsize{10.000000}{12.000000}\selectfont 332}%
\end{pgfscope}%
\begin{pgfscope}%
\definecolor{textcolor}{rgb}{0.000000,0.000000,0.000000}%
\pgfsetstrokecolor{textcolor}%
\pgfsetfillcolor{textcolor}%
\pgftext[x=1.114176in,y=1.525284in,,base]{\color{textcolor}\rmfamily\fontsize{10.000000}{12.000000}\selectfont 56}%
\end{pgfscope}%
\begin{pgfscope}%
\definecolor{textcolor}{rgb}{0.000000,0.000000,0.000000}%
\pgfsetstrokecolor{textcolor}%
\pgfsetfillcolor{textcolor}%
\pgftext[x=1.503527in,y=1.314355in,,base]{\color{textcolor}\rmfamily\fontsize{10.000000}{12.000000}\selectfont 5}%
\end{pgfscope}%
\begin{pgfscope}%
\definecolor{textcolor}{rgb}{0.000000,0.000000,0.000000}%
\pgfsetstrokecolor{textcolor}%
\pgfsetfillcolor{textcolor}%
\pgftext[x=1.892878in,y=1.608002in,,base]{\color{textcolor}\rmfamily\fontsize{10.000000}{12.000000}\selectfont 76}%
\end{pgfscope}%
\begin{pgfscope}%
\definecolor{textcolor}{rgb}{0.000000,0.000000,0.000000}%
\pgfsetstrokecolor{textcolor}%
\pgfsetfillcolor{textcolor}%
\pgftext[x=2.282230in,y=1.533556in,,base]{\color{textcolor}\rmfamily\fontsize{10.000000}{12.000000}\selectfont 58}%
\end{pgfscope}%
\begin{pgfscope}%
\definecolor{textcolor}{rgb}{0.000000,0.000000,0.000000}%
\pgfsetstrokecolor{textcolor}%
\pgfsetfillcolor{textcolor}%
\pgftext[x=2.671581in,y=1.475654in,,base]{\color{textcolor}\rmfamily\fontsize{10.000000}{12.000000}\selectfont 44}%
\end{pgfscope}%
\begin{pgfscope}%
\definecolor{textcolor}{rgb}{0.000000,0.000000,0.000000}%
\pgfsetstrokecolor{textcolor}%
\pgfsetfillcolor{textcolor}%
\pgftext[x=3.060932in,y=1.430159in,,base]{\color{textcolor}\rmfamily\fontsize{10.000000}{12.000000}\selectfont 33}%
\end{pgfscope}%
\begin{pgfscope}%
\definecolor{textcolor}{rgb}{0.000000,0.000000,0.000000}%
\pgfsetstrokecolor{textcolor}%
\pgfsetfillcolor{textcolor}%
\pgftext[x=3.450284in,y=1.372257in,,base]{\color{textcolor}\rmfamily\fontsize{10.000000}{12.000000}\selectfont 19}%
\end{pgfscope}%
\begin{pgfscope}%
\definecolor{textcolor}{rgb}{0.000000,0.000000,0.000000}%
\pgfsetstrokecolor{textcolor}%
\pgfsetfillcolor{textcolor}%
\pgftext[x=3.839635in,y=1.372257in,,base]{\color{textcolor}\rmfamily\fontsize{10.000000}{12.000000}\selectfont 19}%
\end{pgfscope}%
\begin{pgfscope}%
\definecolor{textcolor}{rgb}{0.000000,0.000000,0.000000}%
\pgfsetstrokecolor{textcolor}%
\pgfsetfillcolor{textcolor}%
\pgftext[x=4.228986in,y=1.372257in,,base]{\color{textcolor}\rmfamily\fontsize{10.000000}{12.000000}\selectfont 19}%
\end{pgfscope}%
\begin{pgfscope}%
\definecolor{textcolor}{rgb}{0.000000,0.000000,0.000000}%
\pgfsetstrokecolor{textcolor}%
\pgfsetfillcolor{textcolor}%
\pgftext[x=4.618338in,y=1.359849in,,base]{\color{textcolor}\rmfamily\fontsize{10.000000}{12.000000}\selectfont 16}%
\end{pgfscope}%
\begin{pgfscope}%
\definecolor{textcolor}{rgb}{0.000000,0.000000,0.000000}%
\pgfsetstrokecolor{textcolor}%
\pgfsetfillcolor{textcolor}%
\pgftext[x=5.007689in,y=1.355714in,,base]{\color{textcolor}\rmfamily\fontsize{10.000000}{12.000000}\selectfont 15}%
\end{pgfscope}%
\begin{pgfscope}%
\definecolor{textcolor}{rgb}{0.000000,0.000000,0.000000}%
\pgfsetstrokecolor{textcolor}%
\pgfsetfillcolor{textcolor}%
\pgftext[x=5.397040in,y=1.339170in,,base]{\color{textcolor}\rmfamily\fontsize{10.000000}{12.000000}\selectfont 11}%
\end{pgfscope}%
\end{pgfpicture}%
\makeatother%
\endgroup%

%% file: figures/classification/color_bar_performance.pgf
\begingroup%
\makeatletter%
\begin{pgfpicture}%
\pgfpathrectangle{\pgfpointorigin}{\pgfqpoint{4.632554in}{0.900922in}}%
\pgfusepath{use as bounding box, clip}%
\begin{pgfscope}%
\pgfsetbuttcap%
\pgfsetmiterjoin%
\definecolor{currentfill}{rgb}{1.000000,1.000000,1.000000}%
\pgfsetfillcolor{currentfill}%
\pgfsetlinewidth{0.000000pt}%
\definecolor{currentstroke}{rgb}{1.000000,1.000000,1.000000}%
\pgfsetstrokecolor{currentstroke}%
\pgfsetdash{}{0pt}%
\pgfpathmoveto{\pgfqpoint{0.000000in}{0.000000in}}%
\pgfpathlineto{\pgfqpoint{4.632554in}{0.000000in}}%
\pgfpathlineto{\pgfqpoint{4.632554in}{0.900922in}}%
\pgfpathlineto{\pgfqpoint{0.000000in}{0.900922in}}%
\pgfpathlineto{\pgfqpoint{0.000000in}{0.000000in}}%
\pgfpathclose%
\pgfusepath{fill}%
\end{pgfscope}%
\begin{pgfscope}%
\pgfsetbuttcap%
\pgfsetmiterjoin%
\definecolor{currentfill}{rgb}{1.000000,1.000000,1.000000}%
\pgfsetfillcolor{currentfill}%
\pgfsetlinewidth{0.000000pt}%
\definecolor{currentstroke}{rgb}{0.000000,0.000000,0.000000}%
\pgfsetstrokecolor{currentstroke}%
\pgfsetstrokeopacity{0.000000}%
\pgfsetdash{}{0pt}%
\pgfpathmoveto{\pgfqpoint{0.134722in}{0.499691in}}%
\pgfpathlineto{\pgfqpoint{4.532554in}{0.499691in}}%
\pgfpathlineto{\pgfqpoint{4.532554in}{0.800922in}}%
\pgfpathlineto{\pgfqpoint{0.134722in}{0.800922in}}%
\pgfpathlineto{\pgfqpoint{0.134722in}{0.499691in}}%
\pgfpathclose%
\pgfusepath{fill}%
\end{pgfscope}%
\begin{pgfscope}%
\pgfpathrectangle{\pgfqpoint{0.134722in}{0.499691in}}{\pgfqpoint{4.397831in}{0.301231in}}%
\pgfusepath{clip}%
\pgfsetbuttcap%
\pgfsetmiterjoin%
\definecolor{currentfill}{rgb}{1.000000,1.000000,1.000000}%
\pgfsetfillcolor{currentfill}%
\pgfsetlinewidth{0.010037pt}%
\definecolor{currentstroke}{rgb}{1.000000,1.000000,1.000000}%
\pgfsetstrokecolor{currentstroke}%
\pgfsetdash{}{0pt}%
\pgfusepath{stroke,fill}%
\end{pgfscope}%
\begin{pgfscope}%
\pgfsys@transformshift{0.130000in}{0.500922in}%
\pgftext[left,bottom]{\includegraphics[interpolate=true,width=4.400000in,height=0.300000in]{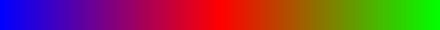}}%
\end{pgfscope}%
\begin{pgfscope}%
\pgfsetbuttcap%
\pgfsetroundjoin%
\definecolor{currentfill}{rgb}{0.000000,0.000000,0.000000}%
\pgfsetfillcolor{currentfill}%
\pgfsetlinewidth{0.803000pt}%
\definecolor{currentstroke}{rgb}{0.000000,0.000000,0.000000}%
\pgfsetstrokecolor{currentstroke}%
\pgfsetdash{}{0pt}%
\pgfsys@defobject{currentmarker}{\pgfqpoint{0.000000in}{-0.048611in}}{\pgfqpoint{0.000000in}{0.000000in}}{%
\pgfpathmoveto{\pgfqpoint{0.000000in}{0.000000in}}%
\pgfpathlineto{\pgfqpoint{0.000000in}{-0.048611in}}%
\pgfusepath{stroke,fill}%
}%
\begin{pgfscope}%
\pgfsys@transformshift{0.134722in}{0.499691in}%
\pgfsys@useobject{currentmarker}{}%
\end{pgfscope}%
\end{pgfscope}%
\begin{pgfscope}%
\definecolor{textcolor}{rgb}{0.000000,0.000000,0.000000}%
\pgfsetstrokecolor{textcolor}%
\pgfsetfillcolor{textcolor}%
\pgftext[x=0.134722in,y=0.402469in,,top]{\color{textcolor}\rmfamily\fontsize{10.000000}{12.000000}\selectfont \(\displaystyle {0}\)}%
\end{pgfscope}%
\begin{pgfscope}%
\pgfsetbuttcap%
\pgfsetroundjoin%
\definecolor{currentfill}{rgb}{0.000000,0.000000,0.000000}%
\pgfsetfillcolor{currentfill}%
\pgfsetlinewidth{0.803000pt}%
\definecolor{currentstroke}{rgb}{0.000000,0.000000,0.000000}%
\pgfsetstrokecolor{currentstroke}%
\pgfsetdash{}{0pt}%
\pgfsys@defobject{currentmarker}{\pgfqpoint{0.000000in}{-0.048611in}}{\pgfqpoint{0.000000in}{0.000000in}}{%
\pgfpathmoveto{\pgfqpoint{0.000000in}{0.000000in}}%
\pgfpathlineto{\pgfqpoint{0.000000in}{-0.048611in}}%
\pgfusepath{stroke,fill}%
}%
\begin{pgfscope}%
\pgfsys@transformshift{0.821884in}{0.499691in}%
\pgfsys@useobject{currentmarker}{}%
\end{pgfscope}%
\end{pgfscope}%
\begin{pgfscope}%
\definecolor{textcolor}{rgb}{0.000000,0.000000,0.000000}%
\pgfsetstrokecolor{textcolor}%
\pgfsetfillcolor{textcolor}%
\pgftext[x=0.821884in,y=0.402469in,,top]{\color{textcolor}\rmfamily\fontsize{10.000000}{12.000000}\selectfont \(\displaystyle {5}\)}%
\end{pgfscope}%
\begin{pgfscope}%
\pgfsetbuttcap%
\pgfsetroundjoin%
\definecolor{currentfill}{rgb}{0.000000,0.000000,0.000000}%
\pgfsetfillcolor{currentfill}%
\pgfsetlinewidth{0.803000pt}%
\definecolor{currentstroke}{rgb}{0.000000,0.000000,0.000000}%
\pgfsetstrokecolor{currentstroke}%
\pgfsetdash{}{0pt}%
\pgfsys@defobject{currentmarker}{\pgfqpoint{0.000000in}{-0.048611in}}{\pgfqpoint{0.000000in}{0.000000in}}{%
\pgfpathmoveto{\pgfqpoint{0.000000in}{0.000000in}}%
\pgfpathlineto{\pgfqpoint{0.000000in}{-0.048611in}}%
\pgfusepath{stroke,fill}%
}%
\begin{pgfscope}%
\pgfsys@transformshift{1.509045in}{0.499691in}%
\pgfsys@useobject{currentmarker}{}%
\end{pgfscope}%
\end{pgfscope}%
\begin{pgfscope}%
\definecolor{textcolor}{rgb}{0.000000,0.000000,0.000000}%
\pgfsetstrokecolor{textcolor}%
\pgfsetfillcolor{textcolor}%
\pgftext[x=1.509045in,y=0.402469in,,top]{\color{textcolor}\rmfamily\fontsize{10.000000}{12.000000}\selectfont \(\displaystyle {10}\)}%
\end{pgfscope}%
\begin{pgfscope}%
\pgfsetbuttcap%
\pgfsetroundjoin%
\definecolor{currentfill}{rgb}{0.000000,0.000000,0.000000}%
\pgfsetfillcolor{currentfill}%
\pgfsetlinewidth{0.803000pt}%
\definecolor{currentstroke}{rgb}{0.000000,0.000000,0.000000}%
\pgfsetstrokecolor{currentstroke}%
\pgfsetdash{}{0pt}%
\pgfsys@defobject{currentmarker}{\pgfqpoint{0.000000in}{-0.048611in}}{\pgfqpoint{0.000000in}{0.000000in}}{%
\pgfpathmoveto{\pgfqpoint{0.000000in}{0.000000in}}%
\pgfpathlineto{\pgfqpoint{0.000000in}{-0.048611in}}%
\pgfusepath{stroke,fill}%
}%
\begin{pgfscope}%
\pgfsys@transformshift{2.196206in}{0.499691in}%
\pgfsys@useobject{currentmarker}{}%
\end{pgfscope}%
\end{pgfscope}%
\begin{pgfscope}%
\definecolor{textcolor}{rgb}{0.000000,0.000000,0.000000}%
\pgfsetstrokecolor{textcolor}%
\pgfsetfillcolor{textcolor}%
\pgftext[x=2.196206in,y=0.402469in,,top]{\color{textcolor}\rmfamily\fontsize{10.000000}{12.000000}\selectfont \(\displaystyle {15}\)}%
\end{pgfscope}%
\begin{pgfscope}%
\pgfsetbuttcap%
\pgfsetroundjoin%
\definecolor{currentfill}{rgb}{0.000000,0.000000,0.000000}%
\pgfsetfillcolor{currentfill}%
\pgfsetlinewidth{0.803000pt}%
\definecolor{currentstroke}{rgb}{0.000000,0.000000,0.000000}%
\pgfsetstrokecolor{currentstroke}%
\pgfsetdash{}{0pt}%
\pgfsys@defobject{currentmarker}{\pgfqpoint{0.000000in}{-0.048611in}}{\pgfqpoint{0.000000in}{0.000000in}}{%
\pgfpathmoveto{\pgfqpoint{0.000000in}{0.000000in}}%
\pgfpathlineto{\pgfqpoint{0.000000in}{-0.048611in}}%
\pgfusepath{stroke,fill}%
}%
\begin{pgfscope}%
\pgfsys@transformshift{2.883367in}{0.499691in}%
\pgfsys@useobject{currentmarker}{}%
\end{pgfscope}%
\end{pgfscope}%
\begin{pgfscope}%
\definecolor{textcolor}{rgb}{0.000000,0.000000,0.000000}%
\pgfsetstrokecolor{textcolor}%
\pgfsetfillcolor{textcolor}%
\pgftext[x=2.883367in,y=0.402469in,,top]{\color{textcolor}\rmfamily\fontsize{10.000000}{12.000000}\selectfont \(\displaystyle {20}\)}%
\end{pgfscope}%
\begin{pgfscope}%
\pgfsetbuttcap%
\pgfsetroundjoin%
\definecolor{currentfill}{rgb}{0.000000,0.000000,0.000000}%
\pgfsetfillcolor{currentfill}%
\pgfsetlinewidth{0.803000pt}%
\definecolor{currentstroke}{rgb}{0.000000,0.000000,0.000000}%
\pgfsetstrokecolor{currentstroke}%
\pgfsetdash{}{0pt}%
\pgfsys@defobject{currentmarker}{\pgfqpoint{0.000000in}{-0.048611in}}{\pgfqpoint{0.000000in}{0.000000in}}{%
\pgfpathmoveto{\pgfqpoint{0.000000in}{0.000000in}}%
\pgfpathlineto{\pgfqpoint{0.000000in}{-0.048611in}}%
\pgfusepath{stroke,fill}%
}%
\begin{pgfscope}%
\pgfsys@transformshift{3.570528in}{0.499691in}%
\pgfsys@useobject{currentmarker}{}%
\end{pgfscope}%
\end{pgfscope}%
\begin{pgfscope}%
\definecolor{textcolor}{rgb}{0.000000,0.000000,0.000000}%
\pgfsetstrokecolor{textcolor}%
\pgfsetfillcolor{textcolor}%
\pgftext[x=3.570528in,y=0.402469in,,top]{\color{textcolor}\rmfamily\fontsize{10.000000}{12.000000}\selectfont \(\displaystyle {25}\)}%
\end{pgfscope}%
\begin{pgfscope}%
\pgfsetbuttcap%
\pgfsetroundjoin%
\definecolor{currentfill}{rgb}{0.000000,0.000000,0.000000}%
\pgfsetfillcolor{currentfill}%
\pgfsetlinewidth{0.803000pt}%
\definecolor{currentstroke}{rgb}{0.000000,0.000000,0.000000}%
\pgfsetstrokecolor{currentstroke}%
\pgfsetdash{}{0pt}%
\pgfsys@defobject{currentmarker}{\pgfqpoint{0.000000in}{-0.048611in}}{\pgfqpoint{0.000000in}{0.000000in}}{%
\pgfpathmoveto{\pgfqpoint{0.000000in}{0.000000in}}%
\pgfpathlineto{\pgfqpoint{0.000000in}{-0.048611in}}%
\pgfusepath{stroke,fill}%
}%
\begin{pgfscope}%
\pgfsys@transformshift{4.257689in}{0.499691in}%
\pgfsys@useobject{currentmarker}{}%
\end{pgfscope}%
\end{pgfscope}%
\begin{pgfscope}%
\definecolor{textcolor}{rgb}{0.000000,0.000000,0.000000}%
\pgfsetstrokecolor{textcolor}%
\pgfsetfillcolor{textcolor}%
\pgftext[x=4.257689in,y=0.402469in,,top]{\color{textcolor}\rmfamily\fontsize{10.000000}{12.000000}\selectfont \(\displaystyle {30}\)}%
\end{pgfscope}%
\begin{pgfscope}%
\definecolor{textcolor}{rgb}{0.000000,0.000000,0.000000}%
\pgfsetstrokecolor{textcolor}%
\pgfsetfillcolor{textcolor}%
\pgftext[x=2.333638in,y=0.223457in,,top]{\color{textcolor}\rmfamily\fontsize{10.000000}{12.000000}\selectfont Number of times the model was outperformed}%
\end{pgfscope}%
\begin{pgfscope}%
\pgfsetrectcap%
\pgfsetmiterjoin%
\pgfsetlinewidth{0.803000pt}%
\definecolor{currentstroke}{rgb}{0.000000,0.000000,0.000000}%
\pgfsetstrokecolor{currentstroke}%
\pgfsetdash{}{0pt}%
\pgfpathmoveto{\pgfqpoint{0.134722in}{0.499691in}}%
\pgfpathlineto{\pgfqpoint{0.134722in}{0.650307in}}%
\pgfpathlineto{\pgfqpoint{0.134722in}{0.800922in}}%
\pgfpathlineto{\pgfqpoint{4.532554in}{0.800922in}}%
\pgfpathlineto{\pgfqpoint{4.532554in}{0.650307in}}%
\pgfpathlineto{\pgfqpoint{4.532554in}{0.499691in}}%
\pgfpathlineto{\pgfqpoint{0.134722in}{0.499691in}}%
\pgfpathclose%
\pgfusepath{stroke}%
\end{pgfscope}%
\end{pgfpicture}%
\makeatother%
\endgroup%

%% file: figures/classification/privacy_output_data_multi_v_bar.pgf
\begingroup%
\makeatletter%
\begin{pgfpicture}%
\pgfpathrectangle{\pgfpointorigin}{\pgfqpoint{5.816193in}{3.111759in}}%
\pgfusepath{use as bounding box, clip}%
\begin{pgfscope}%
\pgfsetbuttcap%
\pgfsetmiterjoin%
\definecolor{currentfill}{rgb}{1.000000,1.000000,1.000000}%
\pgfsetfillcolor{currentfill}%
\pgfsetlinewidth{0.000000pt}%
\definecolor{currentstroke}{rgb}{1.000000,1.000000,1.000000}%
\pgfsetstrokecolor{currentstroke}%
\pgfsetdash{}{0pt}%
\pgfpathmoveto{\pgfqpoint{0.000000in}{0.000000in}}%
\pgfpathlineto{\pgfqpoint{5.816193in}{0.000000in}}%
\pgfpathlineto{\pgfqpoint{5.816193in}{3.111759in}}%
\pgfpathlineto{\pgfqpoint{0.000000in}{3.111759in}}%
\pgfpathlineto{\pgfqpoint{0.000000in}{0.000000in}}%
\pgfpathclose%
\pgfusepath{fill}%
\end{pgfscope}%
\begin{pgfscope}%
\pgfsetbuttcap%
\pgfsetmiterjoin%
\definecolor{currentfill}{rgb}{1.000000,1.000000,1.000000}%
\pgfsetfillcolor{currentfill}%
\pgfsetlinewidth{0.000000pt}%
\definecolor{currentstroke}{rgb}{0.000000,0.000000,0.000000}%
\pgfsetstrokecolor{currentstroke}%
\pgfsetstrokeopacity{0.000000}%
\pgfsetdash{}{0pt}%
\pgfpathmoveto{\pgfqpoint{0.374692in}{2.020142in}}%
\pgfpathlineto{\pgfqpoint{5.716193in}{2.020142in}}%
\pgfpathlineto{\pgfqpoint{5.716193in}{2.758765in}}%
\pgfpathlineto{\pgfqpoint{0.374692in}{2.758765in}}%
\pgfpathlineto{\pgfqpoint{0.374692in}{2.020142in}}%
\pgfpathclose%
\pgfusepath{fill}%
\end{pgfscope}%
\begin{pgfscope}%
\pgfpathrectangle{\pgfqpoint{0.374692in}{2.020142in}}{\pgfqpoint{5.341501in}{0.738623in}}%
\pgfusepath{clip}%
\pgfsetbuttcap%
\pgfsetmiterjoin%
\definecolor{currentfill}{rgb}{0.121569,0.466667,0.705882}%
\pgfsetfillcolor{currentfill}%
\pgfsetlinewidth{0.000000pt}%
\definecolor{currentstroke}{rgb}{0.000000,0.000000,0.000000}%
\pgfsetstrokecolor{currentstroke}%
\pgfsetstrokeopacity{0.000000}%
\pgfsetdash{}{0pt}%
\pgfpathmoveto{\pgfqpoint{0.617487in}{2.020142in}}%
\pgfpathlineto{\pgfqpoint{0.784933in}{2.020142in}}%
\pgfpathlineto{\pgfqpoint{0.784933in}{2.758765in}}%
\pgfpathlineto{\pgfqpoint{0.617487in}{2.758765in}}%
\pgfpathlineto{\pgfqpoint{0.617487in}{2.020142in}}%
\pgfpathclose%
\pgfusepath{fill}%
\end{pgfscope}%
\begin{pgfscope}%
\pgfpathrectangle{\pgfqpoint{0.374692in}{2.020142in}}{\pgfqpoint{5.341501in}{0.738623in}}%
\pgfusepath{clip}%
\pgfsetbuttcap%
\pgfsetmiterjoin%
\definecolor{currentfill}{rgb}{0.121569,0.466667,0.705882}%
\pgfsetfillcolor{currentfill}%
\pgfsetlinewidth{0.000000pt}%
\definecolor{currentstroke}{rgb}{0.000000,0.000000,0.000000}%
\pgfsetstrokecolor{currentstroke}%
\pgfsetstrokeopacity{0.000000}%
\pgfsetdash{}{0pt}%
\pgfpathmoveto{\pgfqpoint{0.952378in}{2.020142in}}%
\pgfpathlineto{\pgfqpoint{1.119823in}{2.020142in}}%
\pgfpathlineto{\pgfqpoint{1.119823in}{2.758765in}}%
\pgfpathlineto{\pgfqpoint{0.952378in}{2.758765in}}%
\pgfpathlineto{\pgfqpoint{0.952378in}{2.020142in}}%
\pgfpathclose%
\pgfusepath{fill}%
\end{pgfscope}%
\begin{pgfscope}%
\pgfpathrectangle{\pgfqpoint{0.374692in}{2.020142in}}{\pgfqpoint{5.341501in}{0.738623in}}%
\pgfusepath{clip}%
\pgfsetbuttcap%
\pgfsetmiterjoin%
\definecolor{currentfill}{rgb}{0.121569,0.466667,0.705882}%
\pgfsetfillcolor{currentfill}%
\pgfsetlinewidth{0.000000pt}%
\definecolor{currentstroke}{rgb}{0.000000,0.000000,0.000000}%
\pgfsetstrokecolor{currentstroke}%
\pgfsetstrokeopacity{0.000000}%
\pgfsetdash{}{0pt}%
\pgfpathmoveto{\pgfqpoint{1.287268in}{2.020142in}}%
\pgfpathlineto{\pgfqpoint{1.454713in}{2.020142in}}%
\pgfpathlineto{\pgfqpoint{1.454713in}{2.729220in}}%
\pgfpathlineto{\pgfqpoint{1.287268in}{2.729220in}}%
\pgfpathlineto{\pgfqpoint{1.287268in}{2.020142in}}%
\pgfpathclose%
\pgfusepath{fill}%
\end{pgfscope}%
\begin{pgfscope}%
\pgfpathrectangle{\pgfqpoint{0.374692in}{2.020142in}}{\pgfqpoint{5.341501in}{0.738623in}}%
\pgfusepath{clip}%
\pgfsetbuttcap%
\pgfsetmiterjoin%
\definecolor{currentfill}{rgb}{0.121569,0.466667,0.705882}%
\pgfsetfillcolor{currentfill}%
\pgfsetlinewidth{0.000000pt}%
\definecolor{currentstroke}{rgb}{0.000000,0.000000,0.000000}%
\pgfsetstrokecolor{currentstroke}%
\pgfsetstrokeopacity{0.000000}%
\pgfsetdash{}{0pt}%
\pgfpathmoveto{\pgfqpoint{1.622159in}{2.020142in}}%
\pgfpathlineto{\pgfqpoint{1.789604in}{2.020142in}}%
\pgfpathlineto{\pgfqpoint{1.789604in}{2.758765in}}%
\pgfpathlineto{\pgfqpoint{1.622159in}{2.758765in}}%
\pgfpathlineto{\pgfqpoint{1.622159in}{2.020142in}}%
\pgfpathclose%
\pgfusepath{fill}%
\end{pgfscope}%
\begin{pgfscope}%
\pgfpathrectangle{\pgfqpoint{0.374692in}{2.020142in}}{\pgfqpoint{5.341501in}{0.738623in}}%
\pgfusepath{clip}%
\pgfsetbuttcap%
\pgfsetmiterjoin%
\definecolor{currentfill}{rgb}{0.121569,0.466667,0.705882}%
\pgfsetfillcolor{currentfill}%
\pgfsetlinewidth{0.000000pt}%
\definecolor{currentstroke}{rgb}{0.000000,0.000000,0.000000}%
\pgfsetstrokecolor{currentstroke}%
\pgfsetstrokeopacity{0.000000}%
\pgfsetdash{}{0pt}%
\pgfpathmoveto{\pgfqpoint{1.957049in}{2.020142in}}%
\pgfpathlineto{\pgfqpoint{2.124494in}{2.020142in}}%
\pgfpathlineto{\pgfqpoint{2.124494in}{2.758765in}}%
\pgfpathlineto{\pgfqpoint{1.957049in}{2.758765in}}%
\pgfpathlineto{\pgfqpoint{1.957049in}{2.020142in}}%
\pgfpathclose%
\pgfusepath{fill}%
\end{pgfscope}%
\begin{pgfscope}%
\pgfpathrectangle{\pgfqpoint{0.374692in}{2.020142in}}{\pgfqpoint{5.341501in}{0.738623in}}%
\pgfusepath{clip}%
\pgfsetbuttcap%
\pgfsetmiterjoin%
\definecolor{currentfill}{rgb}{0.121569,0.466667,0.705882}%
\pgfsetfillcolor{currentfill}%
\pgfsetlinewidth{0.000000pt}%
\definecolor{currentstroke}{rgb}{0.000000,0.000000,0.000000}%
\pgfsetstrokecolor{currentstroke}%
\pgfsetstrokeopacity{0.000000}%
\pgfsetdash{}{0pt}%
\pgfpathmoveto{\pgfqpoint{2.291939in}{2.020142in}}%
\pgfpathlineto{\pgfqpoint{2.459384in}{2.020142in}}%
\pgfpathlineto{\pgfqpoint{2.459384in}{2.751228in}}%
\pgfpathlineto{\pgfqpoint{2.291939in}{2.751228in}}%
\pgfpathlineto{\pgfqpoint{2.291939in}{2.020142in}}%
\pgfpathclose%
\pgfusepath{fill}%
\end{pgfscope}%
\begin{pgfscope}%
\pgfpathrectangle{\pgfqpoint{0.374692in}{2.020142in}}{\pgfqpoint{5.341501in}{0.738623in}}%
\pgfusepath{clip}%
\pgfsetbuttcap%
\pgfsetmiterjoin%
\definecolor{currentfill}{rgb}{0.121569,0.466667,0.705882}%
\pgfsetfillcolor{currentfill}%
\pgfsetlinewidth{0.000000pt}%
\definecolor{currentstroke}{rgb}{0.000000,0.000000,0.000000}%
\pgfsetstrokecolor{currentstroke}%
\pgfsetstrokeopacity{0.000000}%
\pgfsetdash{}{0pt}%
\pgfpathmoveto{\pgfqpoint{2.626830in}{2.020142in}}%
\pgfpathlineto{\pgfqpoint{2.794275in}{2.020142in}}%
\pgfpathlineto{\pgfqpoint{2.794275in}{2.758765in}}%
\pgfpathlineto{\pgfqpoint{2.626830in}{2.758765in}}%
\pgfpathlineto{\pgfqpoint{2.626830in}{2.020142in}}%
\pgfpathclose%
\pgfusepath{fill}%
\end{pgfscope}%
\begin{pgfscope}%
\pgfpathrectangle{\pgfqpoint{0.374692in}{2.020142in}}{\pgfqpoint{5.341501in}{0.738623in}}%
\pgfusepath{clip}%
\pgfsetbuttcap%
\pgfsetmiterjoin%
\definecolor{currentfill}{rgb}{0.121569,0.466667,0.705882}%
\pgfsetfillcolor{currentfill}%
\pgfsetlinewidth{0.000000pt}%
\definecolor{currentstroke}{rgb}{0.000000,0.000000,0.000000}%
\pgfsetstrokecolor{currentstroke}%
\pgfsetstrokeopacity{0.000000}%
\pgfsetdash{}{0pt}%
\pgfpathmoveto{\pgfqpoint{2.961720in}{2.020142in}}%
\pgfpathlineto{\pgfqpoint{3.129165in}{2.020142in}}%
\pgfpathlineto{\pgfqpoint{3.129165in}{2.758765in}}%
\pgfpathlineto{\pgfqpoint{2.961720in}{2.758765in}}%
\pgfpathlineto{\pgfqpoint{2.961720in}{2.020142in}}%
\pgfpathclose%
\pgfusepath{fill}%
\end{pgfscope}%
\begin{pgfscope}%
\pgfpathrectangle{\pgfqpoint{0.374692in}{2.020142in}}{\pgfqpoint{5.341501in}{0.738623in}}%
\pgfusepath{clip}%
\pgfsetbuttcap%
\pgfsetmiterjoin%
\definecolor{currentfill}{rgb}{0.121569,0.466667,0.705882}%
\pgfsetfillcolor{currentfill}%
\pgfsetlinewidth{0.000000pt}%
\definecolor{currentstroke}{rgb}{0.000000,0.000000,0.000000}%
\pgfsetstrokecolor{currentstroke}%
\pgfsetstrokeopacity{0.000000}%
\pgfsetdash{}{0pt}%
\pgfpathmoveto{\pgfqpoint{3.296610in}{2.020142in}}%
\pgfpathlineto{\pgfqpoint{3.464055in}{2.020142in}}%
\pgfpathlineto{\pgfqpoint{3.464055in}{2.758765in}}%
\pgfpathlineto{\pgfqpoint{3.296610in}{2.758765in}}%
\pgfpathlineto{\pgfqpoint{3.296610in}{2.020142in}}%
\pgfpathclose%
\pgfusepath{fill}%
\end{pgfscope}%
\begin{pgfscope}%
\pgfpathrectangle{\pgfqpoint{0.374692in}{2.020142in}}{\pgfqpoint{5.341501in}{0.738623in}}%
\pgfusepath{clip}%
\pgfsetbuttcap%
\pgfsetmiterjoin%
\definecolor{currentfill}{rgb}{0.121569,0.466667,0.705882}%
\pgfsetfillcolor{currentfill}%
\pgfsetlinewidth{0.000000pt}%
\definecolor{currentstroke}{rgb}{0.000000,0.000000,0.000000}%
\pgfsetstrokecolor{currentstroke}%
\pgfsetstrokeopacity{0.000000}%
\pgfsetdash{}{0pt}%
\pgfpathmoveto{\pgfqpoint{3.631501in}{2.020142in}}%
\pgfpathlineto{\pgfqpoint{3.798946in}{2.020142in}}%
\pgfpathlineto{\pgfqpoint{3.798946in}{2.758765in}}%
\pgfpathlineto{\pgfqpoint{3.631501in}{2.758765in}}%
\pgfpathlineto{\pgfqpoint{3.631501in}{2.020142in}}%
\pgfpathclose%
\pgfusepath{fill}%
\end{pgfscope}%
\begin{pgfscope}%
\pgfpathrectangle{\pgfqpoint{0.374692in}{2.020142in}}{\pgfqpoint{5.341501in}{0.738623in}}%
\pgfusepath{clip}%
\pgfsetbuttcap%
\pgfsetmiterjoin%
\definecolor{currentfill}{rgb}{0.121569,0.466667,0.705882}%
\pgfsetfillcolor{currentfill}%
\pgfsetlinewidth{0.000000pt}%
\definecolor{currentstroke}{rgb}{0.000000,0.000000,0.000000}%
\pgfsetstrokecolor{currentstroke}%
\pgfsetstrokeopacity{0.000000}%
\pgfsetdash{}{0pt}%
\pgfpathmoveto{\pgfqpoint{3.966391in}{2.020142in}}%
\pgfpathlineto{\pgfqpoint{4.133836in}{2.020142in}}%
\pgfpathlineto{\pgfqpoint{4.133836in}{2.653248in}}%
\pgfpathlineto{\pgfqpoint{3.966391in}{2.653248in}}%
\pgfpathlineto{\pgfqpoint{3.966391in}{2.020142in}}%
\pgfpathclose%
\pgfusepath{fill}%
\end{pgfscope}%
\begin{pgfscope}%
\pgfpathrectangle{\pgfqpoint{0.374692in}{2.020142in}}{\pgfqpoint{5.341501in}{0.738623in}}%
\pgfusepath{clip}%
\pgfsetbuttcap%
\pgfsetmiterjoin%
\definecolor{currentfill}{rgb}{0.121569,0.466667,0.705882}%
\pgfsetfillcolor{currentfill}%
\pgfsetlinewidth{0.000000pt}%
\definecolor{currentstroke}{rgb}{0.000000,0.000000,0.000000}%
\pgfsetstrokecolor{currentstroke}%
\pgfsetstrokeopacity{0.000000}%
\pgfsetdash{}{0pt}%
\pgfpathmoveto{\pgfqpoint{4.301281in}{2.020142in}}%
\pgfpathlineto{\pgfqpoint{4.468727in}{2.020142in}}%
\pgfpathlineto{\pgfqpoint{4.468727in}{2.687286in}}%
\pgfpathlineto{\pgfqpoint{4.301281in}{2.687286in}}%
\pgfpathlineto{\pgfqpoint{4.301281in}{2.020142in}}%
\pgfpathclose%
\pgfusepath{fill}%
\end{pgfscope}%
\begin{pgfscope}%
\pgfpathrectangle{\pgfqpoint{0.374692in}{2.020142in}}{\pgfqpoint{5.341501in}{0.738623in}}%
\pgfusepath{clip}%
\pgfsetbuttcap%
\pgfsetmiterjoin%
\definecolor{currentfill}{rgb}{0.121569,0.466667,0.705882}%
\pgfsetfillcolor{currentfill}%
\pgfsetlinewidth{0.000000pt}%
\definecolor{currentstroke}{rgb}{0.000000,0.000000,0.000000}%
\pgfsetstrokecolor{currentstroke}%
\pgfsetstrokeopacity{0.000000}%
\pgfsetdash{}{0pt}%
\pgfpathmoveto{\pgfqpoint{4.636172in}{2.020142in}}%
\pgfpathlineto{\pgfqpoint{4.803617in}{2.020142in}}%
\pgfpathlineto{\pgfqpoint{4.803617in}{2.758765in}}%
\pgfpathlineto{\pgfqpoint{4.636172in}{2.758765in}}%
\pgfpathlineto{\pgfqpoint{4.636172in}{2.020142in}}%
\pgfpathclose%
\pgfusepath{fill}%
\end{pgfscope}%
\begin{pgfscope}%
\pgfpathrectangle{\pgfqpoint{0.374692in}{2.020142in}}{\pgfqpoint{5.341501in}{0.738623in}}%
\pgfusepath{clip}%
\pgfsetbuttcap%
\pgfsetmiterjoin%
\definecolor{currentfill}{rgb}{0.121569,0.466667,0.705882}%
\pgfsetfillcolor{currentfill}%
\pgfsetlinewidth{0.000000pt}%
\definecolor{currentstroke}{rgb}{0.000000,0.000000,0.000000}%
\pgfsetstrokecolor{currentstroke}%
\pgfsetstrokeopacity{0.000000}%
\pgfsetdash{}{0pt}%
\pgfpathmoveto{\pgfqpoint{4.971062in}{2.020142in}}%
\pgfpathlineto{\pgfqpoint{5.138507in}{2.020142in}}%
\pgfpathlineto{\pgfqpoint{5.138507in}{2.569815in}}%
\pgfpathlineto{\pgfqpoint{4.971062in}{2.569815in}}%
\pgfpathlineto{\pgfqpoint{4.971062in}{2.020142in}}%
\pgfpathclose%
\pgfusepath{fill}%
\end{pgfscope}%
\begin{pgfscope}%
\pgfpathrectangle{\pgfqpoint{0.374692in}{2.020142in}}{\pgfqpoint{5.341501in}{0.738623in}}%
\pgfusepath{clip}%
\pgfsetbuttcap%
\pgfsetmiterjoin%
\definecolor{currentfill}{rgb}{0.121569,0.466667,0.705882}%
\pgfsetfillcolor{currentfill}%
\pgfsetlinewidth{0.000000pt}%
\definecolor{currentstroke}{rgb}{0.000000,0.000000,0.000000}%
\pgfsetstrokecolor{currentstroke}%
\pgfsetstrokeopacity{0.000000}%
\pgfsetdash{}{0pt}%
\pgfpathmoveto{\pgfqpoint{5.305952in}{2.020142in}}%
\pgfpathlineto{\pgfqpoint{5.473398in}{2.020142in}}%
\pgfpathlineto{\pgfqpoint{5.473398in}{2.758765in}}%
\pgfpathlineto{\pgfqpoint{5.305952in}{2.758765in}}%
\pgfpathlineto{\pgfqpoint{5.305952in}{2.020142in}}%
\pgfpathclose%
\pgfusepath{fill}%
\end{pgfscope}%
\begin{pgfscope}%
\pgfpathrectangle{\pgfqpoint{0.374692in}{2.020142in}}{\pgfqpoint{5.341501in}{0.738623in}}%
\pgfusepath{clip}%
\pgfsetbuttcap%
\pgfsetmiterjoin%
\definecolor{currentfill}{rgb}{1.000000,0.498039,0.054902}%
\pgfsetfillcolor{currentfill}%
\pgfsetlinewidth{0.000000pt}%
\definecolor{currentstroke}{rgb}{0.000000,0.000000,0.000000}%
\pgfsetstrokecolor{currentstroke}%
\pgfsetstrokeopacity{0.000000}%
\pgfsetdash{}{0pt}%
\pgfpathmoveto{\pgfqpoint{0.617487in}{2.758765in}}%
\pgfpathlineto{\pgfqpoint{0.784933in}{2.758765in}}%
\pgfpathlineto{\pgfqpoint{0.784933in}{2.758765in}}%
\pgfpathlineto{\pgfqpoint{0.617487in}{2.758765in}}%
\pgfpathlineto{\pgfqpoint{0.617487in}{2.758765in}}%
\pgfpathclose%
\pgfusepath{fill}%
\end{pgfscope}%
\begin{pgfscope}%
\pgfpathrectangle{\pgfqpoint{0.374692in}{2.020142in}}{\pgfqpoint{5.341501in}{0.738623in}}%
\pgfusepath{clip}%
\pgfsetbuttcap%
\pgfsetmiterjoin%
\definecolor{currentfill}{rgb}{1.000000,0.498039,0.054902}%
\pgfsetfillcolor{currentfill}%
\pgfsetlinewidth{0.000000pt}%
\definecolor{currentstroke}{rgb}{0.000000,0.000000,0.000000}%
\pgfsetstrokecolor{currentstroke}%
\pgfsetstrokeopacity{0.000000}%
\pgfsetdash{}{0pt}%
\pgfpathmoveto{\pgfqpoint{0.952378in}{2.758765in}}%
\pgfpathlineto{\pgfqpoint{1.119823in}{2.758765in}}%
\pgfpathlineto{\pgfqpoint{1.119823in}{2.758765in}}%
\pgfpathlineto{\pgfqpoint{0.952378in}{2.758765in}}%
\pgfpathlineto{\pgfqpoint{0.952378in}{2.758765in}}%
\pgfpathclose%
\pgfusepath{fill}%
\end{pgfscope}%
\begin{pgfscope}%
\pgfpathrectangle{\pgfqpoint{0.374692in}{2.020142in}}{\pgfqpoint{5.341501in}{0.738623in}}%
\pgfusepath{clip}%
\pgfsetbuttcap%
\pgfsetmiterjoin%
\definecolor{currentfill}{rgb}{1.000000,0.498039,0.054902}%
\pgfsetfillcolor{currentfill}%
\pgfsetlinewidth{0.000000pt}%
\definecolor{currentstroke}{rgb}{0.000000,0.000000,0.000000}%
\pgfsetstrokecolor{currentstroke}%
\pgfsetstrokeopacity{0.000000}%
\pgfsetdash{}{0pt}%
\pgfpathmoveto{\pgfqpoint{1.287268in}{2.729220in}}%
\pgfpathlineto{\pgfqpoint{1.454713in}{2.729220in}}%
\pgfpathlineto{\pgfqpoint{1.454713in}{2.758765in}}%
\pgfpathlineto{\pgfqpoint{1.287268in}{2.758765in}}%
\pgfpathlineto{\pgfqpoint{1.287268in}{2.729220in}}%
\pgfpathclose%
\pgfusepath{fill}%
\end{pgfscope}%
\begin{pgfscope}%
\pgfpathrectangle{\pgfqpoint{0.374692in}{2.020142in}}{\pgfqpoint{5.341501in}{0.738623in}}%
\pgfusepath{clip}%
\pgfsetbuttcap%
\pgfsetmiterjoin%
\definecolor{currentfill}{rgb}{1.000000,0.498039,0.054902}%
\pgfsetfillcolor{currentfill}%
\pgfsetlinewidth{0.000000pt}%
\definecolor{currentstroke}{rgb}{0.000000,0.000000,0.000000}%
\pgfsetstrokecolor{currentstroke}%
\pgfsetstrokeopacity{0.000000}%
\pgfsetdash{}{0pt}%
\pgfpathmoveto{\pgfqpoint{1.622159in}{2.758765in}}%
\pgfpathlineto{\pgfqpoint{1.789604in}{2.758765in}}%
\pgfpathlineto{\pgfqpoint{1.789604in}{2.758765in}}%
\pgfpathlineto{\pgfqpoint{1.622159in}{2.758765in}}%
\pgfpathlineto{\pgfqpoint{1.622159in}{2.758765in}}%
\pgfpathclose%
\pgfusepath{fill}%
\end{pgfscope}%
\begin{pgfscope}%
\pgfpathrectangle{\pgfqpoint{0.374692in}{2.020142in}}{\pgfqpoint{5.341501in}{0.738623in}}%
\pgfusepath{clip}%
\pgfsetbuttcap%
\pgfsetmiterjoin%
\definecolor{currentfill}{rgb}{1.000000,0.498039,0.054902}%
\pgfsetfillcolor{currentfill}%
\pgfsetlinewidth{0.000000pt}%
\definecolor{currentstroke}{rgb}{0.000000,0.000000,0.000000}%
\pgfsetstrokecolor{currentstroke}%
\pgfsetstrokeopacity{0.000000}%
\pgfsetdash{}{0pt}%
\pgfpathmoveto{\pgfqpoint{1.957049in}{2.758765in}}%
\pgfpathlineto{\pgfqpoint{2.124494in}{2.758765in}}%
\pgfpathlineto{\pgfqpoint{2.124494in}{2.758765in}}%
\pgfpathlineto{\pgfqpoint{1.957049in}{2.758765in}}%
\pgfpathlineto{\pgfqpoint{1.957049in}{2.758765in}}%
\pgfpathclose%
\pgfusepath{fill}%
\end{pgfscope}%
\begin{pgfscope}%
\pgfpathrectangle{\pgfqpoint{0.374692in}{2.020142in}}{\pgfqpoint{5.341501in}{0.738623in}}%
\pgfusepath{clip}%
\pgfsetbuttcap%
\pgfsetmiterjoin%
\definecolor{currentfill}{rgb}{1.000000,0.498039,0.054902}%
\pgfsetfillcolor{currentfill}%
\pgfsetlinewidth{0.000000pt}%
\definecolor{currentstroke}{rgb}{0.000000,0.000000,0.000000}%
\pgfsetstrokecolor{currentstroke}%
\pgfsetstrokeopacity{0.000000}%
\pgfsetdash{}{0pt}%
\pgfpathmoveto{\pgfqpoint{2.291939in}{2.751228in}}%
\pgfpathlineto{\pgfqpoint{2.459384in}{2.751228in}}%
\pgfpathlineto{\pgfqpoint{2.459384in}{2.758765in}}%
\pgfpathlineto{\pgfqpoint{2.291939in}{2.758765in}}%
\pgfpathlineto{\pgfqpoint{2.291939in}{2.751228in}}%
\pgfpathclose%
\pgfusepath{fill}%
\end{pgfscope}%
\begin{pgfscope}%
\pgfpathrectangle{\pgfqpoint{0.374692in}{2.020142in}}{\pgfqpoint{5.341501in}{0.738623in}}%
\pgfusepath{clip}%
\pgfsetbuttcap%
\pgfsetmiterjoin%
\definecolor{currentfill}{rgb}{1.000000,0.498039,0.054902}%
\pgfsetfillcolor{currentfill}%
\pgfsetlinewidth{0.000000pt}%
\definecolor{currentstroke}{rgb}{0.000000,0.000000,0.000000}%
\pgfsetstrokecolor{currentstroke}%
\pgfsetstrokeopacity{0.000000}%
\pgfsetdash{}{0pt}%
\pgfpathmoveto{\pgfqpoint{2.626830in}{2.758765in}}%
\pgfpathlineto{\pgfqpoint{2.794275in}{2.758765in}}%
\pgfpathlineto{\pgfqpoint{2.794275in}{2.758765in}}%
\pgfpathlineto{\pgfqpoint{2.626830in}{2.758765in}}%
\pgfpathlineto{\pgfqpoint{2.626830in}{2.758765in}}%
\pgfpathclose%
\pgfusepath{fill}%
\end{pgfscope}%
\begin{pgfscope}%
\pgfpathrectangle{\pgfqpoint{0.374692in}{2.020142in}}{\pgfqpoint{5.341501in}{0.738623in}}%
\pgfusepath{clip}%
\pgfsetbuttcap%
\pgfsetmiterjoin%
\definecolor{currentfill}{rgb}{1.000000,0.498039,0.054902}%
\pgfsetfillcolor{currentfill}%
\pgfsetlinewidth{0.000000pt}%
\definecolor{currentstroke}{rgb}{0.000000,0.000000,0.000000}%
\pgfsetstrokecolor{currentstroke}%
\pgfsetstrokeopacity{0.000000}%
\pgfsetdash{}{0pt}%
\pgfpathmoveto{\pgfqpoint{2.961720in}{2.758765in}}%
\pgfpathlineto{\pgfqpoint{3.129165in}{2.758765in}}%
\pgfpathlineto{\pgfqpoint{3.129165in}{2.758765in}}%
\pgfpathlineto{\pgfqpoint{2.961720in}{2.758765in}}%
\pgfpathlineto{\pgfqpoint{2.961720in}{2.758765in}}%
\pgfpathclose%
\pgfusepath{fill}%
\end{pgfscope}%
\begin{pgfscope}%
\pgfpathrectangle{\pgfqpoint{0.374692in}{2.020142in}}{\pgfqpoint{5.341501in}{0.738623in}}%
\pgfusepath{clip}%
\pgfsetbuttcap%
\pgfsetmiterjoin%
\definecolor{currentfill}{rgb}{1.000000,0.498039,0.054902}%
\pgfsetfillcolor{currentfill}%
\pgfsetlinewidth{0.000000pt}%
\definecolor{currentstroke}{rgb}{0.000000,0.000000,0.000000}%
\pgfsetstrokecolor{currentstroke}%
\pgfsetstrokeopacity{0.000000}%
\pgfsetdash{}{0pt}%
\pgfpathmoveto{\pgfqpoint{3.296610in}{2.758765in}}%
\pgfpathlineto{\pgfqpoint{3.464055in}{2.758765in}}%
\pgfpathlineto{\pgfqpoint{3.464055in}{2.758765in}}%
\pgfpathlineto{\pgfqpoint{3.296610in}{2.758765in}}%
\pgfpathlineto{\pgfqpoint{3.296610in}{2.758765in}}%
\pgfpathclose%
\pgfusepath{fill}%
\end{pgfscope}%
\begin{pgfscope}%
\pgfpathrectangle{\pgfqpoint{0.374692in}{2.020142in}}{\pgfqpoint{5.341501in}{0.738623in}}%
\pgfusepath{clip}%
\pgfsetbuttcap%
\pgfsetmiterjoin%
\definecolor{currentfill}{rgb}{1.000000,0.498039,0.054902}%
\pgfsetfillcolor{currentfill}%
\pgfsetlinewidth{0.000000pt}%
\definecolor{currentstroke}{rgb}{0.000000,0.000000,0.000000}%
\pgfsetstrokecolor{currentstroke}%
\pgfsetstrokeopacity{0.000000}%
\pgfsetdash{}{0pt}%
\pgfpathmoveto{\pgfqpoint{3.631501in}{2.758765in}}%
\pgfpathlineto{\pgfqpoint{3.798946in}{2.758765in}}%
\pgfpathlineto{\pgfqpoint{3.798946in}{2.758765in}}%
\pgfpathlineto{\pgfqpoint{3.631501in}{2.758765in}}%
\pgfpathlineto{\pgfqpoint{3.631501in}{2.758765in}}%
\pgfpathclose%
\pgfusepath{fill}%
\end{pgfscope}%
\begin{pgfscope}%
\pgfpathrectangle{\pgfqpoint{0.374692in}{2.020142in}}{\pgfqpoint{5.341501in}{0.738623in}}%
\pgfusepath{clip}%
\pgfsetbuttcap%
\pgfsetmiterjoin%
\definecolor{currentfill}{rgb}{1.000000,0.498039,0.054902}%
\pgfsetfillcolor{currentfill}%
\pgfsetlinewidth{0.000000pt}%
\definecolor{currentstroke}{rgb}{0.000000,0.000000,0.000000}%
\pgfsetstrokecolor{currentstroke}%
\pgfsetstrokeopacity{0.000000}%
\pgfsetdash{}{0pt}%
\pgfpathmoveto{\pgfqpoint{3.966391in}{2.653248in}}%
\pgfpathlineto{\pgfqpoint{4.133836in}{2.653248in}}%
\pgfpathlineto{\pgfqpoint{4.133836in}{2.758765in}}%
\pgfpathlineto{\pgfqpoint{3.966391in}{2.758765in}}%
\pgfpathlineto{\pgfqpoint{3.966391in}{2.653248in}}%
\pgfpathclose%
\pgfusepath{fill}%
\end{pgfscope}%
\begin{pgfscope}%
\pgfpathrectangle{\pgfqpoint{0.374692in}{2.020142in}}{\pgfqpoint{5.341501in}{0.738623in}}%
\pgfusepath{clip}%
\pgfsetbuttcap%
\pgfsetmiterjoin%
\definecolor{currentfill}{rgb}{1.000000,0.498039,0.054902}%
\pgfsetfillcolor{currentfill}%
\pgfsetlinewidth{0.000000pt}%
\definecolor{currentstroke}{rgb}{0.000000,0.000000,0.000000}%
\pgfsetstrokecolor{currentstroke}%
\pgfsetstrokeopacity{0.000000}%
\pgfsetdash{}{0pt}%
\pgfpathmoveto{\pgfqpoint{4.301281in}{2.687286in}}%
\pgfpathlineto{\pgfqpoint{4.468727in}{2.687286in}}%
\pgfpathlineto{\pgfqpoint{4.468727in}{2.758765in}}%
\pgfpathlineto{\pgfqpoint{4.301281in}{2.758765in}}%
\pgfpathlineto{\pgfqpoint{4.301281in}{2.687286in}}%
\pgfpathclose%
\pgfusepath{fill}%
\end{pgfscope}%
\begin{pgfscope}%
\pgfpathrectangle{\pgfqpoint{0.374692in}{2.020142in}}{\pgfqpoint{5.341501in}{0.738623in}}%
\pgfusepath{clip}%
\pgfsetbuttcap%
\pgfsetmiterjoin%
\definecolor{currentfill}{rgb}{1.000000,0.498039,0.054902}%
\pgfsetfillcolor{currentfill}%
\pgfsetlinewidth{0.000000pt}%
\definecolor{currentstroke}{rgb}{0.000000,0.000000,0.000000}%
\pgfsetstrokecolor{currentstroke}%
\pgfsetstrokeopacity{0.000000}%
\pgfsetdash{}{0pt}%
\pgfpathmoveto{\pgfqpoint{4.636172in}{2.758765in}}%
\pgfpathlineto{\pgfqpoint{4.803617in}{2.758765in}}%
\pgfpathlineto{\pgfqpoint{4.803617in}{2.758765in}}%
\pgfpathlineto{\pgfqpoint{4.636172in}{2.758765in}}%
\pgfpathlineto{\pgfqpoint{4.636172in}{2.758765in}}%
\pgfpathclose%
\pgfusepath{fill}%
\end{pgfscope}%
\begin{pgfscope}%
\pgfpathrectangle{\pgfqpoint{0.374692in}{2.020142in}}{\pgfqpoint{5.341501in}{0.738623in}}%
\pgfusepath{clip}%
\pgfsetbuttcap%
\pgfsetmiterjoin%
\definecolor{currentfill}{rgb}{1.000000,0.498039,0.054902}%
\pgfsetfillcolor{currentfill}%
\pgfsetlinewidth{0.000000pt}%
\definecolor{currentstroke}{rgb}{0.000000,0.000000,0.000000}%
\pgfsetstrokecolor{currentstroke}%
\pgfsetstrokeopacity{0.000000}%
\pgfsetdash{}{0pt}%
\pgfpathmoveto{\pgfqpoint{4.971062in}{2.569815in}}%
\pgfpathlineto{\pgfqpoint{5.138507in}{2.569815in}}%
\pgfpathlineto{\pgfqpoint{5.138507in}{2.758765in}}%
\pgfpathlineto{\pgfqpoint{4.971062in}{2.758765in}}%
\pgfpathlineto{\pgfqpoint{4.971062in}{2.569815in}}%
\pgfpathclose%
\pgfusepath{fill}%
\end{pgfscope}%
\begin{pgfscope}%
\pgfpathrectangle{\pgfqpoint{0.374692in}{2.020142in}}{\pgfqpoint{5.341501in}{0.738623in}}%
\pgfusepath{clip}%
\pgfsetbuttcap%
\pgfsetmiterjoin%
\definecolor{currentfill}{rgb}{1.000000,0.498039,0.054902}%
\pgfsetfillcolor{currentfill}%
\pgfsetlinewidth{0.000000pt}%
\definecolor{currentstroke}{rgb}{0.000000,0.000000,0.000000}%
\pgfsetstrokecolor{currentstroke}%
\pgfsetstrokeopacity{0.000000}%
\pgfsetdash{}{0pt}%
\pgfpathmoveto{\pgfqpoint{5.305952in}{2.758765in}}%
\pgfpathlineto{\pgfqpoint{5.473398in}{2.758765in}}%
\pgfpathlineto{\pgfqpoint{5.473398in}{2.758765in}}%
\pgfpathlineto{\pgfqpoint{5.305952in}{2.758765in}}%
\pgfpathlineto{\pgfqpoint{5.305952in}{2.758765in}}%
\pgfpathclose%
\pgfusepath{fill}%
\end{pgfscope}%
\begin{pgfscope}%
\pgfsetbuttcap%
\pgfsetroundjoin%
\definecolor{currentfill}{rgb}{0.000000,0.000000,0.000000}%
\pgfsetfillcolor{currentfill}%
\pgfsetlinewidth{0.803000pt}%
\definecolor{currentstroke}{rgb}{0.000000,0.000000,0.000000}%
\pgfsetstrokecolor{currentstroke}%
\pgfsetdash{}{0pt}%
\pgfsys@defobject{currentmarker}{\pgfqpoint{0.000000in}{-0.048611in}}{\pgfqpoint{0.000000in}{0.000000in}}{%
\pgfpathmoveto{\pgfqpoint{0.000000in}{0.000000in}}%
\pgfpathlineto{\pgfqpoint{0.000000in}{-0.048611in}}%
\pgfusepath{stroke,fill}%
}%
\begin{pgfscope}%
\pgfsys@transformshift{0.701210in}{2.020142in}%
\pgfsys@useobject{currentmarker}{}%
\end{pgfscope}%
\end{pgfscope}%
\begin{pgfscope}%
\definecolor{textcolor}{rgb}{0.000000,0.000000,0.000000}%
\pgfsetstrokecolor{textcolor}%
\pgfsetfillcolor{textcolor}%
\pgftext[x=0.735932in, y=1.042517in, left, base,rotate=90.000000]{\color{textcolor}\rmfamily\fontsize{10.000000}{12.000000}\selectfont Audio (Music)}%
\end{pgfscope}%
\begin{pgfscope}%
\pgfsetbuttcap%
\pgfsetroundjoin%
\definecolor{currentfill}{rgb}{0.000000,0.000000,0.000000}%
\pgfsetfillcolor{currentfill}%
\pgfsetlinewidth{0.803000pt}%
\definecolor{currentstroke}{rgb}{0.000000,0.000000,0.000000}%
\pgfsetstrokecolor{currentstroke}%
\pgfsetdash{}{0pt}%
\pgfsys@defobject{currentmarker}{\pgfqpoint{0.000000in}{-0.048611in}}{\pgfqpoint{0.000000in}{0.000000in}}{%
\pgfpathmoveto{\pgfqpoint{0.000000in}{0.000000in}}%
\pgfpathlineto{\pgfqpoint{0.000000in}{-0.048611in}}%
\pgfusepath{stroke,fill}%
}%
\begin{pgfscope}%
\pgfsys@transformshift{1.036100in}{2.020142in}%
\pgfsys@useobject{currentmarker}{}%
\end{pgfscope}%
\end{pgfscope}%
\begin{pgfscope}%
\definecolor{textcolor}{rgb}{0.000000,0.000000,0.000000}%
\pgfsetstrokecolor{textcolor}%
\pgfsetfillcolor{textcolor}%
\pgftext[x=1.070823in, y=0.985418in, left, base,rotate=90.000000]{\color{textcolor}\rmfamily\fontsize{10.000000}{12.000000}\selectfont Audio (Speech)}%
\end{pgfscope}%
\begin{pgfscope}%
\pgfsetbuttcap%
\pgfsetroundjoin%
\definecolor{currentfill}{rgb}{0.000000,0.000000,0.000000}%
\pgfsetfillcolor{currentfill}%
\pgfsetlinewidth{0.803000pt}%
\definecolor{currentstroke}{rgb}{0.000000,0.000000,0.000000}%
\pgfsetstrokecolor{currentstroke}%
\pgfsetdash{}{0pt}%
\pgfsys@defobject{currentmarker}{\pgfqpoint{0.000000in}{-0.048611in}}{\pgfqpoint{0.000000in}{0.000000in}}{%
\pgfpathmoveto{\pgfqpoint{0.000000in}{0.000000in}}%
\pgfpathlineto{\pgfqpoint{0.000000in}{-0.048611in}}%
\pgfusepath{stroke,fill}%
}%
\begin{pgfscope}%
\pgfsys@transformshift{1.370991in}{2.020142in}%
\pgfsys@useobject{currentmarker}{}%
\end{pgfscope}%
\end{pgfscope}%
\begin{pgfscope}%
\definecolor{textcolor}{rgb}{0.000000,0.000000,0.000000}%
\pgfsetstrokecolor{textcolor}%
\pgfsetfillcolor{textcolor}%
\pgftext[x=1.405713in, y=1.535766in, left, base,rotate=90.000000]{\color{textcolor}\rmfamily\fontsize{10.000000}{12.000000}\selectfont Graph}%
\end{pgfscope}%
\begin{pgfscope}%
\pgfsetbuttcap%
\pgfsetroundjoin%
\definecolor{currentfill}{rgb}{0.000000,0.000000,0.000000}%
\pgfsetfillcolor{currentfill}%
\pgfsetlinewidth{0.803000pt}%
\definecolor{currentstroke}{rgb}{0.000000,0.000000,0.000000}%
\pgfsetstrokecolor{currentstroke}%
\pgfsetdash{}{0pt}%
\pgfsys@defobject{currentmarker}{\pgfqpoint{0.000000in}{-0.048611in}}{\pgfqpoint{0.000000in}{0.000000in}}{%
\pgfpathmoveto{\pgfqpoint{0.000000in}{0.000000in}}%
\pgfpathlineto{\pgfqpoint{0.000000in}{-0.048611in}}%
\pgfusepath{stroke,fill}%
}%
\begin{pgfscope}%
\pgfsys@transformshift{1.705881in}{2.020142in}%
\pgfsys@useobject{currentmarker}{}%
\end{pgfscope}%
\end{pgfscope}%
\begin{pgfscope}%
\definecolor{textcolor}{rgb}{0.000000,0.000000,0.000000}%
\pgfsetstrokecolor{textcolor}%
\pgfsetfillcolor{textcolor}%
\pgftext[x=1.740603in, y=0.990819in, left, base,rotate=90.000000]{\color{textcolor}\rmfamily\fontsize{10.000000}{12.000000}\selectfont Image (Binary)}%
\end{pgfscope}%
\begin{pgfscope}%
\pgfsetbuttcap%
\pgfsetroundjoin%
\definecolor{currentfill}{rgb}{0.000000,0.000000,0.000000}%
\pgfsetfillcolor{currentfill}%
\pgfsetlinewidth{0.803000pt}%
\definecolor{currentstroke}{rgb}{0.000000,0.000000,0.000000}%
\pgfsetstrokecolor{currentstroke}%
\pgfsetdash{}{0pt}%
\pgfsys@defobject{currentmarker}{\pgfqpoint{0.000000in}{-0.048611in}}{\pgfqpoint{0.000000in}{0.000000in}}{%
\pgfpathmoveto{\pgfqpoint{0.000000in}{0.000000in}}%
\pgfpathlineto{\pgfqpoint{0.000000in}{-0.048611in}}%
\pgfusepath{stroke,fill}%
}%
\begin{pgfscope}%
\pgfsys@transformshift{2.040771in}{2.020142in}%
\pgfsys@useobject{currentmarker}{}%
\end{pgfscope}%
\end{pgfscope}%
\begin{pgfscope}%
\definecolor{textcolor}{rgb}{0.000000,0.000000,0.000000}%
\pgfsetstrokecolor{textcolor}%
\pgfsetfillcolor{textcolor}%
\pgftext[x=2.075494in, y=0.324922in, left, base,rotate=90.000000]{\color{textcolor}\rmfamily\fontsize{10.000000}{12.000000}\selectfont Image (More Information)}%
\end{pgfscope}%
\begin{pgfscope}%
\pgfsetbuttcap%
\pgfsetroundjoin%
\definecolor{currentfill}{rgb}{0.000000,0.000000,0.000000}%
\pgfsetfillcolor{currentfill}%
\pgfsetlinewidth{0.803000pt}%
\definecolor{currentstroke}{rgb}{0.000000,0.000000,0.000000}%
\pgfsetstrokecolor{currentstroke}%
\pgfsetdash{}{0pt}%
\pgfsys@defobject{currentmarker}{\pgfqpoint{0.000000in}{-0.048611in}}{\pgfqpoint{0.000000in}{0.000000in}}{%
\pgfpathmoveto{\pgfqpoint{0.000000in}{0.000000in}}%
\pgfpathlineto{\pgfqpoint{0.000000in}{-0.048611in}}%
\pgfusepath{stroke,fill}%
}%
\begin{pgfscope}%
\pgfsys@transformshift{2.375662in}{2.020142in}%
\pgfsys@useobject{currentmarker}{}%
\end{pgfscope}%
\end{pgfscope}%
\begin{pgfscope}%
\definecolor{textcolor}{rgb}{0.000000,0.000000,0.000000}%
\pgfsetstrokecolor{textcolor}%
\pgfsetfillcolor{textcolor}%
\pgftext[x=2.410384in, y=0.934877in, left, base,rotate=90.000000]{\color{textcolor}\rmfamily\fontsize{10.000000}{12.000000}\selectfont Image (Natural)}%
\end{pgfscope}%
\begin{pgfscope}%
\pgfsetbuttcap%
\pgfsetroundjoin%
\definecolor{currentfill}{rgb}{0.000000,0.000000,0.000000}%
\pgfsetfillcolor{currentfill}%
\pgfsetlinewidth{0.803000pt}%
\definecolor{currentstroke}{rgb}{0.000000,0.000000,0.000000}%
\pgfsetstrokecolor{currentstroke}%
\pgfsetdash{}{0pt}%
\pgfsys@defobject{currentmarker}{\pgfqpoint{0.000000in}{-0.048611in}}{\pgfqpoint{0.000000in}{0.000000in}}{%
\pgfpathmoveto{\pgfqpoint{0.000000in}{0.000000in}}%
\pgfpathlineto{\pgfqpoint{0.000000in}{-0.048611in}}%
\pgfusepath{stroke,fill}%
}%
\begin{pgfscope}%
\pgfsys@transformshift{2.710552in}{2.020142in}%
\pgfsys@useobject{currentmarker}{}%
\end{pgfscope}%
\end{pgfscope}%
\begin{pgfscope}%
\definecolor{textcolor}{rgb}{0.000000,0.000000,0.000000}%
\pgfsetstrokecolor{textcolor}%
\pgfsetfillcolor{textcolor}%
\pgftext[x=2.745274in, y=0.209182in, left, base,rotate=90.000000]{\color{textcolor}\rmfamily\fontsize{10.000000}{12.000000}\selectfont Image (Segmentation Mask)}%
\end{pgfscope}%
\begin{pgfscope}%
\pgfsetbuttcap%
\pgfsetroundjoin%
\definecolor{currentfill}{rgb}{0.000000,0.000000,0.000000}%
\pgfsetfillcolor{currentfill}%
\pgfsetlinewidth{0.803000pt}%
\definecolor{currentstroke}{rgb}{0.000000,0.000000,0.000000}%
\pgfsetstrokecolor{currentstroke}%
\pgfsetdash{}{0pt}%
\pgfsys@defobject{currentmarker}{\pgfqpoint{0.000000in}{-0.048611in}}{\pgfqpoint{0.000000in}{0.000000in}}{%
\pgfpathmoveto{\pgfqpoint{0.000000in}{0.000000in}}%
\pgfpathlineto{\pgfqpoint{0.000000in}{-0.048611in}}%
\pgfusepath{stroke,fill}%
}%
\begin{pgfscope}%
\pgfsys@transformshift{3.045443in}{2.020142in}%
\pgfsys@useobject{currentmarker}{}%
\end{pgfscope}%
\end{pgfscope}%
\begin{pgfscope}%
\definecolor{textcolor}{rgb}{0.000000,0.000000,0.000000}%
\pgfsetstrokecolor{textcolor}%
\pgfsetfillcolor{textcolor}%
\pgftext[x=3.080165in, y=1.386653in, left, base,rotate=90.000000]{\color{textcolor}\rmfamily\fontsize{10.000000}{12.000000}\selectfont Molecule}%
\end{pgfscope}%
\begin{pgfscope}%
\pgfsetbuttcap%
\pgfsetroundjoin%
\definecolor{currentfill}{rgb}{0.000000,0.000000,0.000000}%
\pgfsetfillcolor{currentfill}%
\pgfsetlinewidth{0.803000pt}%
\definecolor{currentstroke}{rgb}{0.000000,0.000000,0.000000}%
\pgfsetstrokecolor{currentstroke}%
\pgfsetdash{}{0pt}%
\pgfsys@defobject{currentmarker}{\pgfqpoint{0.000000in}{-0.048611in}}{\pgfqpoint{0.000000in}{0.000000in}}{%
\pgfpathmoveto{\pgfqpoint{0.000000in}{0.000000in}}%
\pgfpathlineto{\pgfqpoint{0.000000in}{-0.048611in}}%
\pgfusepath{stroke,fill}%
}%
\begin{pgfscope}%
\pgfsys@transformshift{3.380333in}{2.020142in}%
\pgfsys@useobject{currentmarker}{}%
\end{pgfscope}%
\end{pgfscope}%
\begin{pgfscope}%
\definecolor{textcolor}{rgb}{0.000000,0.000000,0.000000}%
\pgfsetstrokecolor{textcolor}%
\pgfsetfillcolor{textcolor}%
\pgftext[x=3.415055in, y=0.396682in, left, base,rotate=90.000000]{\color{textcolor}\rmfamily\fontsize{10.000000}{12.000000}\selectfont Text (Natural Language)}%
\end{pgfscope}%
\begin{pgfscope}%
\pgfsetbuttcap%
\pgfsetroundjoin%
\definecolor{currentfill}{rgb}{0.000000,0.000000,0.000000}%
\pgfsetfillcolor{currentfill}%
\pgfsetlinewidth{0.803000pt}%
\definecolor{currentstroke}{rgb}{0.000000,0.000000,0.000000}%
\pgfsetstrokecolor{currentstroke}%
\pgfsetdash{}{0pt}%
\pgfsys@defobject{currentmarker}{\pgfqpoint{0.000000in}{-0.048611in}}{\pgfqpoint{0.000000in}{0.000000in}}{%
\pgfpathmoveto{\pgfqpoint{0.000000in}{0.000000in}}%
\pgfpathlineto{\pgfqpoint{0.000000in}{-0.048611in}}%
\pgfusepath{stroke,fill}%
}%
\begin{pgfscope}%
\pgfsys@transformshift{3.715223in}{2.020142in}%
\pgfsys@useobject{currentmarker}{}%
\end{pgfscope}%
\end{pgfscope}%
\begin{pgfscope}%
\definecolor{textcolor}{rgb}{0.000000,0.000000,0.000000}%
\pgfsetstrokecolor{textcolor}%
\pgfsetfillcolor{textcolor}%
\pgftext[x=3.749945in, y=0.581097in, left, base,rotate=90.000000]{\color{textcolor}\rmfamily\fontsize{10.000000}{12.000000}\selectfont Text (Representation)}%
\end{pgfscope}%
\begin{pgfscope}%
\pgfsetbuttcap%
\pgfsetroundjoin%
\definecolor{currentfill}{rgb}{0.000000,0.000000,0.000000}%
\pgfsetfillcolor{currentfill}%
\pgfsetlinewidth{0.803000pt}%
\definecolor{currentstroke}{rgb}{0.000000,0.000000,0.000000}%
\pgfsetstrokecolor{currentstroke}%
\pgfsetdash{}{0pt}%
\pgfsys@defobject{currentmarker}{\pgfqpoint{0.000000in}{-0.048611in}}{\pgfqpoint{0.000000in}{0.000000in}}{%
\pgfpathmoveto{\pgfqpoint{0.000000in}{0.000000in}}%
\pgfpathlineto{\pgfqpoint{0.000000in}{-0.048611in}}%
\pgfusepath{stroke,fill}%
}%
\begin{pgfscope}%
\pgfsys@transformshift{4.050114in}{2.020142in}%
\pgfsys@useobject{currentmarker}{}%
\end{pgfscope}%
\end{pgfscope}%
\begin{pgfscope}%
\definecolor{textcolor}{rgb}{0.000000,0.000000,0.000000}%
\pgfsetstrokecolor{textcolor}%
\pgfsetfillcolor{textcolor}%
\pgftext[x=4.084836in, y=0.424460in, left, base,rotate=90.000000]{\color{textcolor}\rmfamily\fontsize{10.000000}{12.000000}\selectfont Time Series (Univariate)}%
\end{pgfscope}%
\begin{pgfscope}%
\pgfsetbuttcap%
\pgfsetroundjoin%
\definecolor{currentfill}{rgb}{0.000000,0.000000,0.000000}%
\pgfsetfillcolor{currentfill}%
\pgfsetlinewidth{0.803000pt}%
\definecolor{currentstroke}{rgb}{0.000000,0.000000,0.000000}%
\pgfsetstrokecolor{currentstroke}%
\pgfsetdash{}{0pt}%
\pgfsys@defobject{currentmarker}{\pgfqpoint{0.000000in}{-0.048611in}}{\pgfqpoint{0.000000in}{0.000000in}}{%
\pgfpathmoveto{\pgfqpoint{0.000000in}{0.000000in}}%
\pgfpathlineto{\pgfqpoint{0.000000in}{-0.048611in}}%
\pgfusepath{stroke,fill}%
}%
\begin{pgfscope}%
\pgfsys@transformshift{4.385004in}{2.020142in}%
\pgfsys@useobject{currentmarker}{}%
\end{pgfscope}%
\end{pgfscope}%
\begin{pgfscope}%
\definecolor{textcolor}{rgb}{0.000000,0.000000,0.000000}%
\pgfsetstrokecolor{textcolor}%
\pgfsetfillcolor{textcolor}%
\pgftext[x=4.419726in, y=0.308719in, left, base,rotate=90.000000]{\color{textcolor}\rmfamily\fontsize{10.000000}{12.000000}\selectfont Time Series (Multivariate)}%
\end{pgfscope}%
\begin{pgfscope}%
\pgfsetbuttcap%
\pgfsetroundjoin%
\definecolor{currentfill}{rgb}{0.000000,0.000000,0.000000}%
\pgfsetfillcolor{currentfill}%
\pgfsetlinewidth{0.803000pt}%
\definecolor{currentstroke}{rgb}{0.000000,0.000000,0.000000}%
\pgfsetstrokecolor{currentstroke}%
\pgfsetdash{}{0pt}%
\pgfsys@defobject{currentmarker}{\pgfqpoint{0.000000in}{-0.048611in}}{\pgfqpoint{0.000000in}{0.000000in}}{%
\pgfpathmoveto{\pgfqpoint{0.000000in}{0.000000in}}%
\pgfpathlineto{\pgfqpoint{0.000000in}{-0.048611in}}%
\pgfusepath{stroke,fill}%
}%
\begin{pgfscope}%
\pgfsys@transformshift{4.719894in}{2.020142in}%
\pgfsys@useobject{currentmarker}{}%
\end{pgfscope}%
\end{pgfscope}%
\begin{pgfscope}%
\definecolor{textcolor}{rgb}{0.000000,0.000000,0.000000}%
\pgfsetstrokecolor{textcolor}%
\pgfsetfillcolor{textcolor}%
\pgftext[x=4.754616in, y=0.100000in, left, base,rotate=90.000000]{\color{textcolor}\rmfamily\fontsize{10.000000}{12.000000}\selectfont Time Series (Symbolic Music)}%
\end{pgfscope}%
\begin{pgfscope}%
\pgfsetbuttcap%
\pgfsetroundjoin%
\definecolor{currentfill}{rgb}{0.000000,0.000000,0.000000}%
\pgfsetfillcolor{currentfill}%
\pgfsetlinewidth{0.803000pt}%
\definecolor{currentstroke}{rgb}{0.000000,0.000000,0.000000}%
\pgfsetstrokecolor{currentstroke}%
\pgfsetdash{}{0pt}%
\pgfsys@defobject{currentmarker}{\pgfqpoint{0.000000in}{-0.048611in}}{\pgfqpoint{0.000000in}{0.000000in}}{%
\pgfpathmoveto{\pgfqpoint{0.000000in}{0.000000in}}%
\pgfpathlineto{\pgfqpoint{0.000000in}{-0.048611in}}%
\pgfusepath{stroke,fill}%
}%
\begin{pgfscope}%
\pgfsys@transformshift{5.054785in}{2.020142in}%
\pgfsys@useobject{currentmarker}{}%
\end{pgfscope}%
\end{pgfscope}%
\begin{pgfscope}%
\definecolor{textcolor}{rgb}{0.000000,0.000000,0.000000}%
\pgfsetstrokecolor{textcolor}%
\pgfsetfillcolor{textcolor}%
\pgftext[x=5.089507in, y=1.102702in, left, base,rotate=90.000000]{\color{textcolor}\rmfamily\fontsize{10.000000}{12.000000}\selectfont Tabular Data}%
\end{pgfscope}%
\begin{pgfscope}%
\pgfsetbuttcap%
\pgfsetroundjoin%
\definecolor{currentfill}{rgb}{0.000000,0.000000,0.000000}%
\pgfsetfillcolor{currentfill}%
\pgfsetlinewidth{0.803000pt}%
\definecolor{currentstroke}{rgb}{0.000000,0.000000,0.000000}%
\pgfsetstrokecolor{currentstroke}%
\pgfsetdash{}{0pt}%
\pgfsys@defobject{currentmarker}{\pgfqpoint{0.000000in}{-0.048611in}}{\pgfqpoint{0.000000in}{0.000000in}}{%
\pgfpathmoveto{\pgfqpoint{0.000000in}{0.000000in}}%
\pgfpathlineto{\pgfqpoint{0.000000in}{-0.048611in}}%
\pgfusepath{stroke,fill}%
}%
\begin{pgfscope}%
\pgfsys@transformshift{5.389675in}{2.020142in}%
\pgfsys@useobject{currentmarker}{}%
\end{pgfscope}%
\end{pgfscope}%
\begin{pgfscope}%
\definecolor{textcolor}{rgb}{0.000000,0.000000,0.000000}%
\pgfsetstrokecolor{textcolor}%
\pgfsetfillcolor{textcolor}%
\pgftext[x=5.424397in, y=1.571839in, left, base,rotate=90.000000]{\color{textcolor}\rmfamily\fontsize{10.000000}{12.000000}\selectfont Video}%
\end{pgfscope}%
\begin{pgfscope}%
\pgfsetbuttcap%
\pgfsetroundjoin%
\definecolor{currentfill}{rgb}{0.000000,0.000000,0.000000}%
\pgfsetfillcolor{currentfill}%
\pgfsetlinewidth{0.803000pt}%
\definecolor{currentstroke}{rgb}{0.000000,0.000000,0.000000}%
\pgfsetstrokecolor{currentstroke}%
\pgfsetdash{}{0pt}%
\pgfsys@defobject{currentmarker}{\pgfqpoint{-0.048611in}{0.000000in}}{\pgfqpoint{-0.000000in}{0.000000in}}{%
\pgfpathmoveto{\pgfqpoint{-0.000000in}{0.000000in}}%
\pgfpathlineto{\pgfqpoint{-0.048611in}{0.000000in}}%
\pgfusepath{stroke,fill}%
}%
\begin{pgfscope}%
\pgfsys@transformshift{0.374692in}{2.020142in}%
\pgfsys@useobject{currentmarker}{}%
\end{pgfscope}%
\end{pgfscope}%
\begin{pgfscope}%
\definecolor{textcolor}{rgb}{0.000000,0.000000,0.000000}%
\pgfsetstrokecolor{textcolor}%
\pgfsetfillcolor{textcolor}%
\pgftext[x=0.100000in, y=1.971917in, left, base]{\color{textcolor}\rmfamily\fontsize{10.000000}{12.000000}\selectfont \(\displaystyle {0.0}\)}%
\end{pgfscope}%
\begin{pgfscope}%
\pgfsetbuttcap%
\pgfsetroundjoin%
\definecolor{currentfill}{rgb}{0.000000,0.000000,0.000000}%
\pgfsetfillcolor{currentfill}%
\pgfsetlinewidth{0.803000pt}%
\definecolor{currentstroke}{rgb}{0.000000,0.000000,0.000000}%
\pgfsetstrokecolor{currentstroke}%
\pgfsetdash{}{0pt}%
\pgfsys@defobject{currentmarker}{\pgfqpoint{-0.048611in}{0.000000in}}{\pgfqpoint{-0.000000in}{0.000000in}}{%
\pgfpathmoveto{\pgfqpoint{-0.000000in}{0.000000in}}%
\pgfpathlineto{\pgfqpoint{-0.048611in}{0.000000in}}%
\pgfusepath{stroke,fill}%
}%
\begin{pgfscope}%
\pgfsys@transformshift{0.374692in}{2.389454in}%
\pgfsys@useobject{currentmarker}{}%
\end{pgfscope}%
\end{pgfscope}%
\begin{pgfscope}%
\definecolor{textcolor}{rgb}{0.000000,0.000000,0.000000}%
\pgfsetstrokecolor{textcolor}%
\pgfsetfillcolor{textcolor}%
\pgftext[x=0.100000in, y=2.341228in, left, base]{\color{textcolor}\rmfamily\fontsize{10.000000}{12.000000}\selectfont \(\displaystyle {0.5}\)}%
\end{pgfscope}%
\begin{pgfscope}%
\pgfsetbuttcap%
\pgfsetroundjoin%
\definecolor{currentfill}{rgb}{0.000000,0.000000,0.000000}%
\pgfsetfillcolor{currentfill}%
\pgfsetlinewidth{0.803000pt}%
\definecolor{currentstroke}{rgb}{0.000000,0.000000,0.000000}%
\pgfsetstrokecolor{currentstroke}%
\pgfsetdash{}{0pt}%
\pgfsys@defobject{currentmarker}{\pgfqpoint{-0.048611in}{0.000000in}}{\pgfqpoint{-0.000000in}{0.000000in}}{%
\pgfpathmoveto{\pgfqpoint{-0.000000in}{0.000000in}}%
\pgfpathlineto{\pgfqpoint{-0.048611in}{0.000000in}}%
\pgfusepath{stroke,fill}%
}%
\begin{pgfscope}%
\pgfsys@transformshift{0.374692in}{2.758765in}%
\pgfsys@useobject{currentmarker}{}%
\end{pgfscope}%
\end{pgfscope}%
\begin{pgfscope}%
\definecolor{textcolor}{rgb}{0.000000,0.000000,0.000000}%
\pgfsetstrokecolor{textcolor}%
\pgfsetfillcolor{textcolor}%
\pgftext[x=0.100000in, y=2.710540in, left, base]{\color{textcolor}\rmfamily\fontsize{10.000000}{12.000000}\selectfont \(\displaystyle {1.0}\)}%
\end{pgfscope}%
\begin{pgfscope}%
\pgfsetrectcap%
\pgfsetmiterjoin%
\pgfsetlinewidth{0.803000pt}%
\definecolor{currentstroke}{rgb}{0.000000,0.000000,0.000000}%
\pgfsetstrokecolor{currentstroke}%
\pgfsetdash{}{0pt}%
\pgfpathmoveto{\pgfqpoint{0.374692in}{2.020142in}}%
\pgfpathlineto{\pgfqpoint{0.374692in}{2.758765in}}%
\pgfusepath{stroke}%
\end{pgfscope}%
\begin{pgfscope}%
\pgfsetrectcap%
\pgfsetmiterjoin%
\pgfsetlinewidth{0.803000pt}%
\definecolor{currentstroke}{rgb}{0.000000,0.000000,0.000000}%
\pgfsetstrokecolor{currentstroke}%
\pgfsetdash{}{0pt}%
\pgfpathmoveto{\pgfqpoint{5.716193in}{2.020142in}}%
\pgfpathlineto{\pgfqpoint{5.716193in}{2.758765in}}%
\pgfusepath{stroke}%
\end{pgfscope}%
\begin{pgfscope}%
\pgfsetrectcap%
\pgfsetmiterjoin%
\pgfsetlinewidth{0.803000pt}%
\definecolor{currentstroke}{rgb}{0.000000,0.000000,0.000000}%
\pgfsetstrokecolor{currentstroke}%
\pgfsetdash{}{0pt}%
\pgfpathmoveto{\pgfqpoint{0.374692in}{2.020142in}}%
\pgfpathlineto{\pgfqpoint{5.716193in}{2.020142in}}%
\pgfusepath{stroke}%
\end{pgfscope}%
\begin{pgfscope}%
\pgfsetrectcap%
\pgfsetmiterjoin%
\pgfsetlinewidth{0.803000pt}%
\definecolor{currentstroke}{rgb}{0.000000,0.000000,0.000000}%
\pgfsetstrokecolor{currentstroke}%
\pgfsetdash{}{0pt}%
\pgfpathmoveto{\pgfqpoint{0.374692in}{2.758765in}}%
\pgfpathlineto{\pgfqpoint{5.716193in}{2.758765in}}%
\pgfusepath{stroke}%
\end{pgfscope}%
\begin{pgfscope}%
\pgfsetbuttcap%
\pgfsetmiterjoin%
\definecolor{currentfill}{rgb}{1.000000,1.000000,1.000000}%
\pgfsetfillcolor{currentfill}%
\pgfsetfillopacity{0.800000}%
\pgfsetlinewidth{1.003750pt}%
\definecolor{currentstroke}{rgb}{0.800000,0.800000,0.800000}%
\pgfsetstrokecolor{currentstroke}%
\pgfsetstrokeopacity{0.800000}%
\pgfsetdash{}{0pt}%
\pgfpathmoveto{\pgfqpoint{0.455678in}{2.816613in}}%
\pgfpathlineto{\pgfqpoint{5.635207in}{2.816613in}}%
\pgfpathquadraticcurveto{\pgfqpoint{5.658346in}{2.816613in}}{\pgfqpoint{5.658346in}{2.839751in}}%
\pgfpathlineto{\pgfqpoint{5.658346in}{2.988620in}}%
\pgfpathquadraticcurveto{\pgfqpoint{5.658346in}{3.011759in}}{\pgfqpoint{5.635207in}{3.011759in}}%
\pgfpathlineto{\pgfqpoint{0.455678in}{3.011759in}}%
\pgfpathquadraticcurveto{\pgfqpoint{0.432539in}{3.011759in}}{\pgfqpoint{0.432539in}{2.988620in}}%
\pgfpathlineto{\pgfqpoint{0.432539in}{2.839751in}}%
\pgfpathquadraticcurveto{\pgfqpoint{0.432539in}{2.816613in}}{\pgfqpoint{0.455678in}{2.816613in}}%
\pgfpathlineto{\pgfqpoint{0.455678in}{2.816613in}}%
\pgfpathclose%
\pgfusepath{stroke,fill}%
\end{pgfscope}%
\begin{pgfscope}%
\pgfsetbuttcap%
\pgfsetmiterjoin%
\definecolor{currentfill}{rgb}{0.121569,0.466667,0.705882}%
\pgfsetfillcolor{currentfill}%
\pgfsetlinewidth{0.000000pt}%
\definecolor{currentstroke}{rgb}{0.000000,0.000000,0.000000}%
\pgfsetstrokecolor{currentstroke}%
\pgfsetstrokeopacity{0.000000}%
\pgfsetdash{}{0pt}%
\pgfpathmoveto{\pgfqpoint{0.478817in}{2.884495in}}%
\pgfpathlineto{\pgfqpoint{0.710206in}{2.884495in}}%
\pgfpathlineto{\pgfqpoint{0.710206in}{2.965481in}}%
\pgfpathlineto{\pgfqpoint{0.478817in}{2.965481in}}%
\pgfpathlineto{\pgfqpoint{0.478817in}{2.884495in}}%
\pgfpathclose%
\pgfusepath{fill}%
\end{pgfscope}%
\begin{pgfscope}%
\definecolor{textcolor}{rgb}{0.000000,0.000000,0.000000}%
\pgfsetstrokecolor{textcolor}%
\pgfsetfillcolor{textcolor}%
\pgftext[x=0.802761in,y=2.884495in,left,base]{\color{textcolor}\rmfamily\fontsize{8.330000}{9.996000}\selectfont No Privacy}%
\end{pgfscope}%
\begin{pgfscope}%
\pgfsetbuttcap%
\pgfsetmiterjoin%
\definecolor{currentfill}{rgb}{1.000000,0.498039,0.054902}%
\pgfsetfillcolor{currentfill}%
\pgfsetlinewidth{0.000000pt}%
\definecolor{currentstroke}{rgb}{0.000000,0.000000,0.000000}%
\pgfsetstrokecolor{currentstroke}%
\pgfsetstrokeopacity{0.000000}%
\pgfsetdash{}{0pt}%
\pgfpathmoveto{\pgfqpoint{4.899480in}{2.884495in}}%
\pgfpathlineto{\pgfqpoint{5.130869in}{2.884495in}}%
\pgfpathlineto{\pgfqpoint{5.130869in}{2.965481in}}%
\pgfpathlineto{\pgfqpoint{4.899480in}{2.965481in}}%
\pgfpathlineto{\pgfqpoint{4.899480in}{2.884495in}}%
\pgfpathclose%
\pgfusepath{fill}%
\end{pgfscope}%
\begin{pgfscope}%
\definecolor{textcolor}{rgb}{0.000000,0.000000,0.000000}%
\pgfsetstrokecolor{textcolor}%
\pgfsetfillcolor{textcolor}%
\pgftext[x=5.223424in,y=2.884495in,left,base]{\color{textcolor}\rmfamily\fontsize{8.330000}{9.996000}\selectfont Privacy}%
\end{pgfscope}%
\end{pgfpicture}%
\makeatother%
\endgroup%

%% file: relatedwork.tex

\section{Related Work}\label{ch:relatedwork}

This section presents prior papers that compare or classify generative models for synthetic data. In \autoref{sec:rw_model_comparisons}, we provide an overview of previous works that selectively comprise and compare models in specific domains like healthcare privacy or graph generation. In \autoref{sec:rw_model_classifications}, we focus on literature that aims to organize approaches for \ac{SDG} comprehensively by specific aspects to provide an overview or guidance to novice users.

\subsection{Domain- or Model-Specific Overviews and Comparisons}\label{sec:rw_model_comparisons}

Fernández and Vico \cite{fernandez2013ai} summarize the research done in the field of algorithmic music composition to provide a comprehensive survey of various generation approaches: Grammars, knowledge-based systems, Markov chains, artificial neural networks, evolutionary (genetic) algorithms, and cellular automata.

Goodfellow et al. \cite{goodfellow2016nips} create a tutorial on \acp{GAN}. They show how these models work and how they can be improved for specific tasks. Further, different specialized applications of \acp{GAN} in literature are shown and explained. \acp{GAN} are also compared to other approaches, such as belief networks, autoencoders, and Boltzmann machines, regarding how the likelihood of generated samples can be computed.

Briot et al. \cite{briot2017deep} issue a large-scale survey on music generation via deep learning. They cover different types of musical content (melody, polyphony, performed by humans/machines), representations (Formats: MIDI, piano roll, text. Encodings: Scalar, one-hot, many-hot), strategies (single-step/iterative feed-forward, sampling, etc.), challenges (e.g., variability, interactivity, originality) and deep neural network architectures. In this work,  standard feed-forward and recurrent neural networks, autoencoders, and \ac{RBM} architectures are covered and compared with the five criteria above.

He et al. \cite{he2017deep} provide an overview of different models and benchmarks for image captioning. They capture end-to-end encoder-decoder frameworks like \ac{CNN}-\ac{RNN} combinations and also cover an attention mechanism applied to subregions of the image to improve the decoding. Other approaches are compositional frameworks that generate and arrange tags to generate captions, \acp{GAN}, autoencoders, and \ac{RL}. The authors aim to highlight the importance of image-to-text generation and encourage newcomers to contribute to this topic.

J{\o}rgensen et al. \cite{jorgensen2018deep} review the \ac{VAE} as an alternative to quantum-mechanical computations with lower computational cost to generate new molecular structures and predict their properties. They also discuss approaches to improve the realism of generated data by using grammar-based instead of character-based encoders and decoders and propose other models like \acp{GAN} and \ac{RL} agents.

Korakakis et al. \cite{korakakis2018short} provide an overview of virtual environments and their usage in literature. They illustrate how synthetic data obtained from CAD renders and video games can be used to improve object detection or classification models or train \ac{RL} agents. The authors identify a trend of steadily increasing usage of virtual environments for cheap yet effective model learning, especially for computer vision tasks.

Gaidon et al. \cite{gaidon2018reasonable} present nine papers exploring novel ways of generating and using synthetic data for computer vision tasks. They summarize these nine works and also cover some of the problems encountered by the literature, such as generation challenges or the ``sim2real'' domain gap, which describes the problem of fitting models trained on synthetic data to real applications despite the generated data often being different in some way (e.g., lack of photo-realism). Nevertheless, more and higher-quality synthetic data might help overcome the limitations of current computer vision algorithms, according to the authors.

Kulkarni et al. \cite{kulkarni2018generative} evaluate the performance of \acp{RNN}, \acp{GAN}, and copulas on the generation of synthetic human mobility trajectories in terms of privacy preservation, long-range dependencies, the statistical similarity of the distributions, training and generation time, circadian rhythms, and semantic and geographic similarity. They conclude that copulas have preferable statistical and semantic properties over the neural network models, which also consume more time and are less computationally efficient. Further, they assess that a utility metric to measure and maximize privacy and statistical similarity jointly could improve the usability of synthetic trajectories, but is not yet available.

Hong et al. \cite{hong2019generative} give a detailed introduction to \acp{GAN} and various recently proposed objective functions for them. Further, the combination of a \ac{GAN} with an autoencoder is discussed. Finally, multiple applications of \acp{GAN} in different tasks and fields are covered, and the pros and cons of this model type, such as convergence towards an optimal solution, are highlighted.

Yi et al. \cite{yi2019generative} review \acp{GAN} and their applications in medical imaging. They collect literature using \acp{GAN} for medical purposes such as image synthesis, segmentation, and reconstruction/repair and classify them according to \ac{GAN} method (e.g., pix2pix \cite{isola2017image}), adversarial loss type and quantitative measures (e.g., Wasserstein distance \cite{arjovsky2017wasserstein}) used.

Iqbal et al. \cite{iqbal2020text} survey deep learning text generation models and the progress made from 2015 onwards. They focus on \acp{RNN} such as \acp{LSTM}, \acp{GRU} and bidirectional \acp{RNN}, \acp{CNN}, \acp{VAE}, and \acp{GAN} in combination with \ac{RL}. Various representations (Word2Vec, Glove, FastText), optimization techniques (stochastic gradient descent, RMSProp, AdaGrad, Adam), activation functions (Sigmoid, ReLu), and evaluation methods (Rouge, BLEU) for text generation models are also introduced.

Tsirikoglou et al. \cite{tsirikoglou2020survey} collect and compare different image synthesis and augmentation methods. They identify that the visual data generation pipeline consists of two parts: \textit{Content/scene generation}, which means generating the features of the virtual environment, and \textit{rendering}, which simulates the light transport and perception of sensors. Further, synthetic visual training data has four requirements to be useful: Feature variation and coverage, domain realism, automatic generation of annotations and meta-data, and scalability to large numbers of data points. Over 40 generative models from recent literature are categorized by their modeling and rendering approach and compared regarding image quality and what tasks they are applied to.

Guo et al. \cite{guo2020systematic} extensively cover and analyze the recent literature of deep generative models for graph generation, including \acp{BN}, \acp{VAE}, \acp{GAN}, \acp{RNN}, flow-based learning and \ac{RL}. They provide taxonomies of models for conditional and unconditional graph generation and describe the evaluation metrics applicable in this domain. Finally, the application fields of deep graph generation, such as the analysis of interaction dynamics in social networks, the creation of molecules, or anomaly detection, are discussed.

Seib et al. \cite{seib2020mixing} discuss different techniques to improve neural network training results on computer vision tasks in urban and traffic environments without acquiring additional real-world data. The topics explored are data augmentation, transfer learning, which describes the fine-tuning of pre-trained models for another task with few data samples, and approaches to generating synthetic data. Different 3D engines (Unreal Engine and Unity) and video games (GTA V) and their usage in literature are covered. A future outlook towards \acp{GAN} and their image-to-image translation capabilities is also provided.

Abufadda et al. \cite{abufadda2021survey} examine and summarize related works about \ac{SDG} in the healthcare domain, especially presenting many \ac{GAN} approaches. For each paper, they highlight the research field, used methods and results, and indicate, whether the models are suitable for their task. The authors conclude that generalized solutions for utility and efficiency evaluations would improve the model selection process.

Dankar et al. \cite{dankar2022multi} propose an overall utility score for masked synthetic data sets, where features of the original data have to be kept secret. For that, available metrics are categorized by the measure they aim to preserve, and from each of the four categories (attribute, bivariate, population, and application fidelity), one suitable metric is chosen to determine the final utility of a generative model. The utility measurement approach is evaluated on four recent models and 19 data sets with different sizes and features. They conclude that each privacy-enhancing technology decreases data utility, and the acceptable or necessary decrease to achieve privacy is unknown.

All the literature above provides useful insight into specific application fields of synthetic data and model types. However, the scope of the individual works is usually limited to one domain and a small selection of models and architectures, making them rather unsuitable as a comprehensive introduction to \ac{SDG}.

\subsection{Comprehensive Reviews}\label{sec:rw_model_classifications}

Turhan et al. \cite{turhan2018recent} conduct a comprehensive review of generative models, especially focussing on deep learning models like \acp{GAN} and autoencoders for image generation. They highlight use cases for these models and classify them into five categories: Unsupervised fundamental models (\ac{RBM}, \ac{DBN} and \ac{DBM}), autoencoder-based models, autoregressive models (\acp{CNN} and \acp{RNN}), \ac{GAN}-based models and autoencoder-\ac{GAN} hybrid models. Further, a relation diagram of all presented models is compiled that is particularly useful to beginners in this topic.

Oussidi et al. \cite{oussidi2018deep} consolidate promising types of generative models like \acp{RBM}, \acp{DBN}, \acp{DBM}, \acp{VAE} and \acp{GAN} and describe the three models PixelRNN \cite{oord2016pixel}, DRAW \cite{gregor2015draw} and NADE \cite{larochelle2011neural} in detail. They also put generative models into two categories:

\begin{description}
    \item[Cost function-based models] Models that optimize parameters based on cost/loss, like autoencoders and \acp{GAN}.
    \item[Energy-based models] The joint probability is defined by an energy function, which measures the compatibility of variable configurations \cite{lecun2006tutorial}. This approach is used by Boltzmann machines and their derivatives.
\end{description}

They study their advantages, limitations, and potential for the future. They find that energy-based models are more complex to combine than directed graphical models like feed-forward neural networks, and deep networks often suffer from vanishing or exploding gradients during training. Hence, the bottom layers barely learn anything, while the top layers quickly reach an optimal state.

Wang et al. \cite{wang2019generative} propose an architecture- and loss-based taxonomy for \acp{GAN} and highlight the significant advances made with them in recent years in the field of computer vision. They further discuss the challenges in these tasks, such as image quality, diversity, and training stability, but also the risks involved in being able to generate high-quality fake data, such as fake evidence of crimes or events.

Harshvardhan et al. \cite{harshvardhan2020comprehensive} compile a comprehensive survey of generative models, highlight some noteworthy contributions from literature, and implement and evaluate each presented model to help the readers pick the best solution for their use case. Their high-level review incorporates \acp{GMM}, \acp{HMM}, \ac{LDA}, Boltzmann machines, \acp{VAE} and \acp{GAN}. Further, the models are also classified by their learning type (un-/semi- and supervised learning) and their model architecture (machine learning or subset deep learning).

Eigenschink et al. \cite{eigenschink2021deep} present a data-driven framework that evaluates synthetic data generation models independently of their internal workings. The authors exclusively cover models for synthetic sequential data to complement previous reviews that only focused on particular data (e.g., time-series, videos, text) or model (e.g., \acp{GAN}) types. The models compared in this work are \acp{RNN}, \acp{CNN}, transformers, autoencoders, autoregressive neural networks, and \acp{GAN}. They are compared in terms of

\begin{description}
    \item[Representativeness] How well the synthetic data captures distributions and dependencies between the distributions, for instance, hair colors and eye distances in a face image data set. 
    \item[Novelty] Do new samples resemble samples from the original data set, or are they new observations from the latent distribution? This is especially important for use cases concerned with privacy, like healthcare, because original samples must not be leaked.
    \item[Realism] Similar to Representativeness, a statistical measure, but on a per-sample level: Often, human judgment is used to determine whether synthetic samples are realistic.
    \item[Diversity] Measure of similarity between individual synthetic data points, for example, the average Euclidean distance to their nearest neighbors \cite{donahue2018adversarial}.
    \item[Coherence] Often implicitly evaluated with realism, coherence describes whether the internal structure of single synthetic data points is consistent. Varying Notes in music, for example, are natural and necessary, but random genre changes are not and indicate bad coherence.
\end{description}

Further, the importance of individual criteria is evaluated in different domains such as \ac{NLP}, audio processing, or anonymization of healthcare data, concluding that representativeness and realism are most important for \ac{NLP}. Speech, music, and video tasks mostly rely on realism and coherence, private synthetic \acp{EHR} in healthcare applications need representativeness, novelty, and realism, and mobility trajectories additionally require coherence.

Nikolenko \cite{nikolenko2021synthetic} released a book about synthetic data and how it is currently used for \ac{ML}. The book mainly covers the computer vision topic, especially the collaboration of deep learning models for classification and \acp{GAN} as synthetic data generators, and also provides a historical perspective on it, but other application fields like computer security, bioinformatics and \ac{NLP} also make an appearance. Common problems of generative models, such as privacy concerns and the challenge of domain shift, are also treated in separate sections, and potential solutions are discussed. Finally, an outlook on potential future improvements for \ac{SDG} is given in the form of \ac{RL} or the incorporation of domain knowledge in generative models.

The works presented here have a more comprehensive view on \ac{SDG} methods, some even providing a historical perspective \cite{nikolenko2021synthetic}. Even though these surveys and reviews are all recent, released in 2018 or later, new approaches like transformers \cite{vaswani2017attention} are often missed entirely, and many of them focus primarily on \acp{GAN} or autoencoders. Also, essential domains like music, which could play an important future role in the film and video game industry, are often forgotten. Another important use of a comprehensive overview is the comparison of approaches and guidance of users to select an appropriate model for their use case, which is only provided extensively by \cite{harshvardhan2020comprehensive} and coarsely by \cite{eigenschink2021deep}. The main difference between our work and the presented works is listed in \autoref{tab:rw:comparision}.

\begin{table}[htb!]
    \centering
    \begin{tabular}{|l|l|r|r|l|l|}
    \hline
    \textbf{Approach} & \textbf{Data Type(s)} & \textbf{Model(s)} & \textbf{\#Works} & \textbf{Investigated Aspects} \\ \hline\hline
    \cite{fernandez2013ai} & Music & 6 & 267 & Creativity \\ \hline
    \cite{briot2017deep} & Music & 7 & 12 & Creativity, Model capabilities \\ \hline
    \cite{iqbal2020text} & Text & 5 & 30 & Similarity to real data \\ \hline
    \cite{turhan2018recent} & Image & 2 & 45 & Model relationships \\ \hline
    \cite{oussidi2018deep} & Image & 5 & 13 & Limitations/Advantages \\ \hline
    \cite{wang2019generative} & Image & 1 & 36 & Performance, Architecture \\ \hline
    \cite{harshvardhan2020comprehensive} & Comprehensive & 6 & 20-30 & Lim./Adv., Perf., Implementation \\ \hline
    \cite{eigenschink2021deep} & Sequential Data & 6 & 17 & Creativity, Data Quality \\ \hline\hline
    \textbf{This article} & \textbf{8} types (15 sub-types) & \textbf{20} (42 sub-types) & \textbf{\works{}} & \textbf{9} (see \autoref{sec:classification_criteria}) \\ \hline
\end{tabular} 
    \caption{Comparison of this survey to others.}
    \label{tab:rw:comparision}
\end{table}

%% file: conclusion.tex
\section{Conclusion}
\label{ch:conclusion}

The surge in applications in \ac{ML} has transformed various fields, but limited training data, expensive acquisition, and privacy laws hinder \ac{ML} model efficacy. \ac{SDG} emerges as a solution, yet the diverse and rapidly evolving landscape of \ac{SDG} models, spanning decades of development, poses a challenge for decision-makers. To this end, we conducted a comprehensive survey of \works{} \ac{SDG} model papers, resulting in the categorization of these models into 20 distinct types (42 sub-types). The classification was performed based on criteria extracted from related work and identified during our survey. The key findings from our comprehensive classification are as follows: 

\begin{itemize}
    \item Computer vision is the most popular application field, and \acp{GAN} are the most popular \ac{SDG} models.
    \item Different data types require different model types. \acp{RNN} and transformers are more suitable for sequential data, while \acp{CNN}, \acp{GAN} or autoencoders are mostly used for data with static size.
    \item We also observed a trend of combining different model types. \acp{GAN} and autoencoders often act as frameworks with \acp{RNN} or \acp{CNN} as building blocks. \acp{RNN} are often combined with \ac{RL} components to guide the generation process.
    \item Our analysis revealed challenges in performance evaluation, citing the absence of standardized metrics and datasets. That is, a common set of metrics, datasets, and reference models is needed to enhance comparability among \ac{SDG} models.
    \item We identified a nascent stage in the development of privacy-preserving data generation, where simplistic models like Markov chains, \ac{BN}, and genetic algorithms are prevalent, with \ac{GAN} being the only more complex neural network-based model.
\end{itemize}

We are convinced that our work provides a valuable resource for researchers entering the field, aiding them in selecting suitable approaches for their specific purposes. Furthermore, it is crucial to underscore the necessity for future research to (i) delve into the training and sampling costs of \ac{SDG} models for a more comprehensive classification and (ii) establish a systematic evaluation approach that enhances the overall understanding, the usability, and the comparability among \ac{SDG} models.

%% file: acronyms_sorted.tex

\section{Acronyms}
\label{sec:acronyms}

\begin{acronym}
	\acro{AAE}[AAE]{Adversarial Autoencoder}
	\acro{AF}[AF]{Autoregressive Flow}
	\acro{AI}[AI]{Artificial Intelligence}
	\acro{AIC}[AIC]{Akaike information criterion}
	\acro{ALI}[ALI]{Adversarially Learned Inference}
	\acro{BIC}[BIC]{Bayesian information criterion}
	\acro{BN}[BN]{Bayesian Network}
	\acro{BPTT}[BPTT]{Backpropagation Through Time}
	\acro{BiGAN}[BiGAN]{Bidirectional GAN}
	\acro{C2ST}[C2ST]{Classifier Two Sample Test}
	\acro{CAD}[CAD]{Computer Aided Design}
	\acro{CAE}[CAE]{Contractive Autoencoder}
	\acro{CAN}[CAN]{Creative Adversarial Network}
	\acro{CDBN}[CDBN]{Conditional Deep Belief Network}
	\acro{CNN}[CNN]{Convolutional Neural Network}
	\acro{CRBM}[CRBM]{Conditional Restricted Boltzmann Machine}
	\acro{CT}[CT]{Computed Tomography}
	\acro{DAE}[DAE]{Denoising Autoencoder}
	\acro{DAG}[DAG]{Directed Acyclic Graph}
    \acro{DL}[DL]{Deep Learning}
	\acro{DBM}[DBM]{Deep Boltzmann Machine}
	\acro{DBN}[DBN]{Deep Belief Network}
	\acro{DCGAN}[DCGAN]{Deep Convolutional GAN}
	\acro{DDPM}[DDPM]{Denoising Diffusion Probabilistic Model}
	\acro{DLGM}[DLGM]{Deep Latent Gaussian Model}
    \acro{ECG}[ECG]{Electrocardiogram}
	\acro{EHR}[EHR]{Electronic Health Record}
	\acro{ELBO}[ELBO]{Evidence Lower-Bound}
	\acro{FID}[FID]{Fréchet Inception Distance}
	\acro{GA}[GA]{Genetic Algorithm}
	\acro{GAE}[GAE]{Gated Autoencoder}
	\acro{GAN}[GAN]{Generative Adversarial Net}
	\acro{GMM}[GMM]{Gaussian Mixture Model}
	\acro{GMMN}[GMMN]{Generative Moment Matching Network}
	\acro{GNN}[GNN]{Graph Neural Network}
	\acro{GPU}[GPU]{Graphics Processing Unit}
	\acro{GRU}[GRU]{Gated Recurrent Unit}
	\acro{GSN}[GSN]{Generative Stochastic Network}
	\acro{HMM}[HMM]{Hidden Markov Model}
	\acro{HSI}[HSI]{Hyperspectral Image}
	\acro{IS}[IS]{Inception Score}
	\acro{KDE}[KDE]{Kernel Density Estimators}
	\acro{LDA}[LDA]{Latent Dirichlet Allocation}
	\acro{LRCN}[LRCN]{Long-term Recurrent Convolutional Network}
	\acro{LSTM}[LSTM]{Long Short-Term Memory}
	\acro{MCMC}[MCMC]{Markov chain Monte Carlo}
	\acro{ML}[ML]{Machine Learning}
	\acro{MLP}[MLP]{Multilayer Perceptron}
	\acro{MMD}[MMD]{Maximum Mean Discrepancy}
	\acro{NADE}[NADE]{Neural Autoregressive Distribution Estimator}
	\acro{NLL}[NLL]{Negative Log-Likelihood}
	\acro{NLP}[NLP]{Natural Language Processing}
	\acro{ODE}[ODE]{Ordinary Differential Equation}
	\acro{PPG}[PPG]{Photoplethysmogram}
	\acro{PPGN}[PPGN]{Plug \& Play Generative Network}
	\acro{PSD}[PSD]{Private Spatial Decomposition}
	\acro{RBM}[RBM]{Restricted Boltzmann Machine}
	\acro{RL}[RL]{Reinforcement Learning}
	\acro{RNN}[RNN]{Recurrent Neural Network}
	\acro{RTRBM}[RTRBM]{Recurrent Temporal Boltzmann Machine}
	\acro{ReLU}[ReLU]{rectified linear unit}
	\acro{SDG}[SDG]{Synthetic Data Generation}
	\acro{SLAM}[SLAM]{Simultaneous Localization and Mapping}
	\acro{SRTRBM}[SRTRBM]{Structured Recurrent Temporal Boltzmann Machine}
	\acro{SVHN}[SVHN]{Street View House Numbers}
	\acro{TFD}[TFD]{Toronto Face Database}
	\acro{TRBM}[TRBM]{Temporal Restricted Boltzmann Machine}
	\acro{VAE}[VAE]{Variational Autoencoder}
	\acro{VCAE}[VCAE]{Vine Copula Autoencoder}
	\acro{WAE}[WAE]{Wasserstein Autoencoder}
\end{acronym}